\documentclass[runningheads]{llncs}
\usepackage{graphicx}

\usepackage{tikz}
\usepackage{comment}
\usepackage{amsmath,amssymb} 
\usepackage{color}
\usepackage[T1]{fontenc}
\def\doi#1{\href{https://doi.org/\detokenize{#1}}{\url{https://doi.org/\detokenize{#1}}}}
\usepackage{multirow}

\usepackage{hyperref}
\usepackage{nicematrix}
\usepackage{caption}
\usepackage{subcaption} 
\usepackage{mathtools} 
\usepackage[accsupp]{axessibility}  

\newcommand{\myparagraph}[1]{\smallskip\noindent\textbf{#1}}

\begin{document}
\pagestyle{headings}
\mainmatter
\def\ECCVSubNumber{6180}  

\title{Learning Topological Interactions for Multi-Class Medical Image Segmentation} 

\titlerunning{Topological Interactions for Image Segmentation}
%



\author{Saumya Gupta\inst{\star} \and
Xiaoling Hu\thanks{Equal contribution.}\and
James Kaan \and 
Michael Jin \and
Mutshipay Mpoy \and
Katherine Chung \and
Gagandeep Singh \and
Mary Saltz \and
Tahsin Kurc \and
Joel Saltz \and
Apostolos Tassiopoulos \and
Prateek Prasanna \and
Chao Chen
}

\authorrunning{S. Gupta et al.}
%


\institute{Stony Brook University, Stony Brook, New York, USA\\
\email{\{saumya.gupta, xiaoling.hu, chao.chen.1\}@stonybrook.edu}\\
}

\maketitle

\begin{abstract}
Deep learning methods have achieved impressive performance for multi-class medical image segmentation. However, they are limited in their ability to encode topological interactions among different classes (e.g., containment and exclusion). These constraints naturally arise in biomedical images and can be crucial in improving segmentation quality. In this paper, we introduce a novel \textit{topological interaction module} to encode the topological interactions into a deep neural network. The implementation is completely convolution-based and thus can be very efficient. This empowers us to incorporate the constraints into end-to-end training and enrich the feature representation of neural networks. The efficacy of the proposed method is validated on different types of interactions. We also demonstrate the generalizability of the method on both proprietary and public challenge datasets, in both 2D and 3D settings, as well as across different modalities such as CT and Ultrasound. Code is available at: \url{https://github.com/TopoXLab/TopoInteraction}

\end{abstract}

\begin{figure}[t]
\centering 

 \begin{subfigure}{0.14\linewidth}
  \includegraphics[width=1\textwidth]{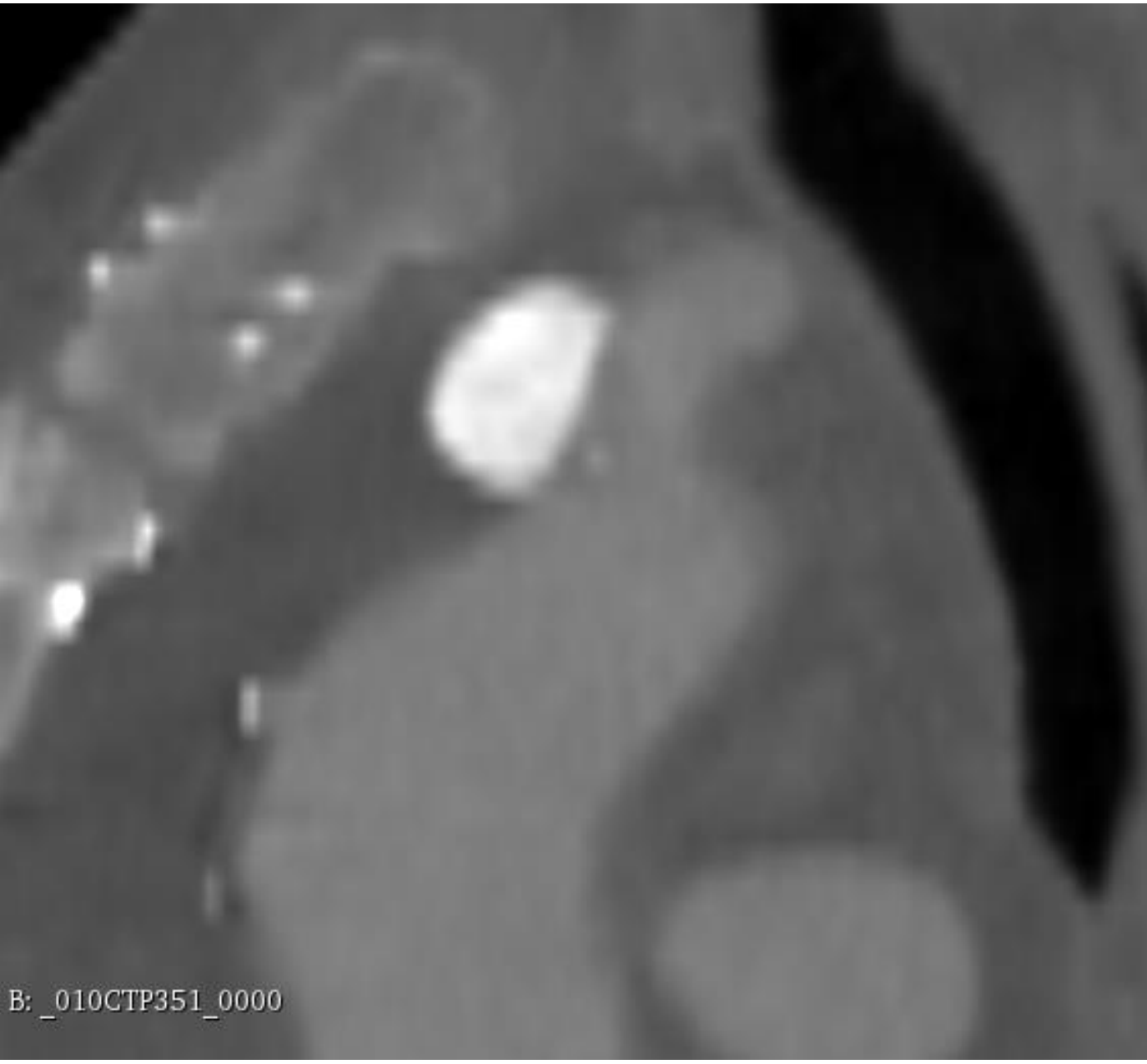}
  \end{subfigure}
  \begin{subfigure}{0.14\linewidth}
     \includegraphics[width=1\textwidth]{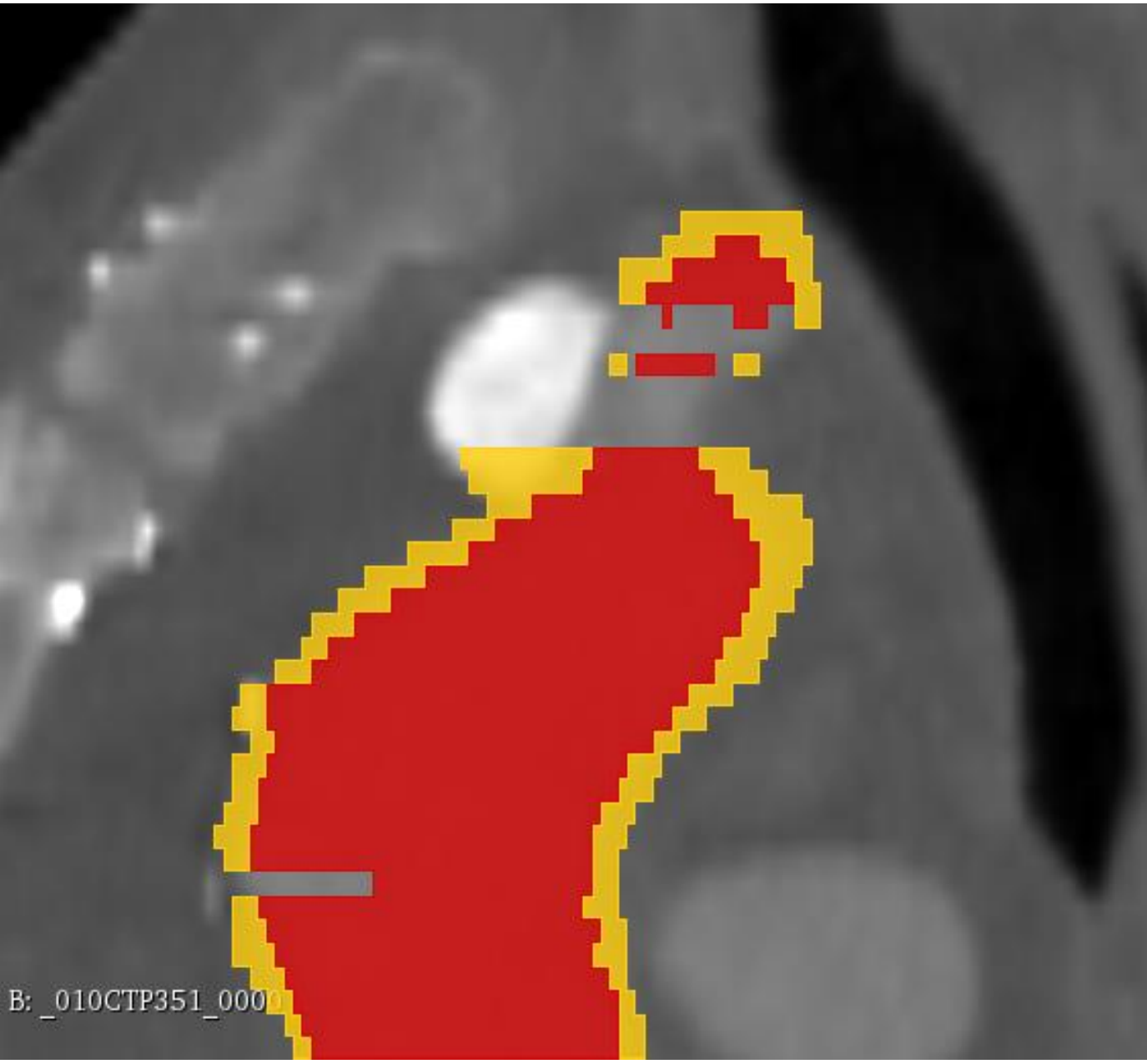}
  \end{subfigure}
    \begin{subfigure}{0.14\linewidth}
     \includegraphics[width=1\textwidth]{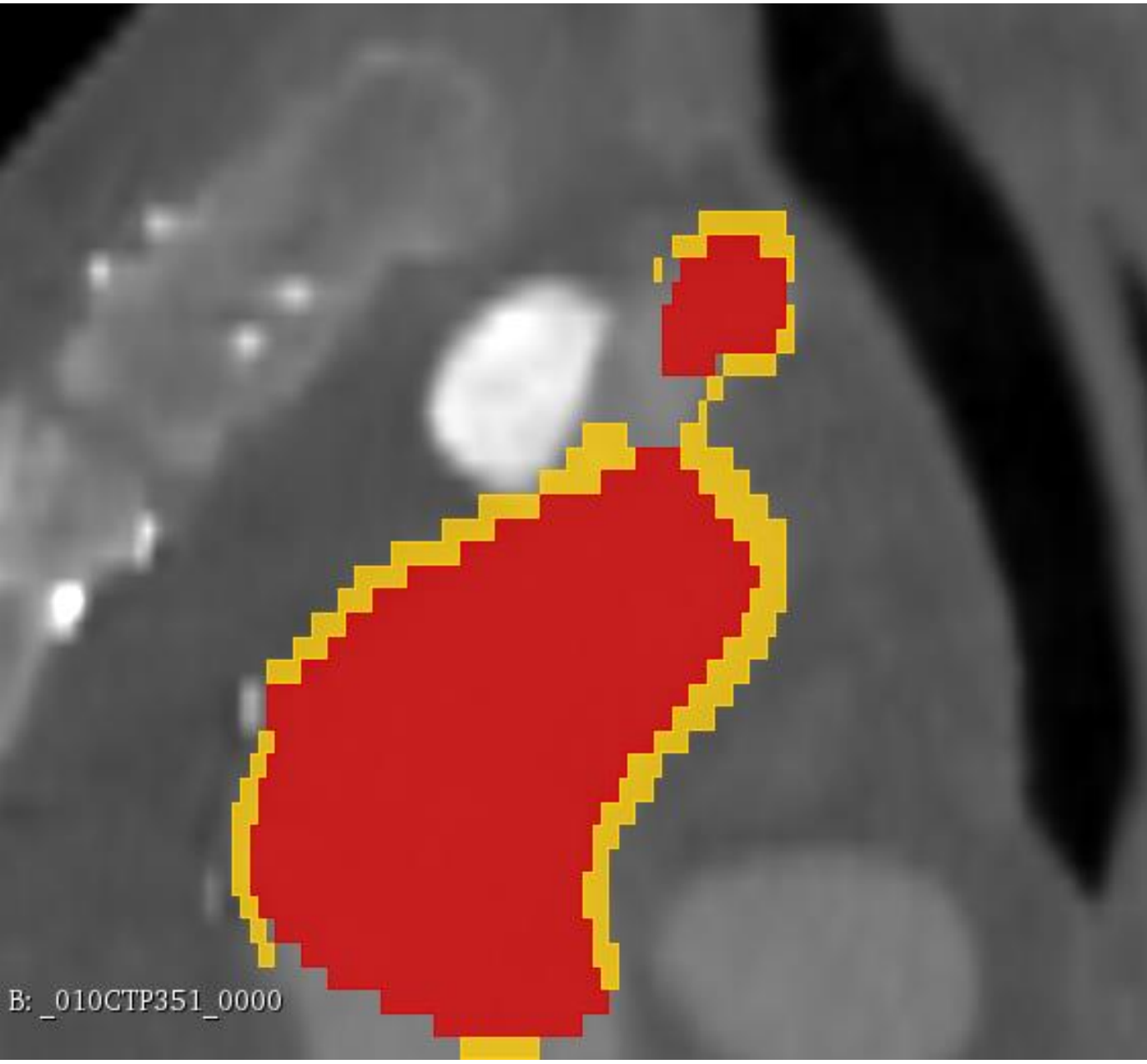}
  \end{subfigure}
    \begin{subfigure}{0.14\linewidth}
     \includegraphics[width=1\textwidth]{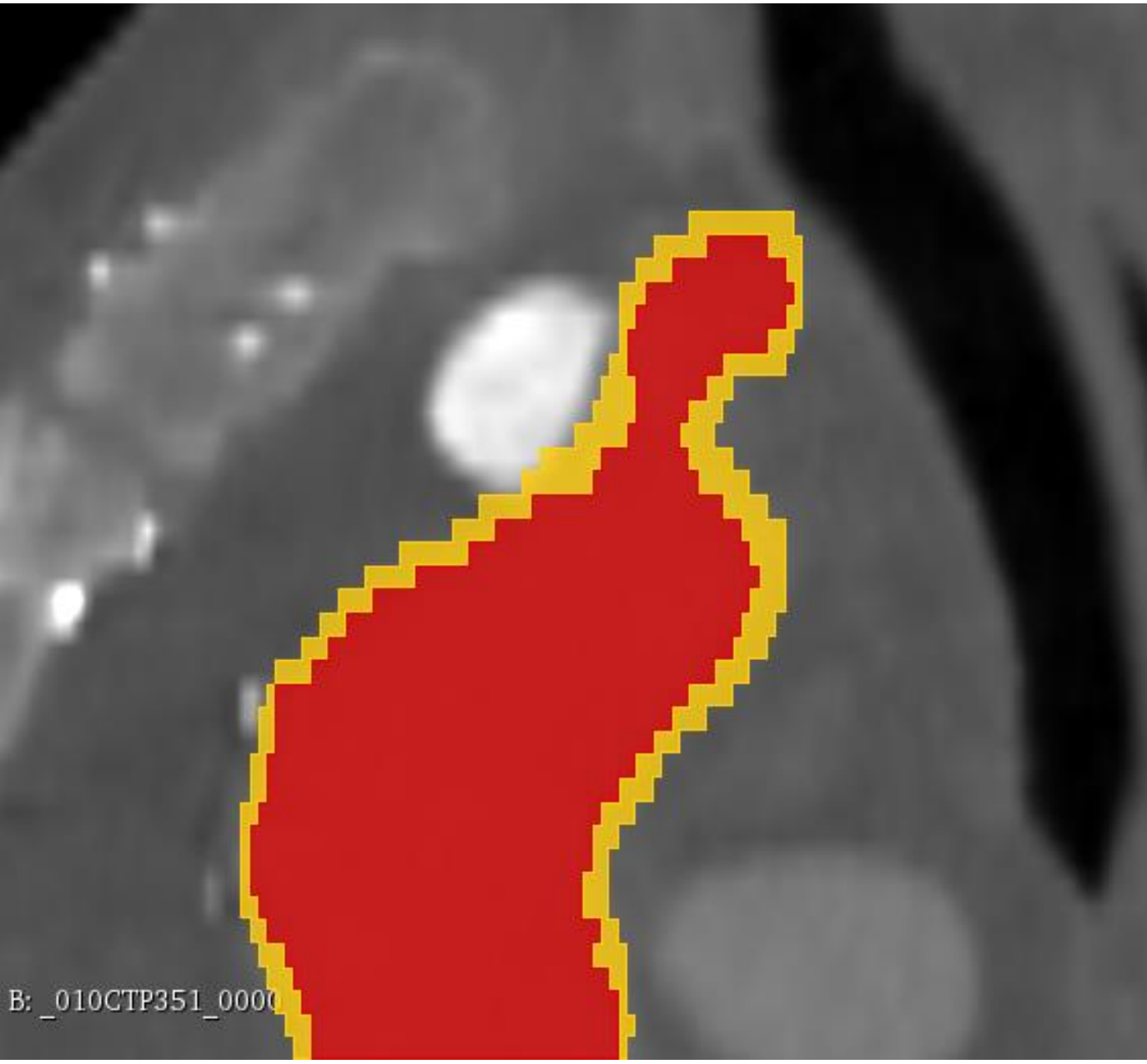}
  \end{subfigure}
  \begin{subfigure}{0.14\linewidth}
     \includegraphics[width=1\textwidth]{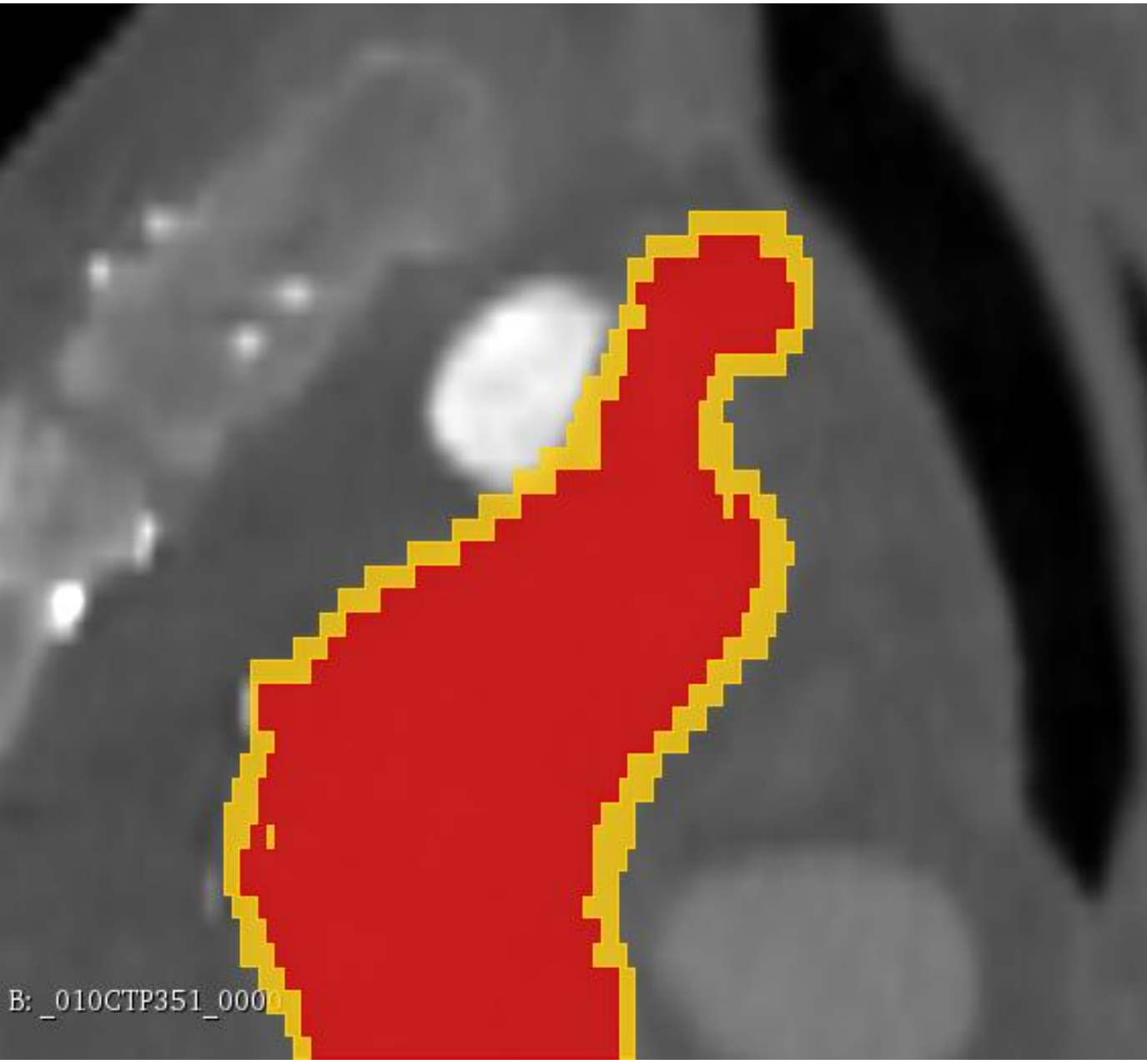}
  \end{subfigure}
  

\begin{subfigure}{0.14\linewidth}
  \includegraphics[width=1\textwidth]{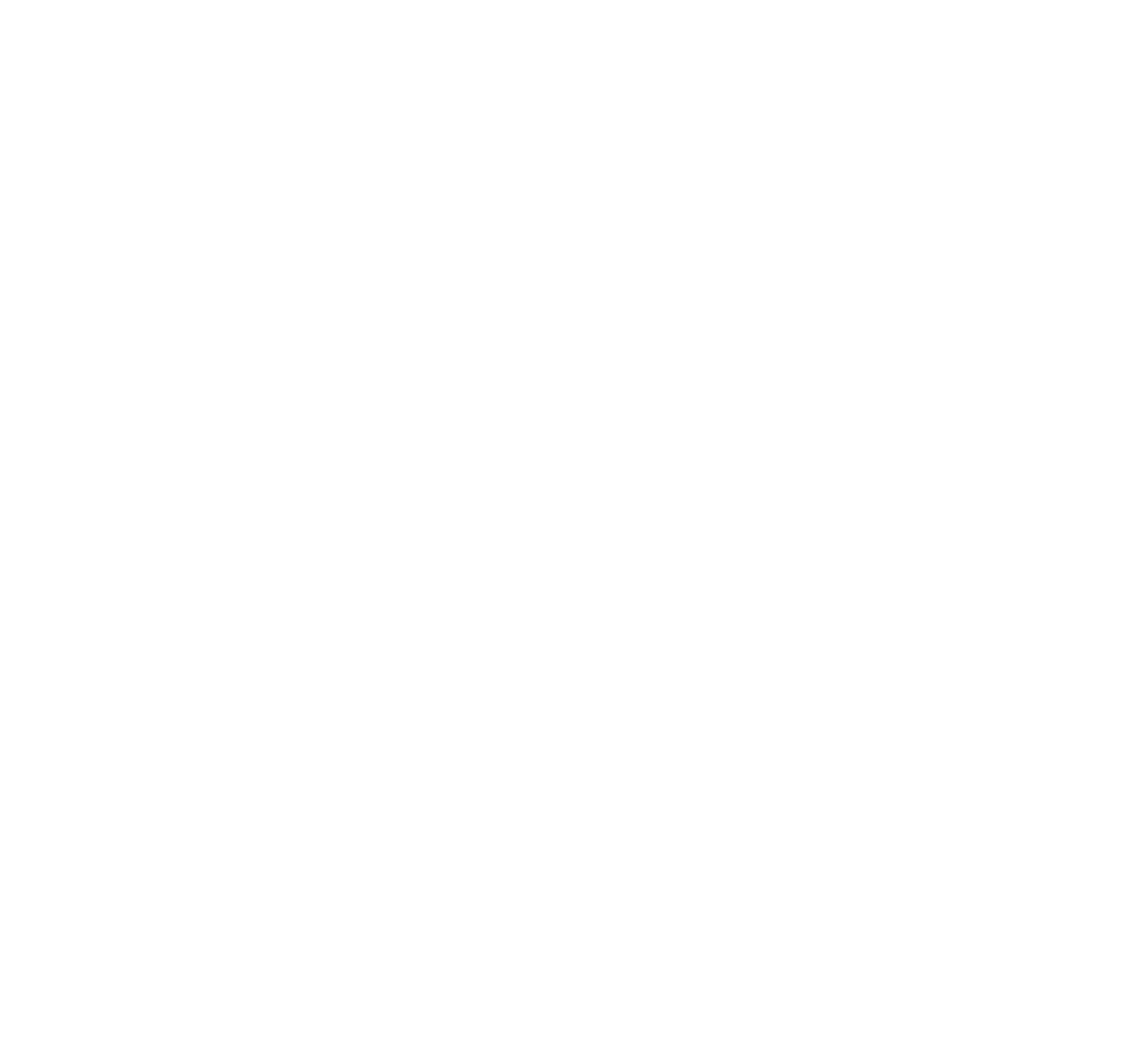}
      \caption{Input}
  \end{subfigure}
  \begin{subfigure}{0.14\linewidth}
     \includegraphics[width=1\textwidth]{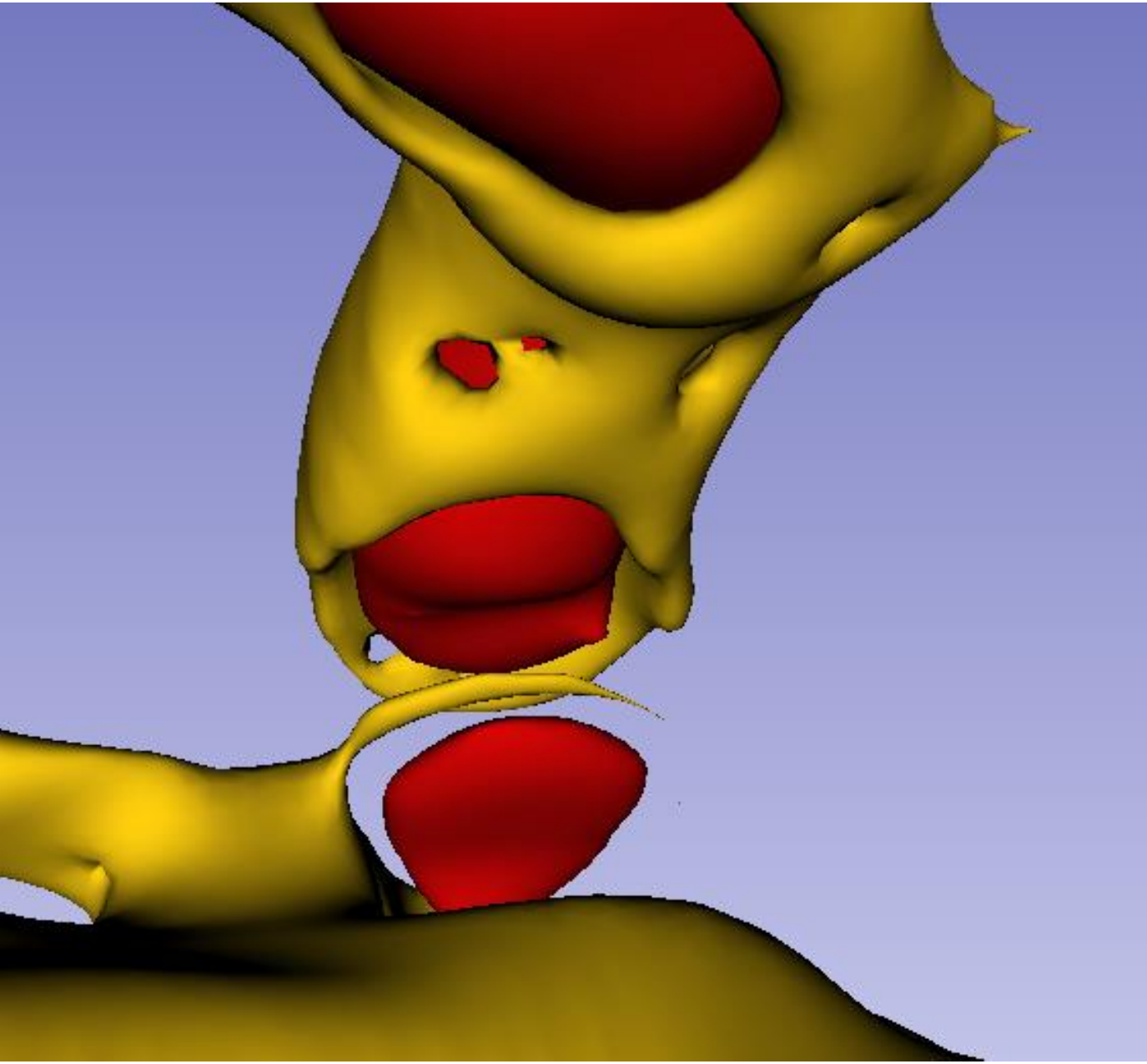}
         \caption{UNet}
  \end{subfigure}
    \begin{subfigure}{0.14\linewidth}
     \includegraphics[width=1\textwidth]{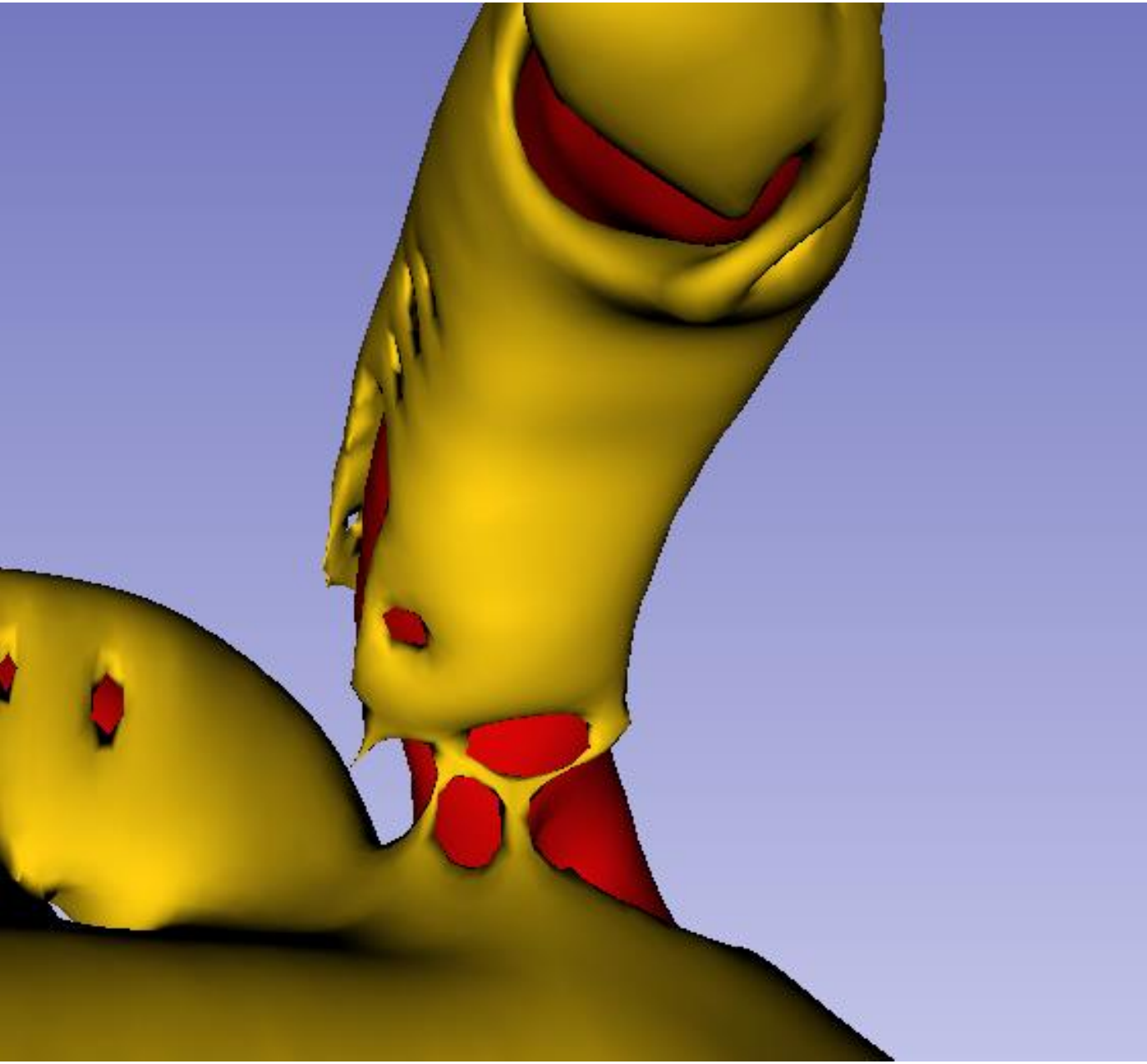}
         \caption{nnUNet}
  \end{subfigure}
    \begin{subfigure}{0.14\linewidth}
     \includegraphics[width=1\textwidth]{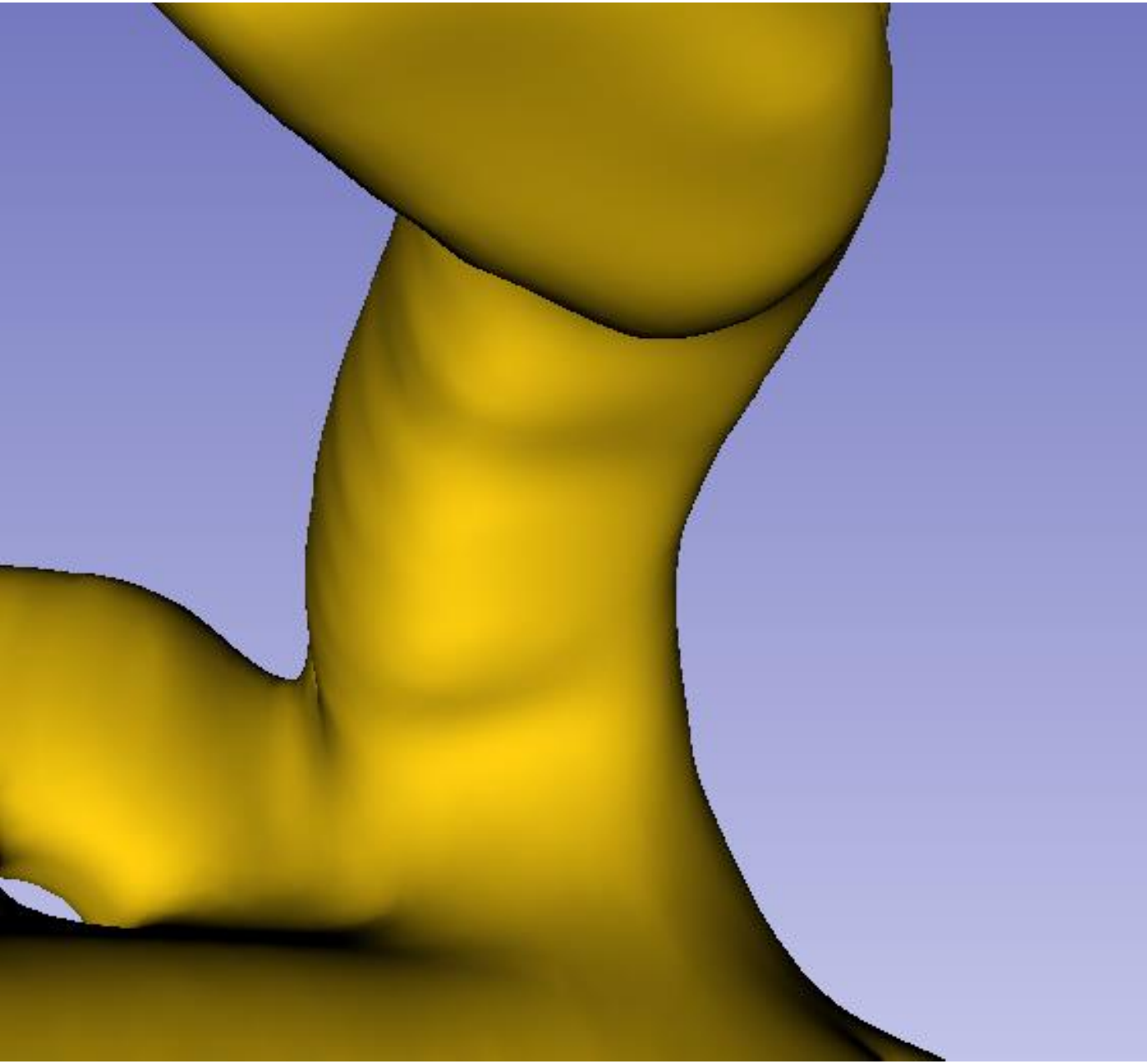}
         \caption{Ours}
  \end{subfigure}
  \begin{subfigure}{0.14\linewidth}
     \includegraphics[width=1\textwidth]{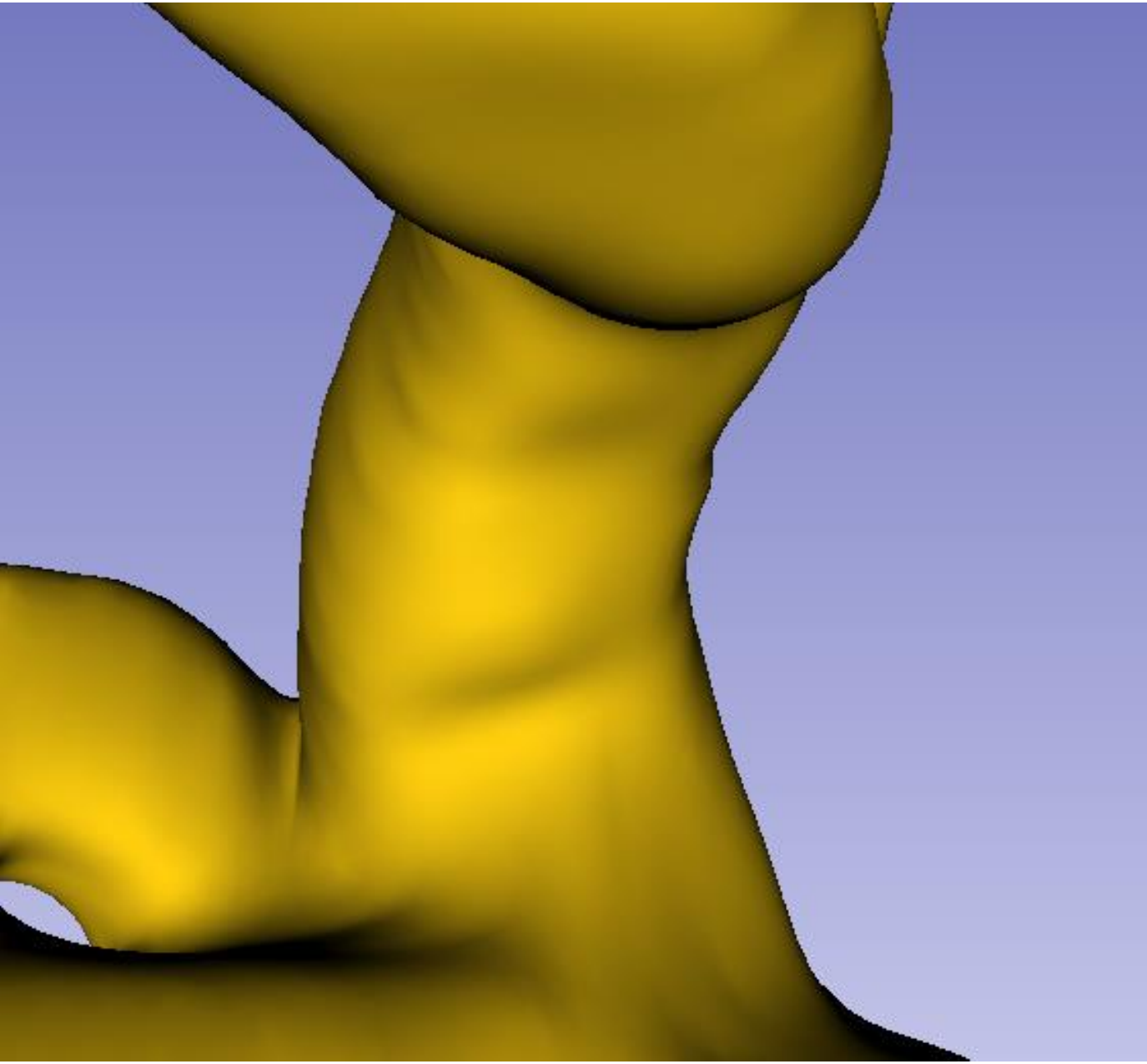}
         \caption{GT}
  \end{subfigure}

\caption{Motivating examples for aorta segmentation. Red and yellow represent aortic lumen and wall, respectively. Anatomically, the lumen is always enclosed by the wall, separated from the background (illustrated in (e) ground truth \textit{GT}). Even strong baselines, e.g., (b) \textit{UNet} and (c) \textit{nnUNet}, fail to respect this anatomically important topological constraint because often the intensity of the wall in the input is similar to that of the background. Our proposed method explicitly encodes the constraint, thereby improving the segmentation quality.  } 
\label{fig:teaser}
\end{figure}

\section{Introduction}

Instead of using hand-crafted features, state-of-the-art deep segmentation methods~\cite{chen2014semantic,chen2017deeplab,chen2017rethinking,he2017mask,long2015fully} learn powerful feature representations automatically and achieve satisfactory performances. However, standard deep neural networks cannot learn global structural constraints regarding semantic labels, which can often be critical in biomedical domains. While existing works mostly focus on encoding the topology of a single label \cite{hu2019topology,hu2021topology,clough2020topological,shit2021cldice}, limited progress has been made addressing the constraints regarding interactions between different labels. Even strong methods (e.g., nnUNet \cite{nnUNet}) may fail to preserve the constraints as they only optimize per-pixel accuracy. For example, in the segmentation of abdominal aorta, we know a priori that the aorta wall always encloses the lumen. Exploiting this constraint can help us segment the wall correctly, providing accurate geometric measures (e.g., wall thickness and aorta volume) for the prediction of aortic aneurysm eruption risk \cite{doweidar2019advances}. See Fig.~\ref{fig:teaser} for an illustration. Another kind of global constraint is mutual exclusion of different labels. For example, in multi-organ segmentation, ensuring different organs to not touch each other can help improve the segmentation quality.

In this paper, we investigate how to help deep neural networks learn these global structural constraints, which we call \emph{topological interactions}, between different semantic labels. 
To encode such interaction constraints into convolutional neural networks is challenging; it is hard to directly encode hard constraints into kernels while keeping them learnable.
Traditional methods~\cite{delong2009globally,ulen2012efficient,nosrati2014local,leon2017multi,kappes2016higher,chen2011enforcing} solve the segmentation problem as a combinatorial optimization problem (e.g., graph-cut or multicut) and encode these topological interactions as constraints of the solution. However, these approaches do not apply to deep neural networks, which do not rely on a global optimization for the inference. Even if one can encode the constrained optimization as a post-processing step, it will be very inefficient. More importantly, the optimization is not differentiable and thus cannot be incorporated into training. 

We propose a novel method to learn the topological interactions for multi-class segmentation tasks. A desirable solution should be efficient. Furthermore, it should be incorporated into training to help the network learn. Our key observation is that a broad class of topological interactions, namely, enclosing and exclusion, boils down to certain impermissible label combinations of adjacent pixels/voxels. Inspired by such observation, we propose a \emph{topological interaction module} that encodes the constraints into a neural network through a series of convolutional operations. Instead of directly encoding the constraint into the convolutional kernels, the proposed module directly identifies locations where the constraints are violated. Our module is extremely efficient due to the convolution-based design. Furthermore, it can naturally be incorporated into the training of neural networks, e.g., through an extra loss penalizing the constraint-violating pixels/voxels.
As shown in Fig.~\ref{fig:teaser}, incorporated with our module, the network can learn to segment aortic walls correctly even when strong baselines, such as nnUNet, fail.

We evaluate the proposed method by performing experiments on both proprietary and public challenge datasets, in both 2D and 3D settings, and across different modalities. The results show that our method is generalizable and can be employed in various scenarios where topological interactions apply. It not only enforces the constraints, but also improves the segmentation quality significantly in standard metrics such as DICE, Hausdorff distance, etc. This is as expected; a network that encodes the constraints also learns a better representation for segmentation.
In summary, our contributions are as follows:

\begin{itemize}
  \item We propose an efficient convolution-based module to encode the topological interactions in a multi-class segmentation setting. 
  \item The proposed module is very efficient and generic. It can be incorporated into any backbone to encode the constraints in an end-to-end training pipeline.
  \item Through extensive experiments on multiple medical imaging datasets, we show our method effectively improves the segmentation quality without increasing computational cost.
\end{itemize}

\section{Related Work}
\label{related}

\myparagraph{Multi-Class Image Segmentation.}
Numerous graph or energy based methods have been proposed to deal with multi-class image segmentation in the pre-deep learning era. Some of these methods integrate fuzzy spatial relations~\cite{colliot2006integration} or encode spatial interactions via inter-object distances~\cite{litvin2005coupled}. Others encode spatial relationships for hierarchical segmentation~\cite{felzenszwalb2010tiered,strekalovskiy2011generalized}. For example, Strekalovskiy et al.~\cite{strekalovskiy2011generalized} enforce geometric constraints by introducing a label ordering constraint. Li et al.~\cite{li2005optimal} propose to segment nested objects with graph-based approaches. Delong et al.~\cite{delong2009globally} propose to encode geometric constraints between different regions into a graph cut framework for multi-class image segmentation.

\myparagraph{Geometric and Topological Constraints.}
Early works, using classic frameworks such as level set or Markov random field, enforce topological or geometric constraints while solving the energy minimization problem~\cite{han2003topology,le2008self,chen2011enforcing,delong2009globally,ulen2012efficient,nosrati2014local,leon2017multi,kappes2016higher}. However, these methods cannot be easily incorporated into the training of deep neural networks. 
In recent years, new methods have been proposed to incorporate geometric/topological constraints into the training of deep neural networks (DNNs)~\cite{hu2019topology,hu2021topology,clough2020topological,shit2021cldice,yang2021topological}. 
These methods enable the DNNs to learn geometry-/topology-aware representations and to deliver better segmentation results. However, all these methods are focusing on the topology, e.g., connections, loops and branches, of a single foreground class. They cannot enforce topological interactions between different  classes. For example, in aorta segmentation, forcing the aortic wall to be a tube in 3D cannot guarantee that the wall contains the lumen and separates it from the background. 
This gap motivates our investigation on encoding the inter-class topological interactions in DNN training.

The method closest to ours is~\cite{bentaieb2016topology}, which we refer to as TopoCRF. It encodes the mutual exclusion constraint as a constraint on the posterior probability (softmax layer output) at each pixel/voxel, without taking neighborhoods into account. Therefore, this approach cannot really exclude the case when adjacent pixels have a forbidden label combination. The explicit construction of $2^c$ constraint-encoding priors for a $c$-class problem is also very expensive and does not scale. Additional methods similar to TopoCRF are~\cite{reddy2019brain} which we refer to as MIDL, and ~\cite{ganaye2019removing} which we refer to as NonAdj. MIDL is a direct application of TopoCRF by simply adding a DICE loss term. NonAdj extends TopoCRF by taking the adjacent pixels into consideration, however, it requires a strong pre-trained model to perform well. Both MIDL and NonAdj focus on modeling joint distributions, and thus suffer from similar issues as TopoCRF.

\section{Methodology}
\label{sec:methodology}

\begin{figure}[t]
\centering 

  \begin{subfigure}{0.45\textwidth}
  \centering 
   \includegraphics[width=.9\linewidth]{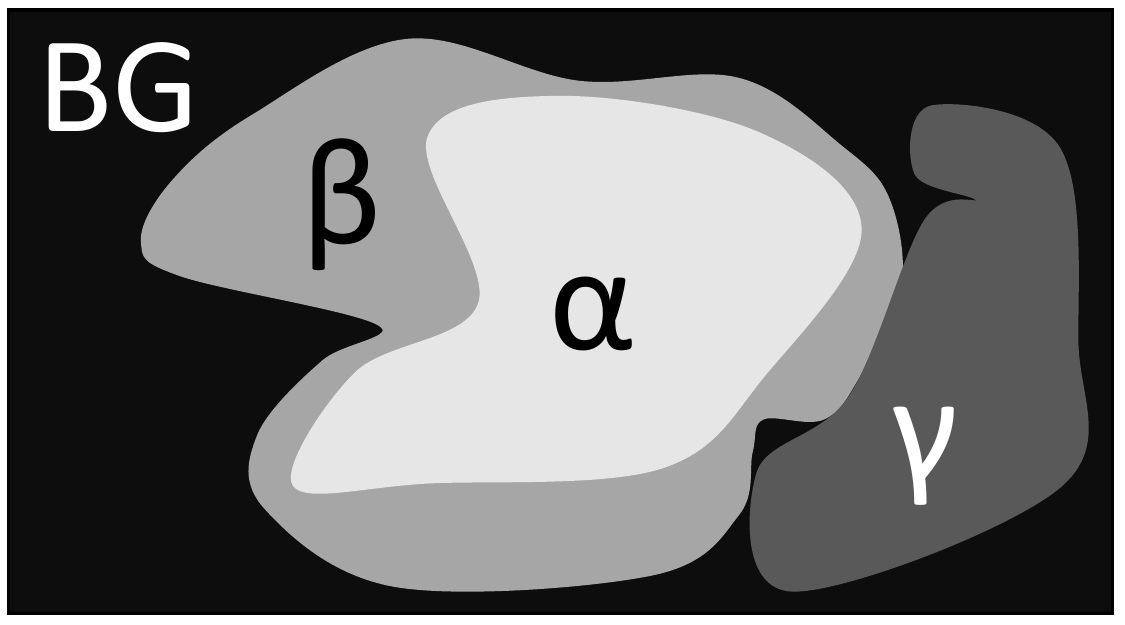}
  \end{subfigure}
  \begin{subfigure}{0.45\textwidth}
  \centering 
     \includegraphics[width=.9\linewidth]{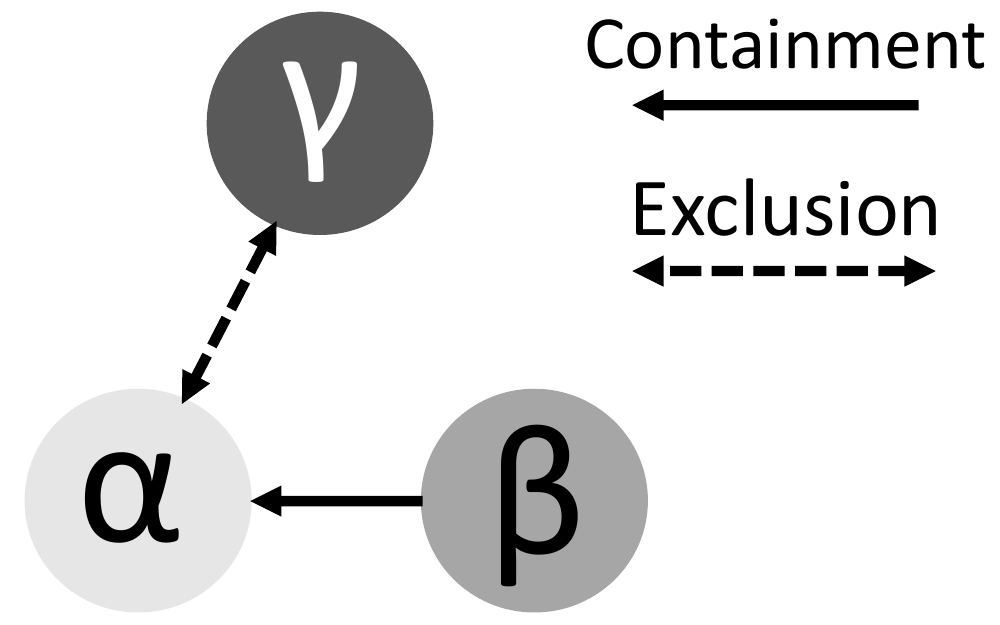}
  \end{subfigure}
\caption{
Schematic illustration of the topological interactions: containment and exclusion. $BG$ denotes the background class. \textbf{Containment}: $\beta$ contains $\alpha$. \textbf{Exclusion}: $\alpha$ and $\gamma$ are mutually exclusive. 
} 
\label{fig:geo-interaction}
\end{figure}

\begin{figure}[t]
\centering 

 \begin{subfigure}{0.23\linewidth}
  \includegraphics[width=1\textwidth]{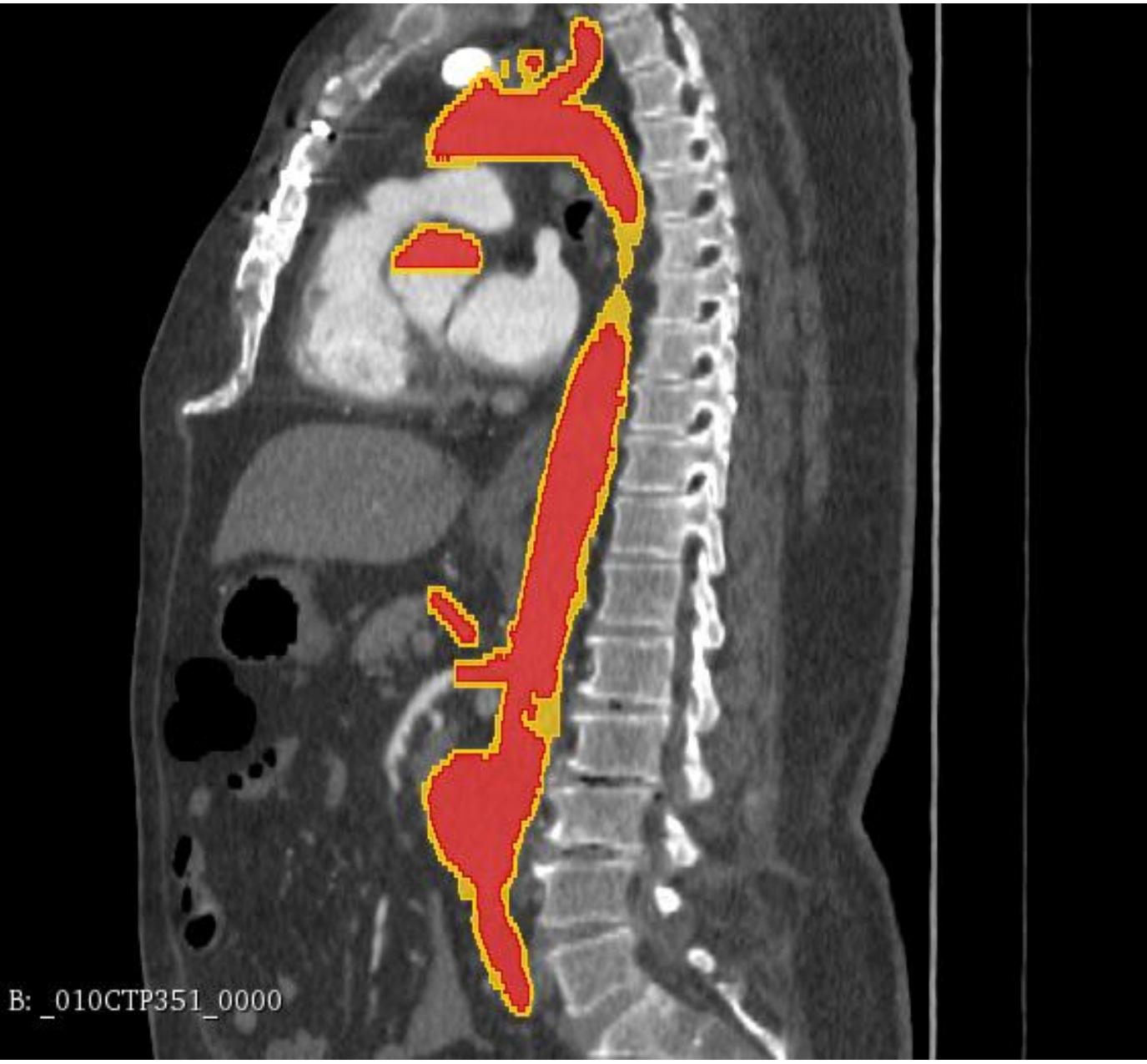}
  \end{subfigure}
  \begin{subfigure}{0.23\linewidth}
     \includegraphics[width=1\textwidth]{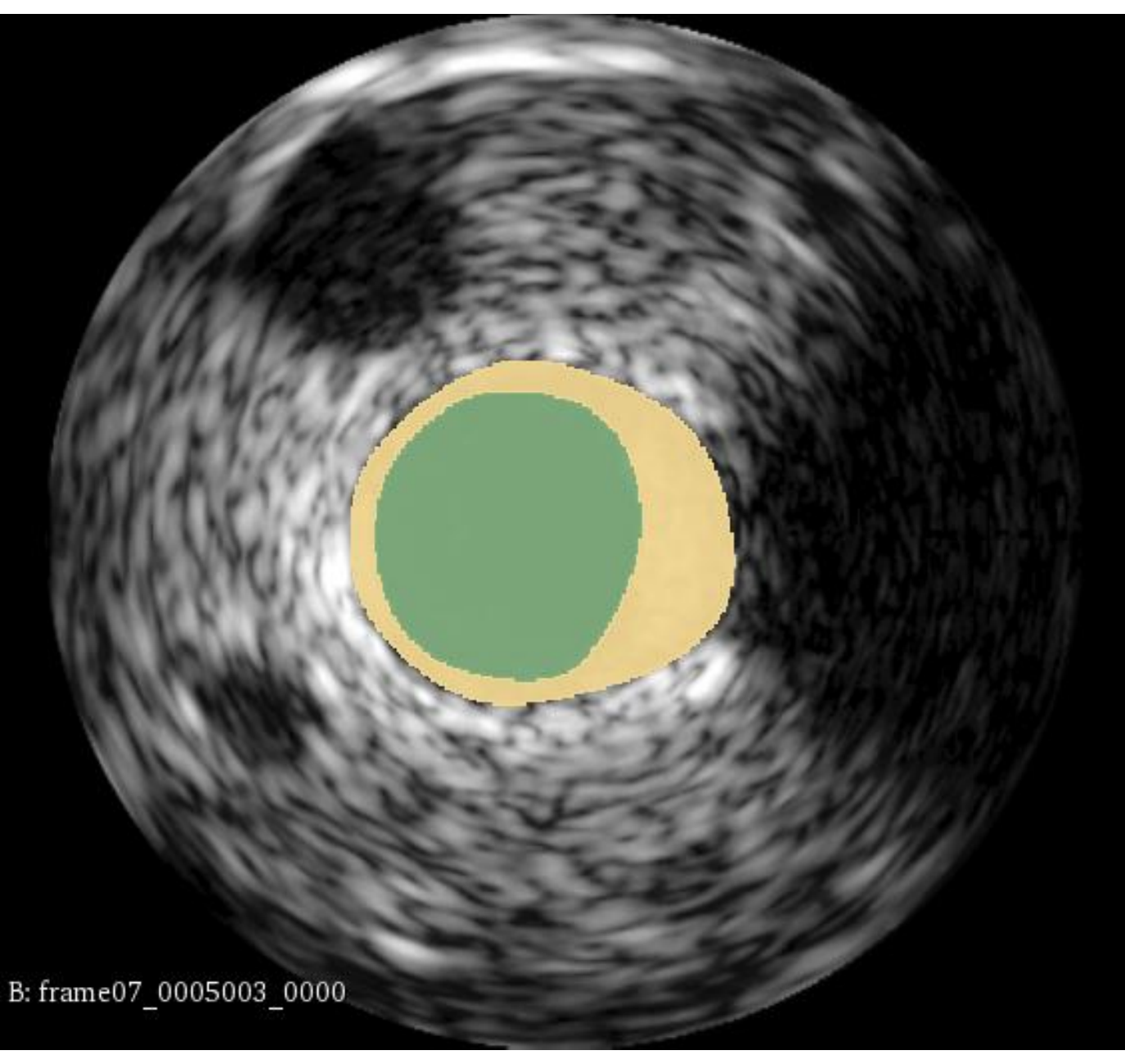}
  \end{subfigure}
    \begin{subfigure}{0.23\linewidth}
     \includegraphics[width=1\textwidth]{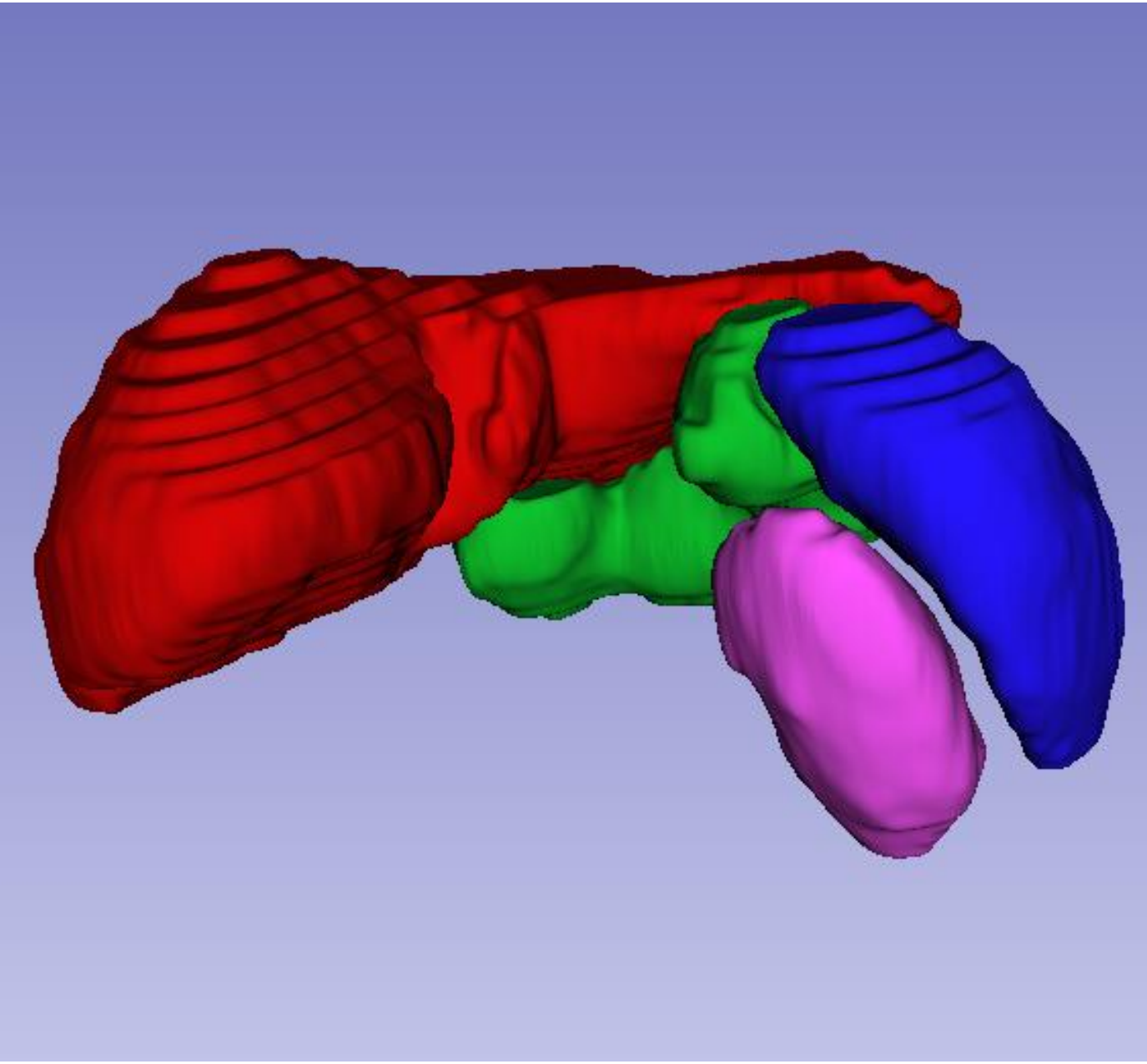}
  \end{subfigure}
    \begin{subfigure}{0.23\linewidth}
     \includegraphics[width=1\textwidth]{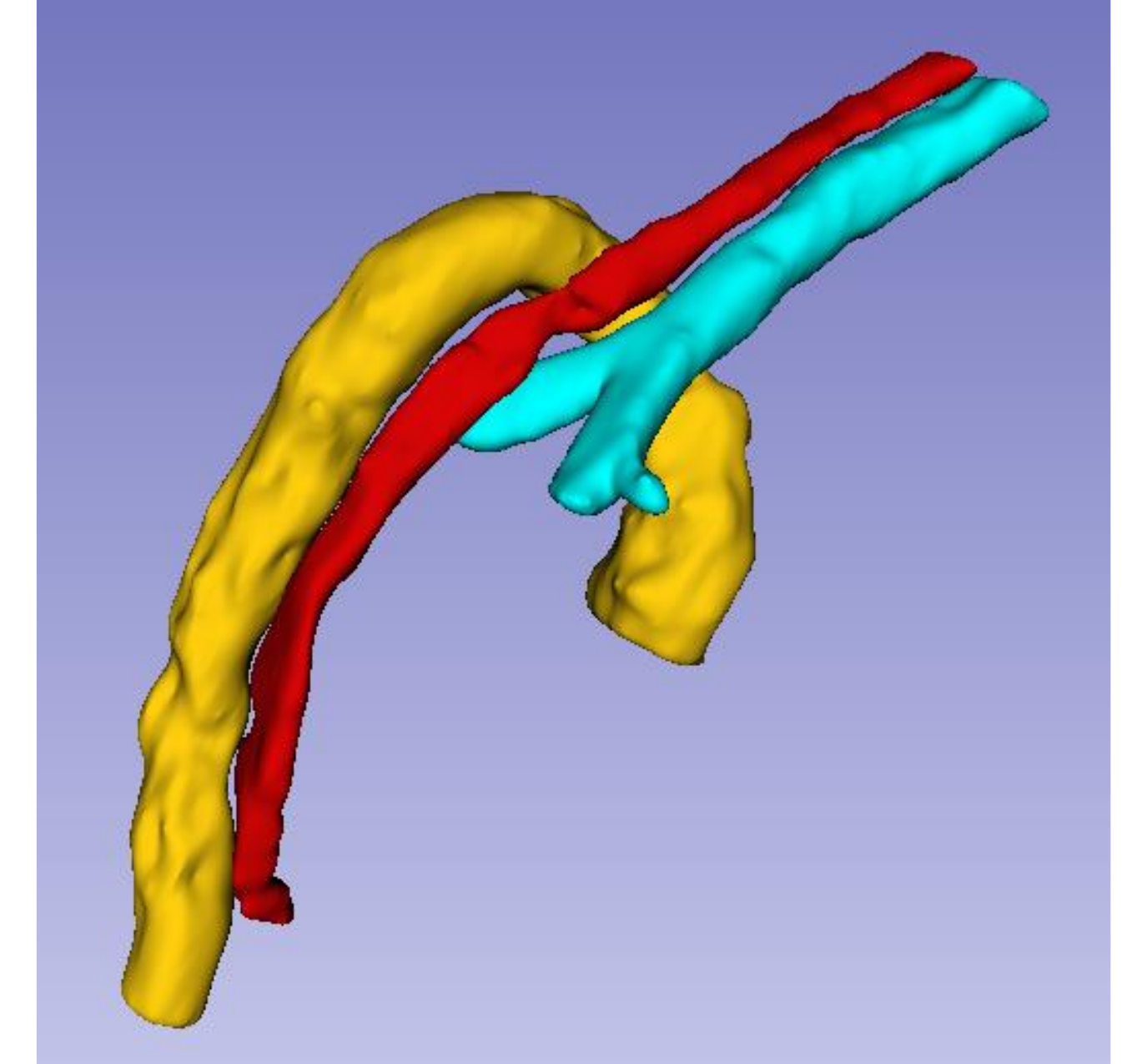}
  \end{subfigure}

\begin{subfigure}{0.23\linewidth}
  \includegraphics[width=1\textwidth]{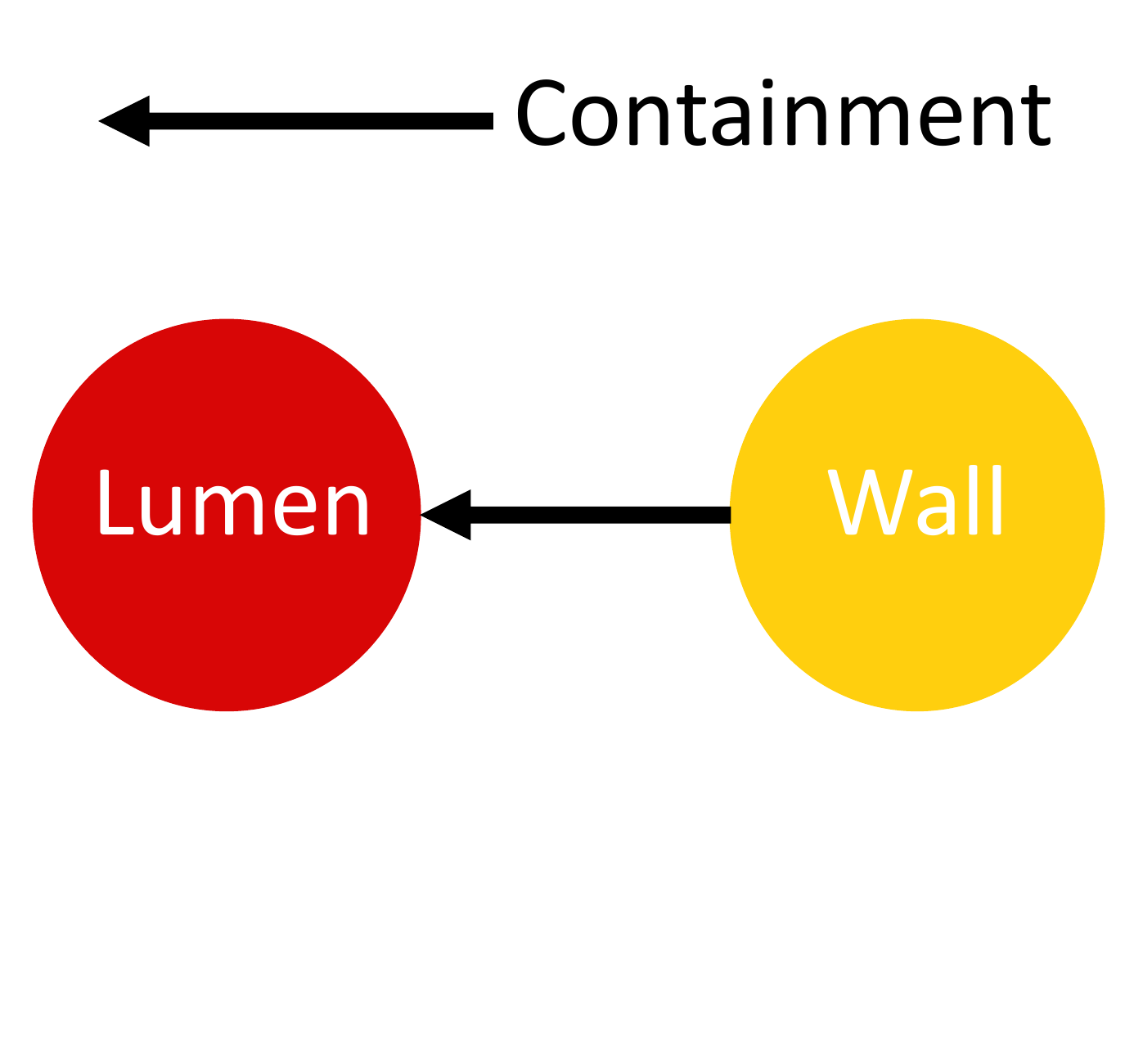}
      \caption{Aorta}
  \end{subfigure}
  \begin{subfigure}{0.23\linewidth}
     \includegraphics[width=1\textwidth]{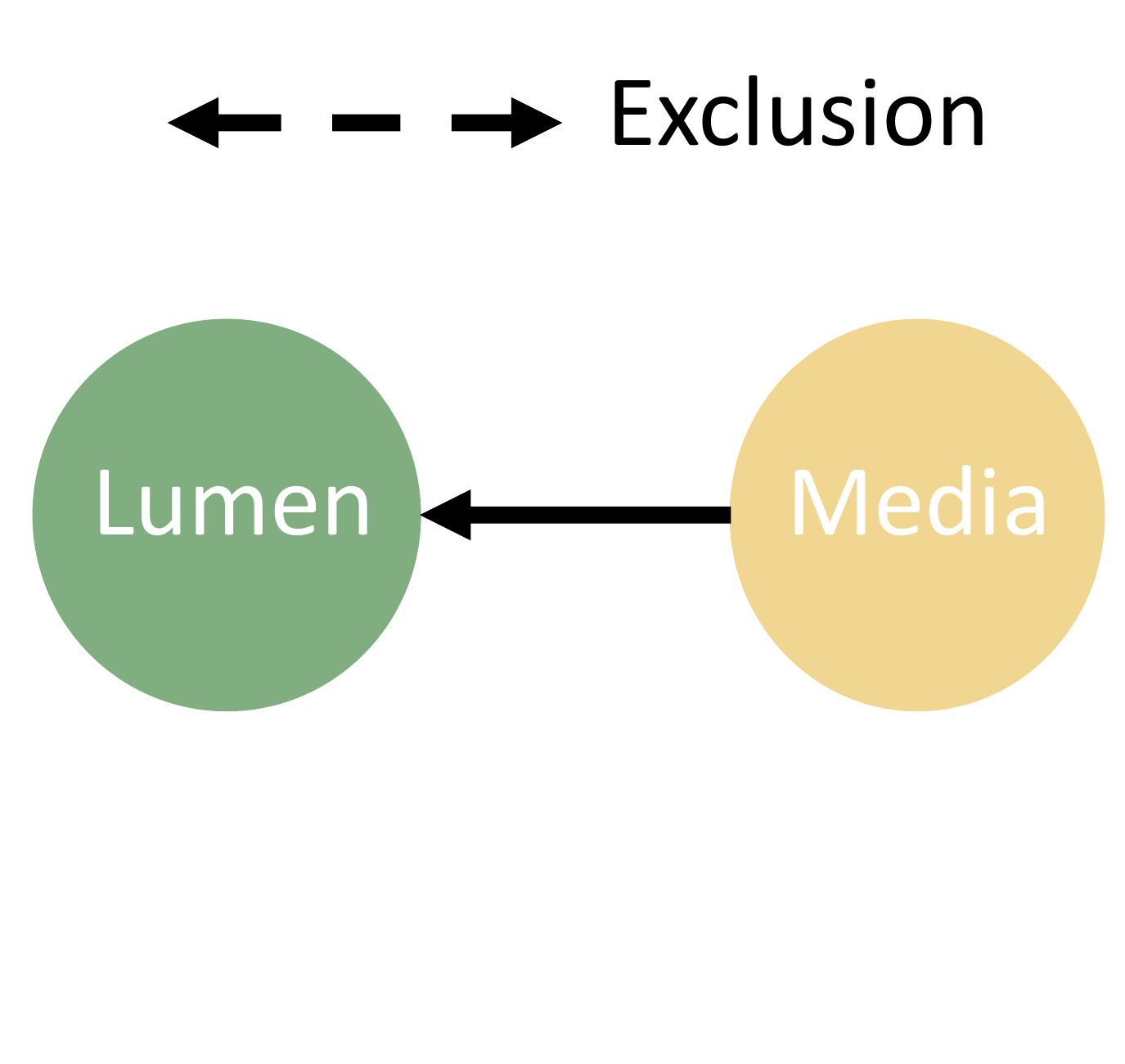}
         \caption{IVUS~\cite{balocco2014standardized}}
  \end{subfigure}
    \begin{subfigure}{0.235\linewidth}
     \includegraphics[width=1\textwidth]{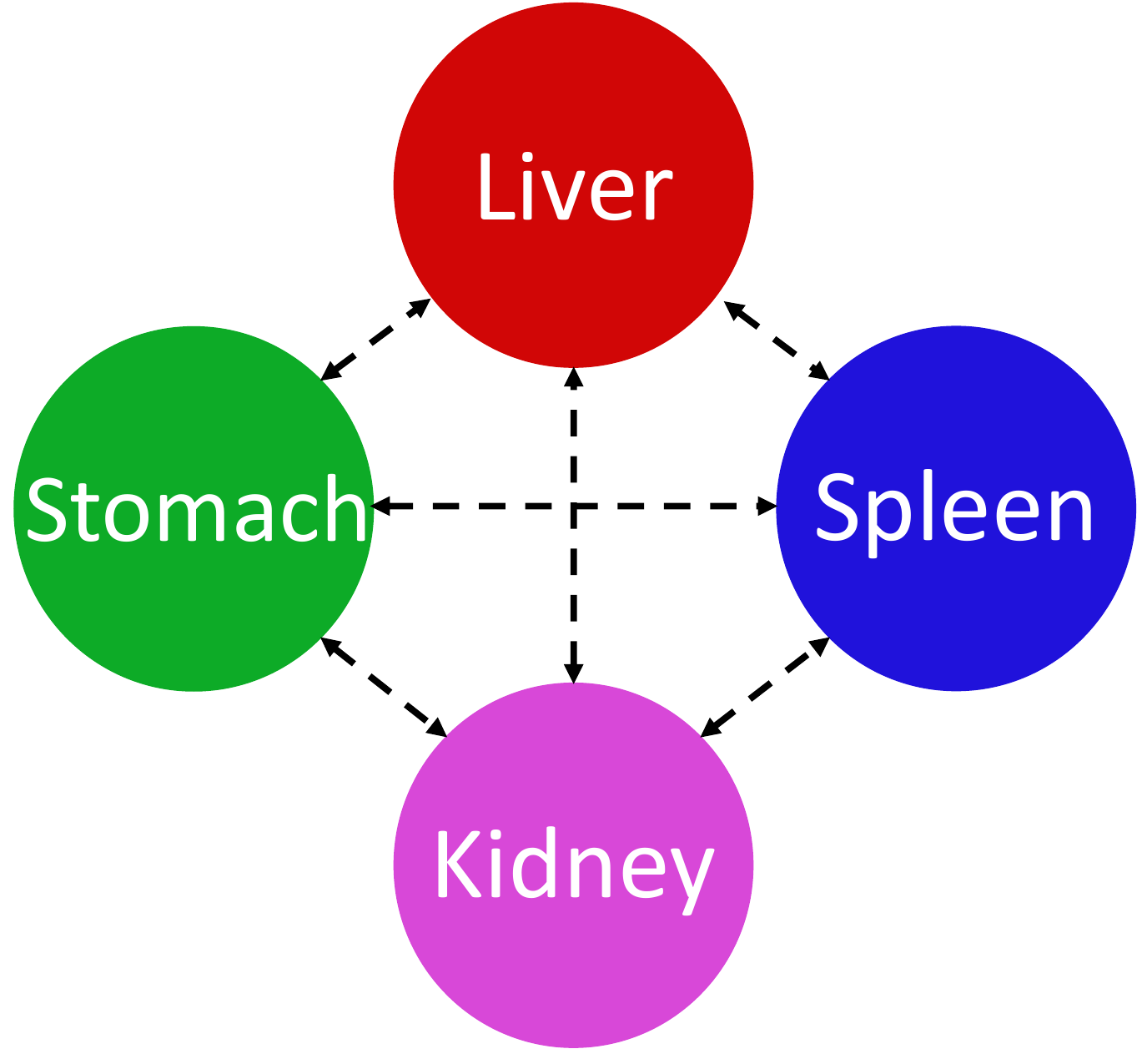}
         \caption{Multi-Atlas~\cite{landman2015miccai}}
  \end{subfigure}
    \begin{subfigure}{0.2251\linewidth}
     \includegraphics[width=1\textwidth]{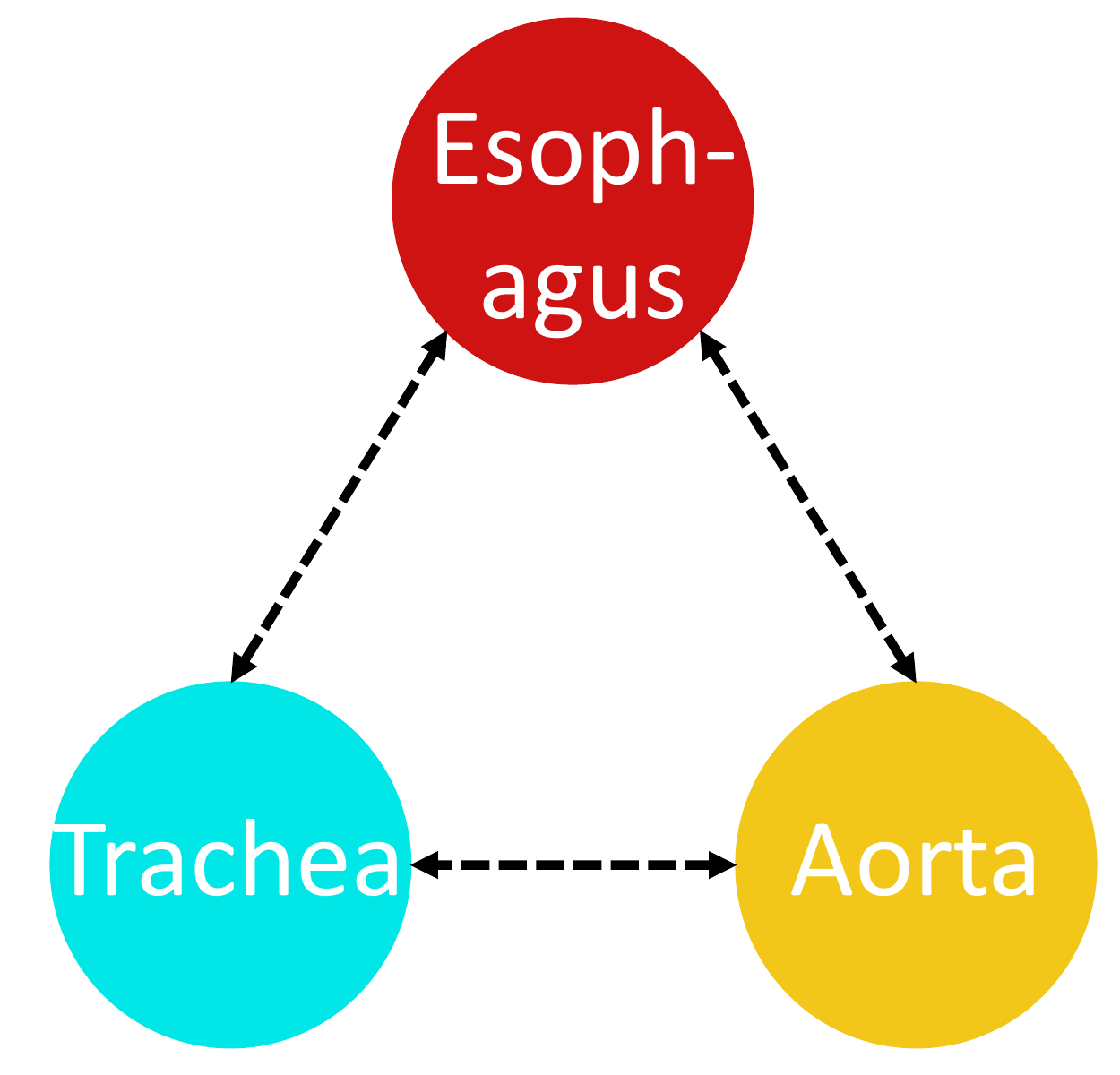}
         \caption{SegTHOR~\cite{lambert2019segthor}}
  \end{subfigure}

\caption{Multi-class topological interactions for each dataset.}
\label{fig:data-interactions}
\end{figure}

Broadly speaking, topological interactions between different foreground classes include two types, containment and exclusion. In Fig.~\ref{fig:geo-interaction}, we illustrate these constraints using three class labels, $\alpha$, $\beta$ and $\gamma$.
\begin{itemize}
  \item \textbf{Containment}: Class $\beta$ contains $\alpha$ if $\beta$ completely surrounds $\alpha$. 
  We use solid arrow from $\beta$ to $\alpha$ to denote the containment relationship. 
  In real applications, e.g., aorta segmentation, the aortic wall contains the lumen. See Fig.~\ref{fig:data-interactions}(a) for an illustration.
  \item \textbf{Exclusion}: Classes $\alpha$ and $\gamma$ are mutually exclusive if the pixels/voxels of class $\alpha$ and class $\gamma$ cannot be adjacent to each other. We use dashed double-arrow to denote the exclusion relationship. In multi-organ segmentation, there is clear separation between stomach and liver. They are mutually exclusive. See Fig.~\ref{fig:data-interactions}(c) for an illustration.
\end{itemize}

These constraints are quite general and can be observed in different medical imaging applications. See Fig.~\ref{fig:data-interactions} for more examples.  
We can also enforce stronger constraints. For containment, we may require the surrounding class ($\beta$ in Fig.~\ref{fig:geo-interaction}) to be at least $d$-pixel thick. For exclusion, we may require the gap between two mutually exclusive classes to be at least $d$-pixel wide. We call these generalized constraints \emph{$d$-containment} and \emph{$d$-exclusion}.


\myparagraph{Overview of the Proposed Method.}
Though the aforementioned topological interactions are global constraints, we observe that they can be encoded in a localized manner. Specifically, both containment and exclusion constraints can be rewritten as forbidding certain label combinations for adjacent pixels/voxels. In the example in Fig.~\ref{fig:geo-interaction}, $\beta$ contains $\alpha$ equals to the constraint that a pixel/voxel of label $\alpha$ cannot be adjacent to a pixel/voxel of any label other than $\beta$ and itself. Exclusion is more straightforward, $\alpha$ and $\gamma$ are mutually exclusive if any two adjacent pixels/voxels do not have the label pair $(\alpha,\gamma)$ or $(\gamma,\alpha)$. 

We enforce these constraints into DNN training by proposing a novel topological interaction module. The idea is to go through all pairs of adjacent pixels/voxels and identify the pairs that violate the desired constraints. Pixels belonging to these pairs are the ones inducing errors into the topological interaction. We will refer to them as \textit{critical pixels}. Our topological interaction module will output these critical pixels. Then, we can incorporate the module into training by designing a loss paying extra penalty to these critical pixels.

\begin{figure}[t]
\centering
\includegraphics[width=.9\columnwidth]{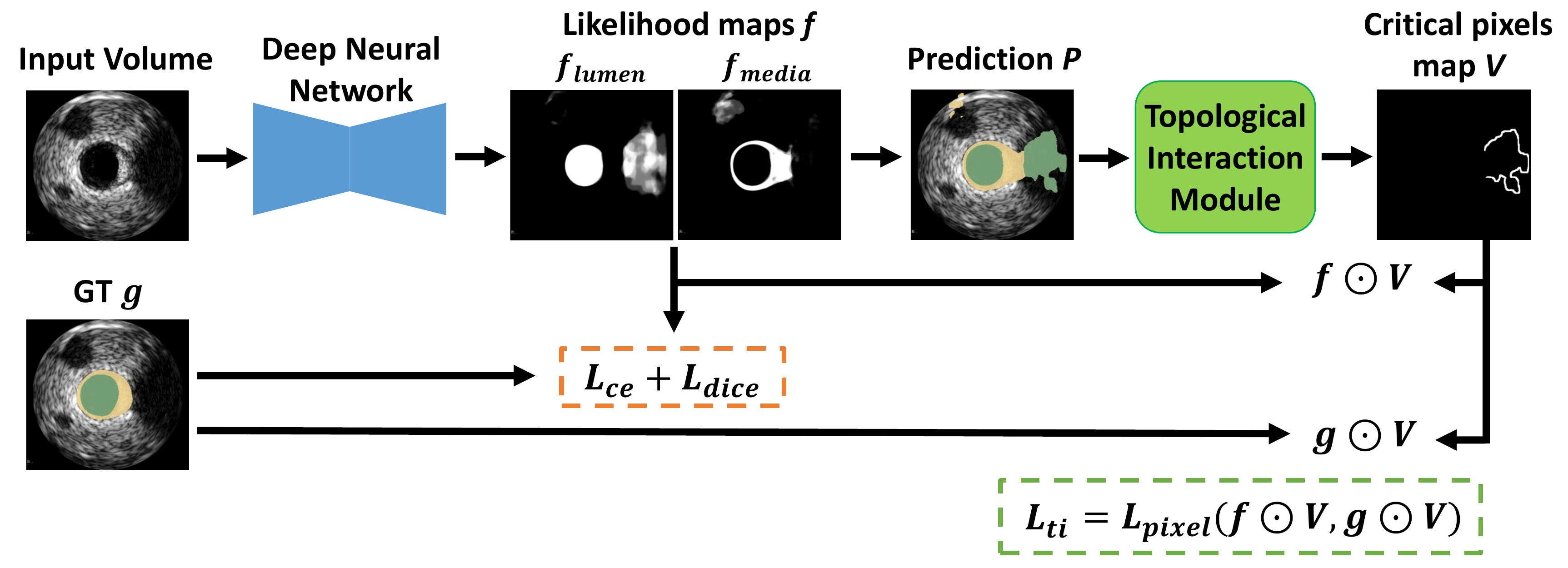}
\caption{An overview of the proposed method. The proposed module encodes the topological interactions between the different classes (e.g., \textit{media} and \textit{lumen} classes in the IVUS dataset follow the containment constraint). Critical pixels are identified and used for the new loss $L_{ti}$.
} \label{method}
\end{figure}

An efficient implementation of the module, however, is not trivial. Simply looping through all pixels is too expensive to serve as a frequent operation during training. 
To this end, we propose an efficient implementation of the constraints purely based on convolutional operations (Sec.~\ref{ssec:methodology_ssec1}). The method is much more efficient and can easily generalize to more challenging $d$-containment and $d$-exclusion without much extra computational expense. Finally, in Sec.~\ref{ssec:methodology_ssec2}, we incorporate the proposed module into training by formulating a loss function penalizing the critical pixels. This ensures the DNNs learn better feature representation while respecting the imposed constraints, as we will demonstrate empirically.
Fig.~\ref{method} provides an overview of the proposed method.

\subsection{Topological Interaction Module}
\label{ssec:methodology_ssec1}
The \textit{topological interaction module} encodes the topological interactions defined above. 
Recall the key is to forbid certain label combinations appearing in any pair of adjacent pixels. Our module identifies the pairs that violate the constraints. 

Next, we explain how to map the constraints into the local constraints regarding two labels that should not appear in adjacent pixels. For exclusion constraint, the forbidding label pair is obvious. In Fig.~\ref{fig:geo-interaction}, labels $\alpha$ and $\gamma$ are mutually exclusive. We create new labels $A=\alpha$ and $C=\gamma$, and forbid them to appear in adjacent pixels. 
For containment constraints, say label $\beta$ contains label $\alpha$ (as in Fig.~\ref{fig:geo-interaction}), we create a new label $A=\alpha$ and a new label $C$ being the union of all other labels except for $\alpha$ and $\beta$. Then $\beta$ containing $\alpha$ is equivalent to $A=\alpha$ not touching $C$.

For the rest of this section, we focus on how to create a module identifying adjacent pixel pair having the label pair $(A,C)$ or $(C,A)$. For ease of exposition, we assume a 2D 4-connectivity neighborhood (i.e., each pixel is only adjacent to 4 neighboring pixels), and so $d=1$. The approach can be naturally generalized to other connectivities as formalized in the classic digital topology \cite{rosenfeld1979digital}. 

\myparagraph{Naive Solution.}
\label{sssection:naive}
Given a discretized segmentation map predicted by the network, 
the naive solution is simply looping over all pixels and 
for each pixel, scan all its neighbors.
For every pair of adjacent pixels with the label pair $(A,C)$ or $(C,A)$, we flag both of the pixels as critical.
The obvious issue with this naive solution is that it is very expensive. 
Furthermore, such computation can only run on a CPU, and so is rather slow; this is detailed in the Supplementary Material.\footnote{There is an alternate way to better implement this naive solution by creating extra maps representing neighboring pixels. The issue of such a method is it does not scale well with larger neighborhood (which is necessary for more general constraints assuming a gap of width $d$ between forbidden label pairs). See the Supplementary Material for more details.}

\begin{figure}[t]
\centering
\includegraphics[width=.75\textwidth]{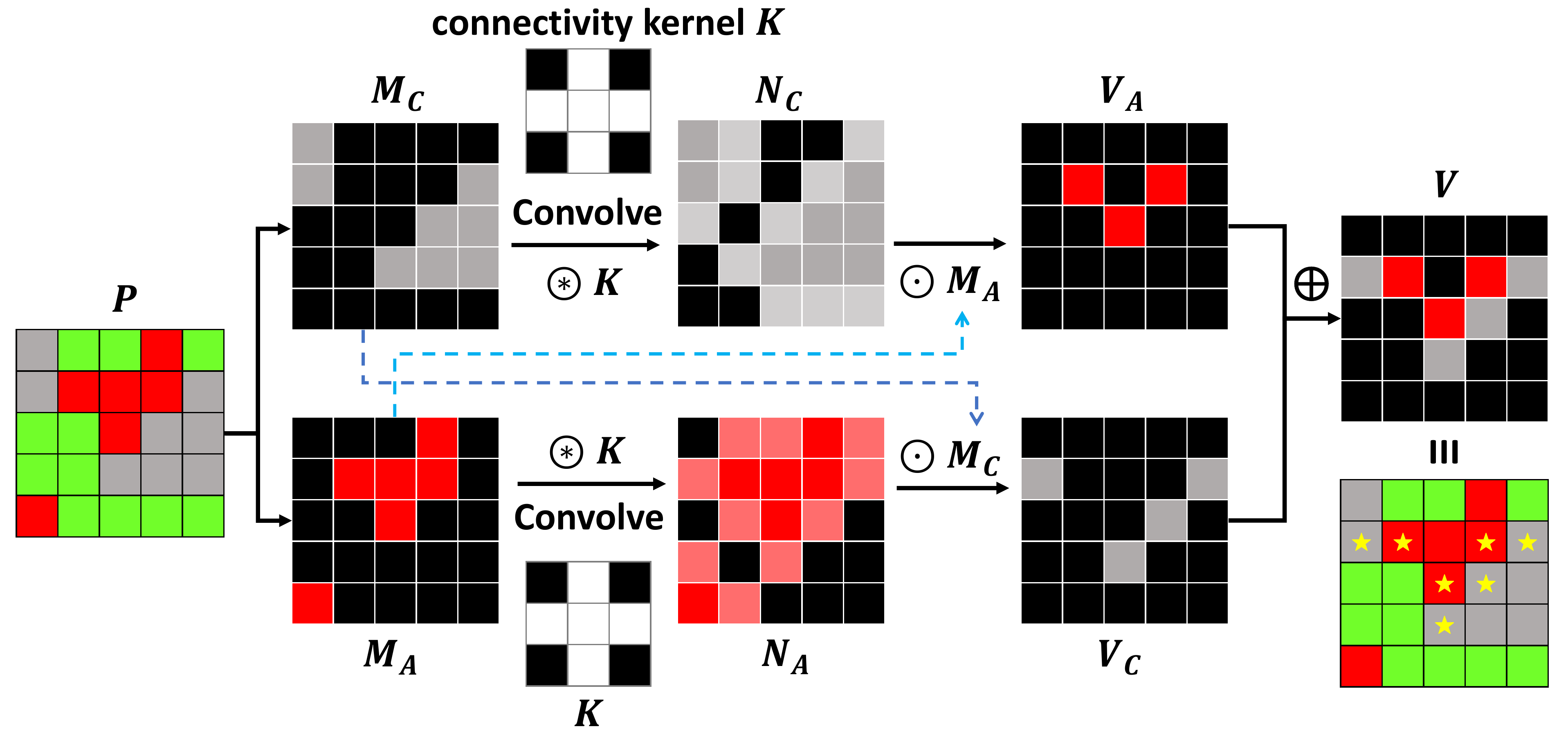}
\caption{2D illustration of the \textbf{proposed} strategy to detect the set $V$ of topological critical pixels. We use 4-connectivity kernel. The entire critical pixel map $V$ is highlighted with $\ast$'s.
}
\label{identify_algorithm}
\end{figure}

\myparagraph{Convolution-Based Solution.} 
\label{sssection:proposed}
Let $P \subseteq \mathbf{R}^{d}$ denote the $d$-dimensional discrete segmentation map predicted by the network. We want to generate a critical pixel map in which only those label-$A$ pixels with a label-$C$ neighbor are activated  and vice-versa. We achieve this goal through manipulations of different semantic masks. First, to determine the critical pixels in $A$, we expand the $C$ mask by $d$ pixels, and then find out the intersection of the expanded mask with the $A$ mask ($d=1$ for 2D 4-connectivity). 
In this way, we obtain the set of all the critical $A$ pixels: they fall within the expanded $C$ mask, and thus must be a neighbor of some $C$ pixels. In top row of Fig.~\ref{identify_algorithm}, second to fourth columns, we show the $C$ mask (denoted by $M_{C}$), its expansion, and the intersection with the $A$ mask (denoted by $M_{A}$), resulting in the critical $A$ pixels. 
In a similar manner, we can obtain the set of critical $C$ pixels by expanding the $A$ mask and finding its intersection with the $C$ mask. This is illustrated in the bottom row, second to fourth columns of Fig.~\ref{identify_algorithm}.

In practice, expanding a mask can be done efficiently using the \textit{dilation} morphological operation~\cite{haralick1987image}. In dilation, we convolve a given binary mask with a kernel $K$. The kernel defines the neighbors of a given voxel. 
Formally, let $M_{A}$ and $M_{C}$ be the class masks for $A$ and $C$ respectively. We then obtain neighborhood information $N_{A}$ and $N_{C}$ via dilation/convolution as follows:
\begin{equation}
    N_{A} \coloneqq M_{A} \circledast K,\quad N_{C} \coloneqq M_{C} \circledast K
    \label{eq:conv}
\end{equation}
where we use $\circledast$ to denote the standard convolution operation. $K$ is the convolution kernel which we refer to as the \textit{connectivity kernel}. As we are dealing with 2D 4-connectivity case, the connectivity kernel used is as shown in Fig.~\ref{identify_algorithm}. Notice that in map $N_{A}$, all the pixels which are in contact with class $A$ get activated. 
We obtain $N_{C}$ in a similar way. Now that we have the expanded neighborhood information, and we use this to find which pixels of $A$ and $C$ fall in each other's neighborhood. 
If $V$ denotes the entire critical pixel map, it can be further divided into $V_{A}$ and $V_{C}$ which contain the critical pixels in class $A$ and $C$ respectively. We can then quantify them as:
\begin{equation}
      V_{A} \coloneqq M_{A} \odot N_{C},\quad V_{C} \coloneqq M_{C} \odot N_{A},\quad V \coloneqq V_{A} \oplus V_{B}
\end{equation}
where $\oplus$ denotes the union operation, and $\odot$ denotes the Hadamard product.

Fig.~\ref{identify_algorithm} gives an overview of our method to compute topological critical pixels in the form of a binary mask $V$. 
Thus through the manipulation of maps obtained via standard convolution, we are able to augment existing information by deriving information relevant to topological interactions.

\myparagraph{Remark on the Connectivity Kernel $K$.} 
We remark that the connectivity kernel $K$ depends on the definition of neighborhood. Our current choice of $K$ corresponds to the 4-connectivity neighborhood (illustrated in Fig.~\ref{identify_algorithm}). In general, we can choose different neighborhood definitions corresponding to different kernels. Following the classic digital topology \cite{rosenfeld1979digital}, in 2D, we can have 4- and 8-connectivities. In 3D, we can have 6- and 26-connectivities. We can also specify different connectivity kernels for classes $A$ and $C$. 
See Supplementary Material for illustrations.\footnote{In digital topology, to ensure the Jordan curve theorem is correct, one needs to have either 4-conn.~for foreground and 8-conn.~for background, or the opposite. 
This is not in conflict with our method. $A$ and $C$ are both considered foreground labels. In 2D, they can use either 4-conn.~or 8-conn.~as long as they are the same. Similar rules apply to 3D.}

We also note it is natural to generalize the neighborhood definition and modify the kernel accordingly to enforce the more general/stronger constraints: $d$-containment and $d$-exclusion. These constraints essentially boil down to the constraint that labels $A$ and $C$ cannot appear on two pixels within distance $d$. To encode such constraints, we simply define the neighborhood of a pixel $p$ to be all pixels within a $(2d+1)\times (2d+1)$ local patch centered at $p$. The connectivity kernel is then an all-one kernel of the same size.

\myparagraph{Computational Efficiency.}
We analyze the computational efficiency of the proposed method by determining its complexity as a function of the input and neighborhood size.
Let the image size be $N \times N$. Suppose we enforce a separation of $d$ pixels, then the neighborhoods to be inspected for each pixel will be $k \times k$, where $k = 2d + 1$. In the naive solution, we require scanning the neighborhood of each pixel via loops and so the time complexity is in the order of $O(N^2k^2)$, not really scalable. This is apart from the fact that such a solution can only run on a CPU. On the contrary, the convolution-based solution has a time complexity $O(N^2 \log N)$. Here $\log N$ is due to the FFT (Fast Fourier Transform) implementation of convolution. While the naive solution's running time is quadratic to $k$, our proposed is independent of $k$ due to FFT. In practice, deep learning frameworks are highly optimized for convolution operations, and so they are several orders of magnitude cheaper than the naive solution. The memory requirement for both methods is similar in the order of $O(N^2)$ to store the map $V$.

\subsection{Incorporating into End-to-End Training}
\label{ssec:methodology_ssec2}
To incorporate the proposed topological interaction module into end-to-end training, we propose a topological interaction loss to correct the violations by penalizing the critical pixels. 

Let $f \in \mathbb{R}^{c \times H \times W}$ be the multi-class likelihood map predicted by the network, where $c$, $H$ and $W$ denote the number of classes, height and width of the image, respectively. $g \in \mathbb{R}^{H \times W}$ is the ground truth segmentation map with discrete labels, $0, 1, ..., c-1$. 
We use $L_{pixel}$ to denote the pixel-wise loss function, 
such as, cross-entropy, mean-squared-error, or dice losses. We use the binary mask $V$ obtained from Sec.~\ref{sssection:proposed}, to define $L_{ti}$, denoting the additional topological interaction loss, as: 

\begin{equation}
\label{loss}
    L_{ti} = L_{pixel}(f\odot V,g\odot V) 
\end{equation}
$L_{ti}$ can essentially encode the topological interactions, correct the topological interaction errors, and eventually produce a segmentation that is topologically correct. The final loss of our method, $L_{total}$, is given by:
\begin{equation}
    L_{total} = L_{ce} + \lambda_{dice} L_{dice} + \lambda_{ti} L_{ti}
\end{equation}
where $L_{ce}$ and $L_{dice}$ denote the cross-entropy and dice loss. The loss is controlled by the weights $\lambda_{dice}$ and $\lambda_{ti}$.

\section{Experiments}
\label{sec:Experiments}

\myparagraph{Datasets.} We validate our method on four datasets: The proprietary \textbf{Aorta} dataset contains 3D CT scans of 28 patients from an institutional database of patients with thoracic and/or abdominal aortic aneurysm. The \textbf{IVUS} (IntraVascular Ultrasound)~\cite{balocco2014standardized} is a 2D dataset of human coronary arteries and contains lumen and media-adventitia labels. 
The \textbf{Multi-Atlas BTCV}~\cite{landman2015miccai} is a multi-organ segmentation challenge, containing 3D CT scans of the cervix and abdomen. We use the abdomen dataset and segment four classes, namely, spleen, left kidney, liver, and stomach which appear in close proximity. We have clinically verified that the exclusion constraint holds among these four classes. The \textbf{SegTHOR}~\cite{lambert2019segthor} 2019 challenge contains 3D CT scans of thoracic organs at risk (OAR). In this dataset, the OARs are the heart, trachea, aorta and esophagus. The exclusion constraint holds among three classes, that is, the trachea, the aorta, and the esophagus do not touch each other. We do not take the heart class into consideration. 

The containment constraint holds for the Aorta and IVUS datasets, while the exclusion constraint holds for the remaining two. Fig.~\ref{fig:data-interactions} gives an overview of the classes in each dataset and the topological interactions among them.

\myparagraph{Baselines and Implementation Details.} We use the PyTorch framework, a single NVIDIA Tesla V100-SXM2 GPU (32G Memory) and a Dual Intel Xeon Silver 4216 CPU@2.1Ghz (16 cores) for all the experiments.
The comparison baselines consist of the UNet~\cite{unet2d,cciccek20163d}, FCN~\cite{FCN8s}, nnUNet~\cite{nnUNet}, TopoCRF~\cite{bentaieb2016topology}, MIDL~\cite{reddy2019brain}, and NonAdj~\cite{ganaye2019removing}.  We use the publicly available codes for UNet, FCN, nnUNet, and NonAdj. For TopoCRF and MIDL, we implemented it by ourselves in PyTorch. Specifically, for TopoCRF, MIDL, NonAdj and our proposed method, we fine-tune the models pre-trained by nnUNet. To support our claim that our method can be incorporated into any backbone, we train our module on FCN and UNet backbones as well.  More details and additional results are included in the Supplementary Material.

The connectivity kernel $K$, in 2D, is a $3 \times 3$ kernel filled with $1$'s to enforce $8$-connectivity. Similarly in 3D, $K$ is a $3 \times 3 \times 3$ kernel filled with $1$'s to enforce $26$-connectivity. We also perform an ablation study on the connectivity kernel in the Supplementary Material.



\begin{figure}[t]
\centering 

      \begin{subfigure}{0.15\linewidth}
  \includegraphics[width=1\textwidth]{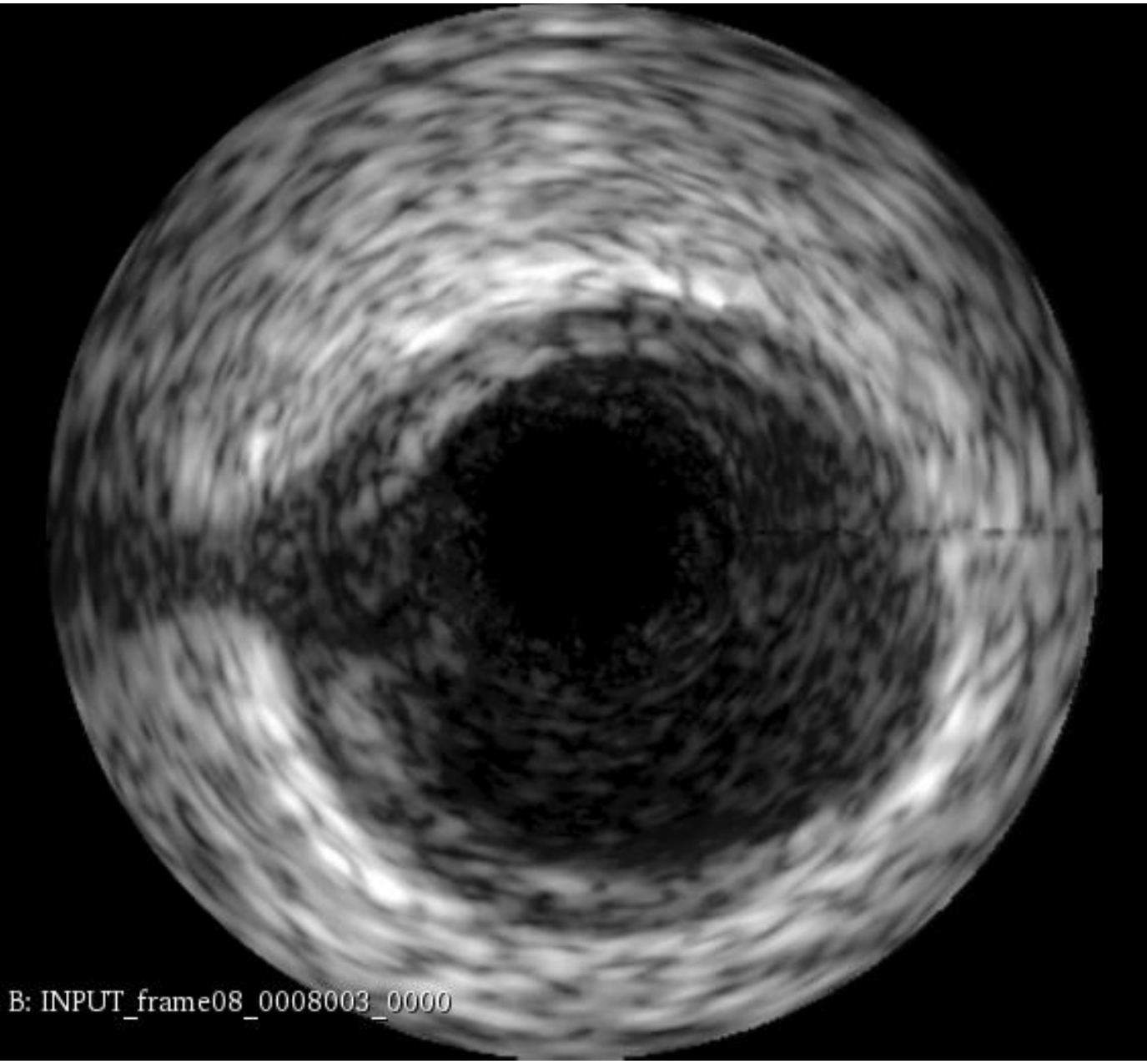}
      \caption{Input}
  \end{subfigure}
  \begin{subfigure}{0.15\linewidth}
     \includegraphics[width=1\textwidth]{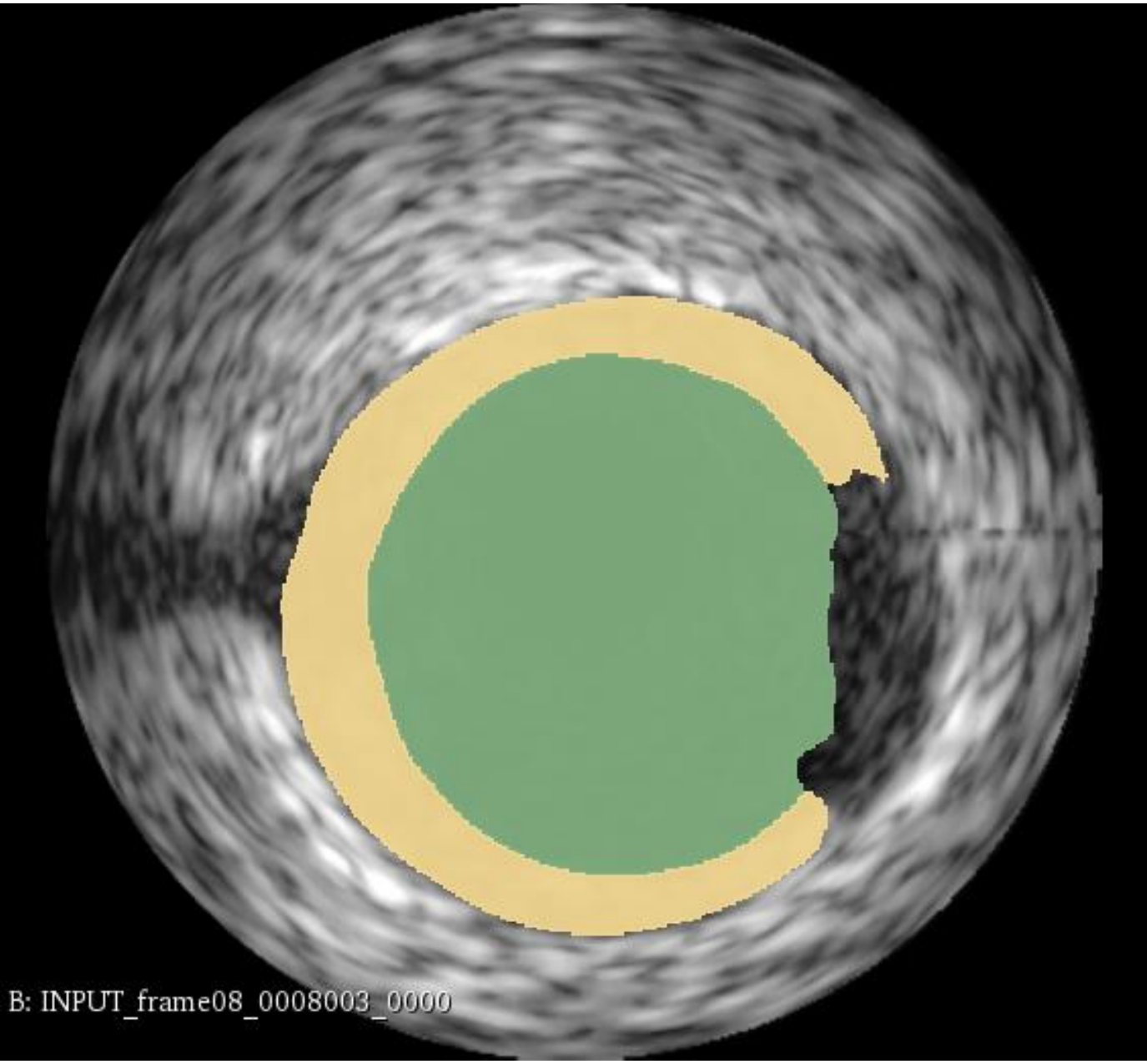}
         \caption{nnUNet}
  \end{subfigure}
    \begin{subfigure}{0.15\linewidth}
     \includegraphics[width=1\textwidth]{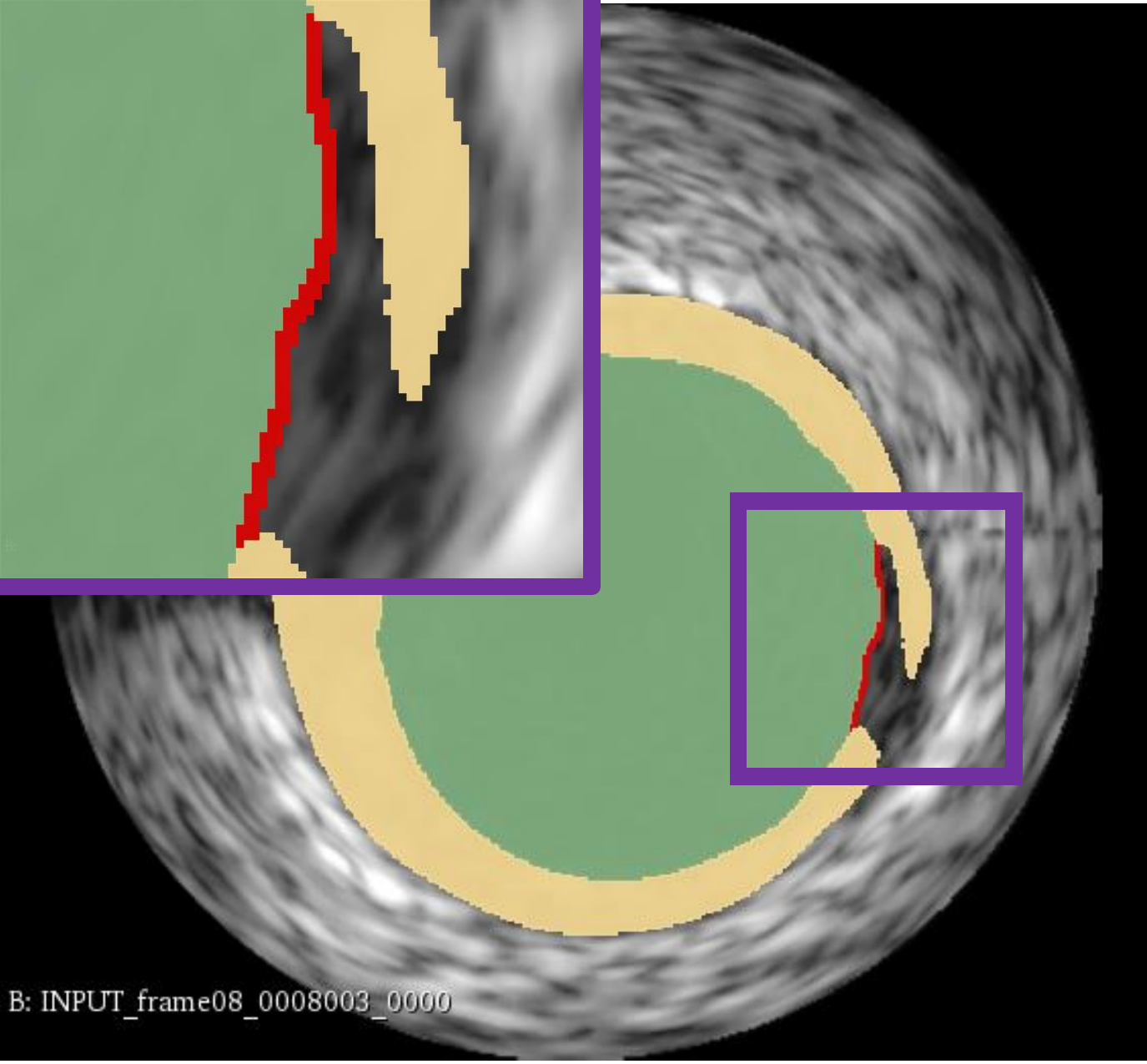}
         \caption{$\approx$Ep. 70}
  \end{subfigure}
    \begin{subfigure}{0.15\linewidth}
     \includegraphics[width=1\textwidth]{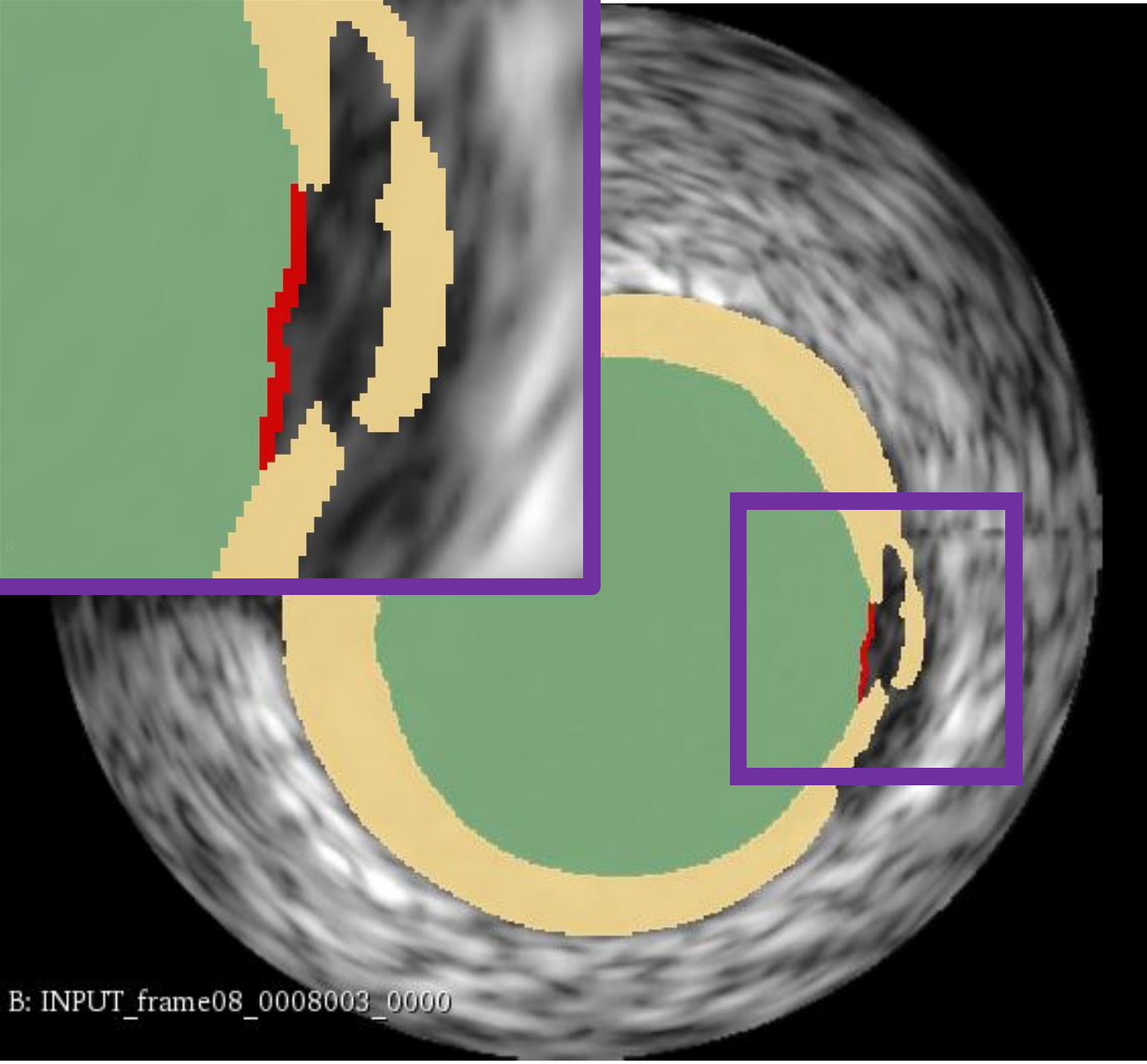}
         \caption{$\approx$Ep. 140}
  \end{subfigure}
      \begin{subfigure}{0.15\linewidth}
     \includegraphics[width=1\textwidth]{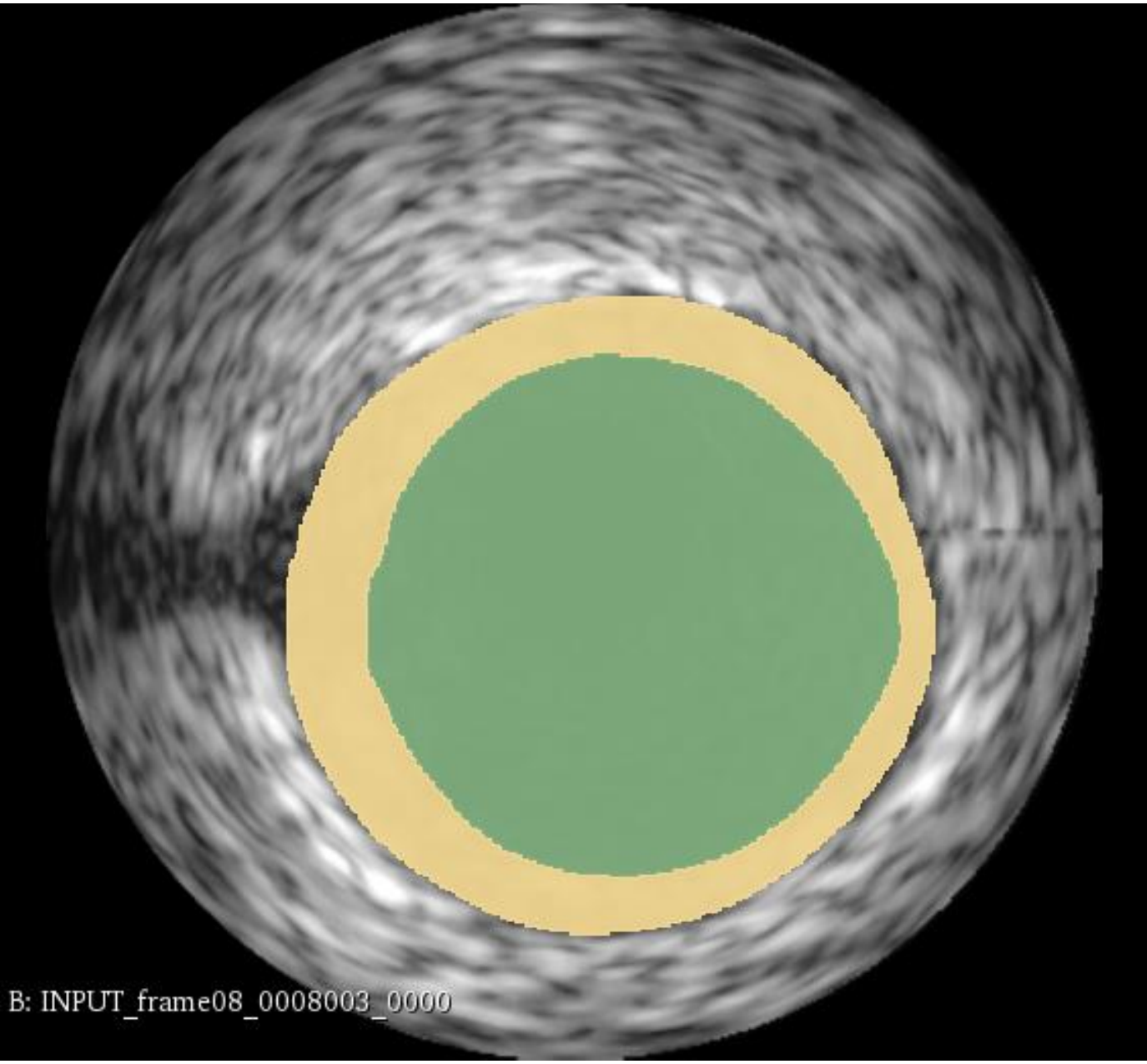}
         \caption{$\approx$Ep. 200}
  \end{subfigure}
      \begin{subfigure}{0.15\linewidth}
     \includegraphics[width=1\textwidth]{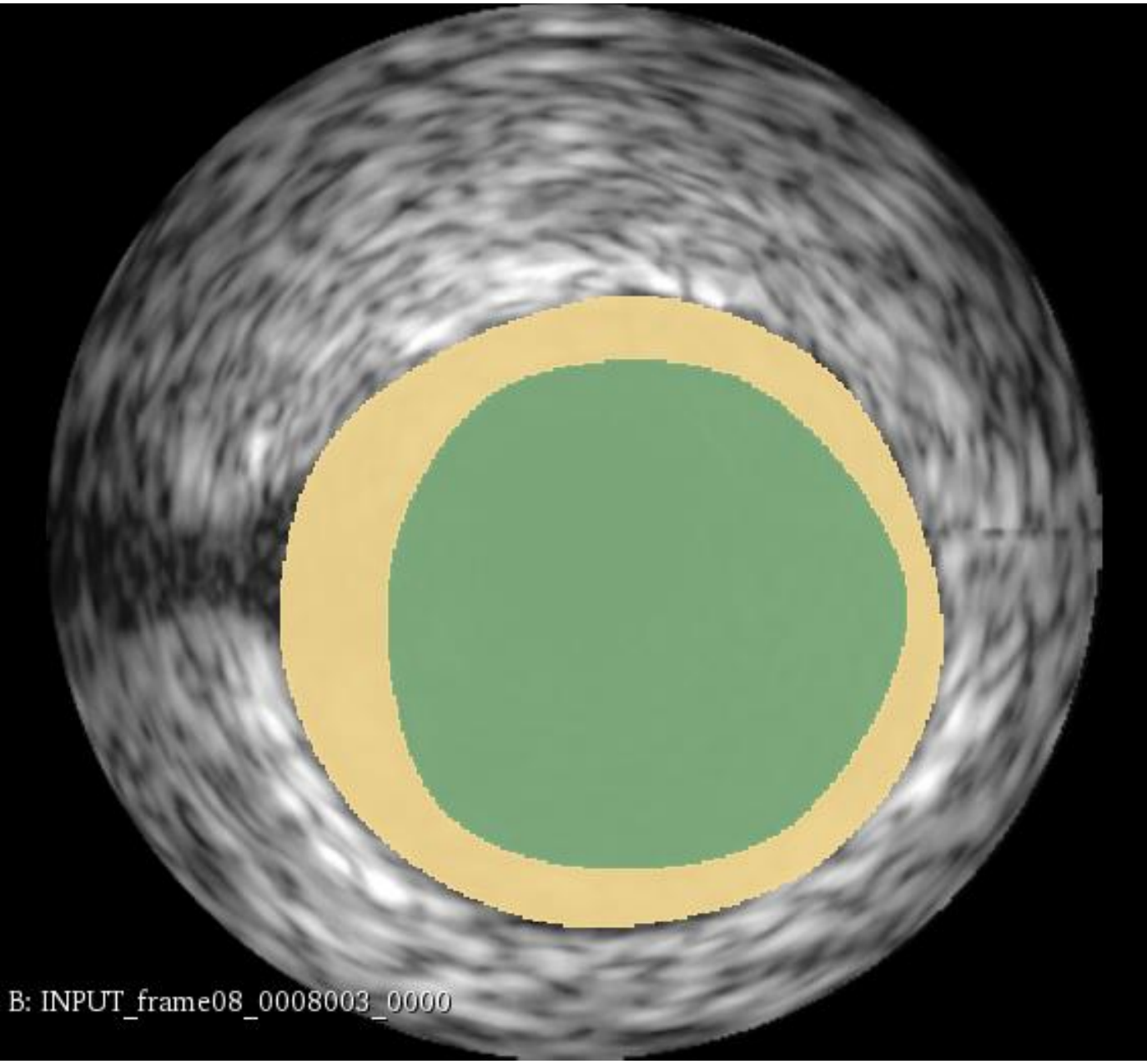}
         \caption{GT}
  \end{subfigure}

\caption{Epoch (\textit{Ep.}) progression of the proposed method. Critical pixel map identified by the module is marked in red.}
\label{fig:violations}
\end{figure}

\begin{figure}[t]
\centering 

 \begin{subfigure}{0.135\linewidth}
  \includegraphics[width=1\textwidth]{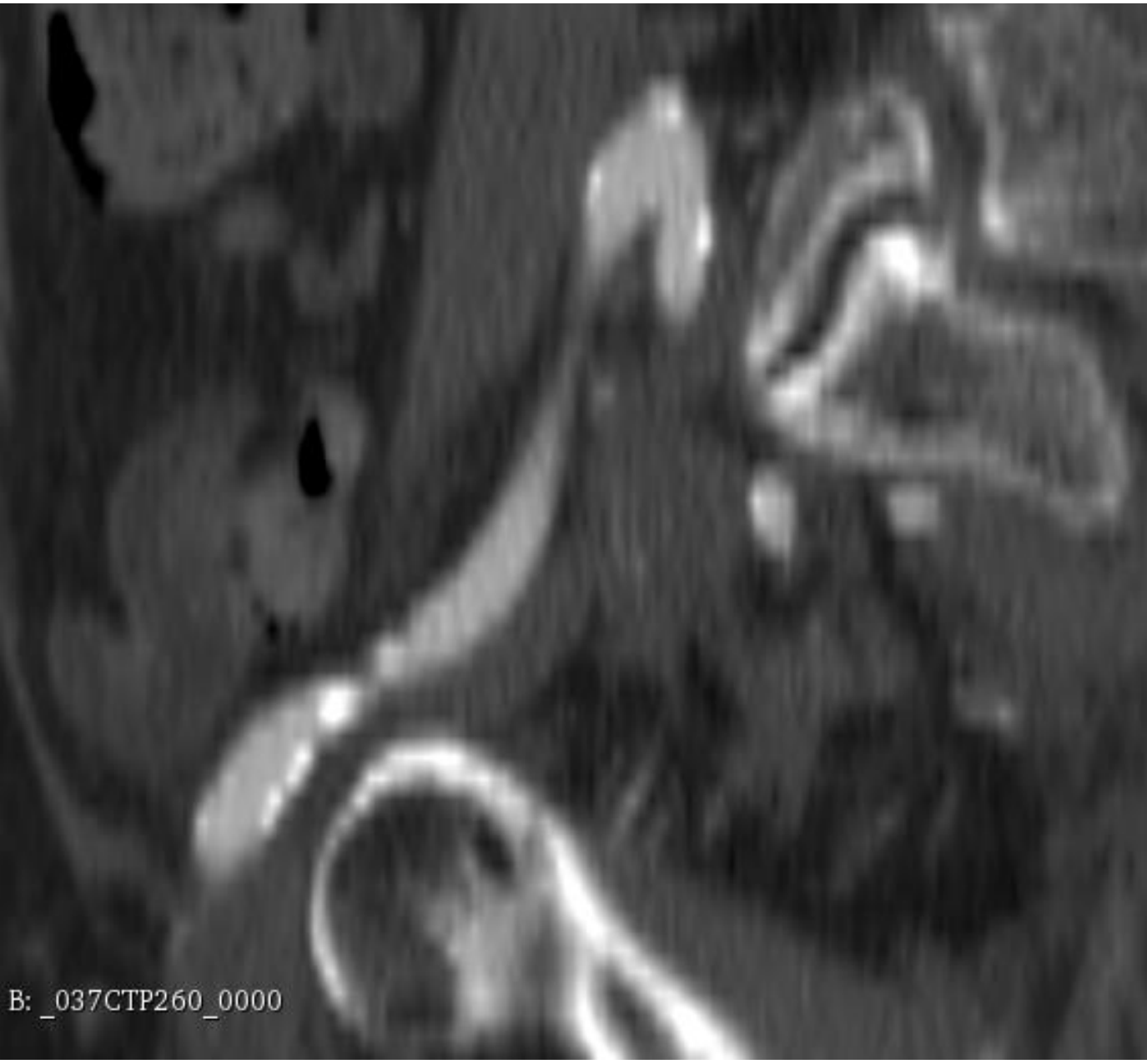}
  \end{subfigure}
    \begin{subfigure}{0.135\linewidth}
     \includegraphics[width=1\textwidth]{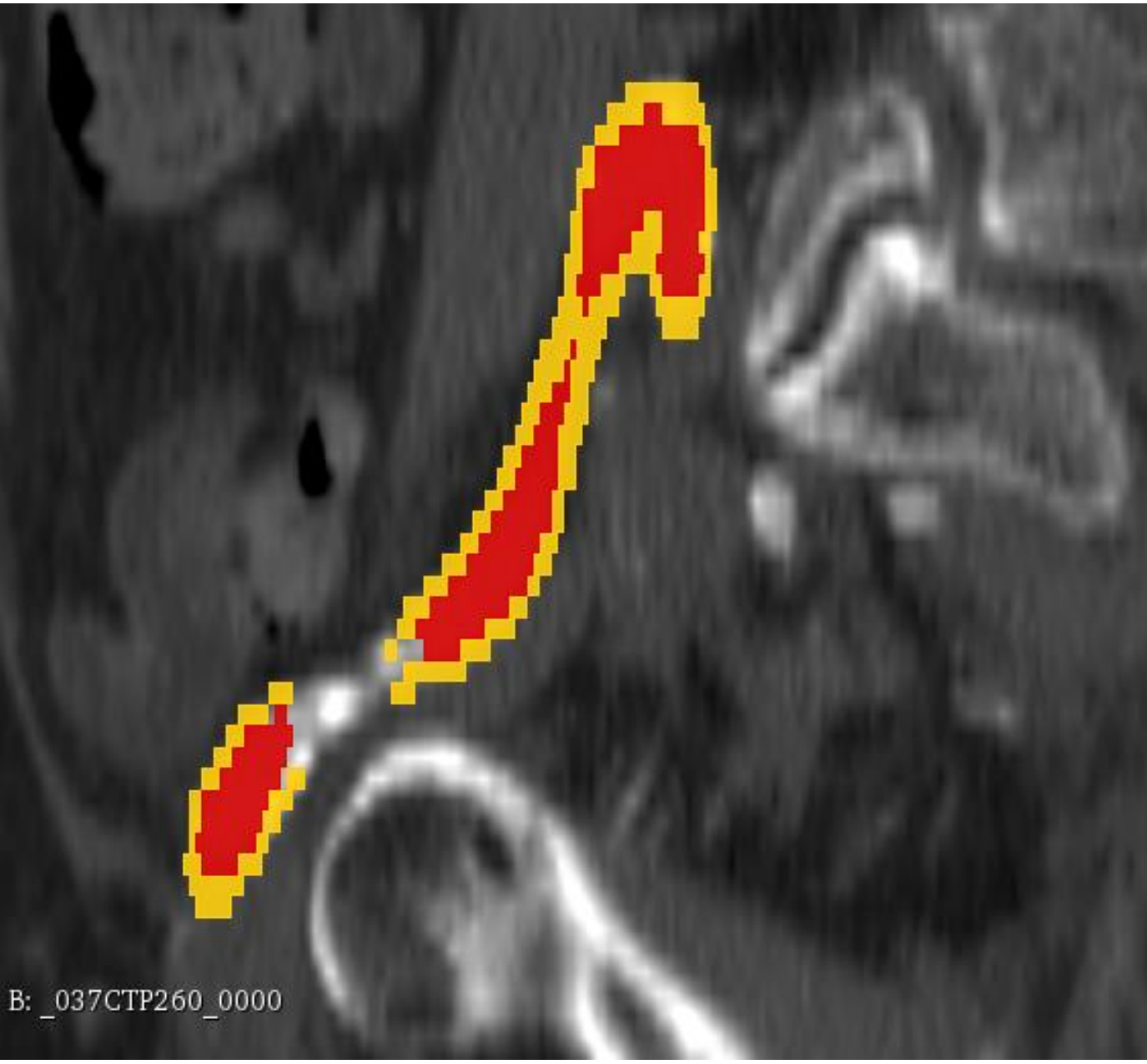}
  \end{subfigure}
    \begin{subfigure}{0.135\linewidth}
     \includegraphics[width=1\textwidth]{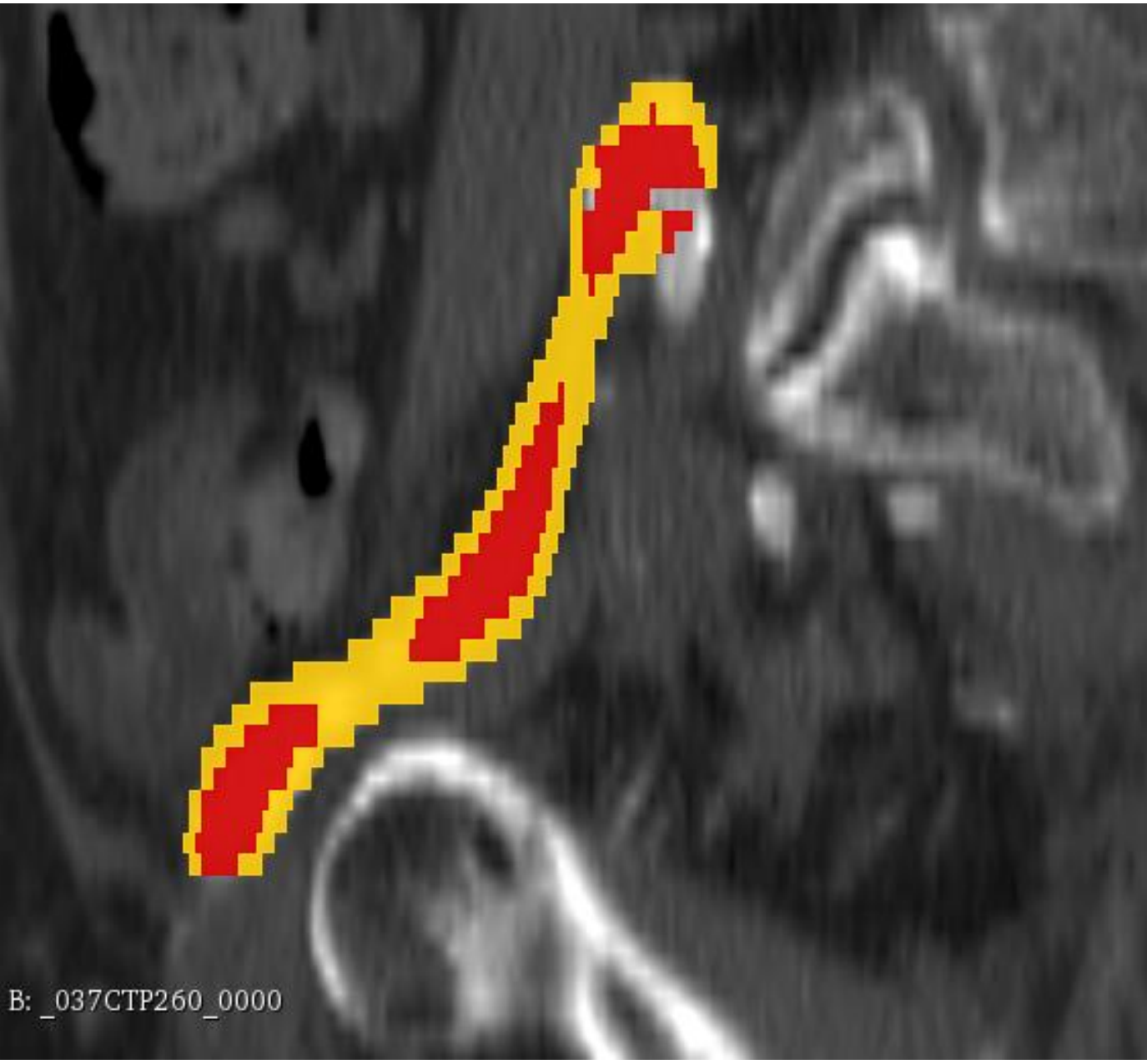}
  \end{subfigure}
    \begin{subfigure}{0.135\linewidth}
     \includegraphics[width=1\textwidth]{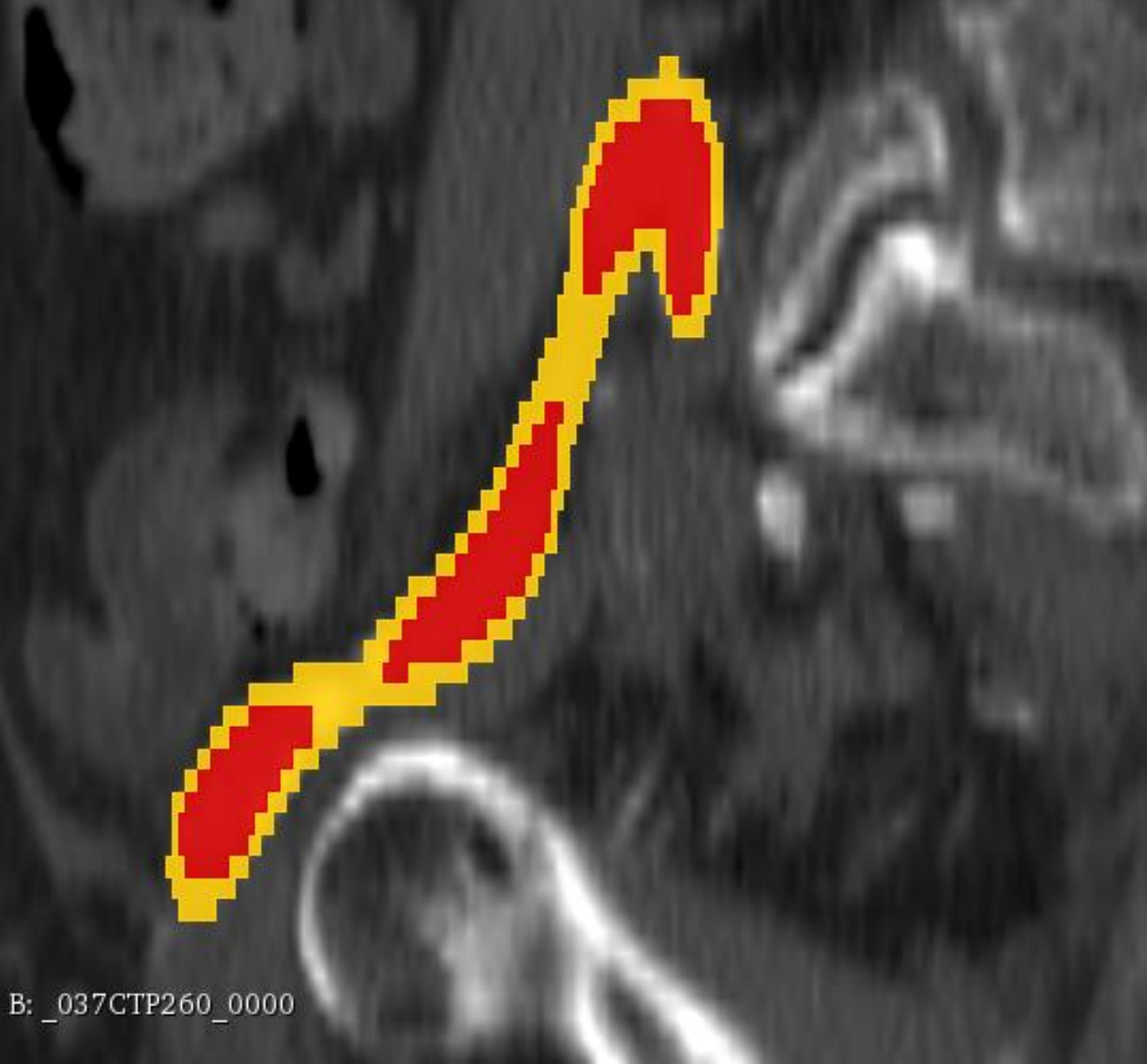}
  \end{subfigure}
    \begin{subfigure}{0.135\linewidth}
     \includegraphics[width=1\textwidth]{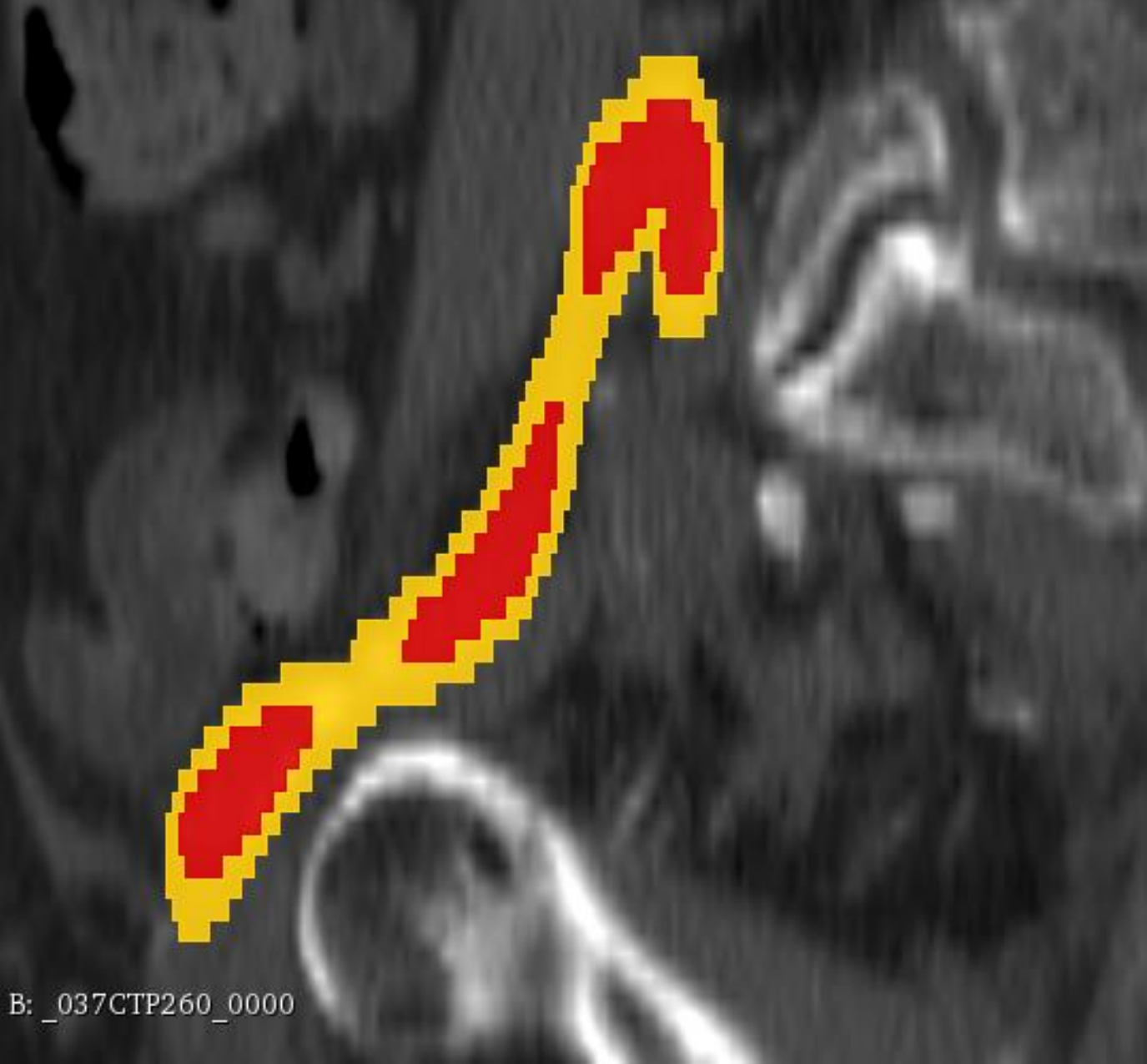}
  \end{subfigure}
      \begin{subfigure}{0.135\linewidth}
     \includegraphics[width=1\textwidth]{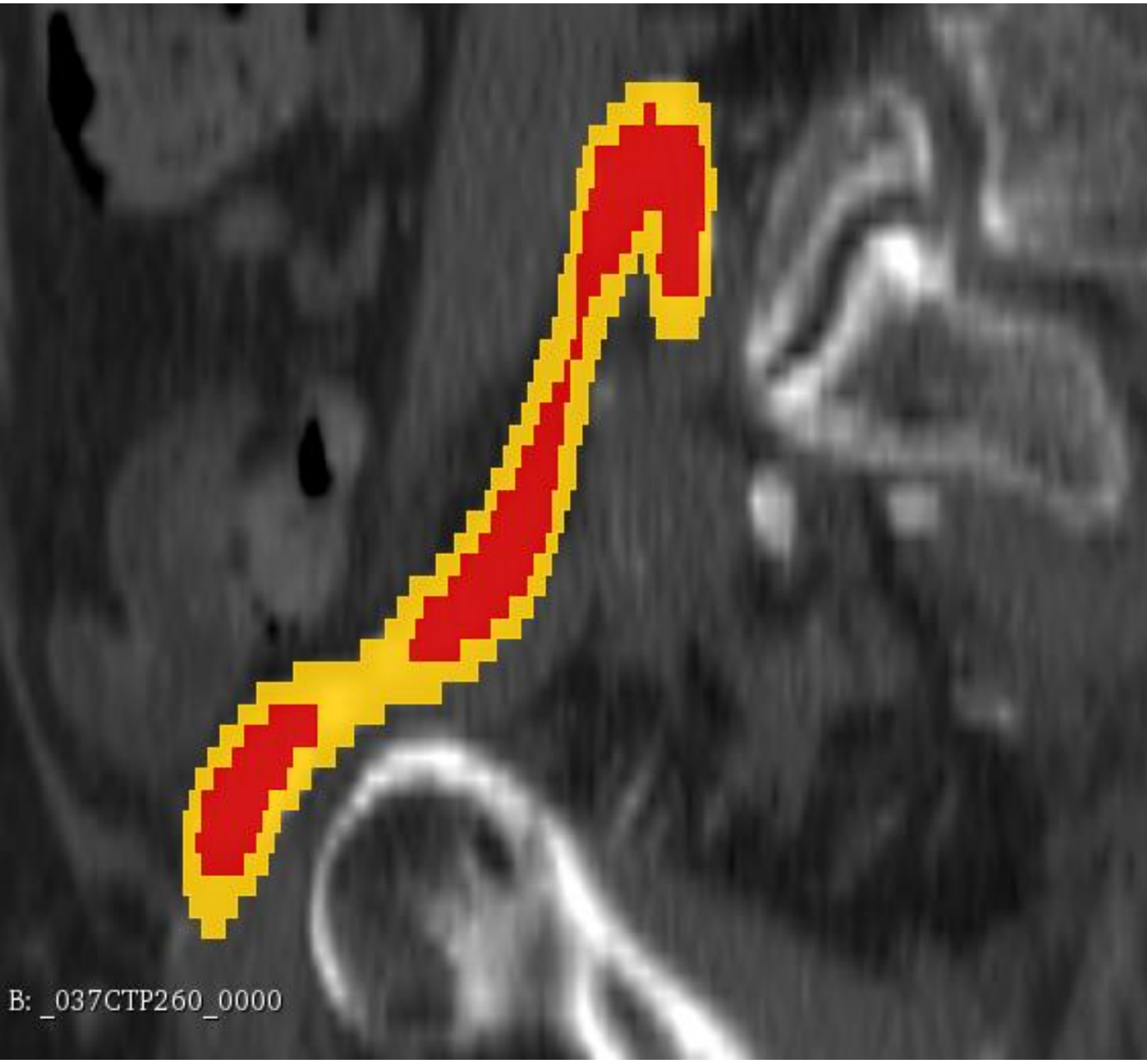}
  \end{subfigure}
      \begin{subfigure}{0.135\linewidth}
     \includegraphics[width=1\textwidth]{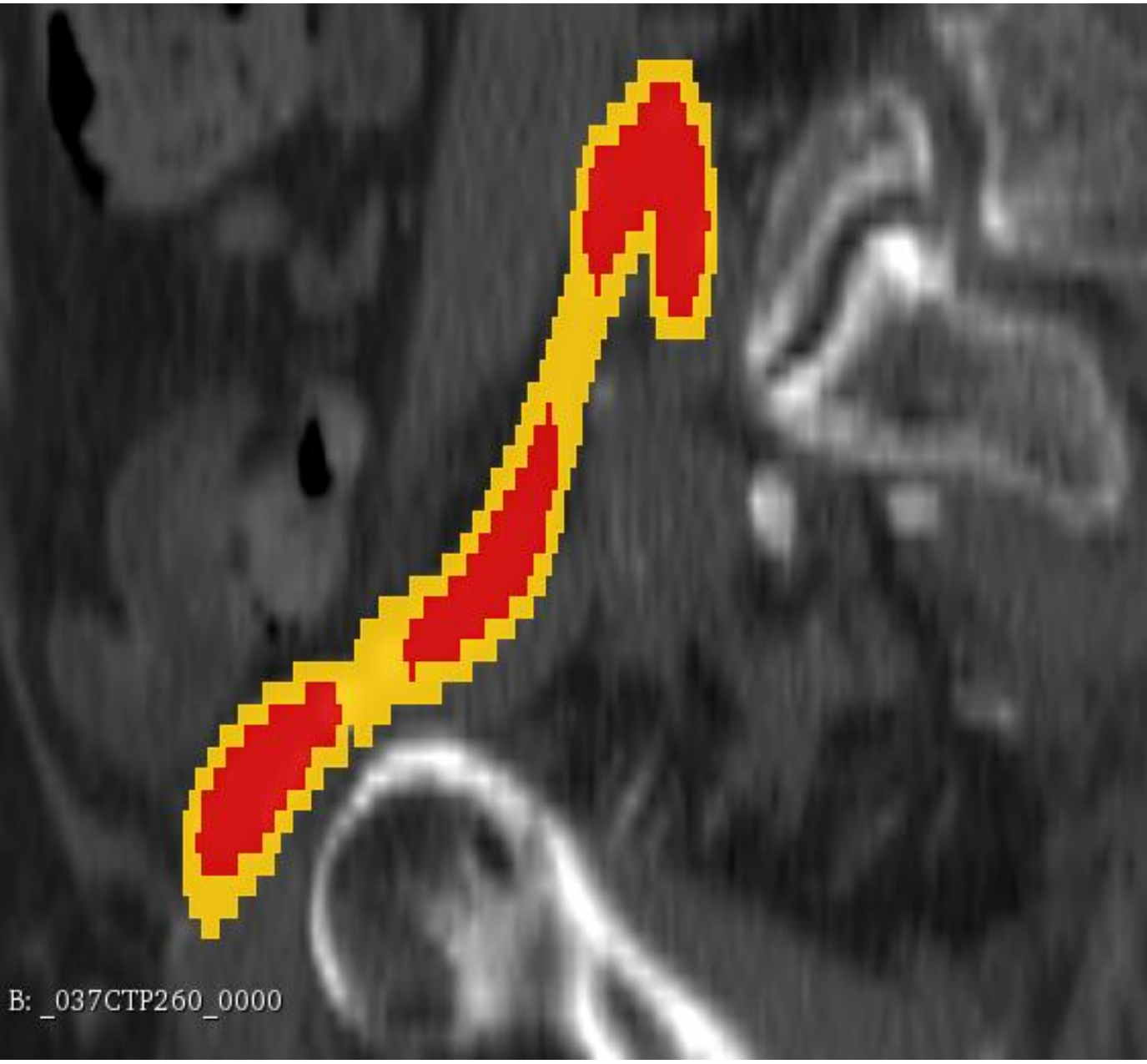}
  \end{subfigure}

\begin{subfigure}{0.135\linewidth}
  \includegraphics[width=1\textwidth]{figures/empty2.pdf}
  \end{subfigure}
    \begin{subfigure}{0.135\linewidth}
     \includegraphics[width=1\textwidth]{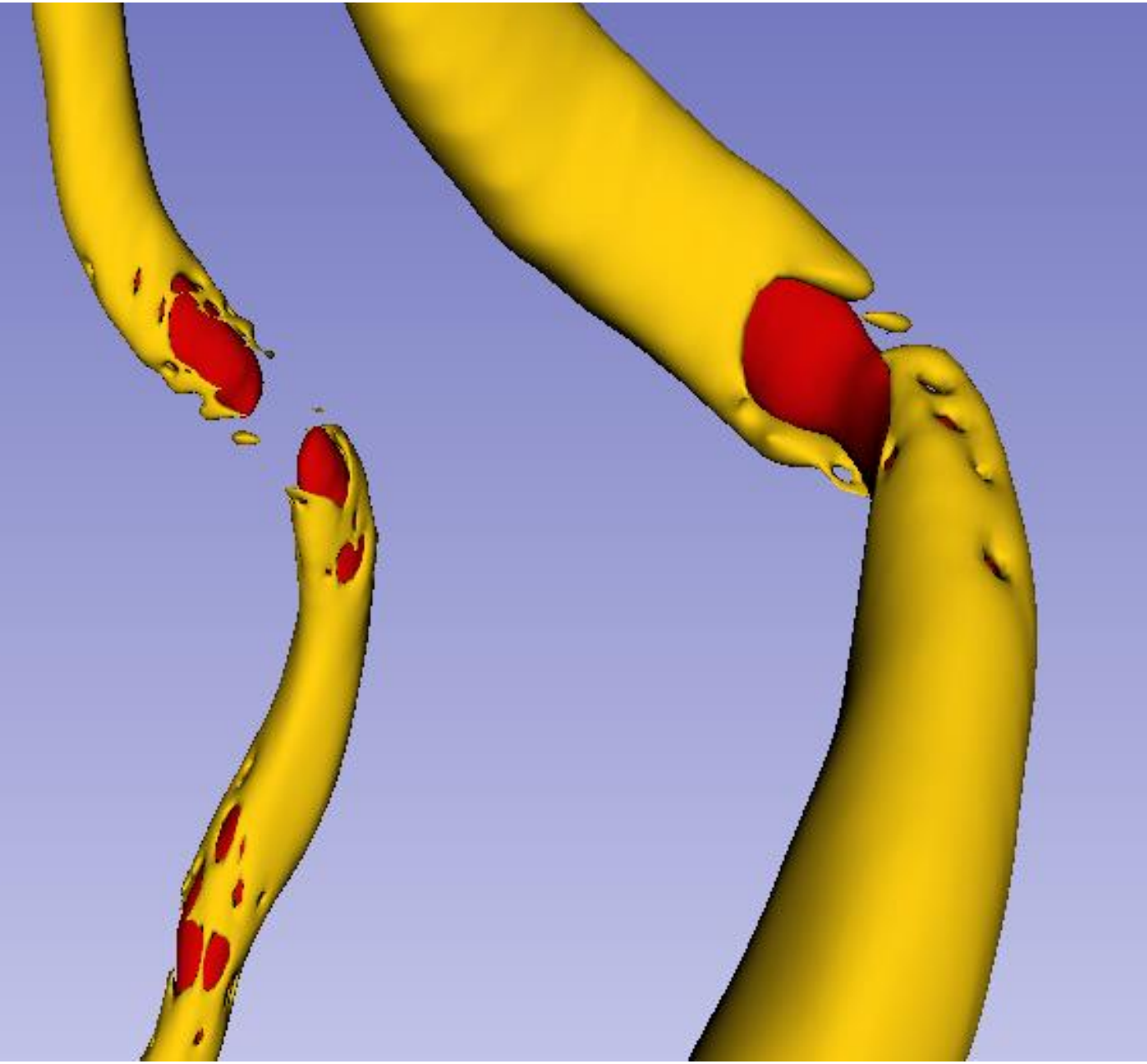}
  \end{subfigure}
  \begin{subfigure}{0.135\linewidth}
     \includegraphics[width=1\textwidth]{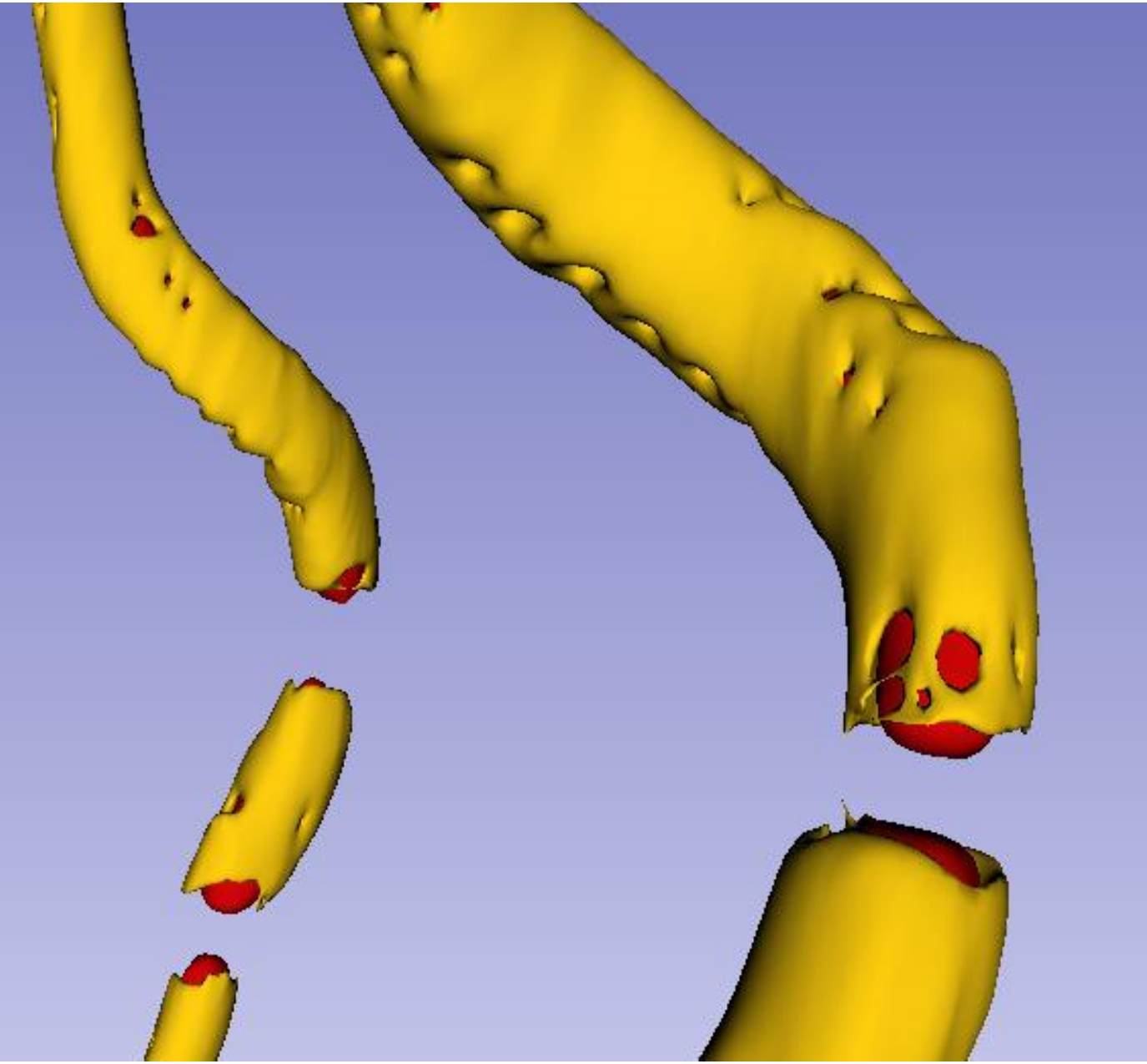}
  \end{subfigure}
      \begin{subfigure}{0.135\linewidth}
     \includegraphics[width=1\textwidth]{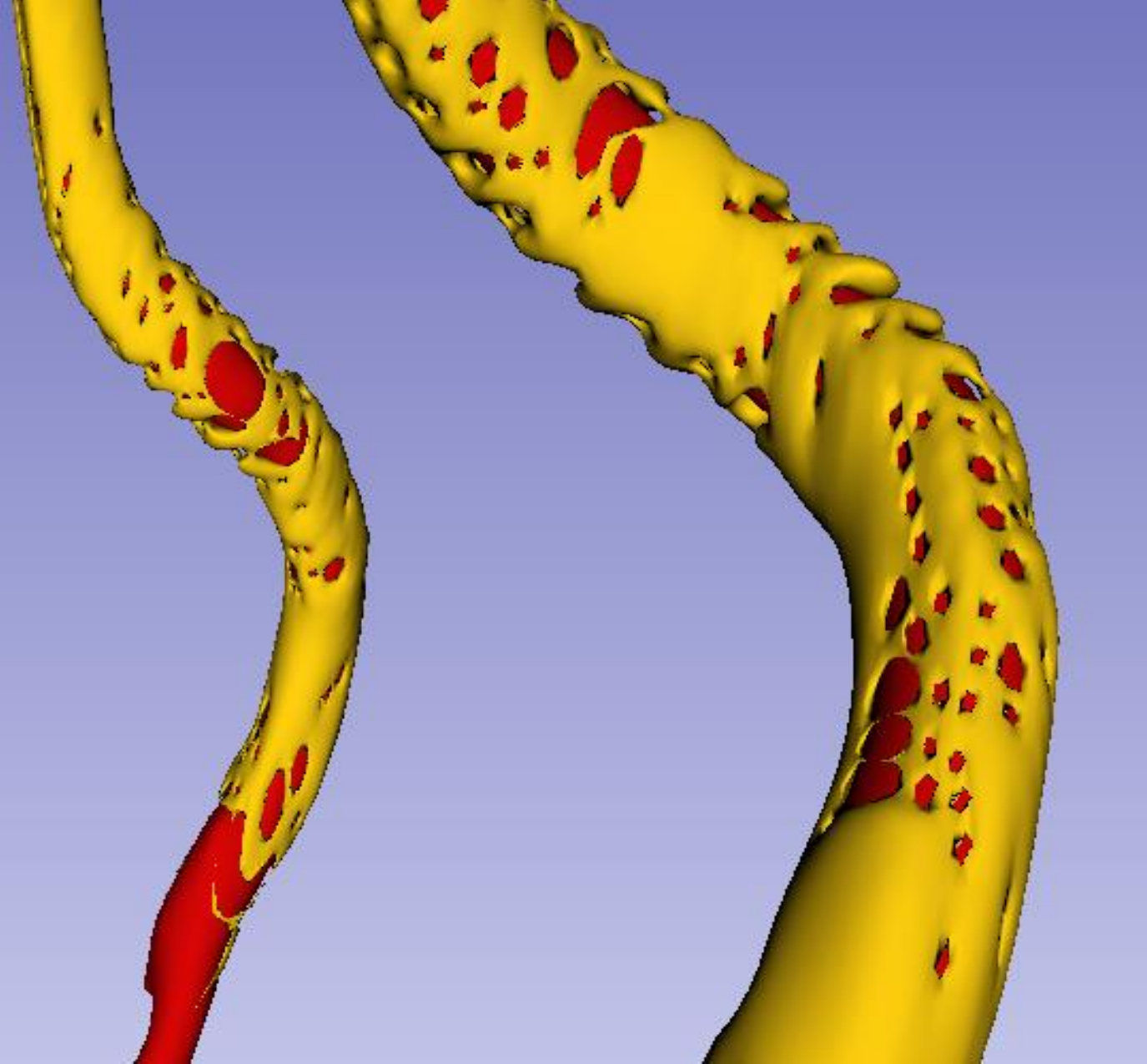}
  \end{subfigure}
    \begin{subfigure}{0.135\linewidth}
     \includegraphics[width=1\textwidth]{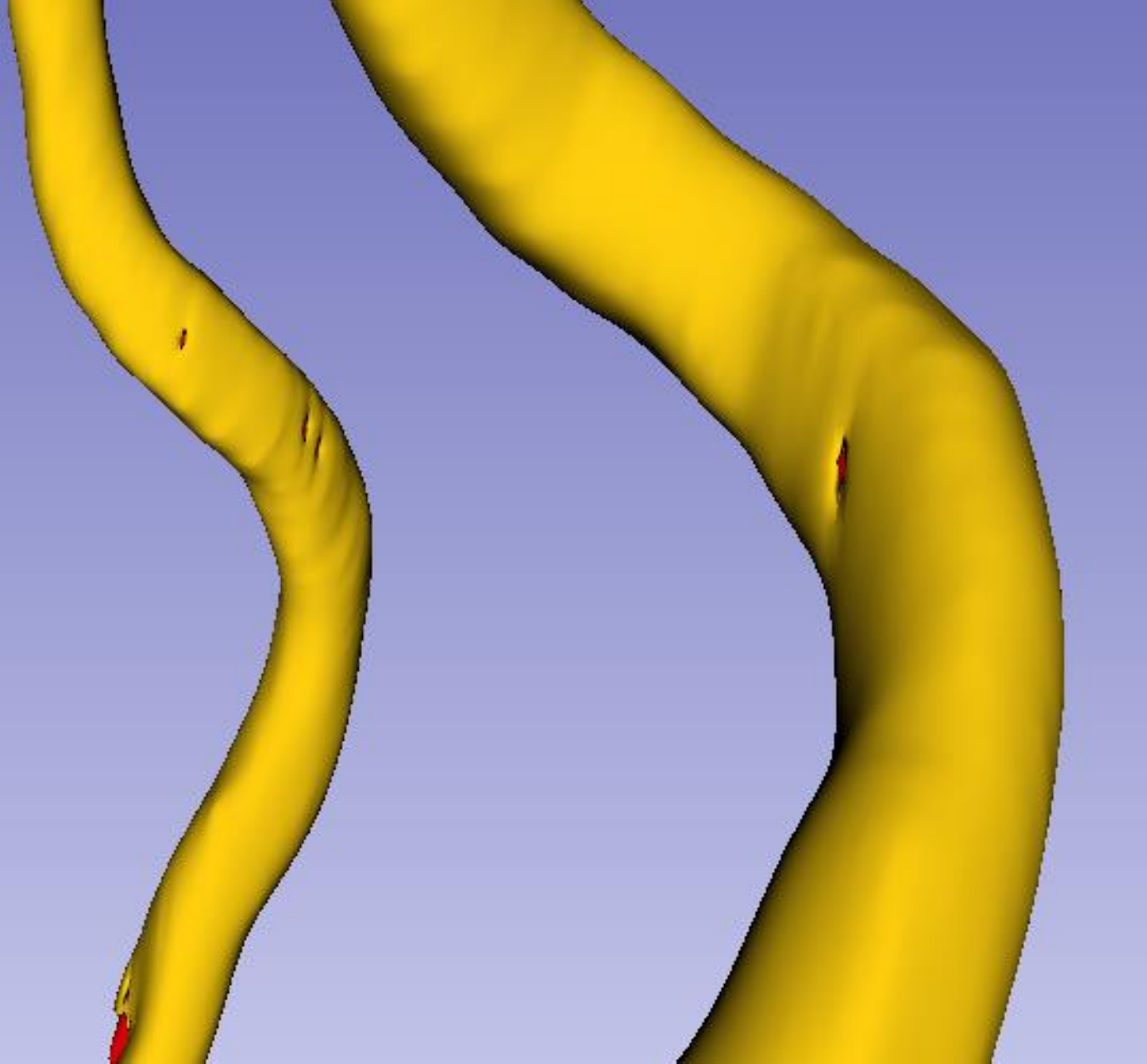}
  \end{subfigure}
      \begin{subfigure}{0.135\linewidth}
     \includegraphics[width=1\textwidth]{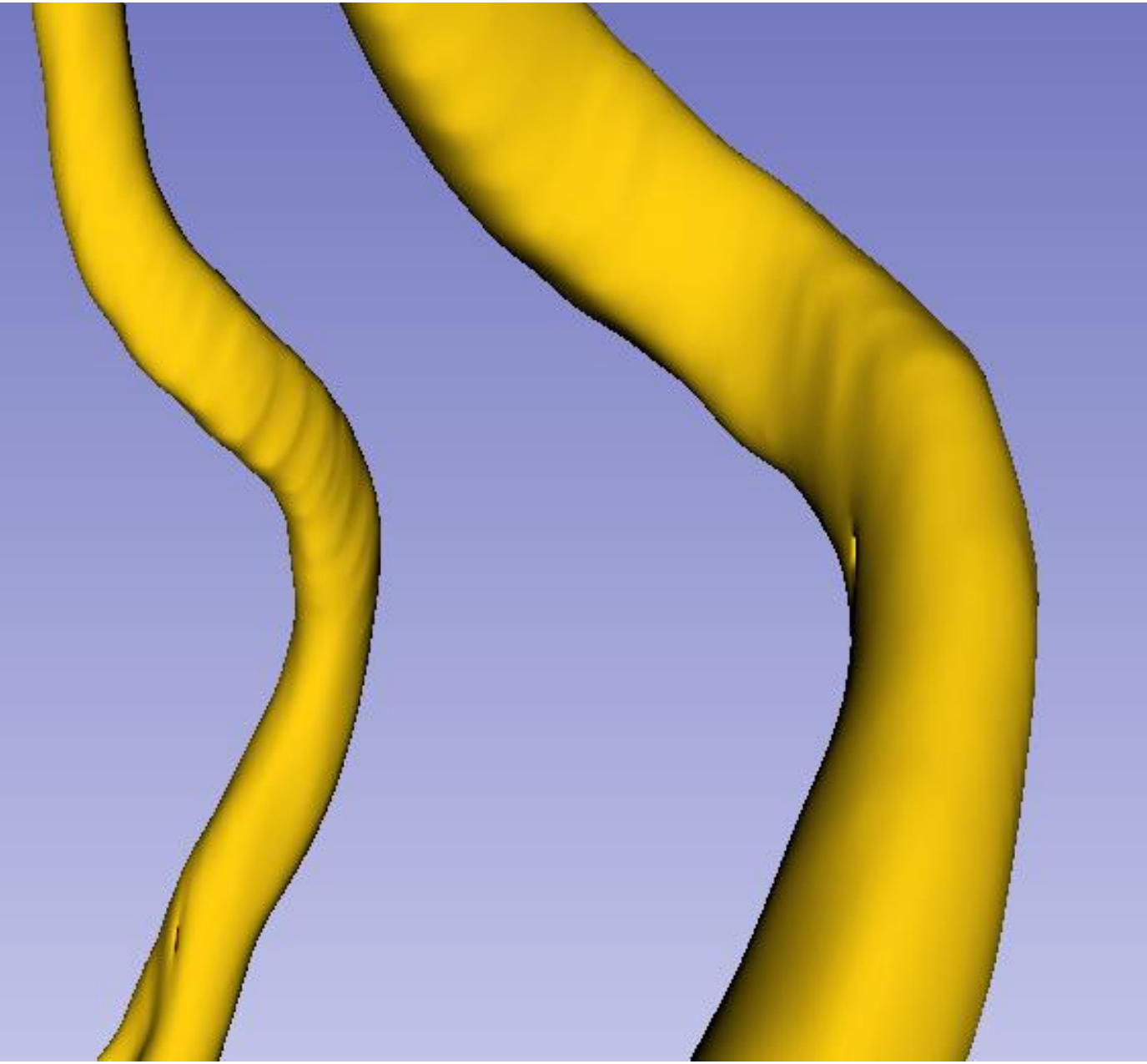}
  \end{subfigure}
      \begin{subfigure}{0.135\linewidth}
     \includegraphics[width=1\textwidth]{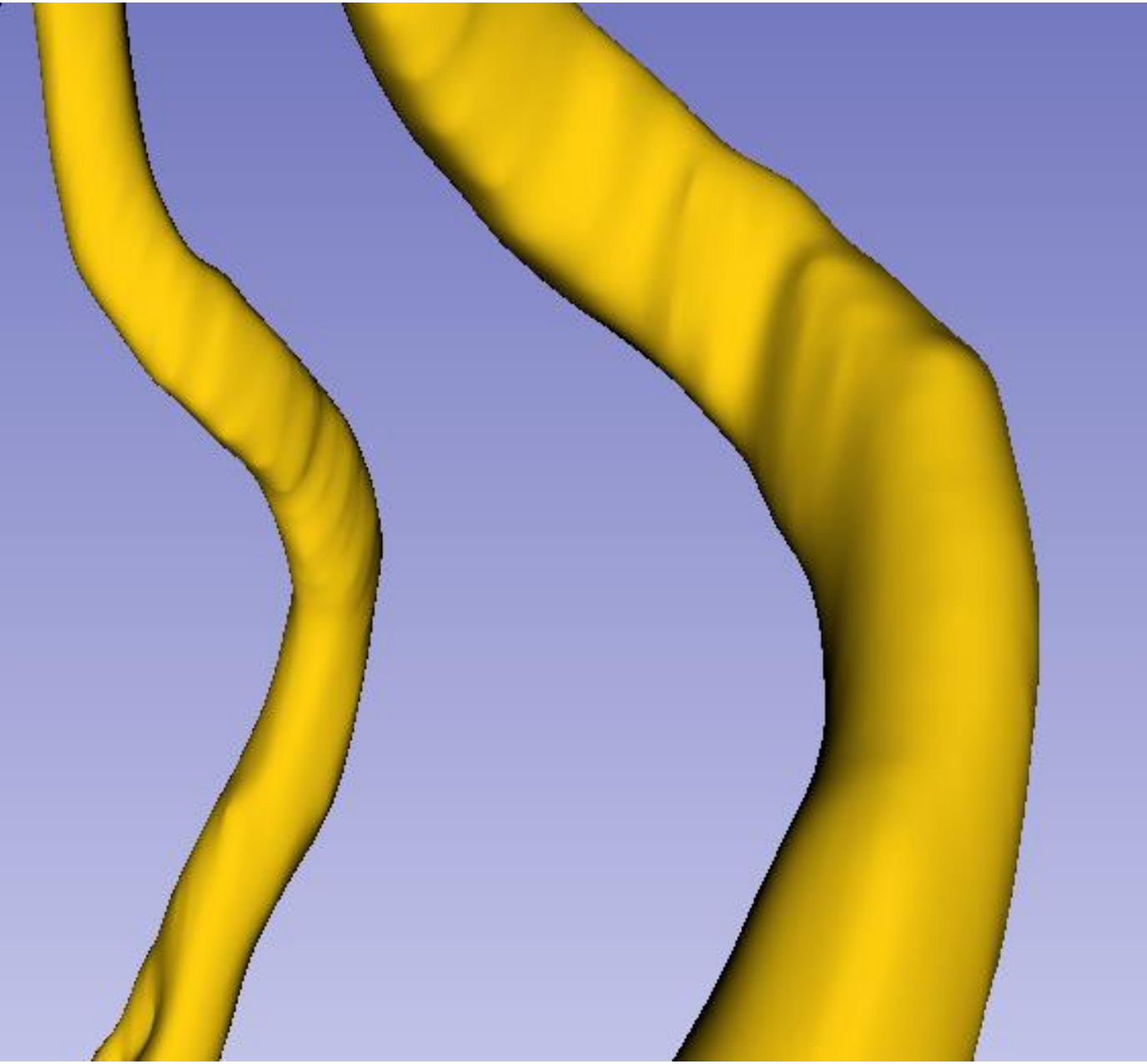}
  \end{subfigure}

      \begin{subfigure}{0.135\linewidth}
  \includegraphics[width=1\textwidth]{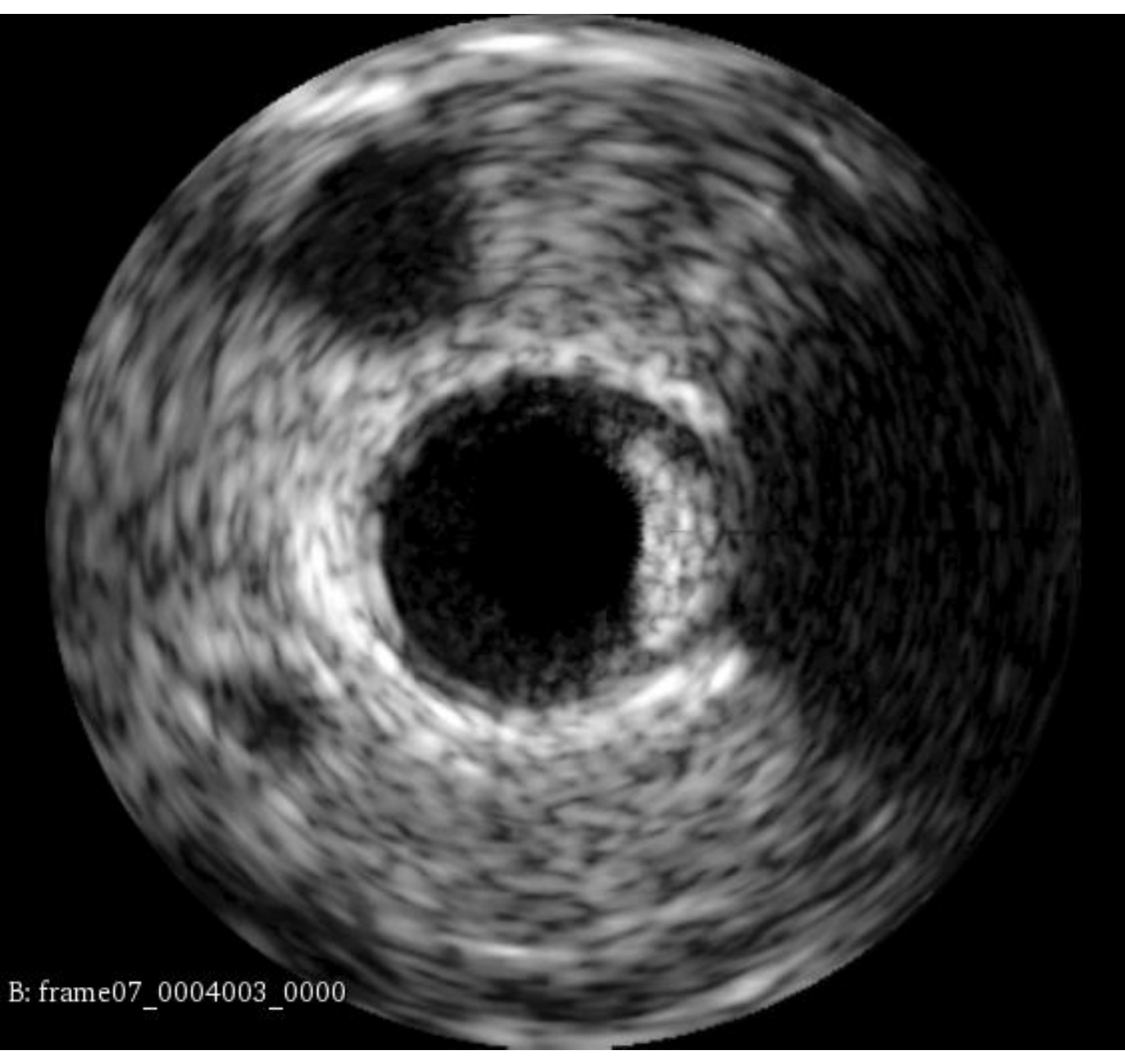}
  \end{subfigure}
    \begin{subfigure}{0.135\linewidth}
     \includegraphics[width=1\textwidth]{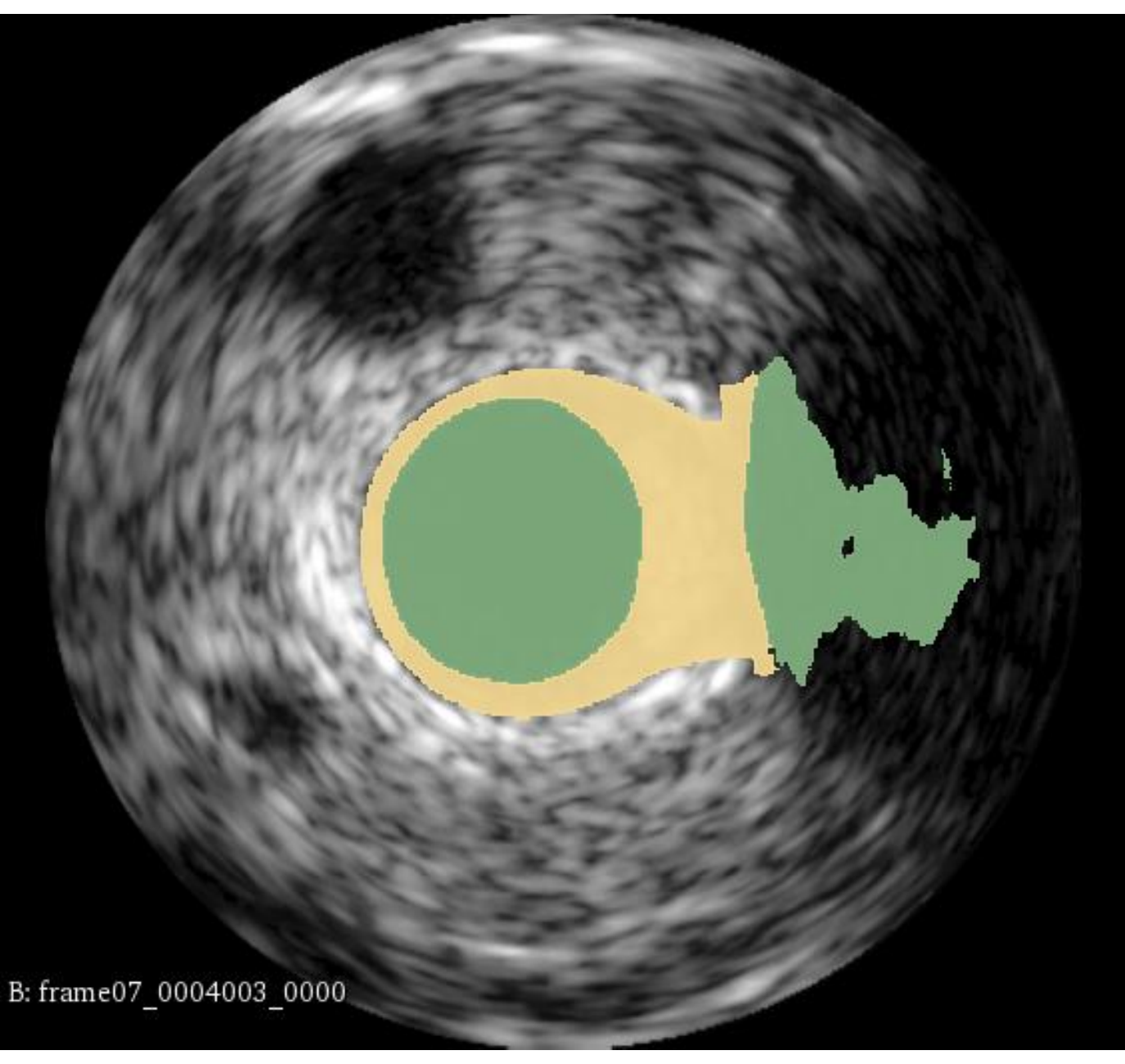}
  \end{subfigure}
      \begin{subfigure}{0.135\linewidth}
     \includegraphics[width=1\textwidth]{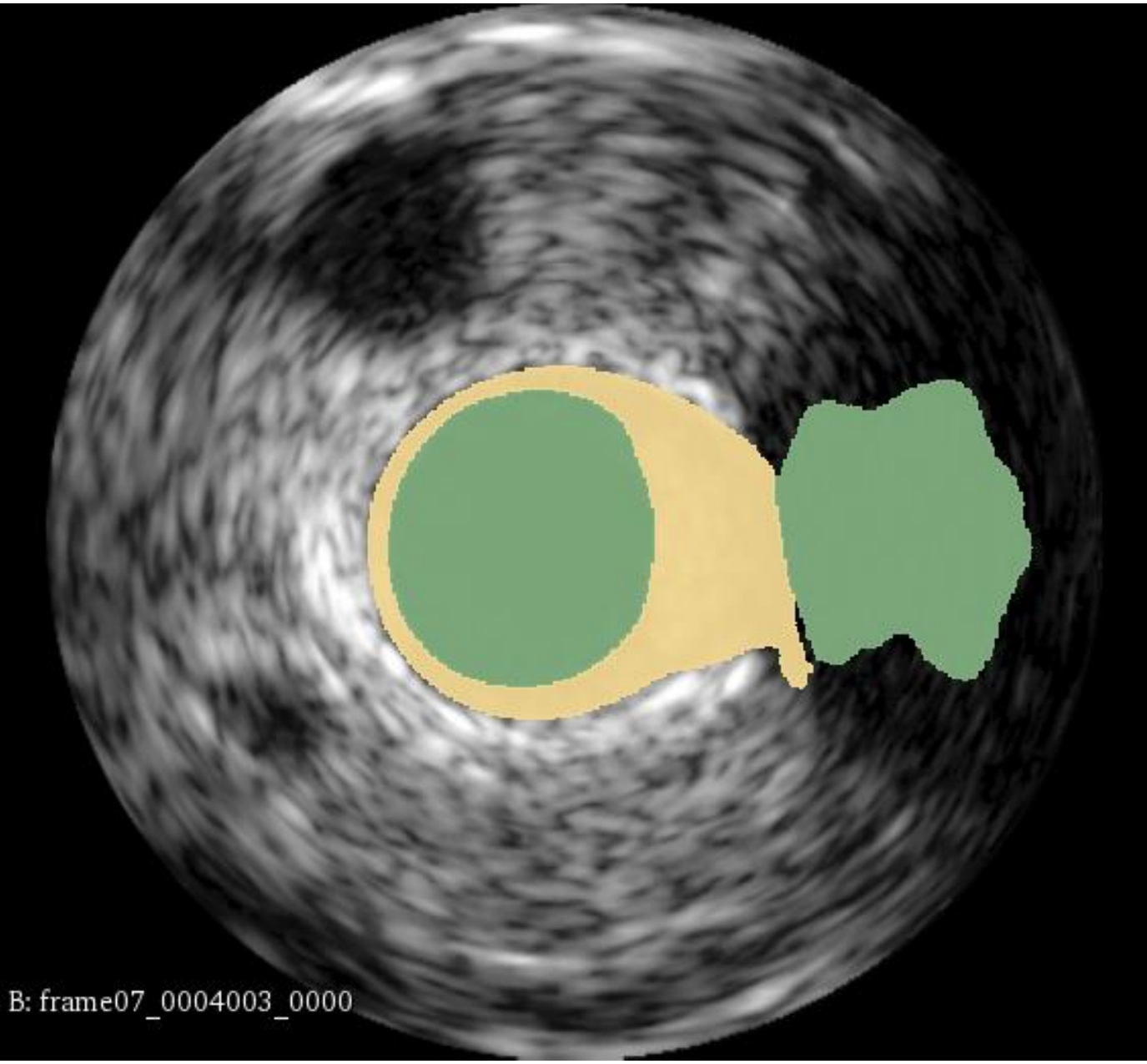}
  \end{subfigure}
      \begin{subfigure}{0.135\linewidth}
     \includegraphics[width=1\textwidth]{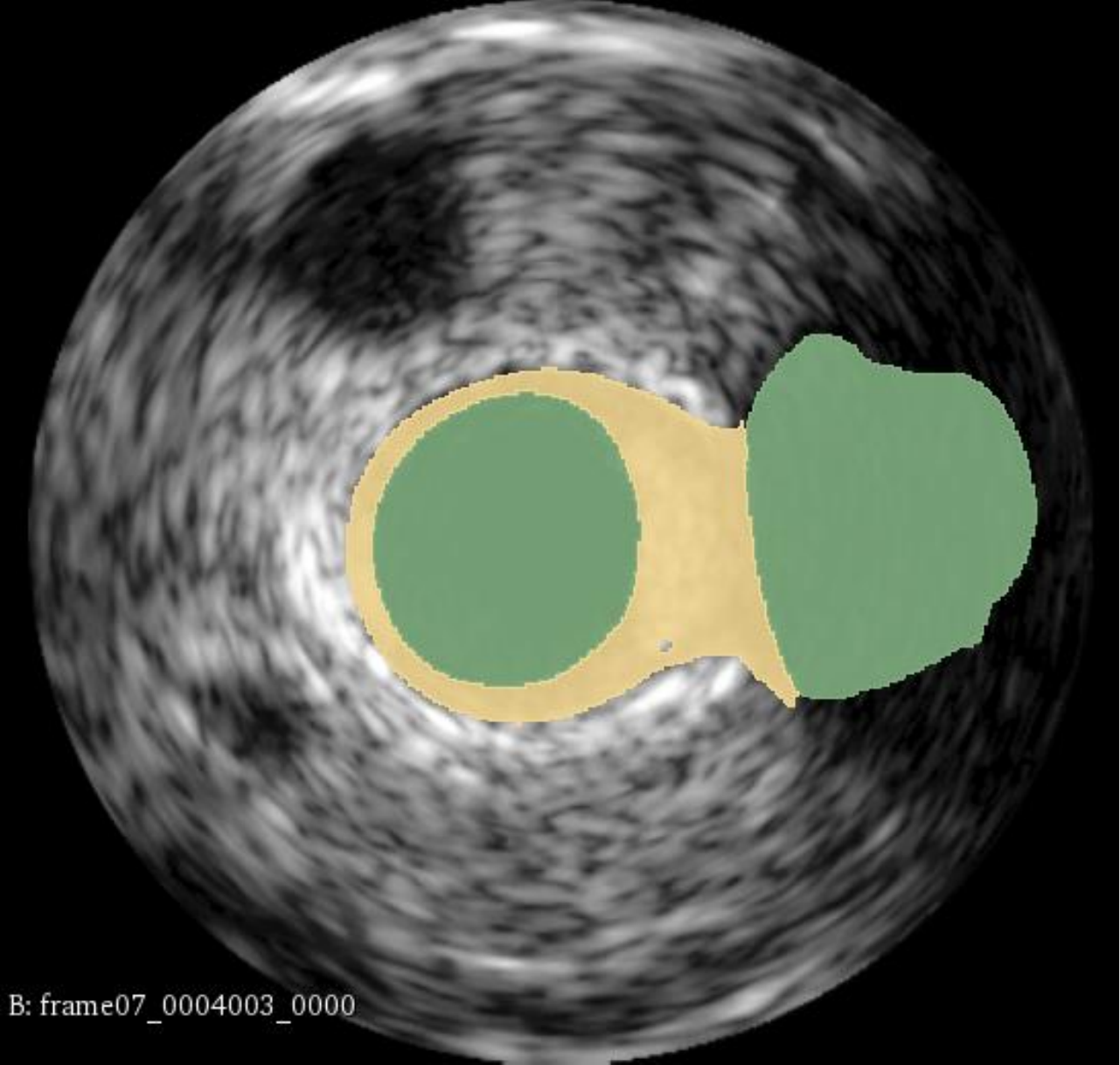}
  \end{subfigure}
    \begin{subfigure}{0.135\linewidth}
     \includegraphics[width=1\textwidth]{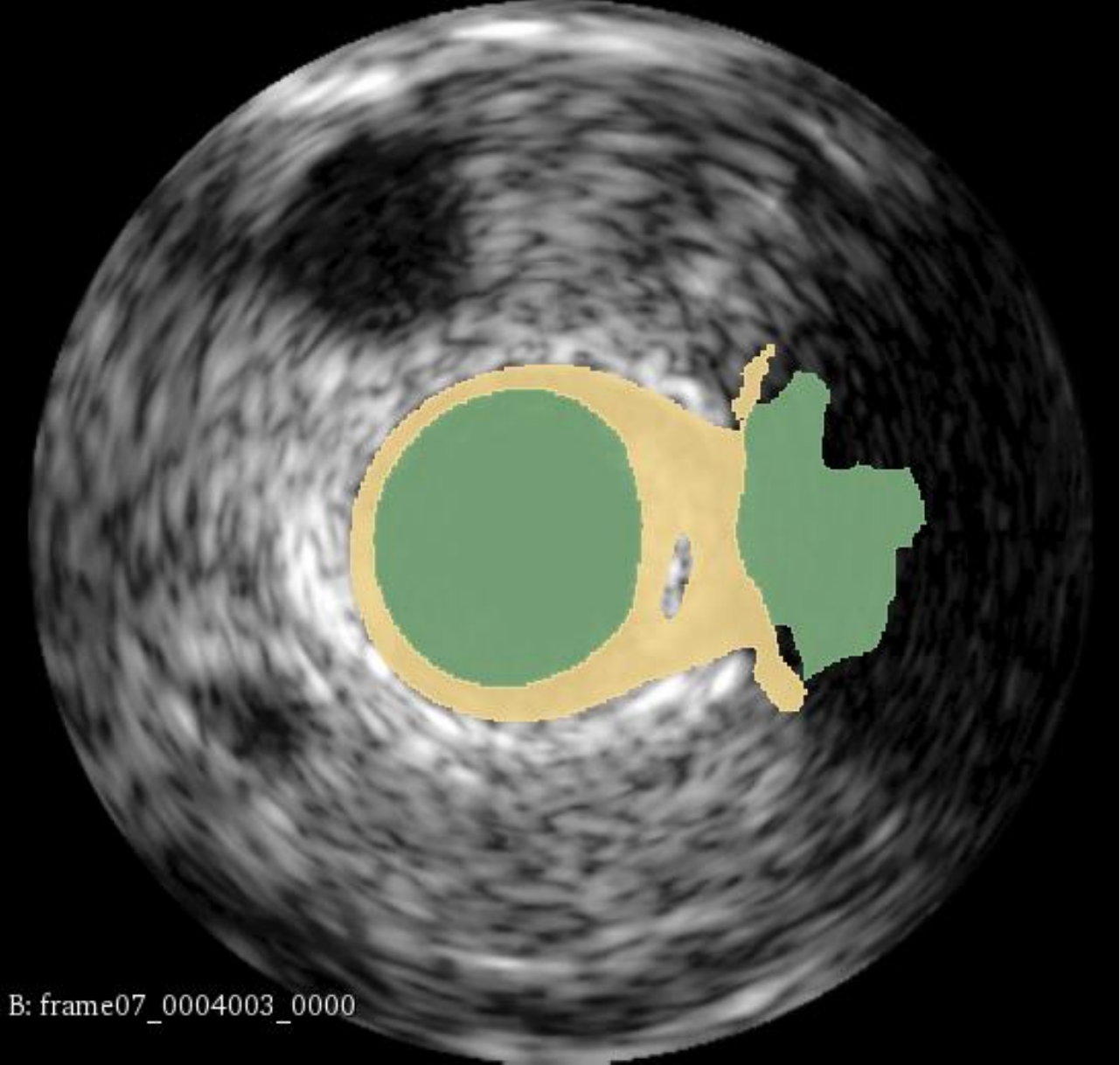}
  \end{subfigure}
      \begin{subfigure}{0.135\linewidth}
     \includegraphics[width=1\textwidth]{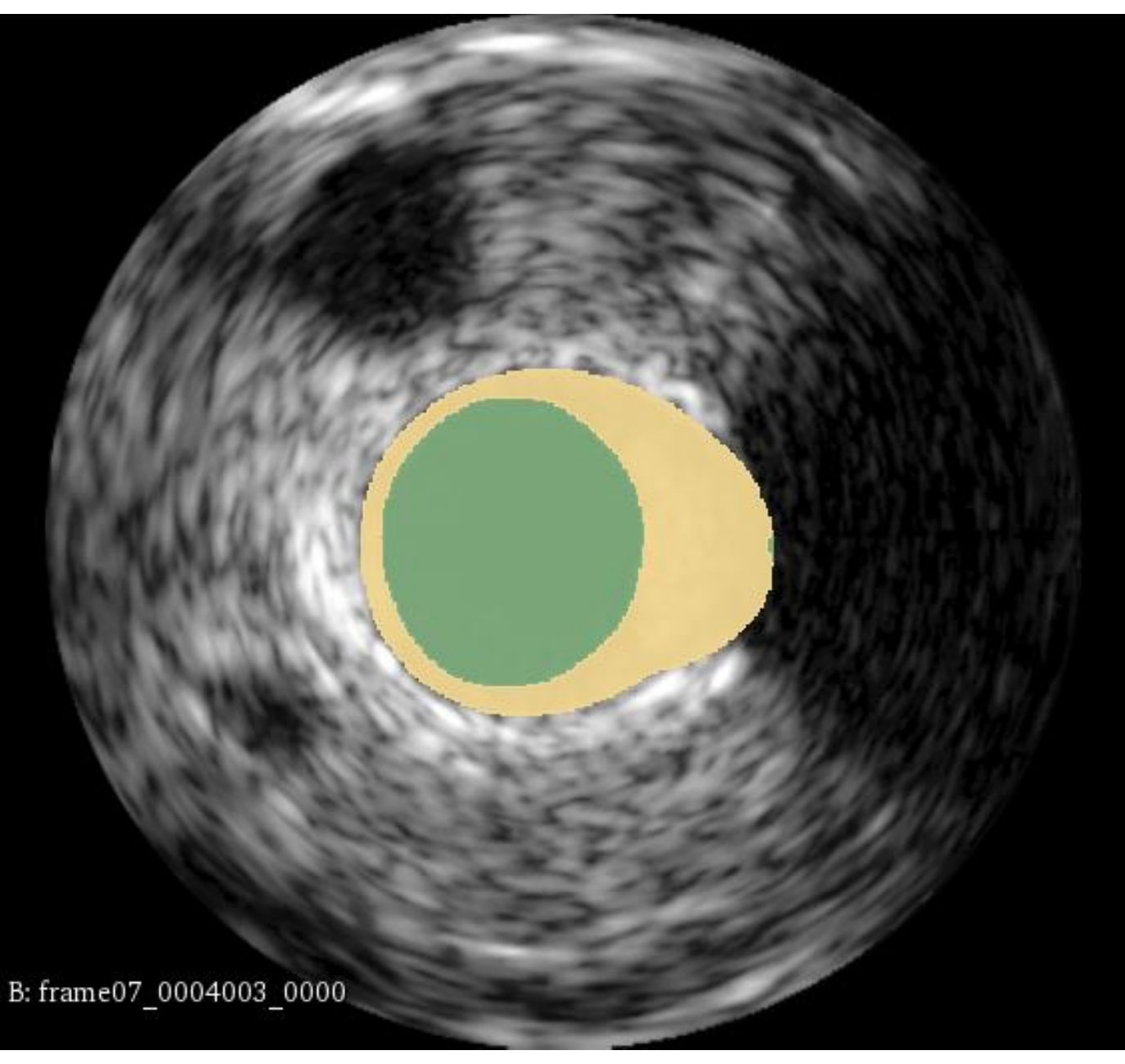}
  \end{subfigure}
      \begin{subfigure}{0.135\linewidth}
     \includegraphics[width=1\textwidth]{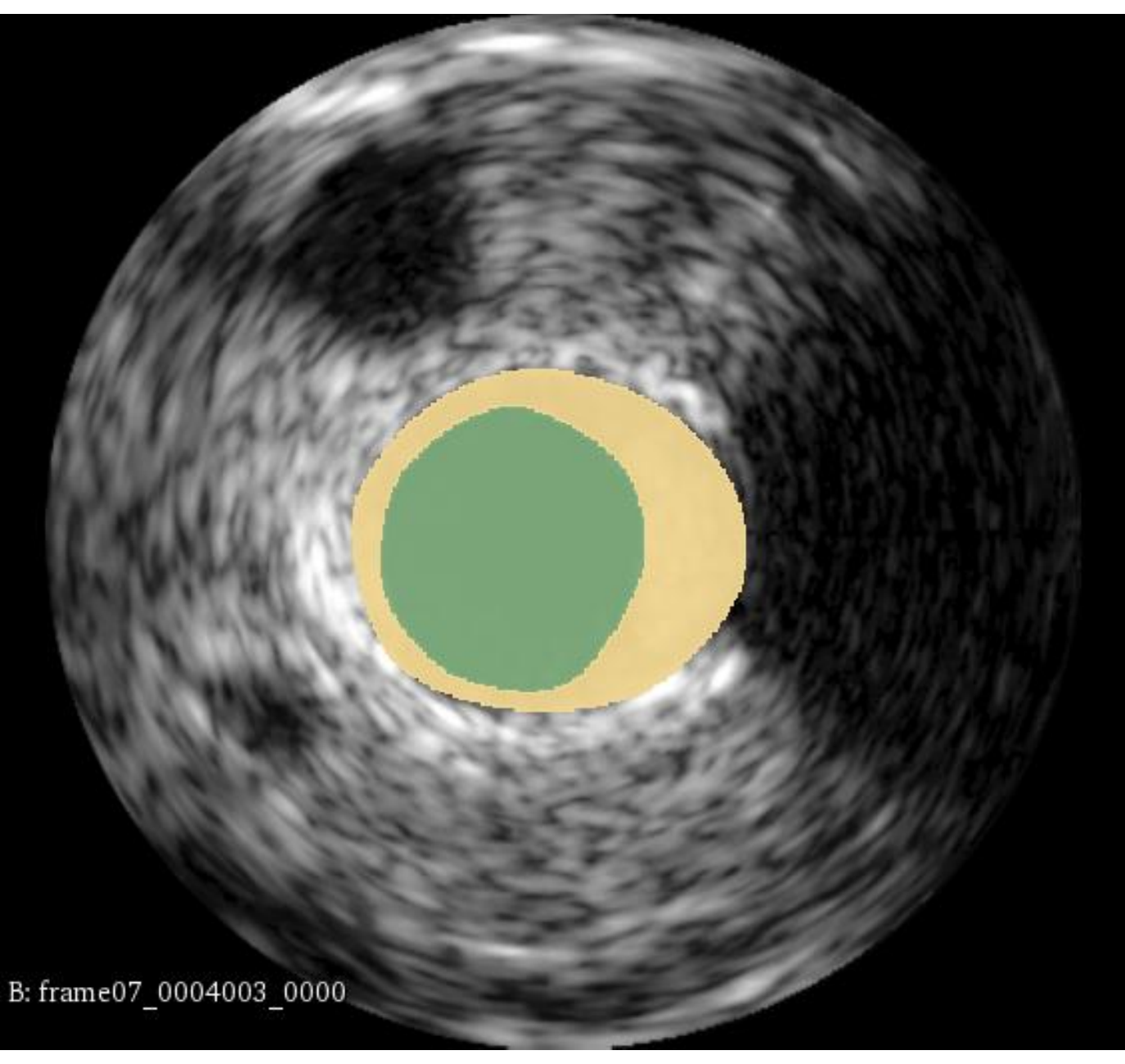}
  \end{subfigure}


 \begin{subfigure}{0.135\linewidth}
  \includegraphics[width=1\textwidth]{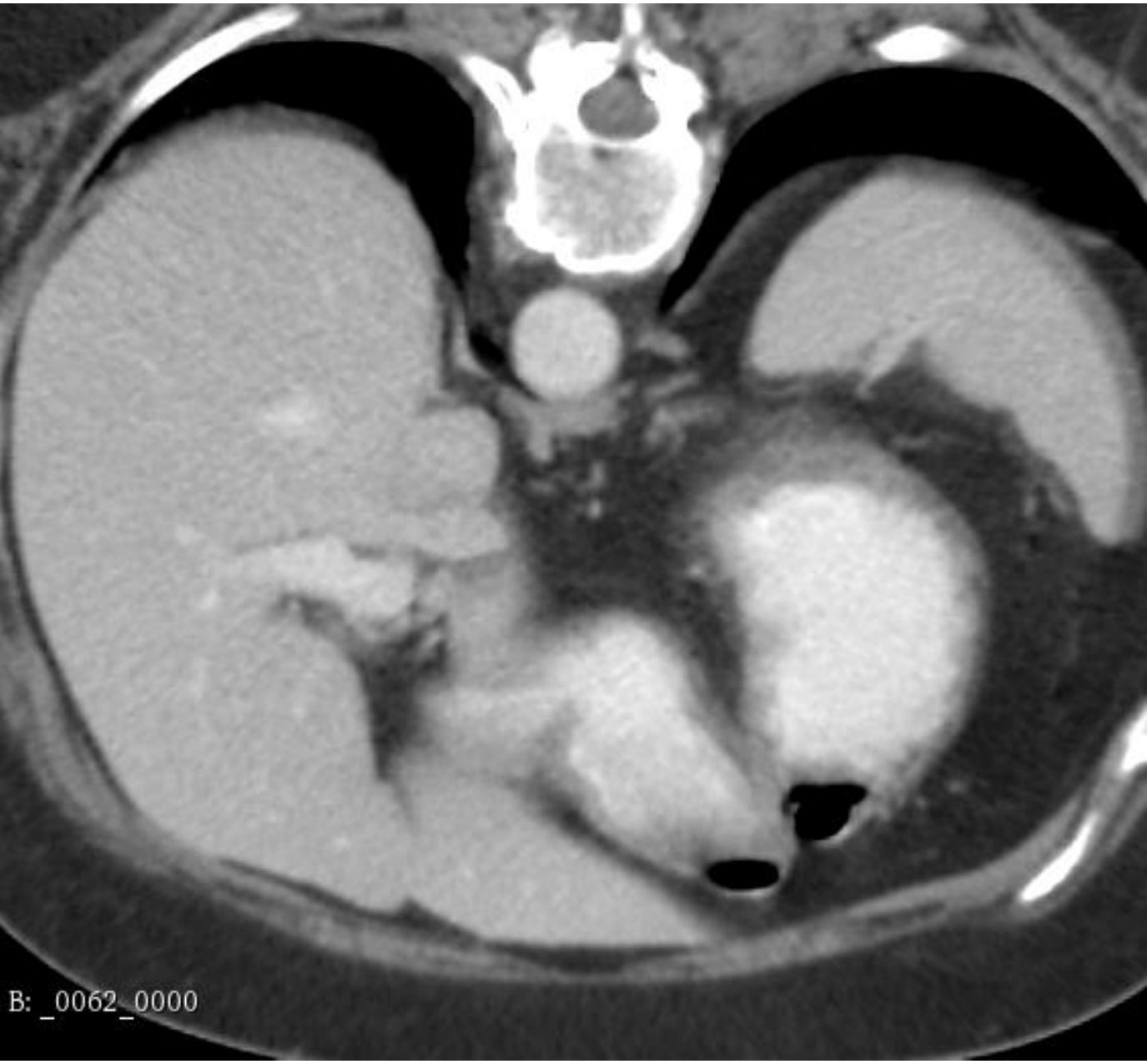}
  \end{subfigure}
    \begin{subfigure}{0.135\linewidth}
     \includegraphics[width=1\textwidth]{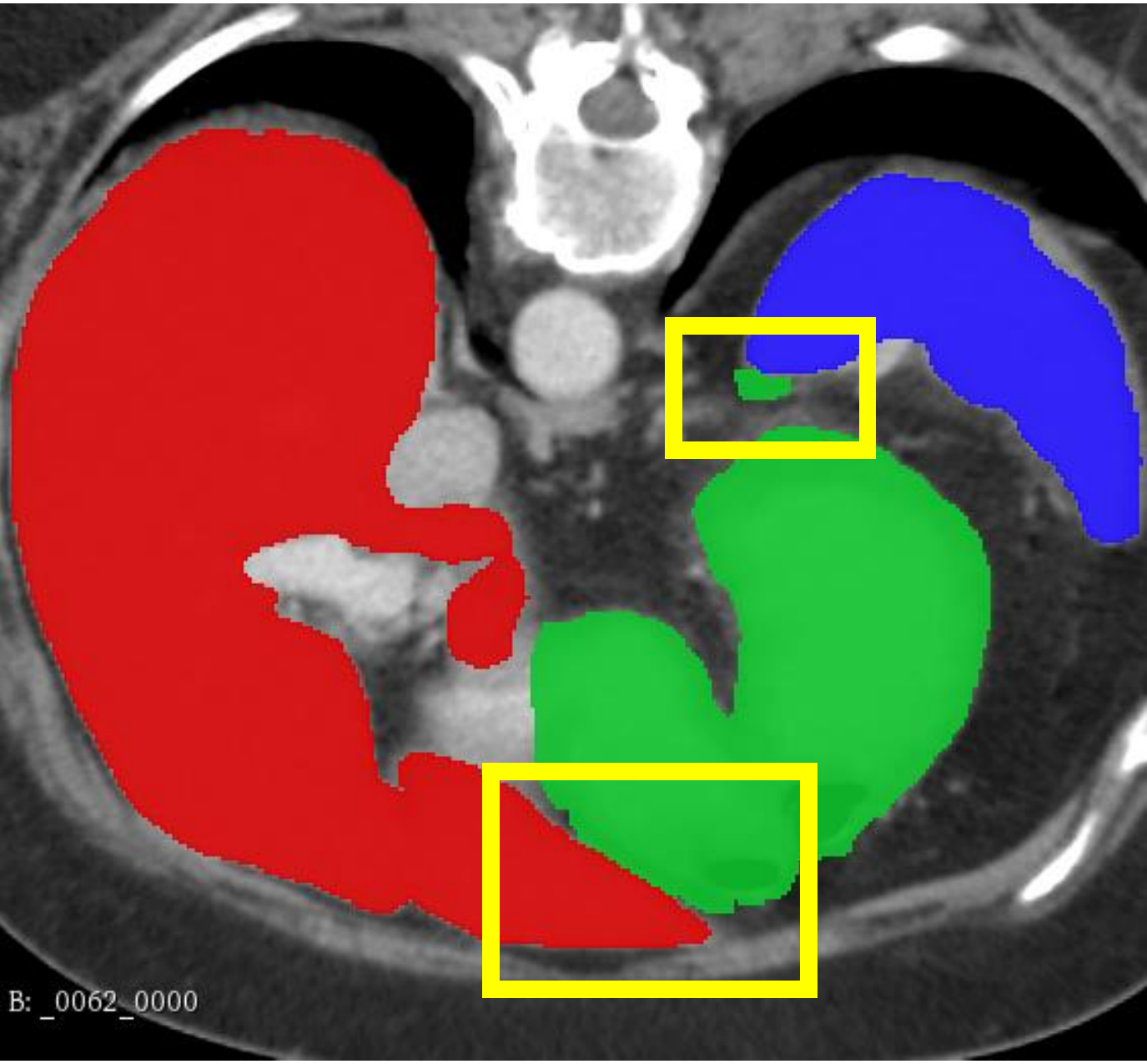}
  \end{subfigure}
  \begin{subfigure}{0.135\linewidth}
     \includegraphics[width=1\textwidth]{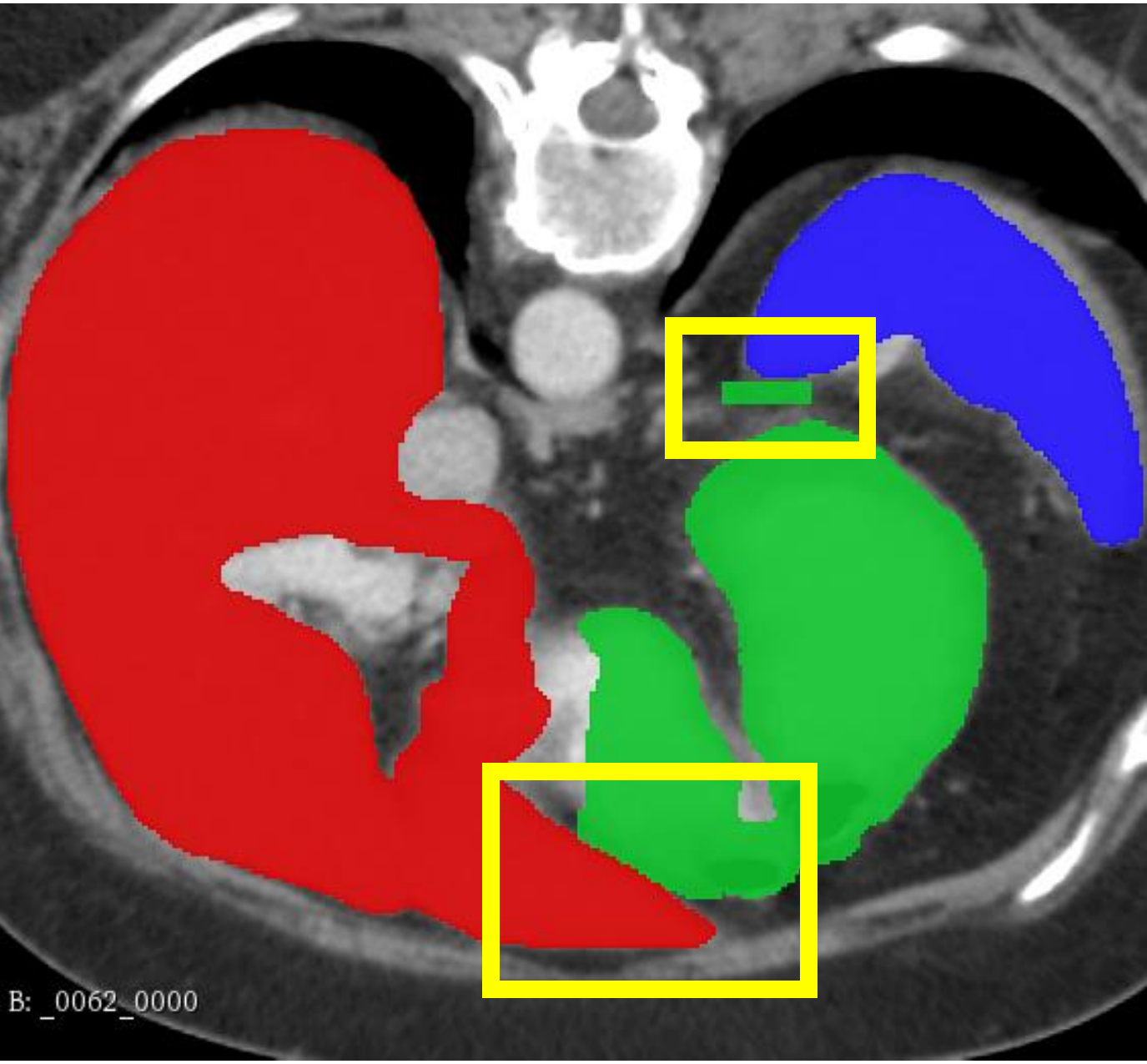}
  \end{subfigure}
      \begin{subfigure}{0.135\linewidth}
     \includegraphics[width=1\textwidth]{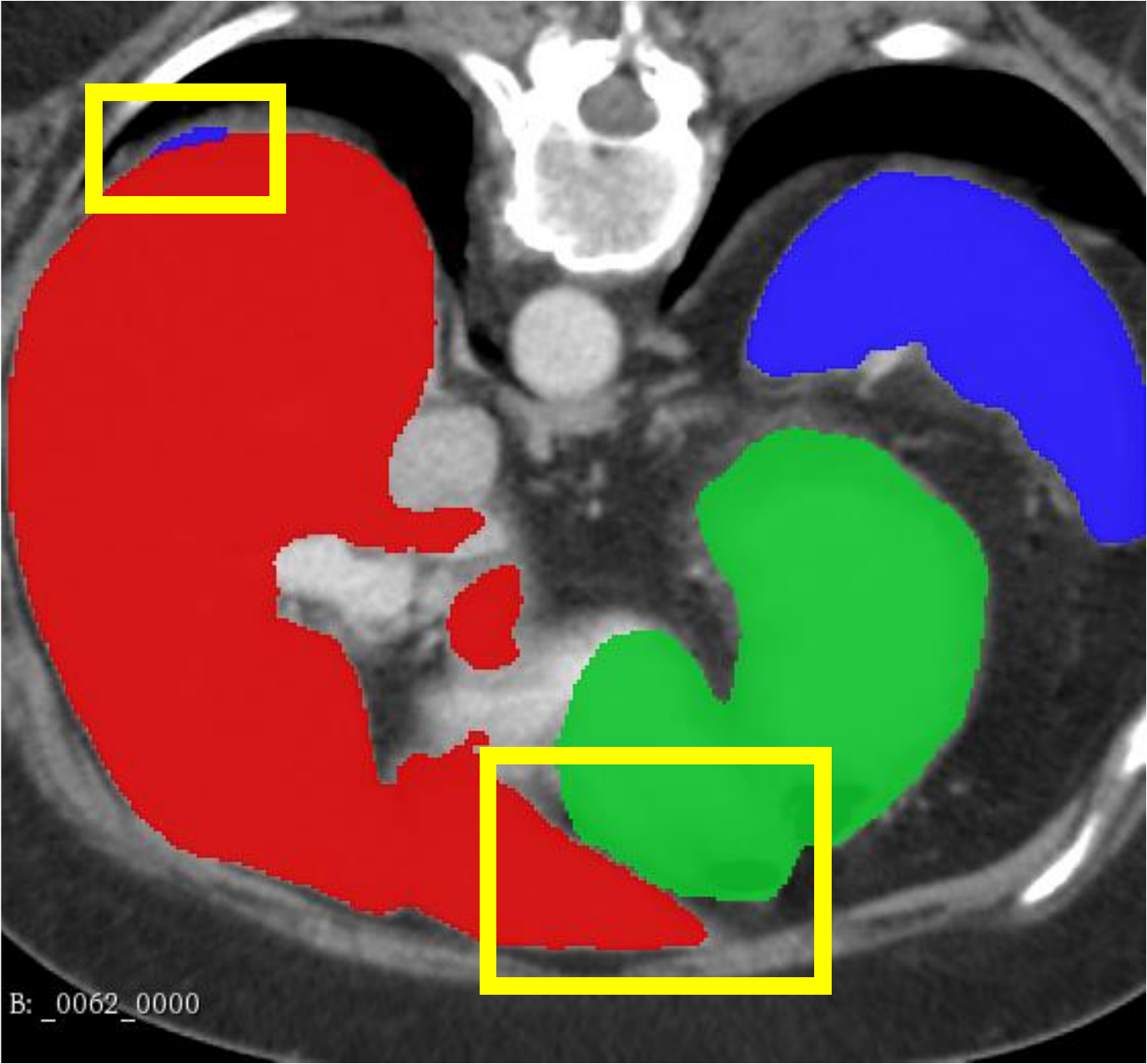}
  \end{subfigure}
  \begin{subfigure}{0.135\linewidth}
     \includegraphics[width=1\textwidth]{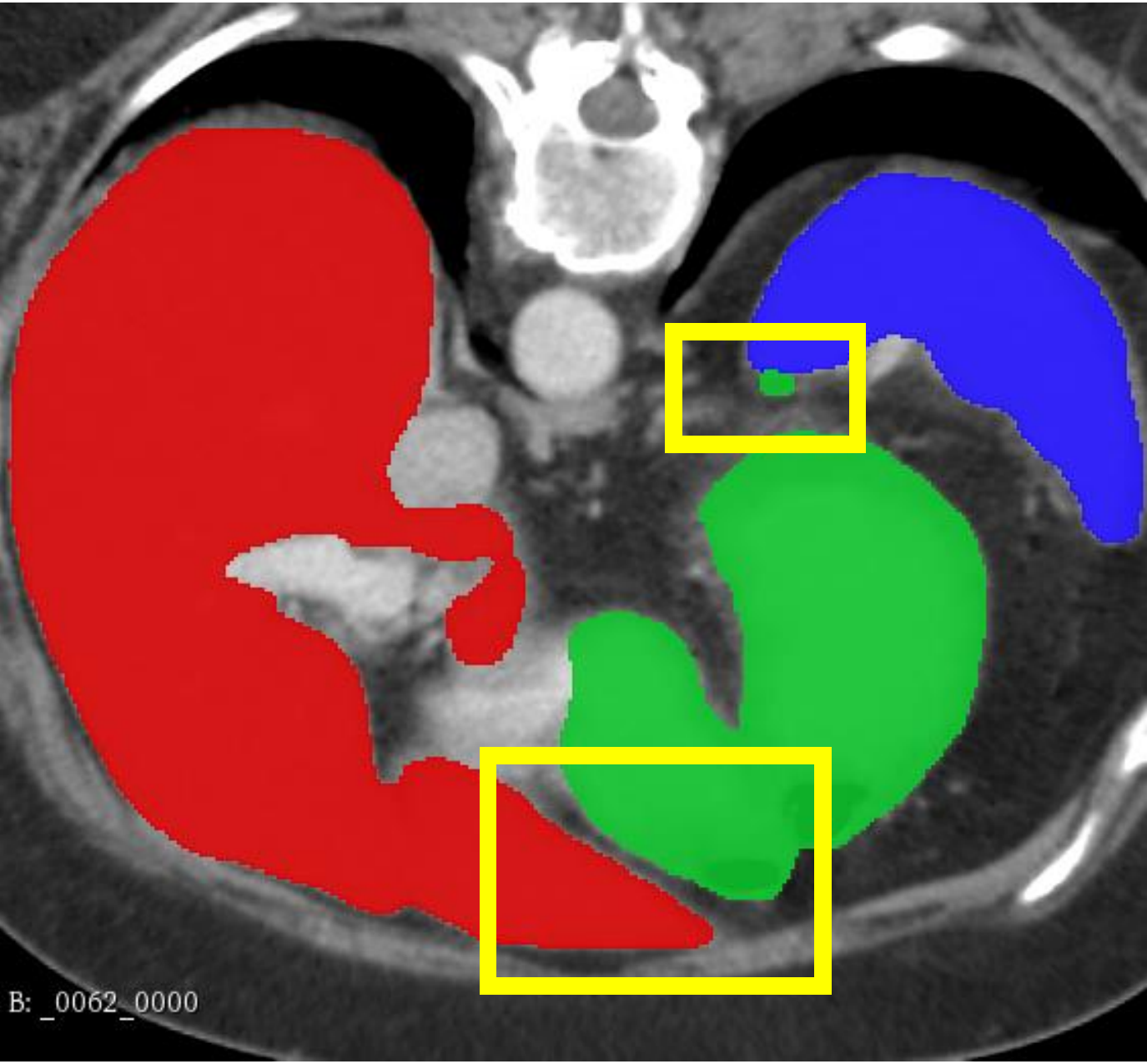}
  \end{subfigure}
      \begin{subfigure}{0.135\linewidth}
     \includegraphics[width=1\textwidth]{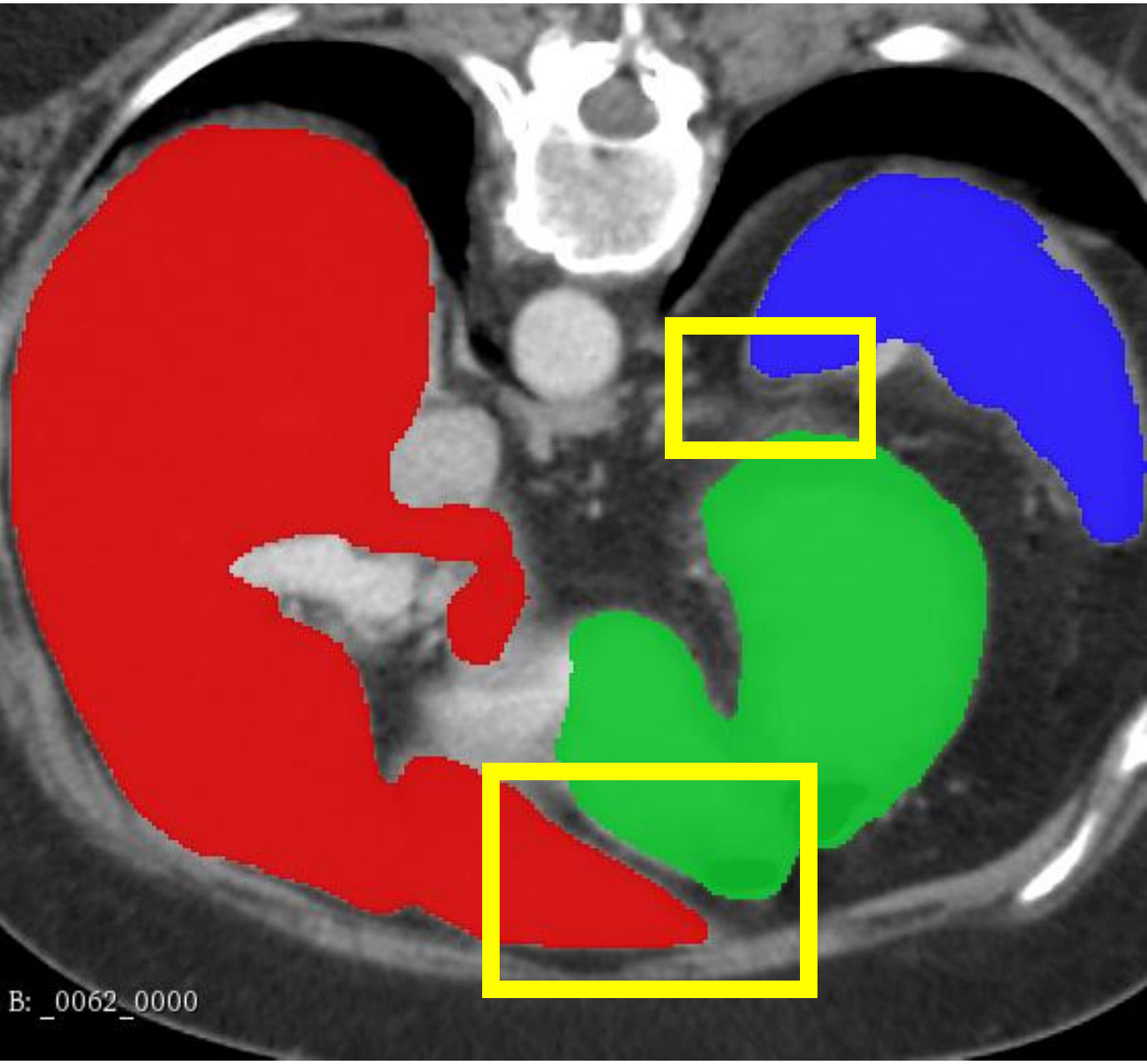}
  \end{subfigure}
      \begin{subfigure}{0.135\linewidth}
     \includegraphics[width=1\textwidth]{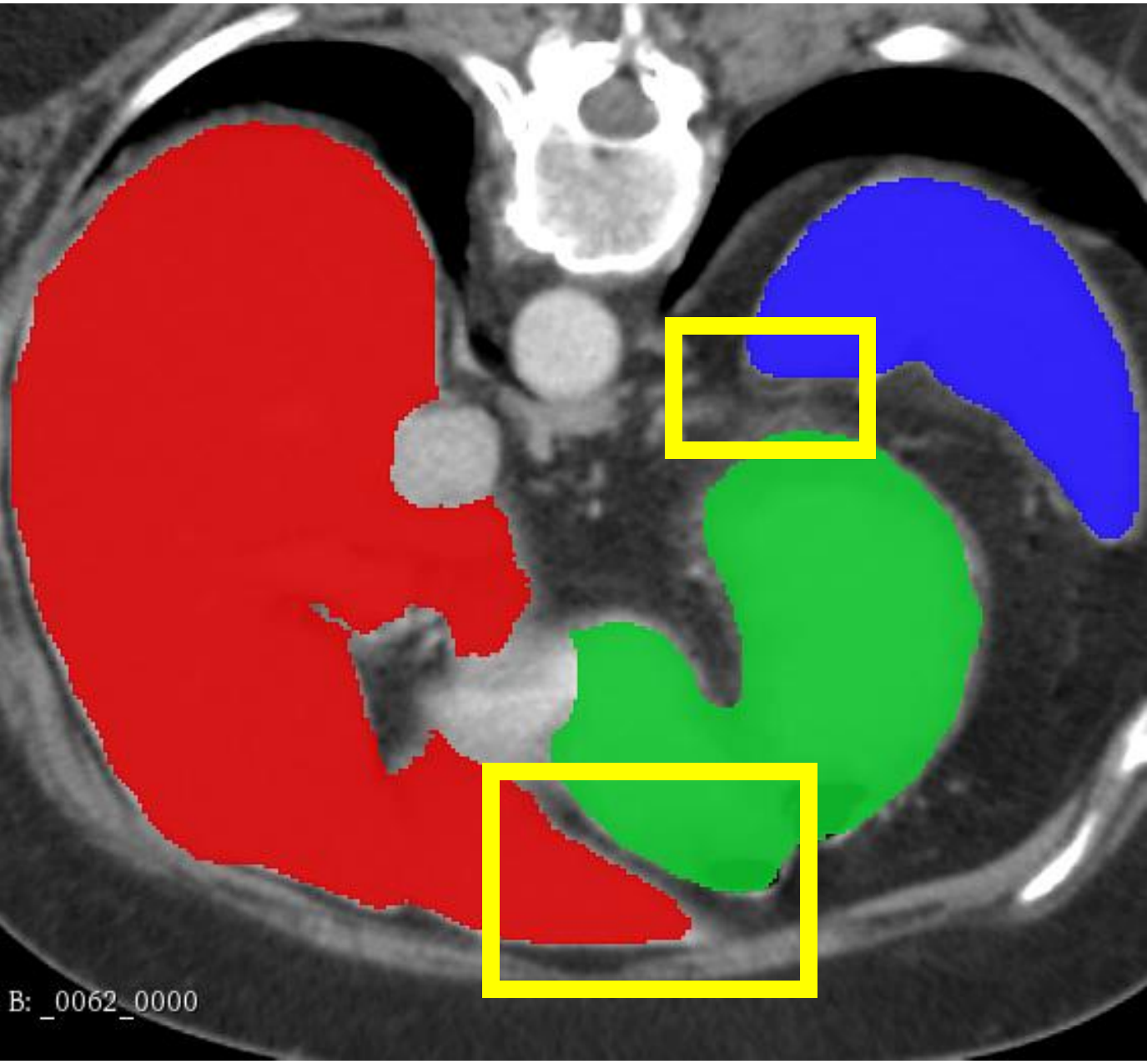}
  \end{subfigure}

        \begin{subfigure}{0.135\linewidth}
  \includegraphics[width=1\textwidth]{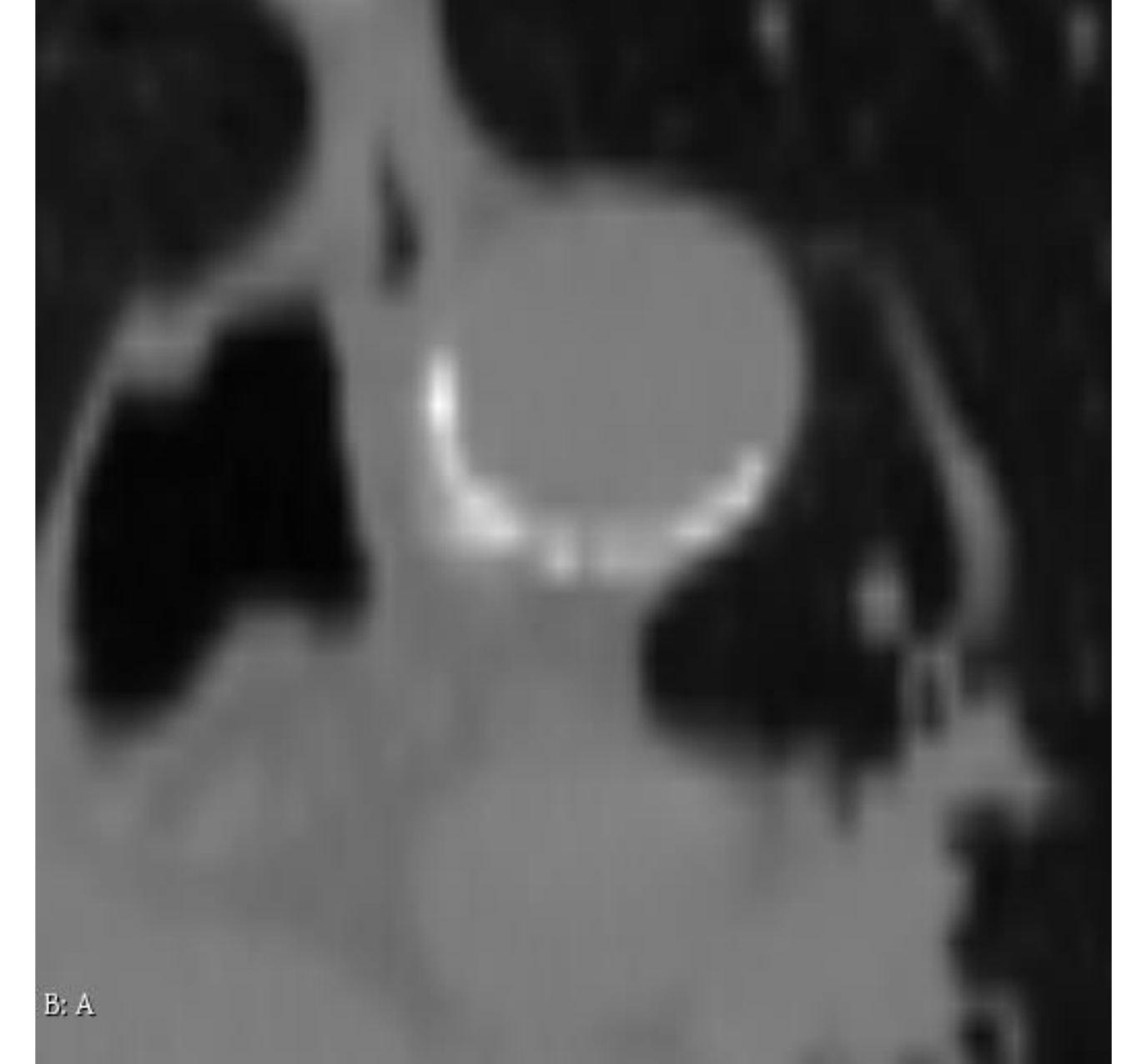}
      \caption{Input}
  \end{subfigure}
    \begin{subfigure}{0.135\linewidth}
     \includegraphics[width=1\textwidth]{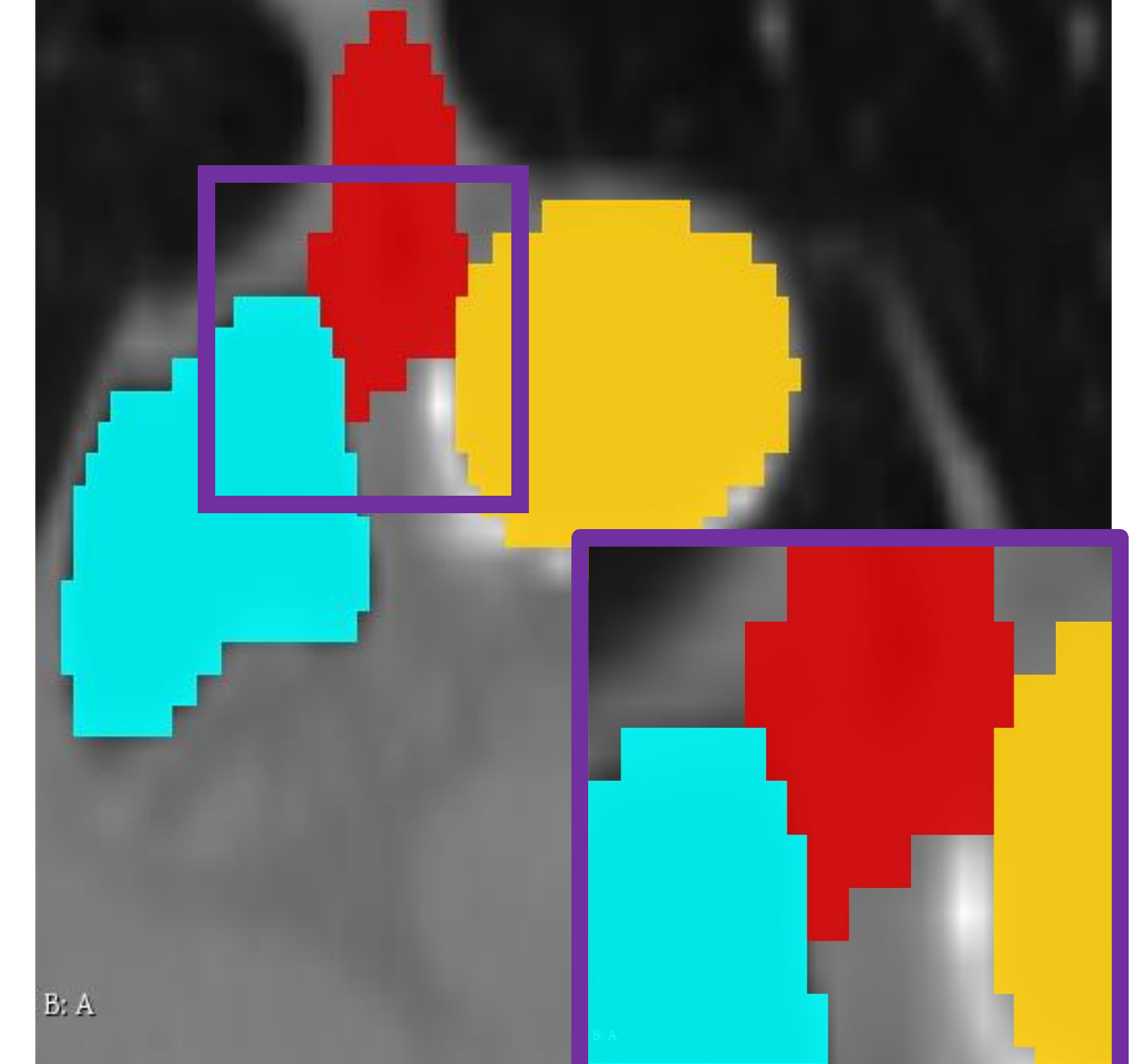}
         \caption{nnUNet}
  \end{subfigure}
      \begin{subfigure}{0.135\linewidth}
     \includegraphics[width=1\textwidth]{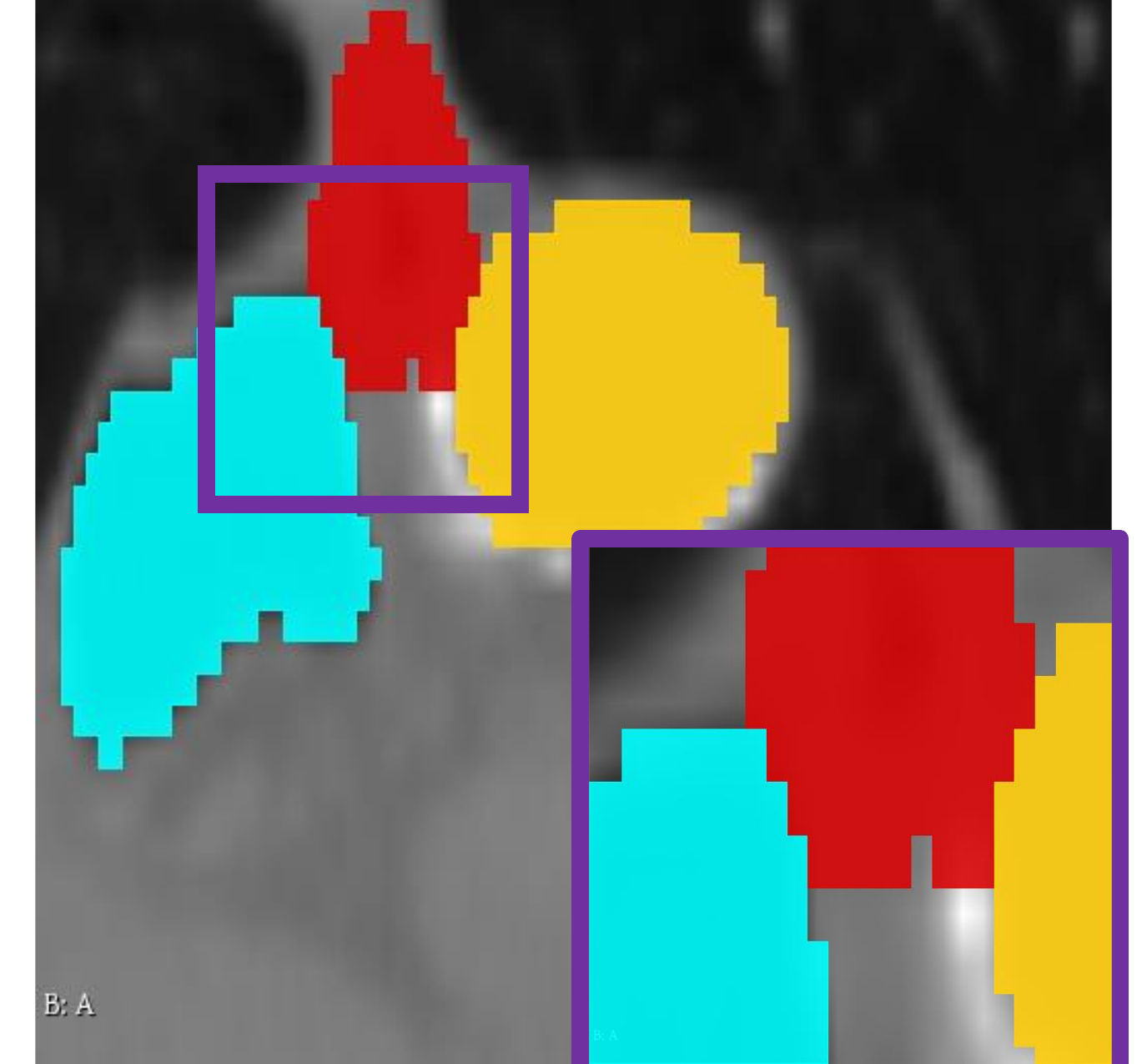}
     \caption{CRF}
  \end{subfigure}
      \begin{subfigure}{0.135\linewidth}
     \includegraphics[width=1\textwidth]{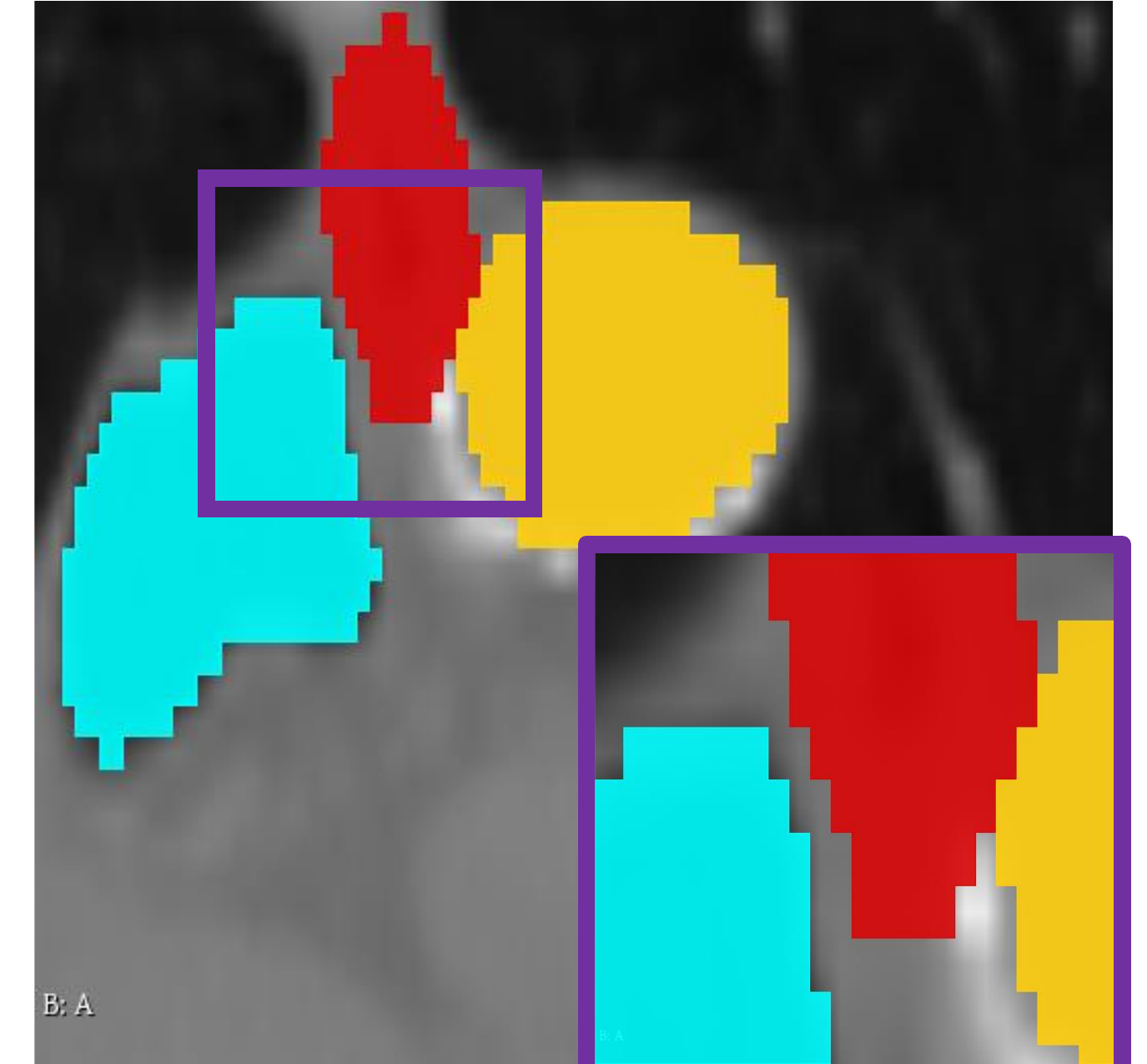}
         \caption{MIDL}
  \end{subfigure}
      \begin{subfigure}{0.135\linewidth}
     \includegraphics[width=1\textwidth]{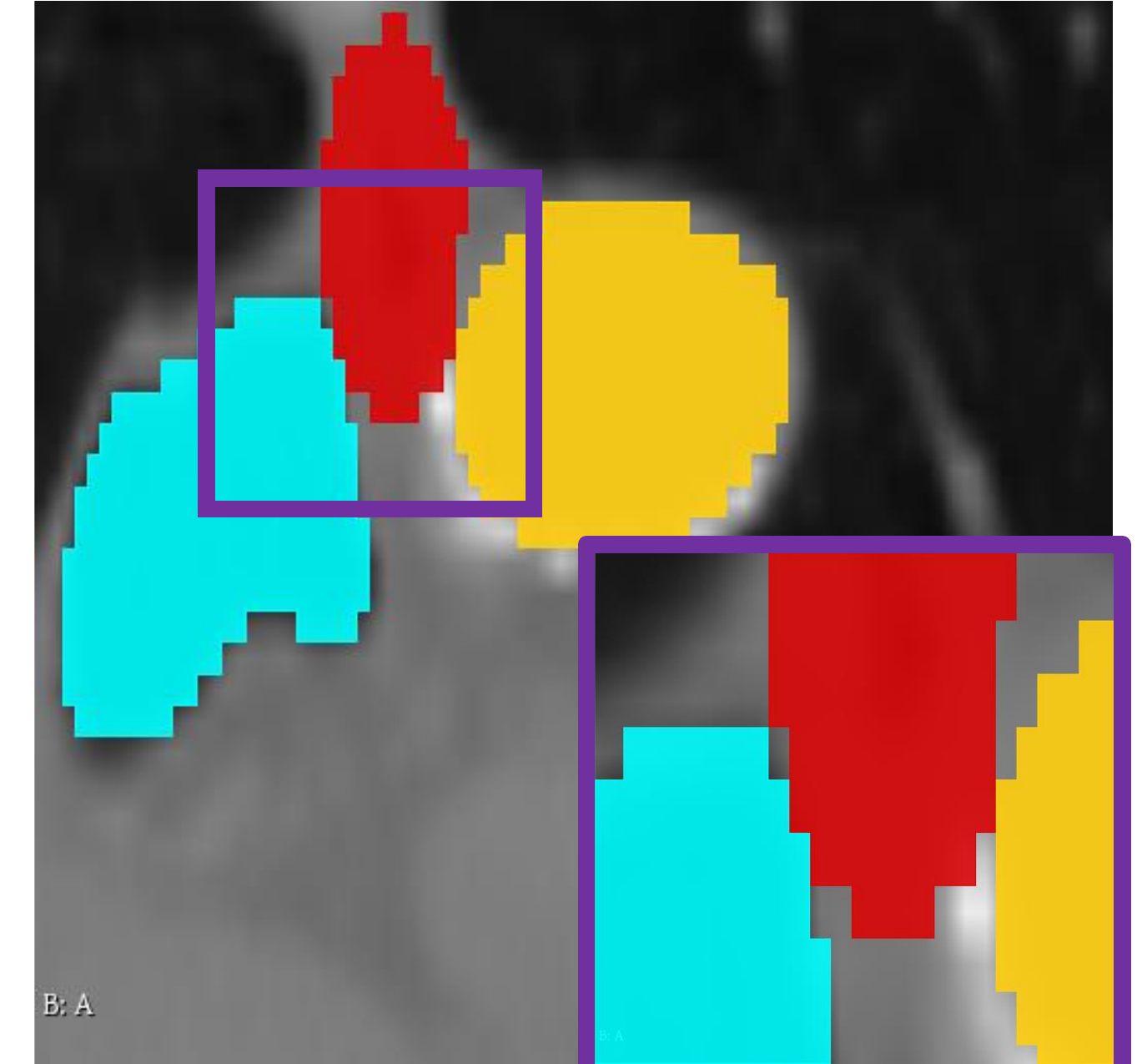}
     \caption{NonAdj}
  \end{subfigure}
      \begin{subfigure}{0.135\linewidth}
     \includegraphics[width=1\textwidth]{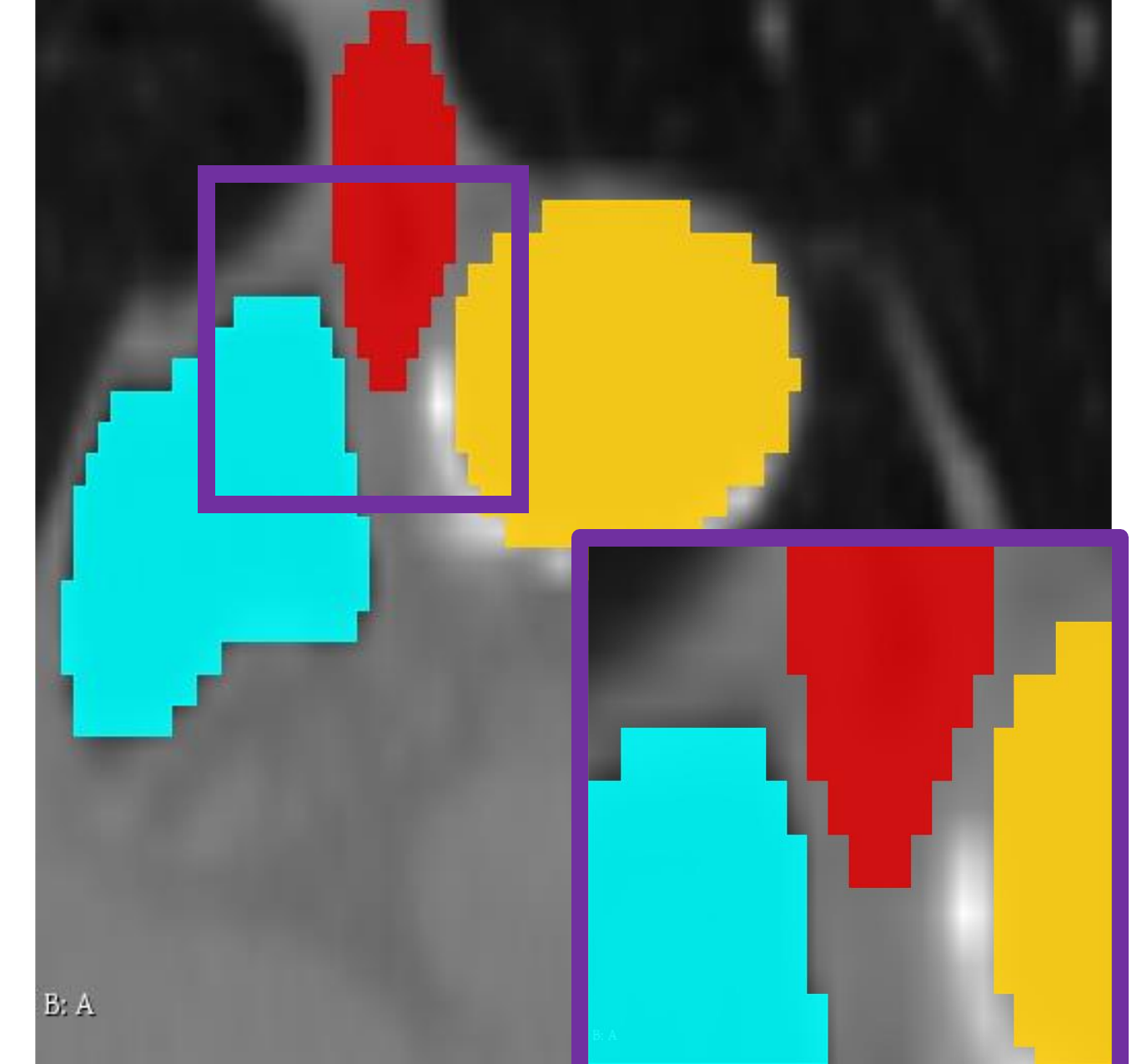}
         \caption{Ours}
  \end{subfigure}
      \begin{subfigure}{0.135\linewidth}
     \includegraphics[width=1\textwidth]{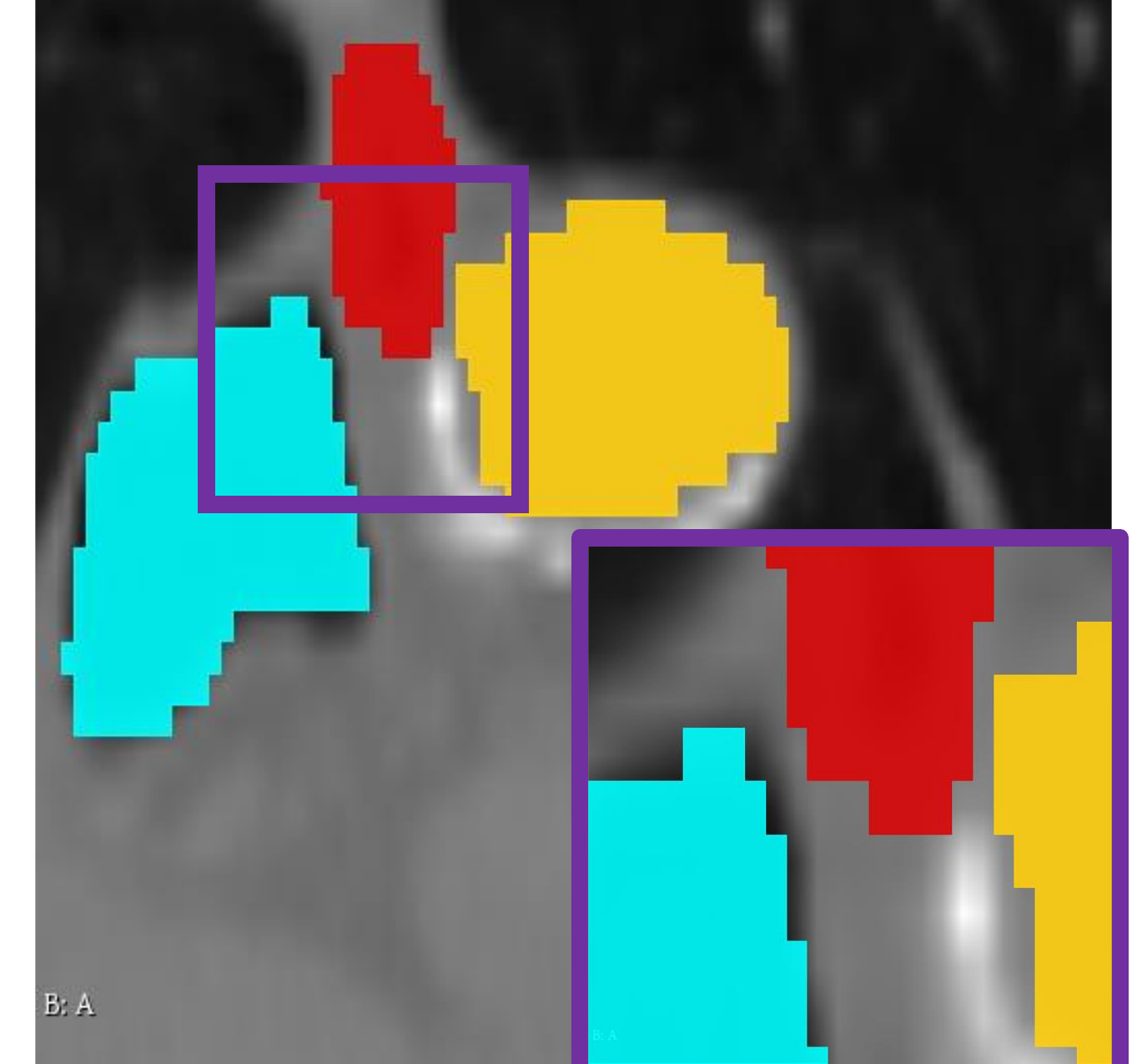}
         \caption{GT}
  \end{subfigure}

\caption{Qualitative results compared with the baselines. Top three rows deal with the containment constraint, while bottom two rows deal with the exclusion constraint. Aorta: rows 1-2, IVUS: row 3, Multi-Atlas: row 4, SegTHOR: row 5. The second row is the 3D view of the first row. It is hard to visualize the input 3D volumetric image and so we leave it blank in the second row. Colors for the classes correspond to the ones used in Fig.~\ref{fig:data-interactions}.}
\label{fig:qualitative_results}
\end{figure}


\myparagraph{Evaluation Metrics}.
Dice score~\cite{zou2004statistical}, Hausdorff distance (HD)~\cite{huttenlocher1993comparing}, and average symmetric surface distance (ASSD)~\cite{heimann2009comparison} are used as the performance metrics. We introduce a new metric called the \textit{\% violations}. The \% violations is calculated by the number of pixels violating the constraint as a fraction of the total number of foreground class pixels/voxels. We report the \% violations for all the pixels/voxels together instead of separately per class. For all metrics, we report the means and standard deviations. We also perform the unpaired t-test~\cite{student1908probable} to determine the statistical significance of the improvement. The statistically significant better performances are highlighted with bold in all the tables. 
The t-test~\cite{student1908probable} used to determine the statistical significance of the improvement has a confidence interval of $95\%$. The best, while not statistically significant, performances are highlighted with italics.


\subsection{Results}
Tab.~\ref{contain} shows the quantitative results for the containment constraint on the Aorta and IVUS datasets, while Tab.~\ref{exclusion} shows the quantitative results for the exclusion constraint on the Multi-Atlas (Abdominal) and SegTHOR datasets. In Fig.~\ref{fig:qualitative_results}, we show the qualitative comparison of different methods. The comprehensive quantitative and qualitative results of our method on the UNet and FCN backbones, along with different connectivity kernels can be found in the Supplementary Material. In general, we observe that learning the topological constraint leads to better feature representation and thus better segmentations both qualitatively and quantitatively. 
We discuss the results for both interactions below.

 \setlength{\tabcolsep}{2pt}
\begin{table*}[t]
  \centering
  \scriptsize
\caption{Quantitative comparison for containment constraint}
\label{contain}
\begin{tabular}{c c c c c c} 
 \hline
 \textbf{Class} & \textbf{Model} & \textbf{Dice}$\uparrow$ & \textbf{HD}$\downarrow$ & \textbf{ASSD}$\downarrow$ & \textbf{\% Violations}$\downarrow$ \\
 \hline \hline
 \multicolumn{6}{c}{\textbf{Aorta dataset}} \\
 \hline
\multirow{7}{*}{\textbf{{\rotatebox[origin=c]{90}{Lumen}}}}  
& UNet~\cite{cciccek20163d} & 0.900 $\pm$ 0.016 & 64.392 $\pm$ 16.874 & 9.315 $\pm$	1.749 & 13.994 $\pm$ 1.809 \\
~& FCN~\cite{FCN8s}  & 0.894 $\pm$ 0.013 & 57.974 $\pm$ 19.756 & 9.77 $\pm$	1.421 & 15.675 $\pm$ 2.409\\
~& nnUNet~\cite{nnUNet} & 0.906 $\pm$ 0.020 & 36.368 $\pm$ 12.559 & 4.563 $\pm$ 0.675 &  5.424 $\pm$  2.461\\
~& Topo-CRF~\cite{bentaieb2016topology} &  0.897 $\pm$ 0.057 &  40.162 $\pm$ 18.687  & 5.952  $\pm$ 0.999 & 8.358 $\pm$ 2.151 \\
~& MIDL~\cite{reddy2019brain} &  0.912 $\pm$ 0.008  & 32.157 $\pm$ 16.270  &  6.405 $\pm$ 0.524  & 6.377 $\pm$ 1.661 \\
~& NonAdj~\cite{ganaye2019removing} &  0.916 $\pm$ 0.030  &  32.465 $\pm$ 18.848  &  4.771 $\pm$ 1.129  & 4.932 $\pm$ 1.479 \\
~& Ours &  \textbf{0.922 $\pm$ 0.009} & \textbf{25.959 $\pm$ 13.574} & \textbf{3.920 $\pm$	0.765} & \textbf{3.526 $\pm$ 1.244}\\
 \hline
\multirow{7}{*}{\textbf{{\rotatebox[origin=c]{90}{Wall}}}}
& UNet~\cite{cciccek20163d} & 0.677 $\pm$ 0.015 & 71.109 $\pm$ 24.653 & 12.497 $\pm$	1.372 & /\\
~& FCN~\cite{FCN8s}  & 0.651 $\pm$ 0.015 & 66.059 $\pm$ 17.188 & 12.339 $\pm$ 0.959 & /\\
~& nnUNet~\cite{nnUNet} & 0.741 $\pm$ 0.026 & 42.486 $\pm$ 15.139 & 8.005 $\pm$	0.811 & /\\
~& Topo-CRF~\cite{bentaieb2016topology} &  0.739 $\pm$ 0.010 &  46.873 $\pm$ 17.636  &  7.914 $\pm$ 0.877 & / \\
~& MIDL~\cite{reddy2019brain} & 0.742 $\pm$ 0.028  & 43.132 $\pm$ 15.624  & 6.420  $\pm$ 1.242  & / \\
~& NonAdj~\cite{ganaye2019removing} & 0.748 $\pm$ 0.017  & 38.197 $\pm$ 19.598  & 4.887 $\pm$ 0.702  & / \\
~& Ours &  \textbf{0.758 $\pm$ 0.017} & \textbf{31.137 $\pm$ 17.772} & \textbf{5.799 $\pm$ 0.737 }& /\\
 \hline
  \multicolumn{6}{c}{\textbf{IVUS dataset}} \\
  \hline
\multirow{7}{*}{\textbf{{\rotatebox[origin=c]{90}{Lumen}}}}
 & UNet~\cite{unet2d} & 0.786 $\pm$ 0.144 & 6.643 $\pm$ 1.936 & 30.944 $\pm$ 11.631 & 5.970 $\pm$ 2.141  \\
~ & FCN~\cite{FCN8s}  &  0.824 $\pm$ 0.071 & 5.319 $\pm$ 1.519 & 22.551 $\pm$ 7.882 & 3.766 $\pm$ 1.444\\
~ & nnUNet~\cite{nnUNet} & 0.893 $\pm$ 0.066 & 3.464 $\pm$ 0.917 & 11.152 $\pm$ 3.954 & 2.708 $\pm$ 1.032\\ 
~ & Topo-CRF~\cite{bentaieb2016topology} &  0.887 $\pm$ 0.096  &  4.138 $\pm$ 1.454  &  10.497 $\pm$ 2.487 & 2.371 $\pm$ 0.960 \\
~& MIDL~\cite{reddy2019brain} &  0.891 $\pm$ 0.073  &  4.226 $\pm$ 1.390  &  10.641 $\pm$ 2.322  &  2.394 $\pm$ 0.918    \\
~& NonAdj~\cite{ganaye2019removing} &  0.897 $\pm$ 0.081  & 3.140  $\pm$ 1.154  &  9.628 $\pm$ 3.221  & 2.173 $\pm$ 0.994    \\
~ & Ours &  \textbf{0.949 $\pm$ 0.070} & \textbf{2.046 $\pm$ 1.079} & \textbf{6.057 $\pm$ 2.746} & \textbf{0.157 $\pm$ 0.808}\\
 \hline
\multirow{7}{*}{\textbf{{\rotatebox[origin=c]{90}{Media}}}}
 & UNet~\cite{unet2d} &  0.651 $\pm$ 0.130 & 7.391 $\pm$ 1.072 & 21.984 $\pm$ 6.634 & /\\
~ & FCN~\cite{FCN8s}  &  0.782 $\pm$ 0.144 & 6.806 $\pm$ 1.147 & 13.863 $\pm$ 4.511 & / \\
~ & nnUNet~\cite{nnUNet} & 0.856 $\pm$ 0.090 & 5.646 $\pm$ 1.228 & 6.491 $\pm$ 2.314 & /\\ 
~ & Topo-CRF~\cite{bentaieb2016topology} &  0.843 $\pm$ 0.106  &  5.409 $\pm$ 1.166  &  5.929 $\pm$ 1.785 & / \\
~& MIDL~\cite{reddy2019brain} &   0.841 $\pm$ 0.121  &  5.461 $\pm$ 1.214  &  6.071 $\pm$ 1.837  & / \\
~& NonAdj~\cite{ganaye2019removing} &  0.848$\pm$ 0.117  & 5.983  $\pm$ 1.342  &  6.615 $\pm$ 1.937  & / \\
~ & Ours &  \textbf{0.910 $\pm$ 0.089} & \textbf{3.873 $\pm$ 0.933} & \textbf{3.171 $\pm$ 1.871} & /\\
 \hline
\end{tabular}
\end{table*}

\myparagraph{Quantitative and Qualitative Results for Containment Constraint.}\\
From Tab.~\ref{contain}, we observe that the proposed method improves the quality of segmentations by improving all the metrics significantly. 
In Fig.~\ref{fig:qualitative_results}, we see that the networks trained with the proposed method have considerably fewer topological violations compared to the other baseline networks. In the top two rows of the figure, we see that the proposed method fixes the topological interaction errors by enforcing the lumen always be enclosed by the wall.
By enforcing this constraint, our method is able to reconstruct the broken lumen and wall structures, thereby significantly improving the segmentation quality. In the third row, we show results on the IVUS dataset. Due to artifacts in the input (like shadow), nnUNet erroneously classifies extraneous lumen regions beyond the media. Due to the smoothness loss component in TopoCRF, the boundaries of its segmentations are a lot smoother compared to nnUNet, however, it also fails to correct the lumen prediction. MIDL performs similarly as TopoCRF, and while NonAdj performs better than both of them, it still fails in several places. By enforcing the containment constraint, our method is able to learn better features and gets rid of such extraneous lumen regions. 

For both the Aorta and IVUS datasets, by identifying the critical pixels, our method improves the learning capability of the network through the epochs. 
In Fig.~\ref{fig:violations}, we show how our method improves the network predictions through the epochs on an IVUS data sample.
Our results demonstrate that our proposed method is able to significantly improve the segmentation quality without the need for any additional post-processing.


\setlength{\tabcolsep}{2pt}
\begin{table*}[!t]
  \centering
  \scriptsize
\caption{Quantitative comparison for exclusion constraint}
\label{exclusion}
\begin{tabular}{c c c c c c} 
 \hline
 \textbf{Class} & \textbf{Model} & \textbf{Dice}$\uparrow$ & \textbf{HD}$\downarrow$ & \textbf{ASSD}$\downarrow$ & \textbf{\% Violations}$\downarrow$ \\
 \hline \hline
  \multicolumn{6}{c}{\textbf{Multi-Atlas dataset}} \\
  \hline
\multirow{7}{*}{\textbf{{\rotatebox[origin=c]{90}{Spleen}}}}
& UNet~\cite{cciccek20163d} &  $0.919 \pm 0.041$ & $47.037 \pm 17.365$ & $4.323 \pm 0.367$  &  $1.857 \pm 0.123$  \\
~& FCN~\cite{FCN8s}  & $0.909 \pm 0.037$ & $134.915 \pm 65.623$ & $17.646 \pm 10.604$ &  $3.041 \pm 0.181$  \\
~& nnUNet~\cite{nnUNet} &  $0.950 \pm 0.041$ & $6.084 \pm 1.078$ & $0.573 \pm 0.131$ & $0.819 \pm 0.064$ \\
~& Topo-CRF~\cite{bentaieb2016topology} & 0.947  $\pm$ 0.028  & 6.403  $\pm$ 1.039  & 1.844   $\pm$ 0.517 & 0.934 $\pm$ 0.032 \\
~& MIDL~\cite{reddy2019brain} &  0.944 $\pm$ 0.015  &  5.597 $\pm$ 1.374  & 0.565  $\pm$ 0.124  & 0.725 $\pm$ 0.151 \\
~& NonAdj~\cite{ganaye2019removing} & 0.952 $\pm$ 0.058  & 5.621 $\pm$ 1.065  &  0.513 $\pm$ 0.175  & 0.521 $\pm$ 0.082  \\
~& Ours & \textit{0.960 $\pm$ 0.009} &  \textbf{5.340 $\pm$ 1.049} & \textbf{0.484 $\pm$ 0.109} & \textbf{0.464 $\pm$ 0.043}\\
 \hline

\multirow{7}{*}{\textbf{{\rotatebox[origin=c]{90}{Kidney}}}}
& UNet~\cite{cciccek20163d} & $0.908 \pm 0.079$ & $61.602 \pm 13.168$ & $9.992 \pm 2.461$ &/\\
~& FCN~\cite{FCN8s}  & $0.892 \pm 0.018$ &	$187.472 \pm 36.096$ & $11.583 \pm 2.396$ &/\\
~& nnUNet~\cite{nnUNet} & $0.931 \pm 0.018$ &	$27.252 \pm 5.406$ &	$5.352 \pm 0.199$ &/\\
~& Topo-CRF~\cite{bentaieb2016topology} & 0.928 $\pm$ 0.059  &  30.209 $\pm$ 5.317  & 6.308  $\pm$ 0.905 & / \\
~& MIDL~\cite{reddy2019brain} & 0.935  $\pm$ 0.071  & 25.208  $\pm$ 5.440  &  4.885 $\pm$ 0.421  & / \\
~& NonAdj~\cite{ganaye2019removing} & 0.934 $\pm$ 0.012  & 24.182  $\pm$ 5.561  & 4.692 $\pm$ 0.657  & / \\
~& Ours & \textit{0.936 $\pm$ 0.026} & \textbf{20.013 $\pm$ 2.785} & \textbf{4.298 $\pm$ 0.798} &/\\
 \hline

\multirow{7}{*}{\textbf{{\rotatebox[origin=c]{90}{Liver}}}}
& UNet~\cite{cciccek20163d} & $0.912 \pm 0.016$ & $64.556 \pm 13.894$ & $2.324 \pm 0.513$ &/\\
~& FCN~\cite{FCN8s} & $0.885 \pm 0.034$	& $183.870 \pm 49.796$ & $29.061 \pm 13.484$ &/\\
~& nnUNet~\cite{nnUNet} & $0.951 \pm 0.008$ & $38.931 \pm 12.161$ & $1.922 \pm 0.506$ &/\\
~& Topo-CRF~\cite{bentaieb2016topology} & 0.949  $\pm$ 0.006  & 46.449 $\pm$ 14.188  &  2.072 $\pm$ 0.313 & / \\
~& MIDL~\cite{reddy2019brain} & 0.955  $\pm$ 0.005  & 34.276 $\pm$ 11.253  & 1.344  $\pm$ 0.431  & / \\
~& NonAdj~\cite{ganaye2019removing} & 0.957  $\pm$ 0.003  & 33.671  $\pm$ 13.543  & 1.185  $\pm$ 0.372  & / \\
~& Ours & \textbf{0.962 $\pm$ 0.005} & \textbf{30.341 $\pm$ 9.111} & \textbf{0.985 $\pm$ 0.386} &/\\
 \hline

\multirow{7}{*}{\textbf{{\rotatebox[origin=c]{90}{Stomach}}}}
& UNet~\cite{cciccek20163d} & $0.846 \pm 0.084$ & $76.000 \pm 24.352$ & $5.023 \pm 1.508$ &/\\
~& FCN~\cite{FCN8s}   &  $0.708 \pm 0.156$ & $172.855 \pm 43.735$ & $11.328 \pm 3.178$ &/\\
~& nnUNet~\cite{nnUNet} & $0.895 \pm 0.015$ & $45.767 \pm 7.960$ & $2.720 \pm 0.430$ &/\\
~& Topo-CRF~\cite{bentaieb2016topology} & 0.888 $\pm$ 0.015  & 46.877  $\pm$ 9.861  &  3.675 $\pm$ 0.358 & / \\
~& MIDL~\cite{reddy2019brain} & 0.899  $\pm$ 0.012  & 40.282  $\pm$ 6.437  &  2.567 $\pm$ 0.431  & / \\
~& NonAdj~\cite{ganaye2019removing} & 0.907  $\pm$ 0.028  & 41.749 $\pm$ 8.630  &  2.184 $\pm$ 0.325  & / \\
~& Ours & \textbf{0.910 $\pm$ 0.018} & \textbf{35.514 $\pm$ 10.295} & \textbf{1.644 $\pm$ 0.311} &/\\
 \hline
  \multicolumn{6}{c}{\textbf{SegTHOR dataset}} \\
  \hline
\multirow{7}{*}{\textbf{{\rotatebox[origin=c]{90}{Esophagus}}}}
& UNet~\cite{cciccek20163d} & 0.827 $\pm$ 0.038 & 11.357	$\pm$ 2.709 & 1.186 $\pm$ 0.113 & 3.212 $\pm$	0.720 \\
~ &FCN~\cite{FCN8s}  &  0.800 $\pm$ 0.031  &  10.770	$\pm$ 2.085  &  1.303 $\pm$ 0.128 & 3.616	$\pm$ 0.709  \\
~ &nnUNet~\cite{nnUNet} &  0.841 $\pm$ 0.014  &  8.018	$\pm$ 2.085  &  0.950 $\pm$ 0.070 & 1.947 $\pm$ 0.525  \\ 
~& Topo-CRF~\cite{bentaieb2016topology} &  0.839 $\pm$ 0.029  & 8.602 $\pm$ 2.363  &  0.991 $\pm$	0.081 & 2.070 $\pm$ 0.687 \\
~& MIDL~\cite{reddy2019brain} & 0.840 $\pm$ 0.020  & 7.266 $\pm$ 2.132  & 0.921 $\pm$ 0.136  & 1.271 $\pm$ 0.912 \\
~& NonAdj~\cite{ganaye2019removing} & 0.843 $\pm$ 0.020  & 6.293 $\pm$ 2.703  & 0.897 $\pm$ 0.078  & 1.215 $\pm$ 0.211 \\
~ & Ours &  \textbf{0.858 $\pm$ 0.019}  &  \textbf{5.582	$\pm$ 2.250}  &  \textbf{0.798 $\pm$ 0.042} & \textbf{0.749 $\pm$ 0.428}  \\
 \hline
 
 \multirow{7}{*}{\textbf{{\rotatebox[origin=c]{90}{Trachea}}}}
& UNet~\cite{cciccek20163d} &  0.897 $\pm$ 0.027  &  10.656 $\pm$	4.047  &  0.728 $\pm$ 	0.146 & /  \\
~ &FCN~\cite{FCN8s}  &  0.891 $\pm$ 	0.031  &  11.789 $\pm$ 	5.291  &  0.953 $\pm$ 0.221 & /  \\
~ &nnUNet~\cite{nnUNet} &  0.910 $\pm$  0.018  &  9.423 $\pm$ 	2.393  &  0.478 $\pm$ 0.152 & /  \\ 
~& Topo-CRF~\cite{bentaieb2016topology} &  0.909 $\pm$ 0.022  &  10.435 $\pm$ 2.334  &  0.473 $\pm$ 0.167 & / \\
~& MIDL~\cite{reddy2019brain} & 0.914 $\pm$ 0.027  & 7.929 $\pm$ 2.305  & 0.456 $\pm$ 0.143  & / \\
~& NonAdj~\cite{ganaye2019removing} & 0.913 $\pm$ 0.028  & 7.866 $\pm$ 2.343  & 0.440 $\pm$ 0.113  & / \\
~ & Ours & \textbf{ 0.929 $\pm$  0.020 }&  \textbf{7.280 $\pm$ 	2.109}  & \textbf{ 0.316 $\pm$ 0.186} & /  \\
 \hline

\end{tabular}
\end{table*}

\myparagraph{Quantitative and Qualitative Results for Exclusion Constraint.}
For the Multi-Atlas dataset, our method brings in the greatest improvement for the stomach and liver classes. As can be seen in fourth row of Fig.~\ref{fig:qualitative_results}, it is correctly able to separate these two classes while the other methods fail to do so. This correlates with the quantitative metrics as well. In the case of the spleen and kidney classes, nnUNet itself predicts separation between these two classes. Our method improves the dice score slightly, but significantly improves other metrics like HD and ASSD.
For the SegTHOR dataset, our method brings in the greatest improvement for the esophagus and trachea classes which tend to come in contact at several points across their lengths. In the final row of Fig.~\ref{fig:qualitative_results}, we show that our proposed method is able to impose the exclusion constraint among the three classes. 
For the aorta class, nnUNet is largely able to separate it from the other classes, and so our method's performance on this class is comparable to nnUNet. We include results for the aorta and heart classes in the Supplementary Material.

\setlength{\tabcolsep}{4pt}
\begin{table*}[t]
  \centering
    \scriptsize
\caption{Ablation study for $L_{pixel}$ and $\lambda_{ti}$ (IVUS)}
\label{table_ablation}
\begin{tabular}{c c c c c c} 
 \hline
 \textbf{Class} & \textbf{$L_{pixel}$} & \textbf{Dice}$\uparrow$  & \textbf{HD}$\downarrow$ & \textbf{ASSD}$\downarrow$ & \textbf{\% Violations}$\downarrow$\\ 
 \hline \hline
 \multirow{4}{*}{\textbf{{\rotatebox[origin=c]{90}{\tiny{Lumen}}}}}
 & None & 0.893 $\pm$ 0.066 & 3.464 $\pm$ 0.917 & 11.152 $\pm$ 3.954 & 2.708 $\pm$ 1.032\\
 & MSE &  0.915 $\pm$ 0.073	& 3.162 $\pm$ 0.937	& 9.963 $\pm$ 3.086	& 0.835 $\pm$ 0.907   \\
 & DICE  &  0.937 $\pm$ 0.067 &	2.385 $\pm$ 1.065 &	6.520 $\pm$ 2.845 &	0.320 $\pm$ 0.811  \\
 & CE  &  \textbf{0.949 $\pm$ 0.070} & \textbf{2.046 $\pm$ 1.079} & \textbf{6.057 $\pm$ 2.746} & \textbf{0.157 $\pm$ 0.808}\\
 \hline
\multirow{4}{*}{\textbf{{\rotatebox[origin=c]{90}{\tiny{Media}}}}}
 & None & 0.856 $\pm$ 0.090 & 5.646 $\pm$ 1.228 & 6.491 $\pm$ 2.314 & /\\ 
 & MSE &  0.893 $\pm$ 0.087 & 4.042 $\pm$ 0.986 & 3.874 $\pm$ 1.912  &  /  \\
 & DICE  &  0.896 $\pm$ 0.088 & 3.964 $\pm$ 1.112 & 3.445 $\pm$ 1.681  &  /  \\
 & CE &  \textbf{0.910 $\pm$ 0.089} & \textbf{3.873 $\pm$ 0.933} & \textbf{3.171 $\pm$ 1.871} & /\\
 \hline
 \hline
 \textbf{Class} & \textbf{$\lambda_{ti}$} & \textbf{Dice}$\uparrow$  & \textbf{HD}$\downarrow$ & \textbf{ASSD}$\downarrow$ & \textbf{\% Violations}$\downarrow$\\ 
 \hline \hline
 \multirow{5}{*}{\textbf{{\rotatebox[origin=c]{90}{\tiny{Lumen}}}}}
  & 0 & 0.893 $\pm$ 0.066 & 3.464 $\pm$ 0.917 & 11.152 $\pm$ 3.954 & 2.708 $\pm$ 1.032\\
  &  5.0e-5 &  0.913 $\pm$ 0.071	&3.249 $\pm$ 0.998	&9.338 $\pm$ 3.649	&0.964 $\pm$ 0.893   \\ 
  & 1.0e-4  &  \textbf{0.949 $\pm$ 0.070} & \textbf{2.046 $\pm$ 1.079} & \textbf{6.057 $\pm$ 2.746} & \textbf{0.157 $\pm$ 0.808}\\
  &  1.5e-4 & 0.941 $\pm$ 0.069	&2.124 $\pm$ 1.062	&6.426 $\pm$ 2.976	&0.187 $\pm$ 0.814   \\ 
 & 2.0e-4 &  0.938 $\pm$ 0.070	&2.428 $\pm$ 1.041	&6.558 $\pm$ 2.780&	0.252 $\pm$ 0.830   \\
 
 \hline
 \multirow{5}{*}{\textbf{{\rotatebox[origin=c]{90}{\tiny{Media}}}}}
   & 0 & 0.856 $\pm$ 0.090 & 5.646 $\pm$ 1.228 & 6.491 $\pm$ 2.314 & /\\ 
& 5.0e-5  &  0.877 $\pm$ 0.088	&5.099 $\pm$ 0.997&	5.024 $\pm$ 2.100 &  /   \\
 & 1.0e-4 &  \textbf{0.910 $\pm$ 0.089} & \textbf{3.873 $\pm$ 0.933} & \textbf{3.171 $\pm$ 1.871} & /\\
 &  1.5e-4 &  0.905 $\pm$ 0.088	&3.889 $\pm$ 0.919	&3.257 $\pm$ 1.877  &  /   \\
 & 2.0e-4 &  0.885 $\pm$ 0.089	&4.319 $\pm$ 1.059&	4.364 $\pm$ 1.943  &  /   \\
 \hline
\end{tabular}
\end{table*}

\subsection{Ablation Studies}
To further demonstrate the efficacy of the proposed method, we conduct several ablation studies. The following ablation studies have been performed on the IVUS dataset (containment constraint). We perform identical ablation studies on the Multi-Atlas dataset (exclusion constraint) in the Supplementary Material.

\myparagraph{Ablation Study for Loss Functions.} Our additional topological interaction loss $L_{ti}$ is a general term, and can adopt any existing pixel-wise loss function. We conduct an ablation study using three different loss functions for $L_{pixel}$, the cross-entropy loss (CE), the mean-squared-error loss (MSE), and the dice loss. The results are tabulated in the top half of Tab.~\ref{table_ablation}, where the \textit{None} entry denotes nnUNet trained without $L_{ti}$. Using CE for $L_{pixel}$ gives the best performance. However, using any of the choices for $L_{pixel}$ 
results in improvement across all metrics compared to the vanilla nnUNet.
Thus $L_{ti}$ is a generic term which works towards its intended purpose of correcting topological errors irrespective of the choice of $L_{pixel}$.

\myparagraph{Ablation Study for Loss Weights.}
Since the topological loss is the main contribution of this paper, we conduct another ablation study in terms of its weight $\lambda_{ti}$. 
We run the experiments with different weights for the additional topological interaction loss and report the results in the bottom half of Tab.~\ref{table_ablation}. When $\lambda_{ti}$=$1e$-$4$, the proposed method achieves the best performance. However, a reasonable range of $\lambda_{ti}$ 
always results in improvement. This demonstrates the efficacy and robustness of the proposed method. 
\section{Conclusion}
\label{sec:conclusion}


We introduce a new convolution-based module for multi-class image segmentation that focuses on topological interactions. The module consists of an efficient algorithm to identify critical pixels which induce topological errors. We also introduce an additional topologically constrained loss function.  By incorporating the module as well as the loss function into the training of deep neural networks, we enforce the network to learn better feature representations, resulting in improved segmentation quality. Results suggest that the method is generalizable to both 2D and 3D settings, and across modalities such as US and CT.

\myparagraph{Acknowledgements.} 
We thank the anonymous reviewers for their constructive feedback.
The reported research was partly supported by grants NSF IIS-1909038 and NIH 1R21CA258493-01A1.

\clearpage


\title{Learning Topological Interactions for Multi-Class Medical Image Segmentation \\--- Supplementary Material ---}

\titlerunning{Topological Interactions for Image Segmentation - Supplementary}
\author{Saumya Gupta\inst{\star} \and
Xiaoling Hu\thanks{Equal contribution.}\and
James Kaan \and 
Michael Jin \and
Mutshipay Mpoy \and
Katherine Chung \and
Gagandeep Singh \and
Mary Saltz \and
Tahsin Kurc \and
Joel Saltz \and
Apostolos Tassiopoulos \and
Prateek Prasanna \and
Chao Chen
}

\authorrunning{S. Gupta et al.}
%


\institute{Stony Brook University, Stony Brook, New York, USA\\
\email{\{saumya.gupta, xiaoling.hu, chao.chen.1\}@stonybrook.edu}\\
}

\maketitle

\setcounter{section}{5}
\setcounter{figure}{7}
\setcounter{table}{3}

In the supplementary material, we begin with an alternate implementation of the naive method in Sec.~\ref{section:naive}. In Sec.~\ref{section:kernels}, we provide illustrations of the connectivity kernel in both 2D and 3D settings. In Sec.~\ref{section:dataset}-\ref{section:qual-quant-add}, we provide detailed descriptions of the experiments, namely the datasets, architectures, implementations, additional ablation studies and qualitative and quantitative results.

\section{Alternate Implementation of Naive Solution}
\label{section:naive}
In Sec.~\ref{ssec:methodology_ssec1}, we discussed the naive solution which involved simply looping over all the pixels and scanning all its neighbors. The obvious issue with loops is that though it takes $O(1)$ time to access the neighborhoods of any single pixel, it takes polynomial time to access the neighborhoods of all the pixels together. Here we discuss an alternate implementation of the same idea. Although it is more efficient than the naive method discussed in Sec.~\ref{ssec:methodology_ssec1}, it is still inferior to the convolution-based method in terms of speed and complexity. We provide details of the alternate naive method here in the supplementary due to space constraints in the main paper.

We continue to use the same terminology for terms $A$, $B$, $C$, $P$, $d$, $k$ etc. as used in Sec.~\ref{ssec:methodology_ssec1}. We assume 2D 4-connectivity scenario.

Since the connectivity defined is constant for every pixel, we can translate the idea of looping over every pixel to that of \emph{shifted} maps instead. A map $P_{r}$ is obtained by shifting every pixel in $P$ to the right by one pixel. Similarly we can obtain maps $P_l$, $P_u$ and $P_d$, which are obtained from $P$ by shifting one pixel to the left, up, and down, respectively. Thus for $i$, we have the 4-connectivity neighbors of the pixel $P[i]$, that is, $P_{r}[i]$, $P_{l}[i]$, $P_{u}[i]$, and $P_{d}[i]$. And with the help of these maps, we have access to the neighborhoods of every pixel simultaneously without loops. We can now use algebraic manipulation to determine whether pixel $P[i] \in A$ has a $C$ neighbor(s) or not.

\begin{figure}
\centering
\includegraphics[width=.8\textwidth]{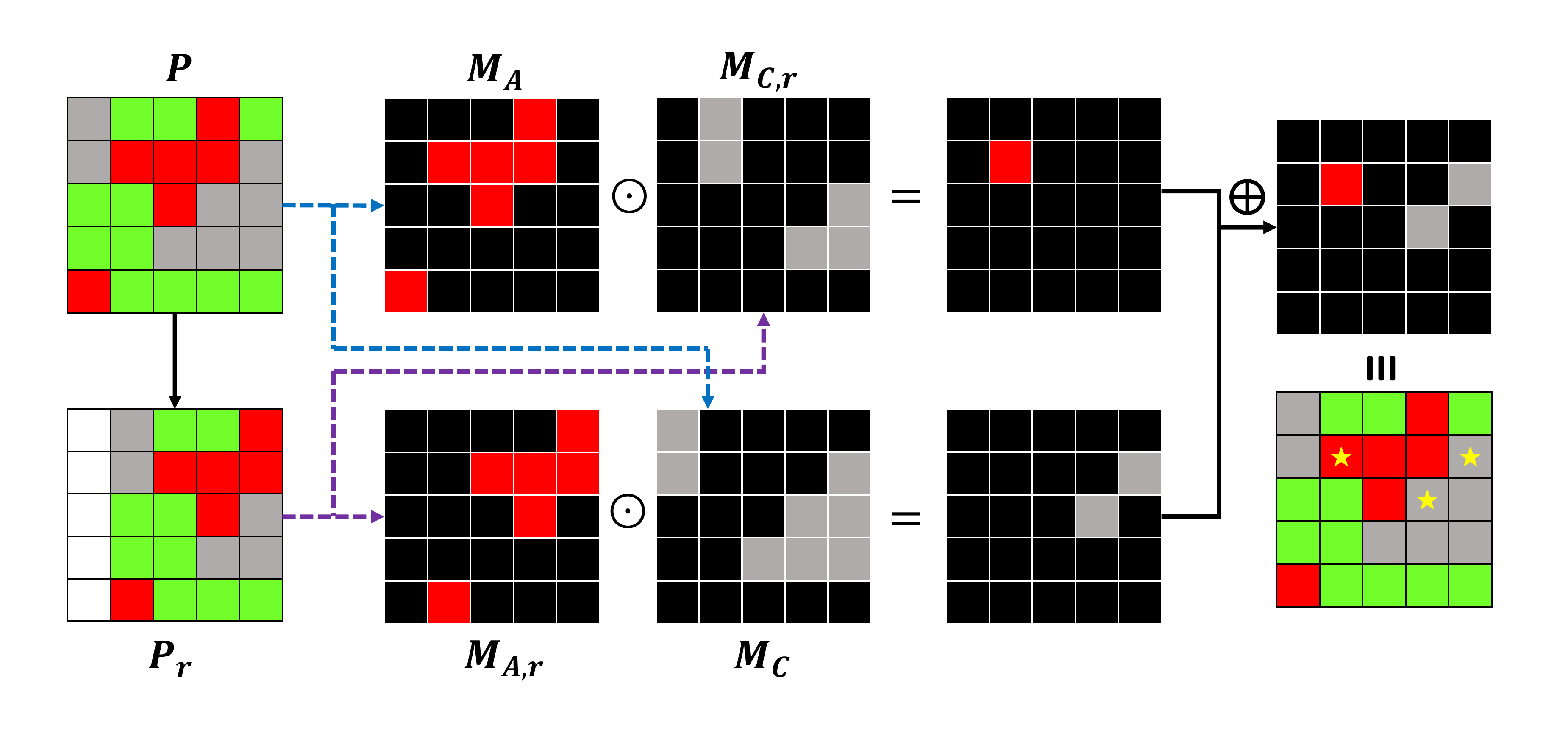}
\caption{2D illustration of the \textbf{alternate naive} algorithm to detect the set of critical pixels. The figure demonstrates the logic to obtain the critical pixels in the right direction using $P_r$. The same logic needs to be extended to $P_{l}$, $P_{u}$, $P_{d}$ to obtain the entire critical pixel set for the 4-connectivity case. Topologically critical pixels are marked with (*).} 
\label{naive_algorithm}
\end{figure}

We need to prune the neighborhood maps so that we are left with only the critical pixel feature map. Let $M_{A}$ be a mask obtained from $P$ such that it contains a $1$ at locations where the pixels are in $A$. Similarly, let $M_{C}$ be a mask obtained from $P$ for $C$. 
We similarly obtain masks $M_{A,w}$ and $M_{C,w}$ from each shifted map $P_{w}$. These masks reduce the context to classes $A$ and $C$ alone while discarding others. Note that here subscript $w$ is used as a generic subscript to denote any of $r$, $l$, $u$, and $d$. 

Now for each neighborhood map $P_{w}$, the term $(M_{A} \odot M_{C,w}) \cup (M_{C} \odot M_{A,w})$ gives the critical pixel map in that direction. Intuitively, it captures all the pixels of $A$ that fall in the neighborhood of $C$ and vice-versa. If we take the union of all these terms constructed from every direction, we obtain all the pixels in $A$ and $C$ which appear in each other's neighborhood. 
Fig.~\ref{naive_algorithm} gives an overview of the algorithm by obtaining the critical pixel map using only $P_{r}$. We can extend the same logic for other $P_{w}$. While intuitive, the disadvantage of this approach is in its scalablility with respect to $d$.

\myparagraph{Computational Efficiency.}
We analyze the computational efficiency by determining complexity as a function of the input and neighborhood size.
Let the image size be $N \times N$ and we enforce a separation of $d$ pixels. In the alternate naive solution, we require $d$ shifted maps along each direction, or $k=2d$ maps along an axis. The time complexity is therefore in the order of $O(N^2k^2)$. The memory requirement will be $O(N^2)$ to store masks $M_{C,w}$ and $M_{A,w}$, and we can optimize this by using an allocated buffer into which we can keep over-writing the masks generated for each direction. As discussed in Sec.~\ref{ssec:methodology_ssec1}, the proposed solution has a time complexity of $O(N^2 \log N)$. Thus, the alternate naive solution is not scalable with respect to $d$ (or $k$), whereas our proposed method has a running time independent of the specified neighborhood size. The memory requirements of both methods are similar. In practice, deep learning frameworks are highly optimized for convolution operations, and so they are much cheaper than computing shifts along axes.

 \myparagraph{Running Times.} For the same network architecture, the inference time remains the same irrespective of the loss functions; the difference is in the training times. We further compare the training times of using the naive method, the alternate naive method and the convolution-based method in the topological interaction module. We report the average time for training one epoch on the IVUS dataset, having a batch size of $5$, input size of $384 \times 384$, and $d=1$. 
For the naive solution, it takes 69.4s to compute the $L_{ti}$ for each epoch. With the alternate naive solution, it takes 5.9s to compute the $L_{ti}$ for each epoch, while it takes only 0.8s for the proposed convolution-based method. The significant difference between the naive and convolution-based methods boils down to the fact that convolutions are highly optimized for GPUs, whereas looping across each pixel in CPU-space incurs huge time. We thus conclude that the convolution-based method is highly efficient compared to both the naive and alternate naive methods, and has negligible timing overhead.

\section{Remark on the Connectivity Kernel $K$}
\label{section:kernels}
\begin{figure}
\centering 

  \begin{subfigure}{\textwidth}
  \centering 
   \includegraphics[width=.65\linewidth]{figures/misc/proposed-conv-2.pdf}
   \caption{2D 4-connectivity}
  \end{subfigure}
  \begin{subfigure}{\textwidth}
  \centering 
     \includegraphics[width=.65\linewidth]{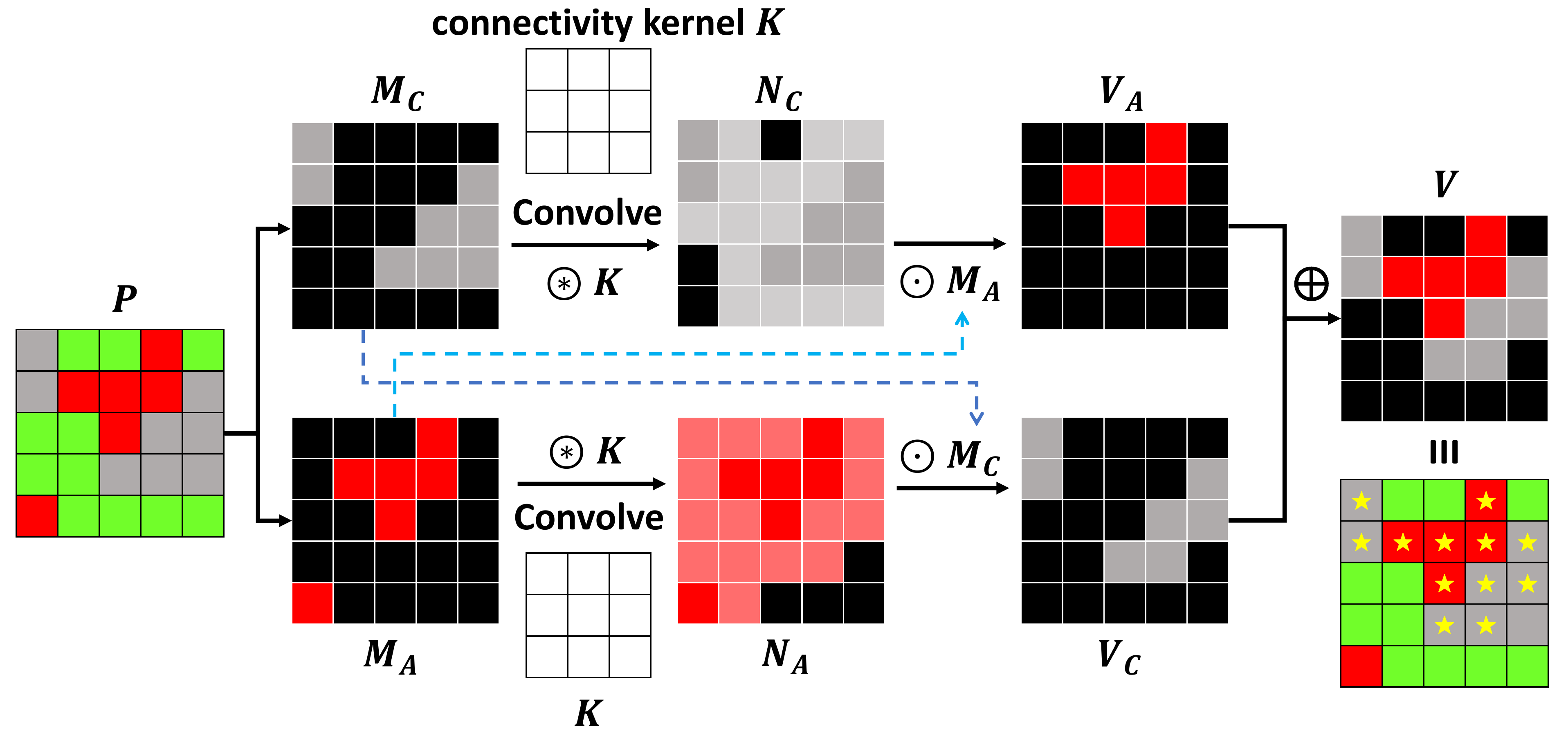}
     \caption{2D 8-connectivity}
  \end{subfigure}
    \begin{subfigure}{\textwidth}
  \centering 
     \includegraphics[width=\linewidth]{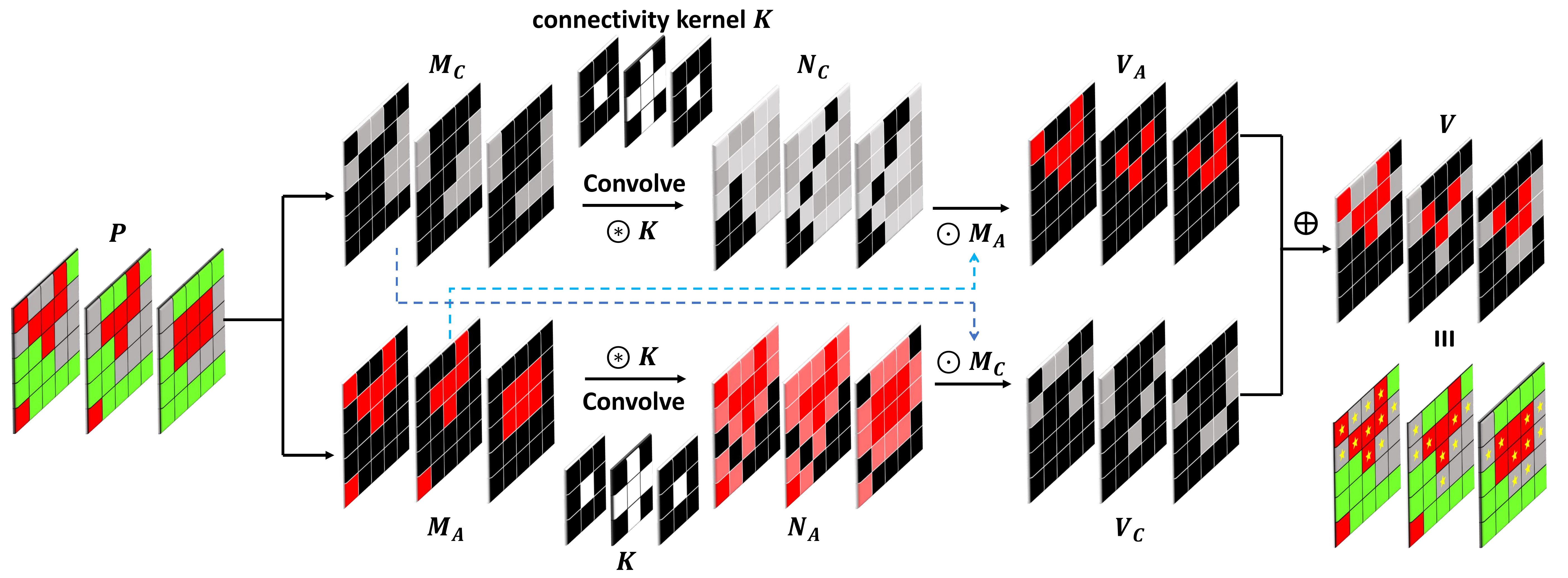}
     \caption{3D 6-connectivity}
  \end{subfigure}
    \begin{subfigure}{\textwidth}
  \centering 
     \includegraphics[width=\linewidth]{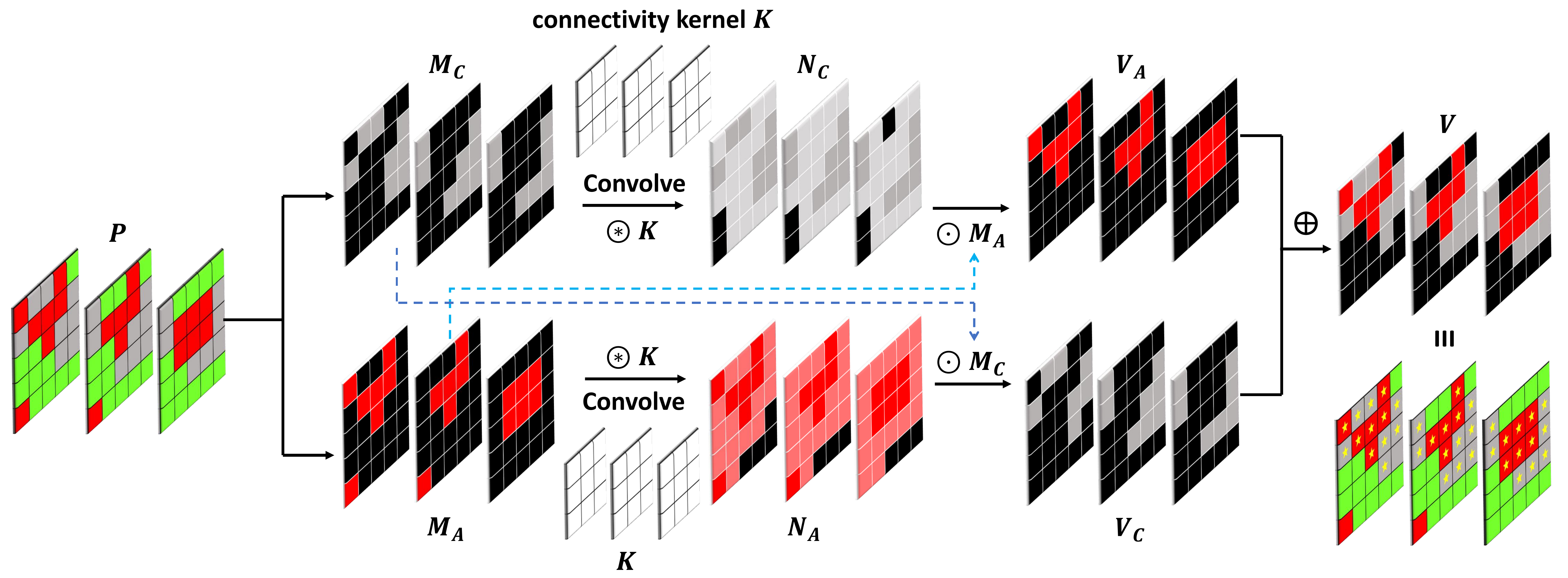}
     \caption{3D 26-connectivity}
  \end{subfigure}
\caption{Illustration of the \textbf{proposed} strategy to detect the set $V$ of topological critical pixels using different connectivity kernels. The entire critical pixel map $V$ is highlighted with $\ast$'s.} 
\label{fig:2d-3d-algo}
\end{figure}

In Fig.~\ref{fig:2d-3d-algo}, we provide illustrations on how (for the same input) the critical pixels map ($V$) changes based on the connectivity kernel ($K$) used. We provide illustrations for the 2D case using 4-connectivity and 8-connectivity kernels, and, for the 3D case using 6-connectivity and 26-connectivity kernels.

\section{Details of the Datasets}
\label{section:dataset}
Fig.~\ref{fig:data-interactions} in the main text gives an overview of the classes in each dataset and the topological interactions among them. 
The datasets are described in more detail as follows.

\myparagraph{Aorta.}
The aorta dataset is a proprietary dataset. 
3D CT scans were obtained from 28 randomly selected patients from an institutional database of patients with thoracic and/or abdominal aortic aneurysm. Inclusion criteria for patients included known aneurysmal disease of the aorta and history of undergoing contrast-enhanced CT with arterial-phase contrast injection. Ground truth annotations of the aortic lumen and wall were obtained by four expert readers, working in consensus. Unlike existing aorta datasets\footnote{https://competitions.codalab.org/competitions/21145}, our dataset contains accurate aortic wall annotations, which have significant clinical implications. The containment constraint holds as the lumen is completely surrounded by the wall.

\myparagraph{IVUS Challenge~\cite{balocco2014standardized}.} The IVUS (IntraVascular UltraSound) challenge is a MICCAI 2011 dataset; we use dataset B in this work. This is a 2D dataset, with each image of dimension $384 \times 384$. It has been created from in-vivo pullbacks of human coronary arteries and contains lumen and media-adventitia labels. There is a predetermined split of 109 training images and 326 test images. The containment constraints holds as the lumen is completely surrounded by the media. The difficulty of this dataset arises due to the imbalanced train-test split, as well as several artifacts (e.g. shadow) in the test set, which causes standard deep neural networks to misclassify the lumen class beyond the media. 

\myparagraph{Multi-Atlas Labeling Beyond the Cranial Vault~\cite{landman2015miccai}.} The MICCAI 2015 challenge `Multi-Atlas Labeling Beyond the Cranial Vault' is a multi-organ segmentation challenge, containing 3D CT scans of the cervix and abdomen. We use the abdomen dataset, which contains thirteen abdominal organ labels. To validate our method, we chose organs that are in close proximity yet exclude each other. We segment four out of the thirteen classes, namely, spleen, left kidney, liver, and stomach. We have clinically verified that the exclusion constraint holds among these four classes, that is, each of these four classes exclude each other. There are 30 volumes available for training, and 20 volumes for testing. The ground truth for the test dataset is available at~\cite{gibsoneli20181169361}. We note that while anatomically, the organs follow the exclusion constraint, the available GT did not adhere to it. With the help of clinicians, we have corrected the GT to follow the exclusion constraint. Thus all the baselines were trained on the corrected GT.

\myparagraph{SegTHOR~\cite{lambert2019segthor}.} The SegTHOR 2019 challenge dataset contains 3D CT scans of thoracic organs at risk (OAR). In this dataset, the OARs are the heart, the trachea, the aorta and the esophagus, which have varying spatial and appearance characteristics. The dataset contains 40 training volumes and 20 testing volumes. 
The exclusion constraint holds among three classes, that is, the trachea, the aorta, and the esophagus do not touch each other. We note that while anatomically, the organs follow the exclusion constraint, the available GT did not adhere to it. With the help of clinicians, we have corrected the GT to follow the exclusion constraint. Thus all the baselines were trained on the corrected GT.


\setlength{\tabcolsep}{2pt}
\begin{table*}[t]
  \centering
    \scriptsize
\caption{Training Configuration.}
\label{table:impl-details}
\begin{tabular}{c | c | c | c | c | c} 
 \hline
 \textbf{Dataset} & \textbf{Model} & \textbf{Patch Size}  & \textbf{Batch Size} & \textbf{LR} & \textbf{Optimizer}\\ 
 \hline \hline
 \multirow{3}{*}{\textbf{{\tiny{Aorta}}}}
  & FCN~\cite{FCN8s} & $512 \times 512$ & $8$ & $0.01$ & SGD \\
  & UNet~\cite{cciccek20163d} & $112 \times 112 \times 80$ & $2$ & $0.01$ & momentum $0.99$\\
  & nnUNet~\cite{nnUNet} & $160 \times 160 \times 80$ & $2$ & $0.01$ & weight decay 3e-5\\
 \hline
 \multirow{3}{*}{\textbf{{\tiny{IVUS}}}}
  & FCN~\cite{FCN8s} & $128 \times 128$  & $8$ & $0.01$ & SGD\\
  & UNet~\cite{unet2d} & $128 \times 128$  & $8$ & $0.01$ &  momentum $0.99$\\
  & nnUNet~\cite{nnUNet} & $384 \times 384$  & $5$ & $0.01$ & weight decay 3e-5\\
\hline
 \multirow{3}{*}{\textbf{{\tiny{Multi-Atlas}}}}
  & FCN~\cite{FCN8s} & $256 \times 256$  & $4$ & $0.01$ & SGD\\
  & UNet~\cite{cciccek20163d} & $64 \times 64 \times 32$ & $4$ & $0.01$ & momentum $0.99$\\
  & nnUNet~\cite{nnUNet} & $192 \times 192 \times 48$ & $2$ & $0.01$ & weight decay 3e-5\\
  \hline
   \multirow{3}{*}{\textbf{{\tiny{SegTHOR}}}}
  & FCN~\cite{FCN8s} & $256 \times 256$  & $4$ & $0.01$ & SGD\\
  & UNet~\cite{cciccek20163d} & $64 \times 64 \times 32$ & $4$ & $0.01$ &  momentum $0.99$\\
  & nnUNet~\cite{nnUNet} & $160 \times 192 \times 64$ & $2$ & $0.01$ & weight decay 3e-5\\
  \hline
\end{tabular}
\end{table*}

\section{Implementation Details}
\label{section:impl}
\begin{figure}[t]
\centering 
    \begin{subfigure}{\textwidth}
  \centering 
     \includegraphics[width=\linewidth]{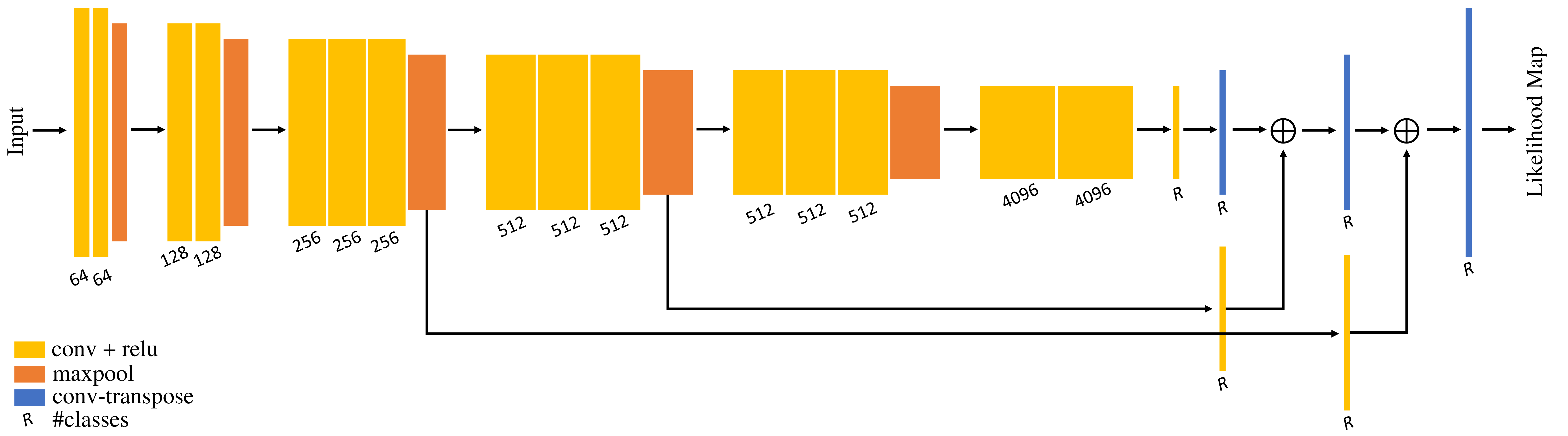}
     \caption{FCN-8s}
  \end{subfigure}
    \begin{subfigure}{\textwidth}
  \centering 
     \includegraphics[width=\linewidth]{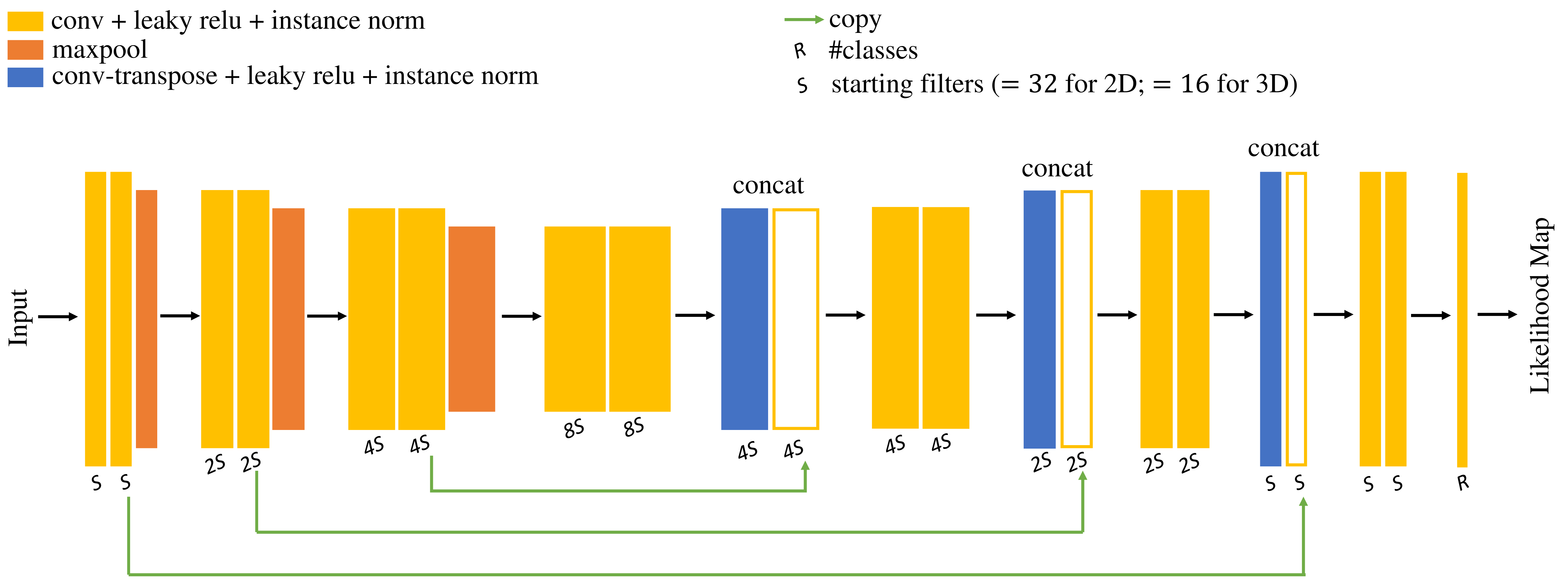}
     \caption{UNet}
  \end{subfigure}
\caption{Baseline network architectures.} 
\label{fig:baseline-archs}
\end{figure}

We use the PyTorch framework, a single NVIDIA Tesla V100-SXM2 GPU (32G Memory) and a Dual Intel Xeon Silver 4216 CPU@2.1Ghz (16 cores) for all the experiments. We use the publicly available codes for UNet \footnote{https://github.com/johschmidt42/PyTorch-2D-3D-UNet-Tutorial}, FCN \footnote{https://github.com/pochih/FCN-pytorch}, nnUNet \footnote{https://github.com/MIC-DKFZ/nnUNet}, and NonAdj \footnote{https://github.com/trypag/NonAdjLoss}. The architecture diagrams for the UNet and FCN networks used are shown in Fig.~\ref{fig:baseline-archs}. The architecture diagram for nnUNet is not shown as nnUNet uses its planning strategy to generate the best architecture for each dataset. 

For the proposed method, the weight term $\lambda_{dice}$ in the loss function is set to 1.0 by default from nnUNet's planning strategy. We obtain the best results with $L_{pixel}$ set to the cross-entropy loss, $\lambda_{ti} = 1e$-$4$ in the 2D setting, and $\lambda_{ti} = 1e$-$6$ in the 3D settings. 

The training hyperparameters for each network on each dataset is as tabulated in Tab.~\ref{table:impl-details}. The loss function used for UNet and FCN is same as that used in vanilla nnUNet, i.e., $L_{ce} + L_{dice}$.

\section{Additional Ablation Studies}
\label{section:abl_add}
In this section we conduct identical ablation studies as Tab~\ref{table_ablation} in the main paper. Here, we conduct this on the Multi-Atlas (exclusion dataset). We report the results in Tab.~\ref{abl1} and Tab.~\ref{abl2}.  The observation is consistent with the ablation studies on IVUS in the main paper. Using cross-entropy as the surrogate loss function for our topological loss gives the best performance. The method is robust to the choice of the loss weight $\lambda_{ti}$. Within a reasonable range,  $\lambda_{ti}$ does impact the performance positively.

 \setlength{\tabcolsep}{2pt}
\begin{table}[t]
  \centering
  \scriptsize
\caption{Ablation study for $L_{pixel}$ (Multi-Atlas)}
\label{abl1}
\begin{tabular}{c c c c c c} 
 \hline
 \textbf{Class} & \textbf{$L_{pixel}$} & \textbf{Dice}$\uparrow$  & \textbf{HD}$\downarrow$ & \textbf{ASSD}$\downarrow$ & \textbf{\% Violations}$\downarrow$\\ 
 \hline \hline
\multirow{4}{*}{\textbf{{\rotatebox[origin=c]{90}{\tiny{Spleen}}}}}
 & None &  0.950 $\pm$ 0.041  &  6.084 $\pm$ 1.078  &  0.573 $\pm$ 0.131  &  0.819 $\pm$ 0.064 \\
 & MSE &  0.952 $\pm$ 0.025  &  5.402 $\pm$ 1.041  &  0.492 $\pm$ 0.118  &  0.552 $\pm$ 0.071  \\
~ & DICE &  0.957 $\pm$ 0.013  &  5.368 $\pm$ 1.042  &  0.488 $\pm$ 0.124  &  0.493 $\pm$ 0.058  \\
~ & CE &  \textit{0.960 $\pm$ 0.009}  &  \textbf{5.340 $\pm$ 1.049}  &  \textbf{0.484 $\pm$ 0.109}  &  \textbf{0.464 $\pm$ 0.043} \\
 \hline
\multirow{4}{*}{\textbf{{\rotatebox[origin=c]{90}{\tiny{Kidney}}}}}
 & None &  0.931 $\pm$ 0.018  &  27.252 $\pm$ 5.406  &  5.352 $\pm$ 0.199  &  /  \\
 & MSE &  0.934 $\pm$ 0.019  &  22.808 $\pm$ 3.186  &  5.089 $\pm$ 0.368  &  /\\
~ & DICE &  0.935 $\pm$ 0.028  & 21.935 $\pm$ 2.772  & 4.610 $\pm$ 0.465  &  /\\
~ & CE & \textit{0.936 $\pm$ 0.026}  &  \textbf{20.013 $\pm$ 2.785} & \textbf{4.298 $\pm$ 0.798}  &  /\\
 \hline
 \multirow{4}{*}{\textbf{{\rotatebox[origin=c]{90}{\tiny{Liver}}}}}
 & None &  0.951 $\pm$ 0.008  & 38.931  $\pm$ 12.161  & 1.922 $\pm$ 0.506  &  /\\
 & MSE & 0.958 $\pm$ 0.009  & 31.672 $\pm$ 10.112  & 1.542 $\pm$ 0.628  &  / \\
~ & DICE & 0.961 $\pm$ 0.009  & 30.941 $\pm$ 9.668  & 1.195 $\pm$ 4.80  &  /\\
~ & CE & \textbf{0.962 $\pm$ 0.005}  & \textbf{30.341 $\pm$ 9.111}  & \textbf{0.985 $\pm$ 0.386}  & /\\
 \hline
 \multirow{4}{*}{\textbf{{\rotatebox[origin=c]{90}{\tiny{Stomach}}}}}
 & None & 0.895 $\pm$ 0.015  &  45.767 $\pm$ 7.960  & 2.720  $\pm$ 0.430  &  / \\
 & MSE & 0.905 $\pm$ 0.014  & 39.608 $\pm$ 9.717  &  2.264 $\pm$ 0.418  &  / \\
~ & DICE & 0.908 $\pm$ 0.016  & 37.763 $\pm$ 9.854  & 1.831 $\pm$ 0.402  &  /\\
~ & CE & \textbf{0.910 $\pm$ 0.018}  &  \textbf{35.514 $\pm$ 10.295}  &  \textbf{1.644 $\pm$ 0.311}  &  / \\
 \hline
\end{tabular}
\end{table}

 \setlength{\tabcolsep}{2pt}
\begin{table}[t]
  \centering
  \scriptsize
\caption{Ablation study for $\lambda_{ti}$ (Multi-Atlas)}
\label{abl2}
\begin{tabular}{c c c c c c} 
 \hline
 \textbf{Class} & \textbf{$\lambda_{ti}$} & \textbf{Dice}$\uparrow$  & \textbf{HD}$\downarrow$ & \textbf{ASSD}$\downarrow$ & \textbf{\% Violations}$\downarrow$\\ 
 \hline \hline
\multirow{4}{*}{\textbf{{\rotatebox[origin=c]{90}{\tiny{Spleen}}}}}
 & 0  &  0.950 $\pm$ 0.041  &  6.084 $\pm$ 1.078  &  0.573 $\pm$ 0.131  &  0.819 $\pm$ 0.064 \\
 & 5.0e-7 &  0.954 $\pm$ 0.029	& 5.399 $\pm$ 1.034 & 0.491 $\pm$ 0.112 & 0.541 $\pm$ 0.049 \\
~ & 1.0e-6 &  \textit{0.960 $\pm$ 0.009}  &  \textbf{5.340 $\pm$ 1.049}  &  \textbf{0.484 $\pm$ 0.109}  &  \textbf{0.464 $\pm$ 0.043} \\
~ & 1.5e-6 &  0.958  $\pm$  0.016 & 5.361  $\pm$  1.025 & 0.487  $\pm$  0.122 & 0.475  $\pm$  0.046 \\
 \hline
\multirow{4}{*}{\textbf{{\rotatebox[origin=c]{90}{\tiny{Kidney}}}}}
 & 0 &  0.931 $\pm$ 0.018  &  27.252 $\pm$ 5.406  &  5.352 $\pm$ 0.199  &  /  \\
 & 5.0e-7 & 0.934 $\pm$ 0.022 & 22.459 $\pm$ 3.625 & 4.936 $\pm$ 0.513 &  / \\
~ & 1.0e-6 & \textit{0.936 $\pm$ 0.026}  &  \textbf{20.013 $\pm$ 2.785} & \textbf{4.298 $\pm$ 0.798}  &  /\\
~ & 1.5e-6 & 0.935 $\pm$ 0.031 & 21.360 $\pm$ 2.909 & 4.380 $\pm$ 0.687 &/  \\
 \hline
 \multirow{4}{*}{\textbf{{\rotatebox[origin=c]{90}{\tiny{Liver}}}}}
 & 0 &  0.951 $\pm$ 0.008  & 38.931  $\pm$ 12.161  & 1.922 $\pm$ 0.506  &  /\\
 & 5.0e-7 & 0.959 $\pm$ 0.010 & 31.390 $\pm$ 10.571 & 1.429 $\pm$ 0.421 &  /  \\
~ & 1.0e-6 & \textbf{0.962 $\pm$ 0.005}  & \textbf{30.341 $\pm$ 9.111}  & 0.985 $\pm$ 0.386  & /\\
~ & 1.5e-6 & 0.961 $\pm$ 0.007 & 30.586 $\pm$ 9.313 & \textbf{0.966 $\pm$ 0.405} &/ \\
 \hline
 \multirow{4}{*}{\textbf{{\rotatebox[origin=c]{90}{\tiny{Stomach}}}}}
 & 0 & 0.895 $\pm$ 0.015  &  45.767 $\pm$ 7.960  & 2.720  $\pm$ 0.430  &  / \\
 & 5.0e-7 & 0.904 $\pm$ 0.013 & 38.984 $\pm$ 9.351 & 2.014 $\pm$ 0.477 &  /  \\
~ & 1.0e-6 & \textbf{0.910 $\pm$ 0.018}  &  \textbf{35.514 $\pm$ 10.295}  &  \textbf{1.644 $\pm$ 0.311}  &  / \\
~ & 1.5e-6 & 0.908 $\pm$ 0.019 & 36.151 $\pm$ 10.192 & 1.721 $\pm$ 0.336 &  /  \\
 \hline
\end{tabular}
\end{table}

\section{Additional Results}
\label{section:qual-quant-add}
In all the tables of the main paper, the statistically significant better performances are highlighted with bold. In the supplementary, we highlight in bold the statistically significant better performances within each backbone class (UNet, FCN, nnUNet). The t-test~\cite{student1908probable} used to determine the statistical significance of the improvement has a confidence interval of $95\%$. The best, while not statistically significant, performances within each backbone class are highlighted with italics.

We provide comprehensive quantitative results for all the datasets in Tab.~\ref{comp-aorta}, ~\ref{comp-ivus}, ~\ref{comp-multi}, and ~\ref{comp-seg}. In the tables, `UNet+Ours' denotes our method trained on the UNet backbone. Similarly, `FCN+Ours' denotes our method trained on the FCN backbone. We observe that the proposed method improves the quality of segmentations by improving all the metrics significantly compared to the backbone. This supports our claim that our method can be incorporated into any backbone.

We also provide results of our method by changing the connectivity kernel. The default connectivity kernel $K$, in 2D, is a $3 \times 3$ kernel filled with $1$'s to enforce $8$-connectivity. Similarly in 3D, $K$ is a $3 \times 3 \times 3$ kernel filled with $1$'s to enforce $26$-connectivity. For the 2D setting, we also provide results on using the $4$-connectivity kernel, which we denote by `Ours (4conn)' in Tab.~\ref{comp-ivus}. For the 3D setting, we also provide results on using the $6$-connectivity kernel, which we denote by `Ours (6conn)' in Tab.~\ref{comp-aorta}, ~\ref{comp-multi}, and ~\ref{comp-seg}. We observe that while using a smaller connectivity kernel does not seem as good as using the default, it is still stronger than other baselines.

\begin{figure}[t]
\centering 

 \begin{subfigure}{0.15\linewidth}
  \includegraphics[width=1\textwidth]{figures/aorta/sample4/input-slice.pdf}
  \caption{Input}
  \end{subfigure}
  \begin{subfigure}{0.15\linewidth}
     \includegraphics[width=1\textwidth]{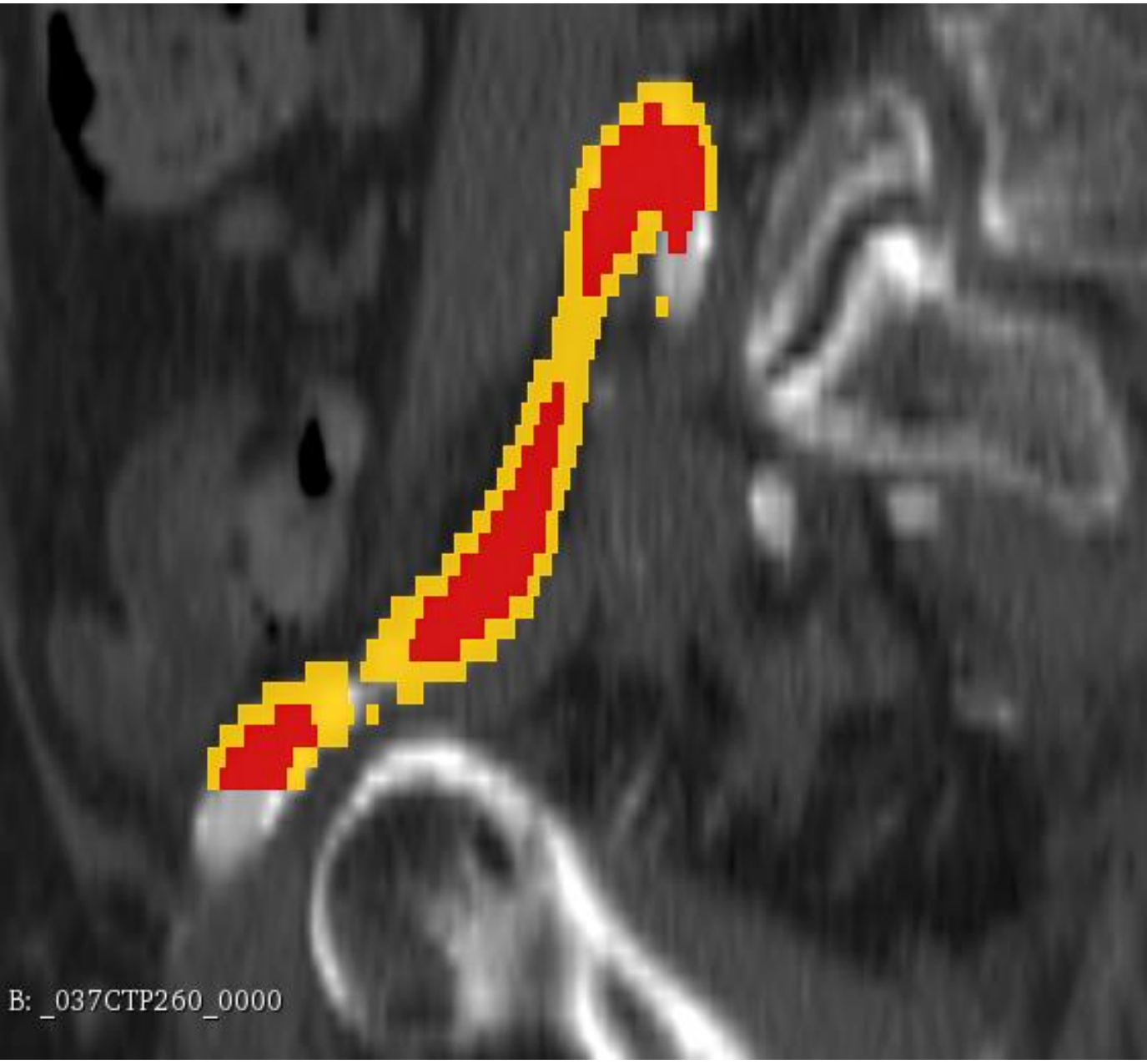}
     \caption{UNet}
  \end{subfigure}
    \begin{subfigure}{0.15\linewidth}
  \includegraphics[width=1\textwidth]{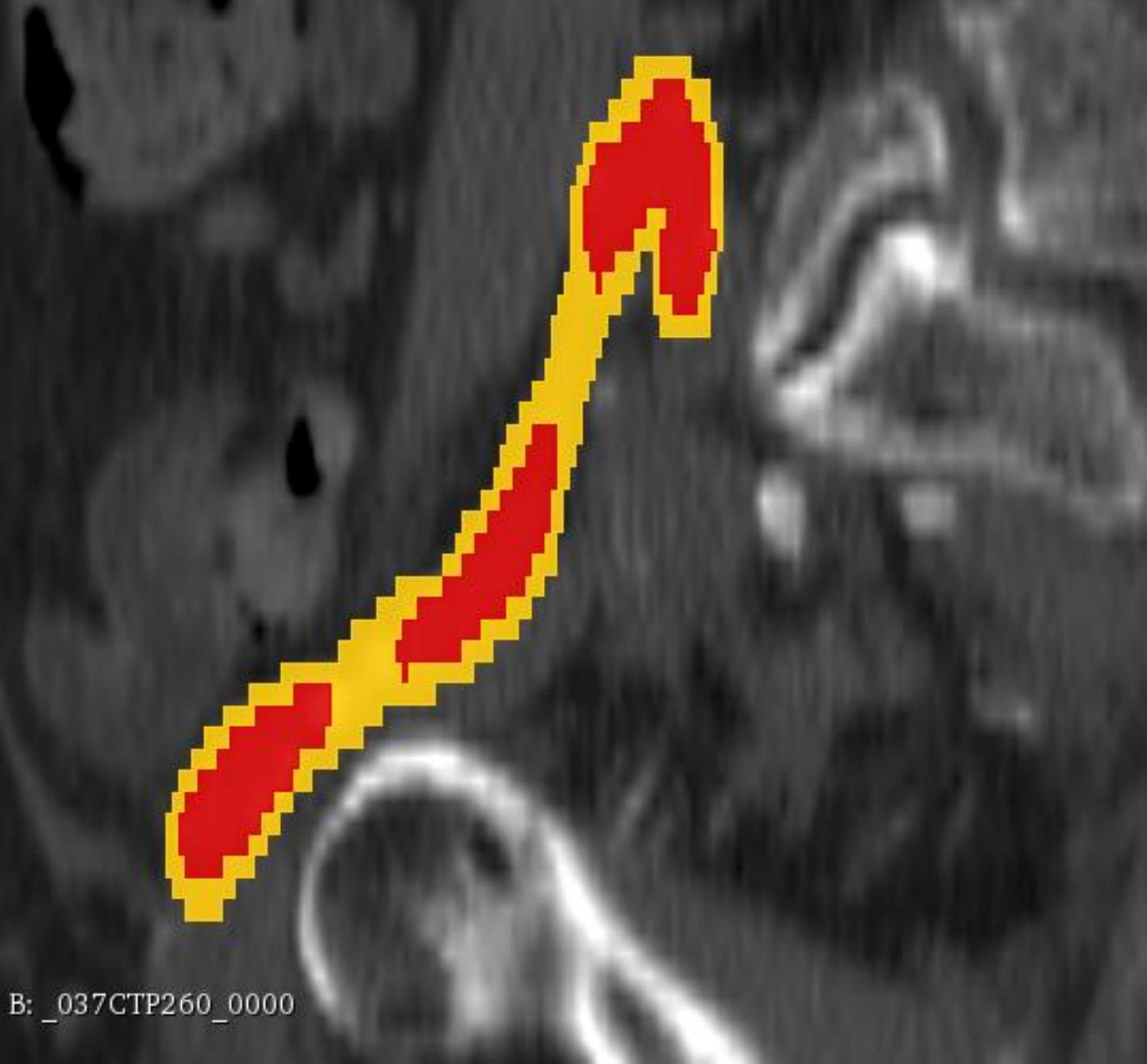}
  \caption{Unet+O}
  \end{subfigure}
    \begin{subfigure}{0.15\linewidth}
     \includegraphics[width=1\textwidth]{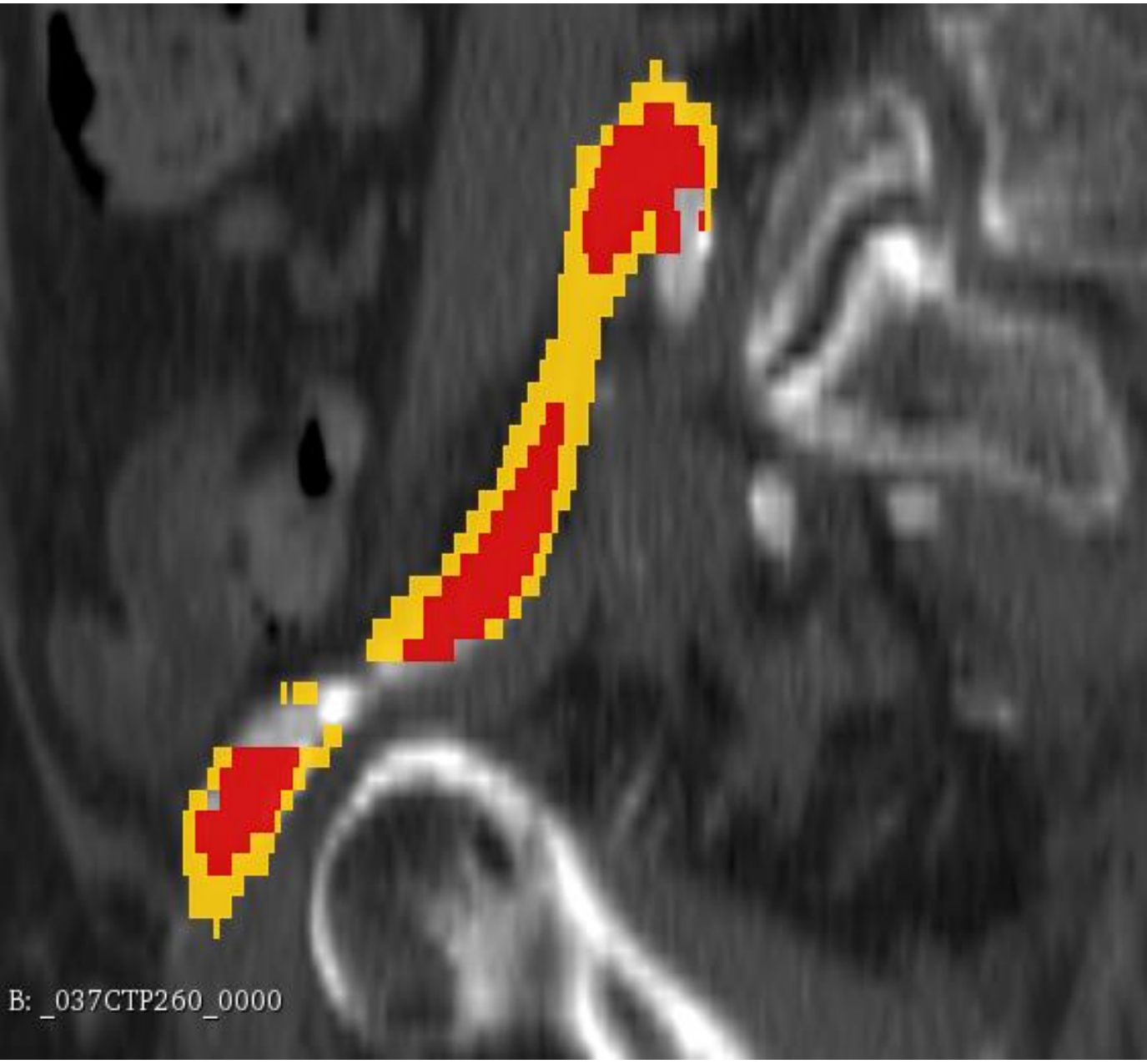}
     \caption{FCN}
  \end{subfigure}
    \begin{subfigure}{0.15\linewidth}
     \includegraphics[width=1\textwidth]{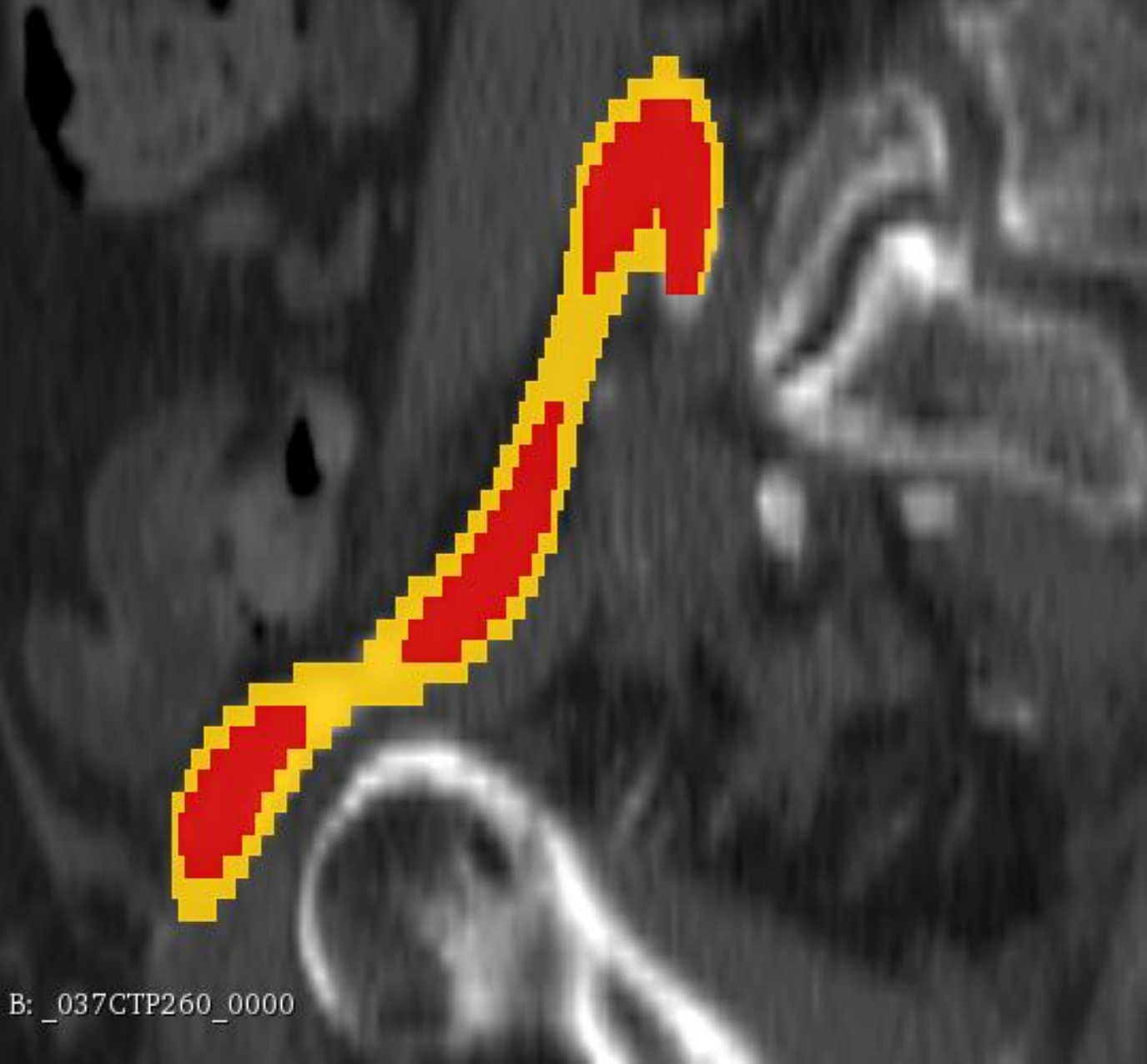}
     \caption{FCN+O}
  \end{subfigure}
    \begin{subfigure}{0.15\linewidth}
     \includegraphics[width=1\textwidth]{figures/aorta/sample4/nnunet-slice.pdf}
     \caption{nnUNet}
  \end{subfigure}

    \begin{subfigure}{0.15\linewidth}
     \includegraphics[width=1\textwidth]{figures/aorta/sample4/crf-slice.pdf}
     \caption{CRF}
  \end{subfigure}
    \begin{subfigure}{0.15\linewidth}
     \includegraphics[width=1\textwidth]{figures/aorta/sample4/midl-slice.pdf}
     \caption{MIDL}
  \end{subfigure}
    \begin{subfigure}{0.15\linewidth}
     \includegraphics[width=1\textwidth]{figures/aorta/sample4/nonadj-slice.pdf}
    \caption{NonAdj}
  \end{subfigure}
    \begin{subfigure}{0.15\linewidth}
     \includegraphics[width=1\textwidth]{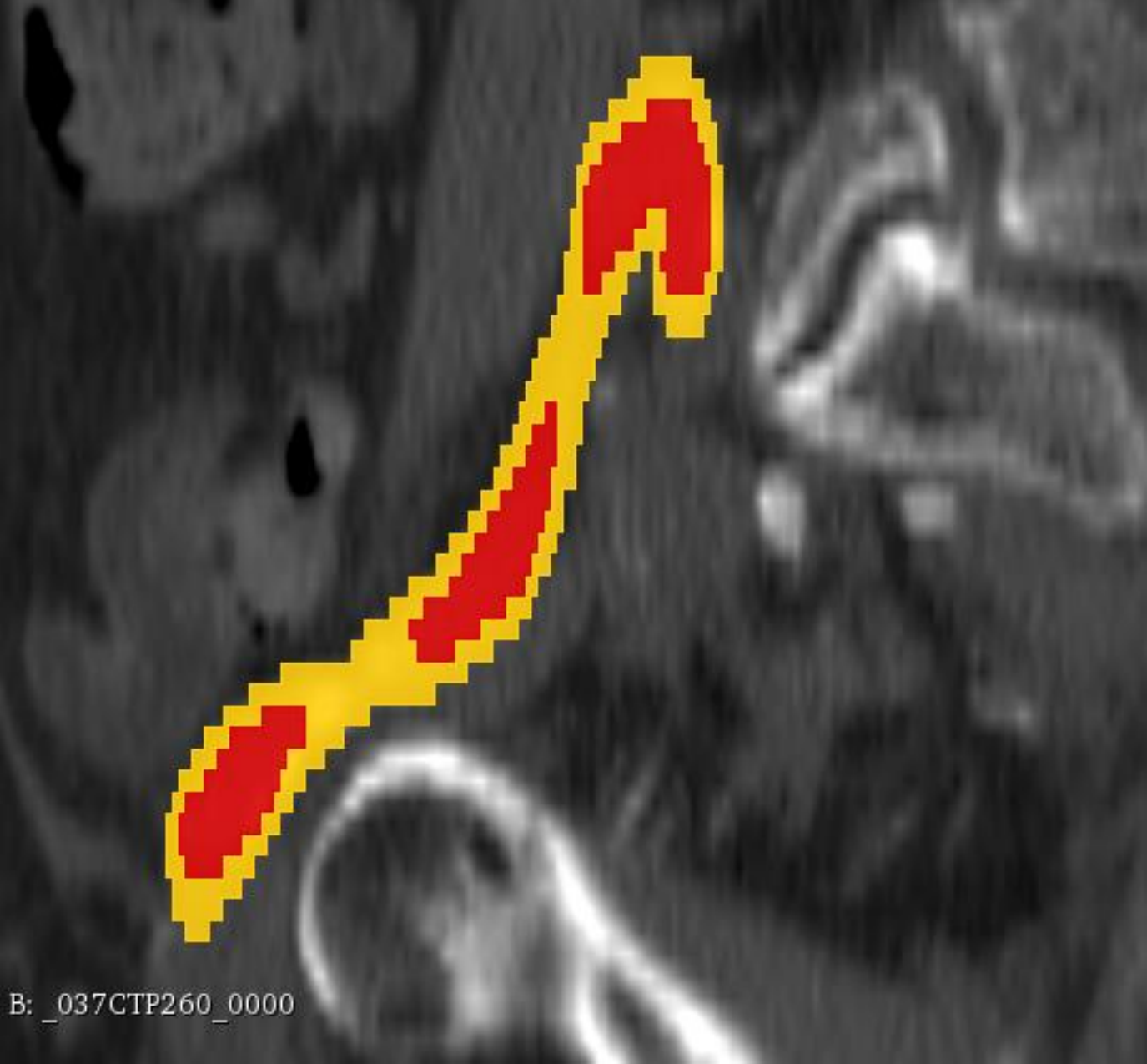}
     \caption{Ours6C}
  \end{subfigure}
      \begin{subfigure}{0.15\linewidth}
     \includegraphics[width=1\textwidth]{figures/aorta/sample4/topo-slice.pdf}
     \caption{Ours}
  \end{subfigure}
      \begin{subfigure}{0.15\linewidth}
     \includegraphics[width=1\textwidth]{figures/aorta/sample4/gt-slice.pdf}
     \caption{GT}
  \end{subfigure}

\begin{subfigure}{0.15\linewidth}
  \includegraphics[width=1\textwidth]{figures/empty2.pdf}
  \end{subfigure}
  \begin{subfigure}{0.15\linewidth}
     \includegraphics[width=1\textwidth]{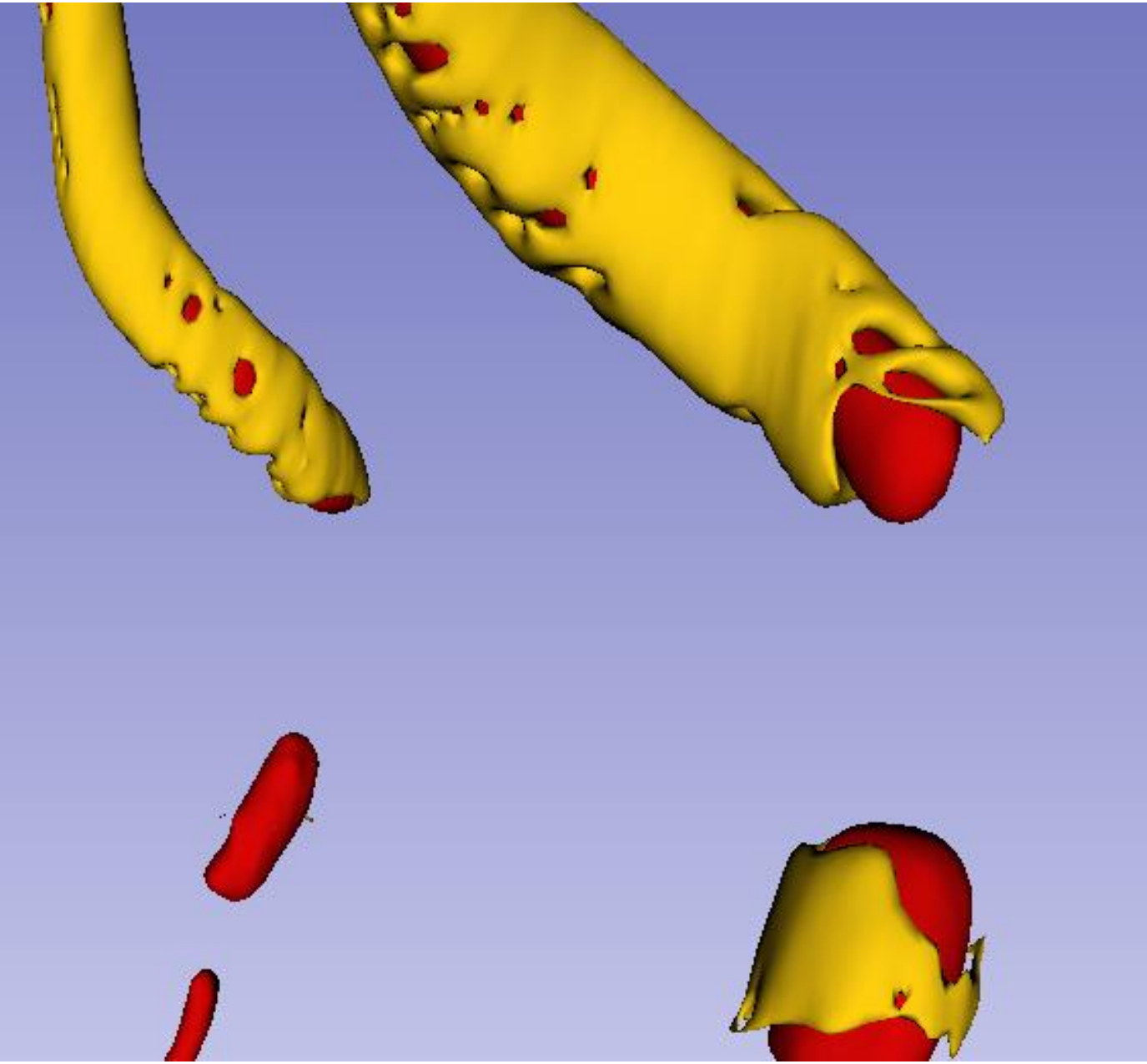}
    \caption{UNet}
  \end{subfigure}
  \begin{subfigure}{0.15\linewidth}
  \includegraphics[width=1\textwidth]{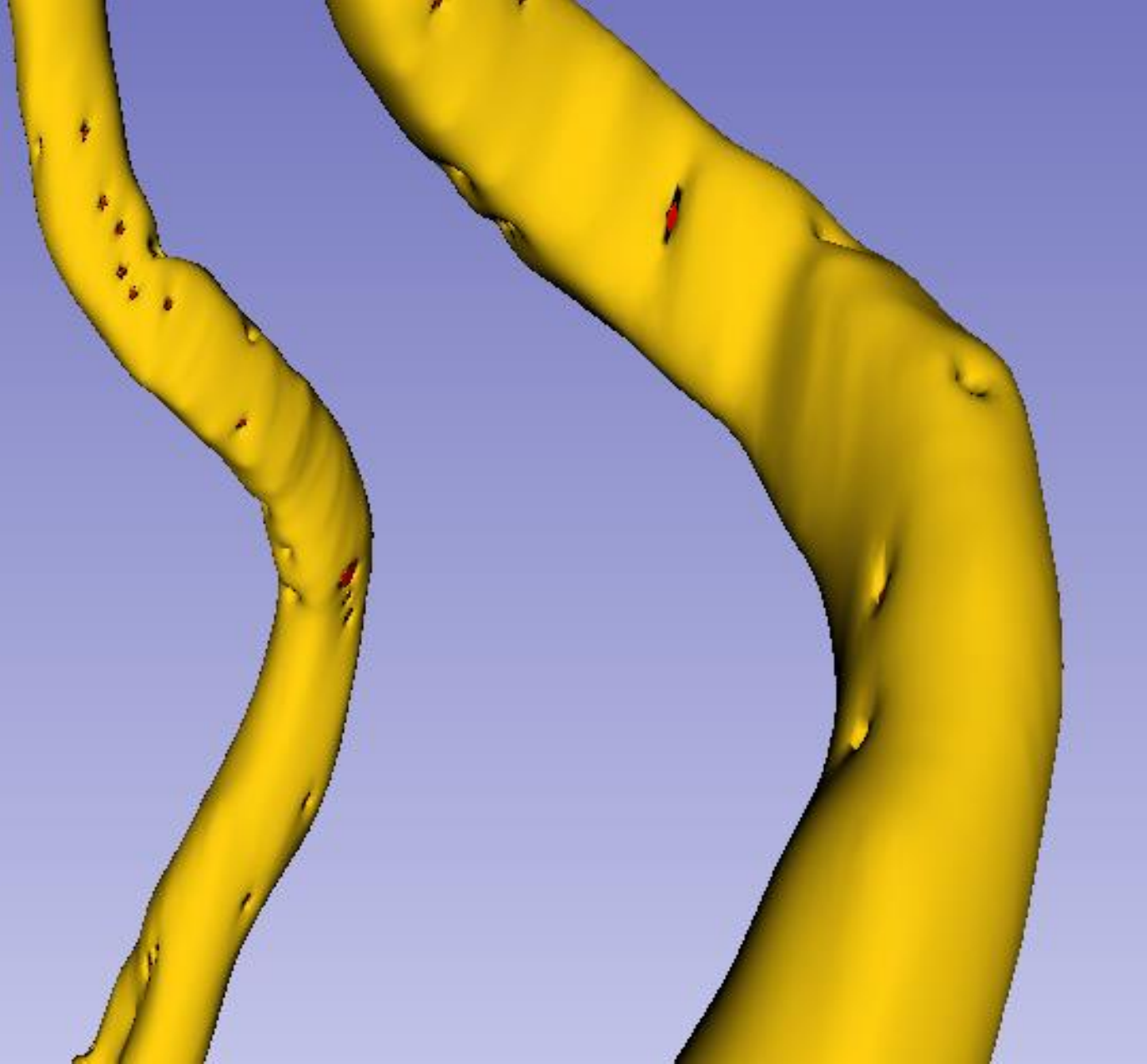}
  \caption{UNet+O}
  \end{subfigure}
    \begin{subfigure}{0.15\linewidth}
     \includegraphics[width=1\textwidth]{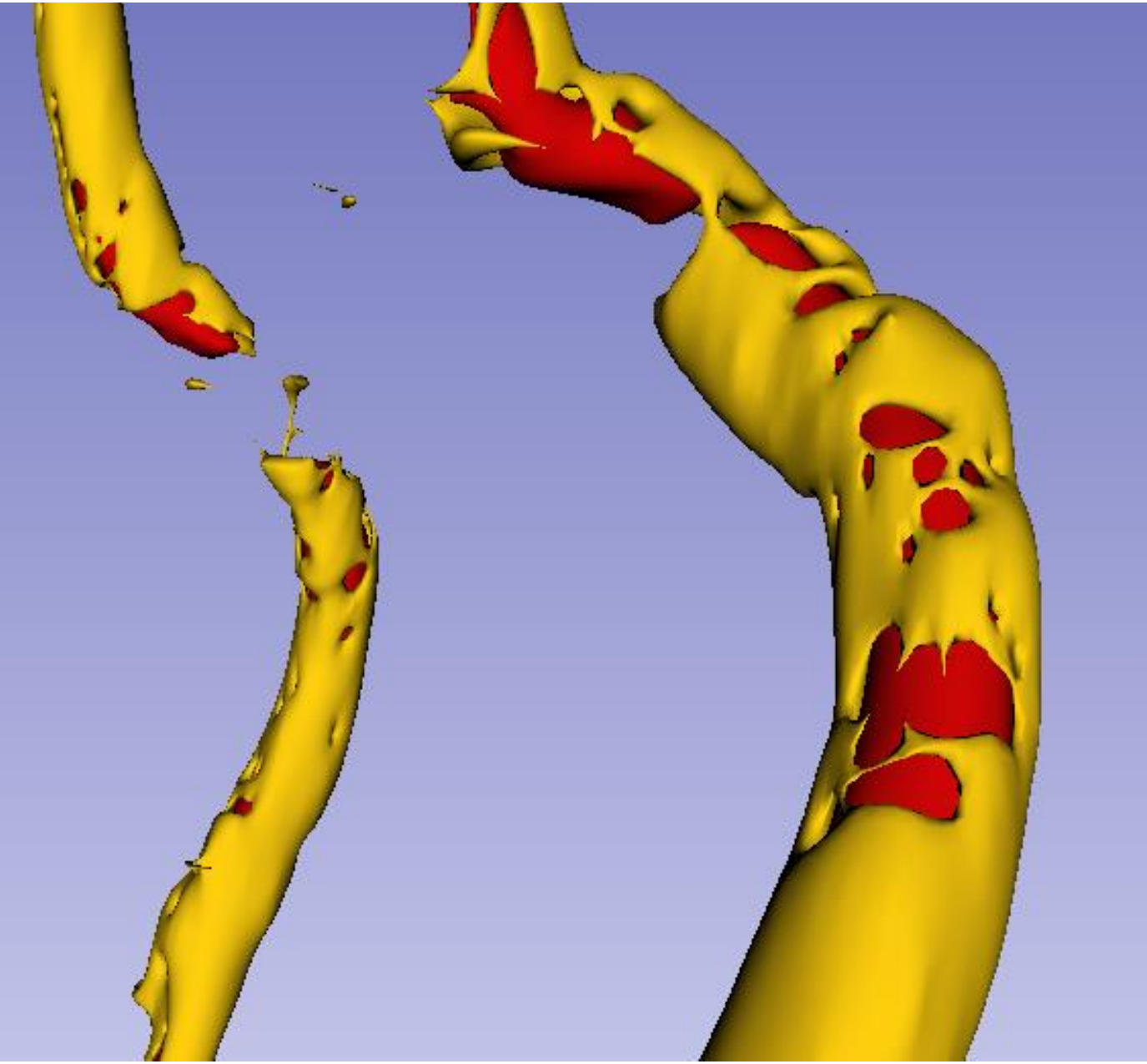}
    \caption{FCN}
  \end{subfigure}
    \begin{subfigure}{0.15\linewidth}
     \includegraphics[width=1\textwidth]{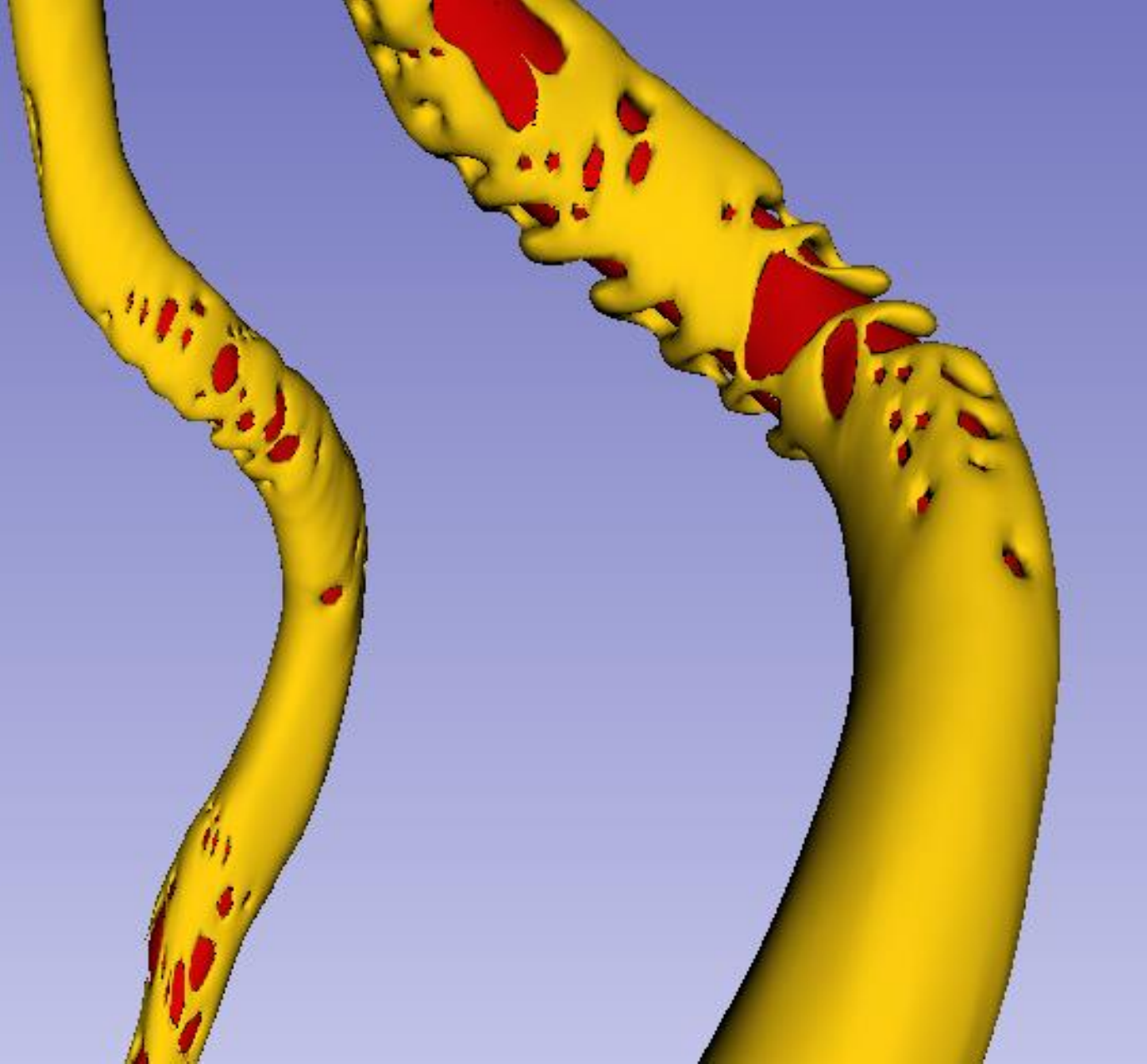}
    \caption{FCN+O}
  \end{subfigure}
    \begin{subfigure}{0.15\linewidth}
     \includegraphics[width=1\textwidth]{figures/aorta/sample4/nnunet-render.pdf}
     \caption{nnUNet}
  \end{subfigure}

  \begin{subfigure}{0.15\linewidth}
     \includegraphics[width=1\textwidth]{figures/aorta/sample4/crf-render.pdf}
     \caption{CRF}
  \end{subfigure}
      \begin{subfigure}{0.15\linewidth}
     \includegraphics[width=1\textwidth]{figures/aorta/sample4/midl-render.pdf}
    \caption{MIDL}
  \end{subfigure}
    \begin{subfigure}{0.15\linewidth}
     \includegraphics[width=1\textwidth]{figures/aorta/sample4/nonadj-render.pdf}
     \caption{NonAdj}
  \end{subfigure}
    \begin{subfigure}{0.15\linewidth}
     \includegraphics[width=1\textwidth]{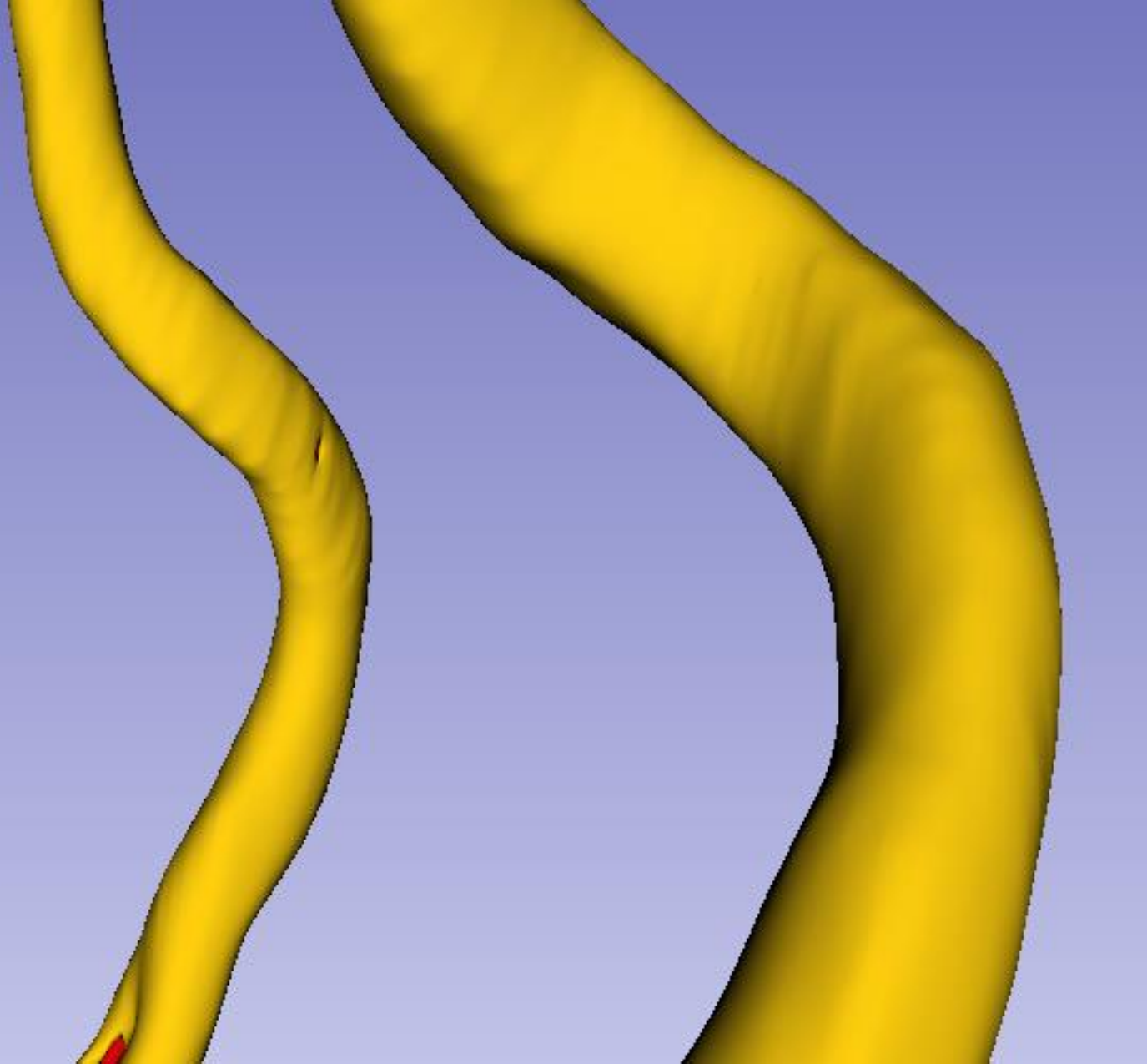}
     \caption{Ours6C}
  \end{subfigure}
      \begin{subfigure}{0.15\linewidth}
     \includegraphics[width=1\textwidth]{figures/aorta/sample4/topo-render.pdf}
     \caption{Ours}
  \end{subfigure}
      \begin{subfigure}{0.15\linewidth}
     \includegraphics[width=1\textwidth]{figures/aorta/sample4/gt-render.pdf}
    \caption{GT}
  \end{subfigure}
\caption{Qualitative Aorta results compared with the baselines. Rows 3-4 are corresponding 3D renderings. It is hard to visualize the input 3D volumetric image and so we leave it blank in the third row. Colors for the classes correspond to the ones used in Fig.~\ref{fig:data-interactions}.}
\label{fig:aorta-add-1}
\end{figure}

\begin{figure}[t]
\centering 

        \begin{subfigure}{0.14\linewidth}
  \includegraphics[width=1\textwidth]{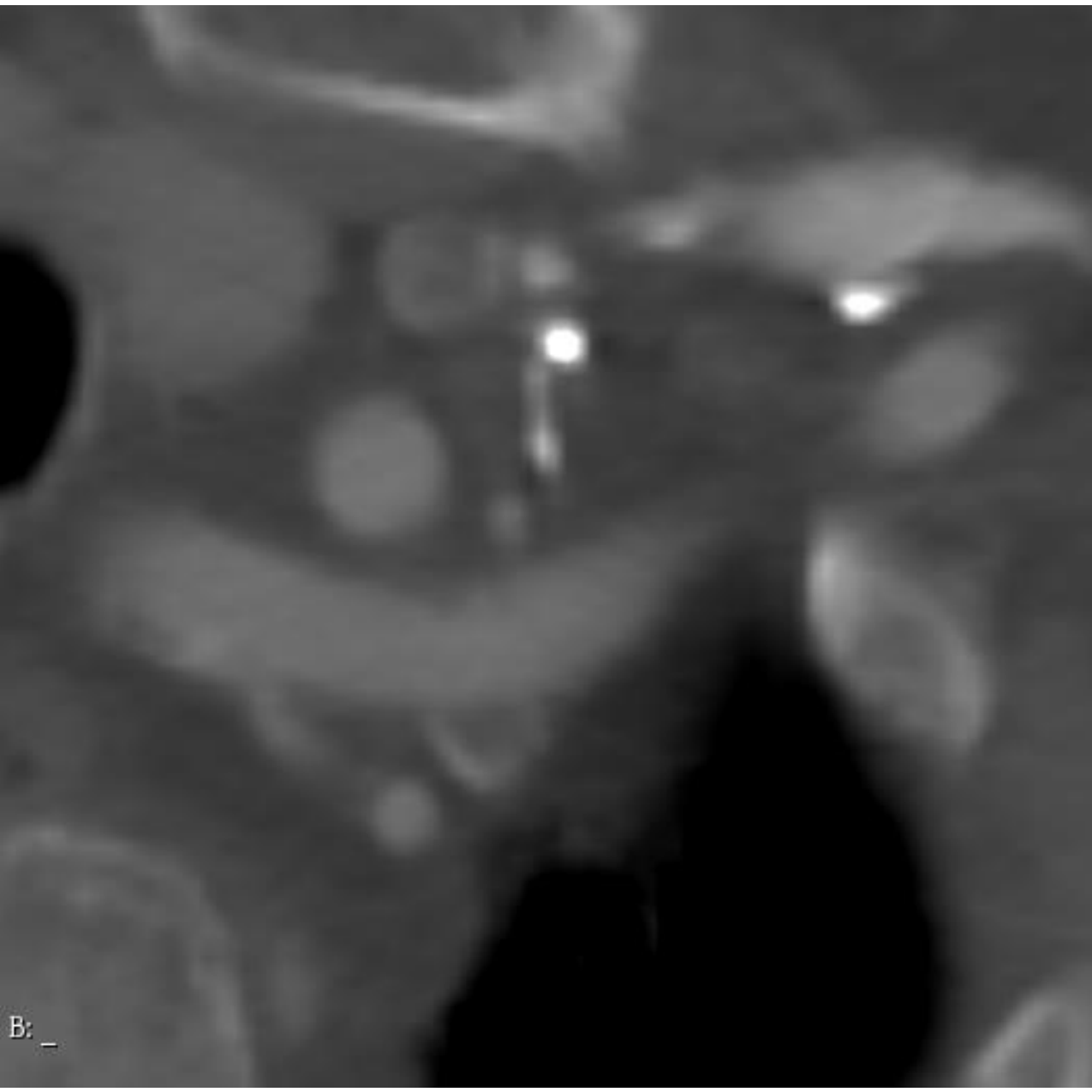}
  \caption{Input}
  \end{subfigure}
  \begin{subfigure}{0.14\linewidth}
     \includegraphics[width=1\textwidth]{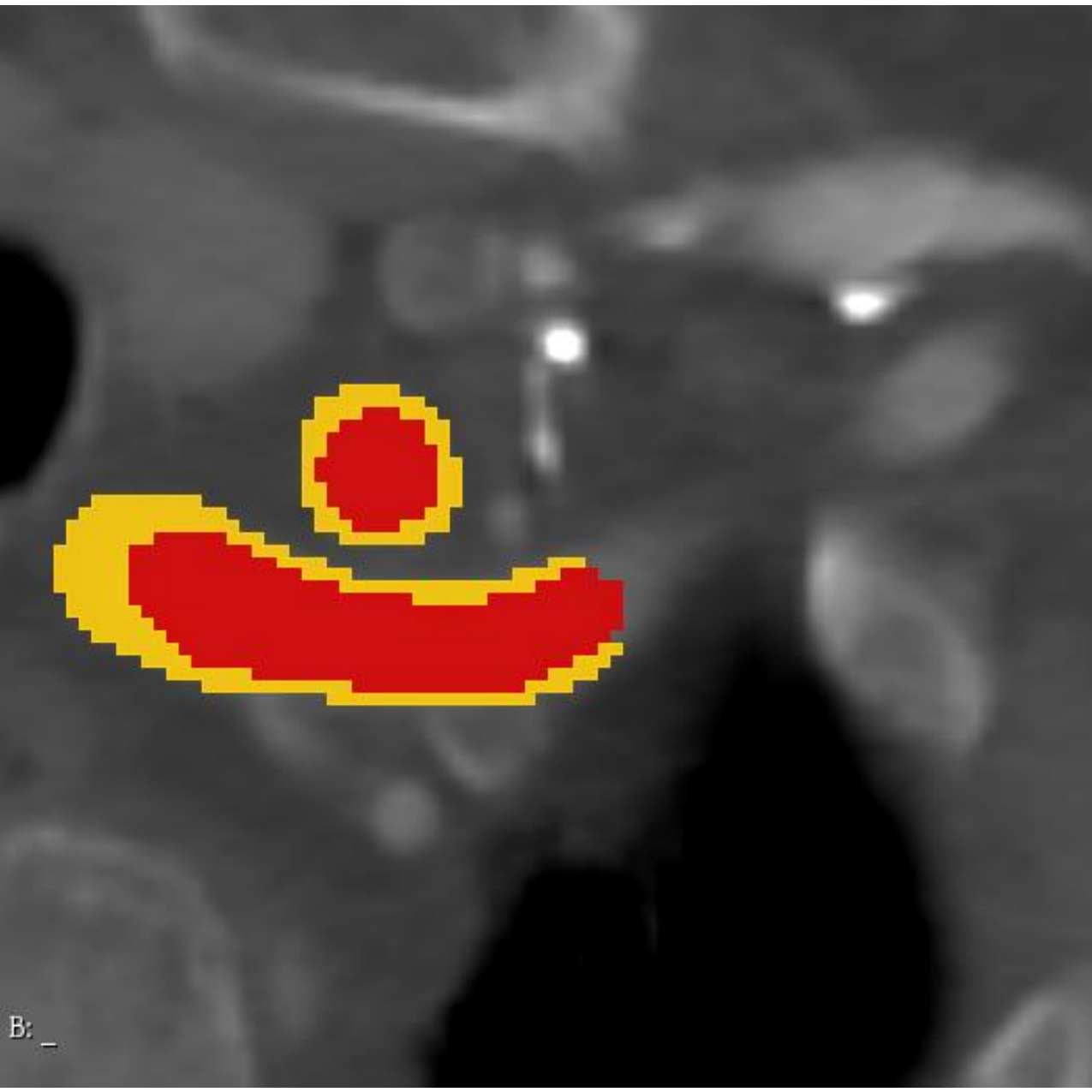}
     \caption{UNet}
  \end{subfigure}
    \begin{subfigure}{0.15\linewidth}
     \includegraphics[width=1\textwidth]{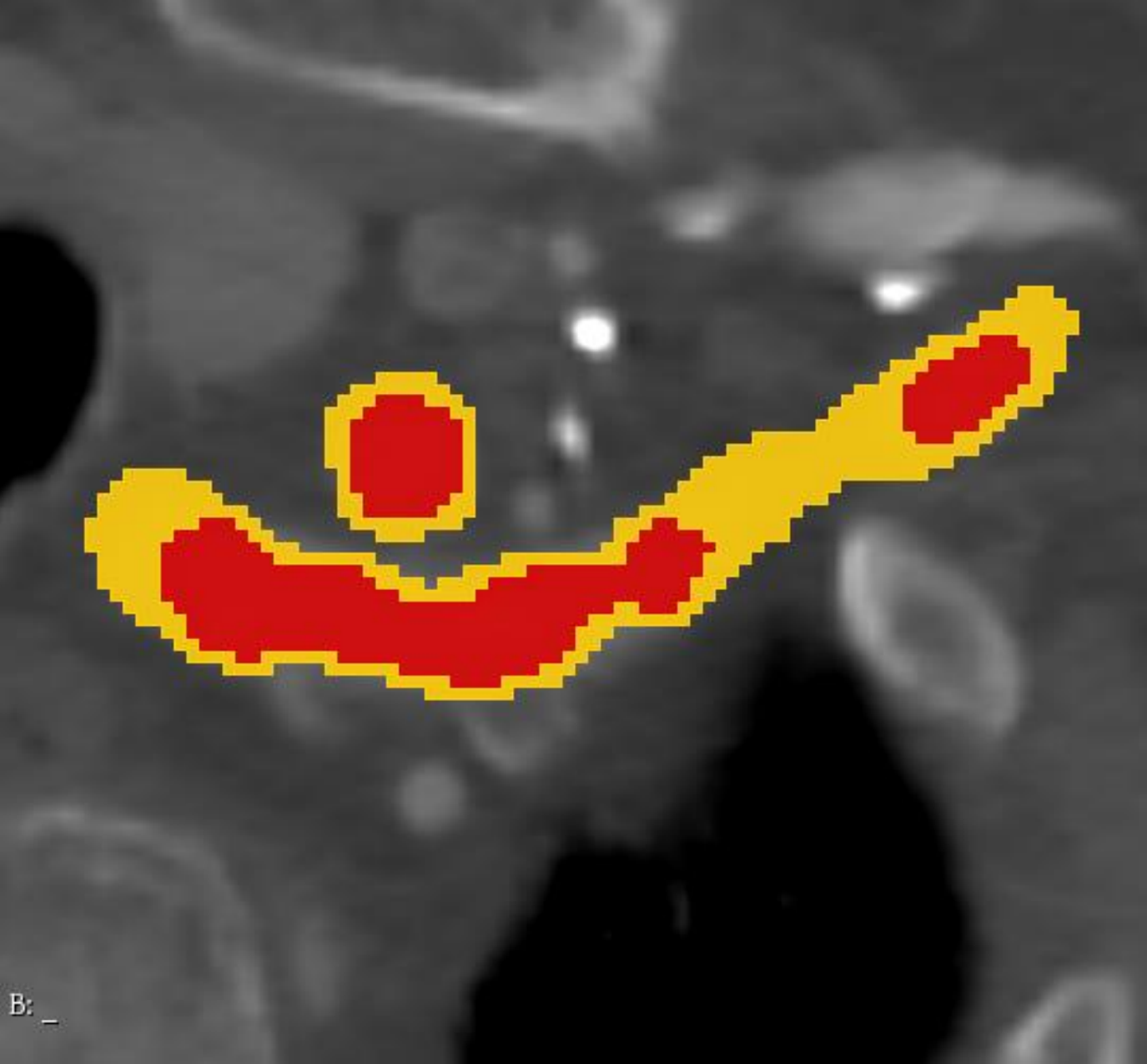}
     \caption{UNet+O}
  \end{subfigure}
    \begin{subfigure}{0.14\linewidth}
     \includegraphics[width=1\textwidth]{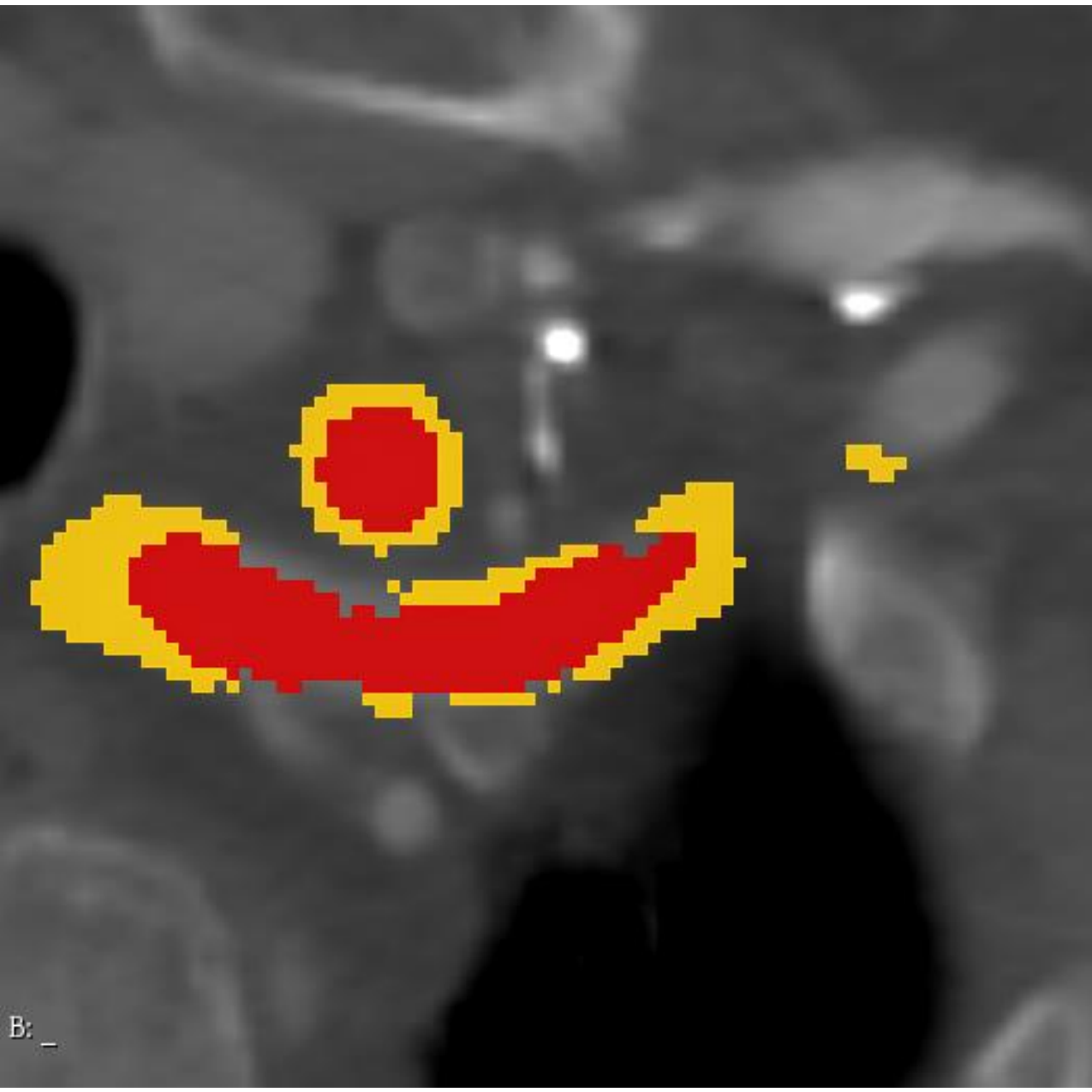}
     \caption{FCN}
  \end{subfigure}
      \begin{subfigure}{0.15\linewidth}
     \includegraphics[width=1\textwidth]{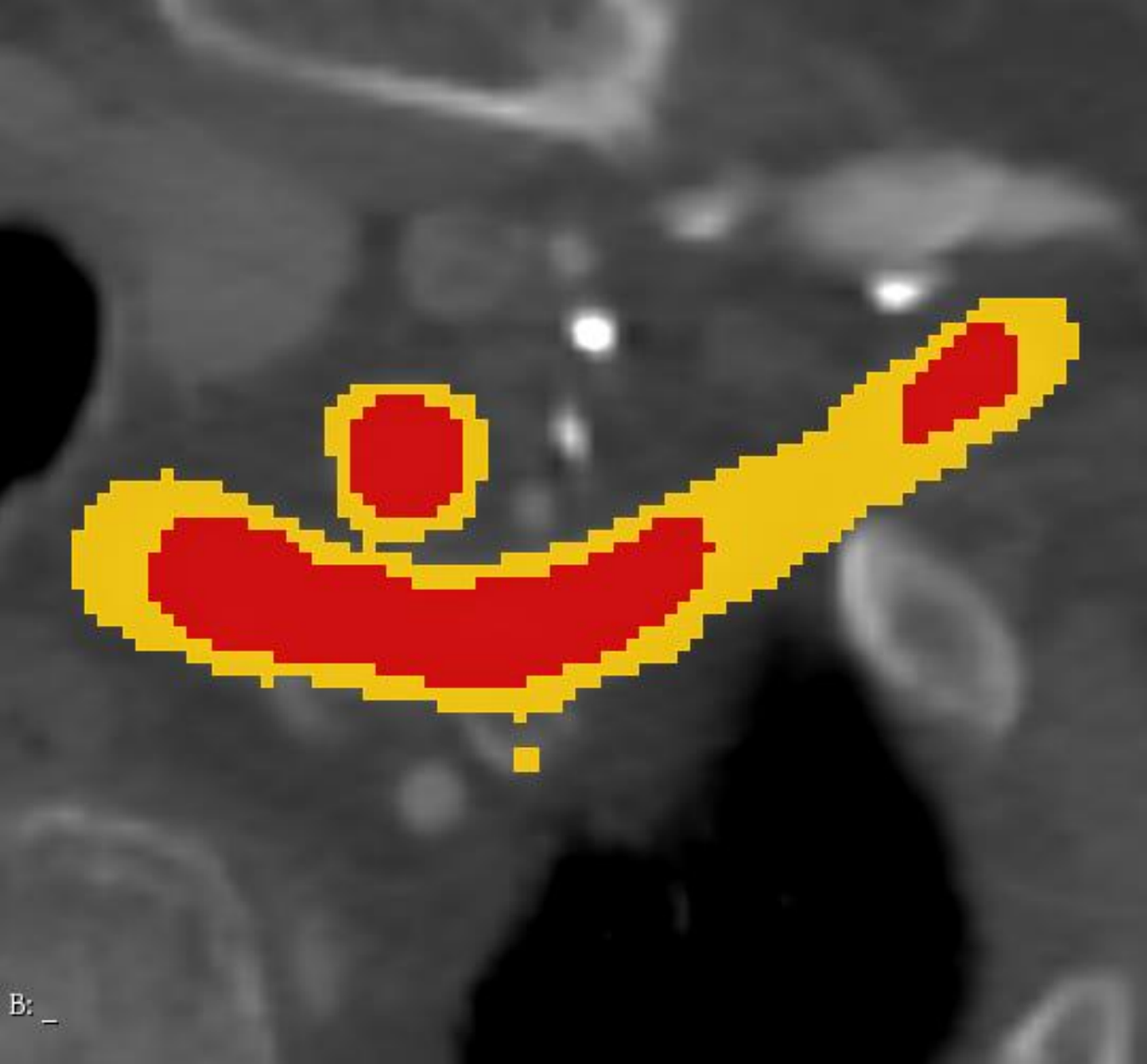}
     \caption{FCN+O}
  \end{subfigure}
    \begin{subfigure}{0.14\linewidth}
     \includegraphics[width=1\textwidth]{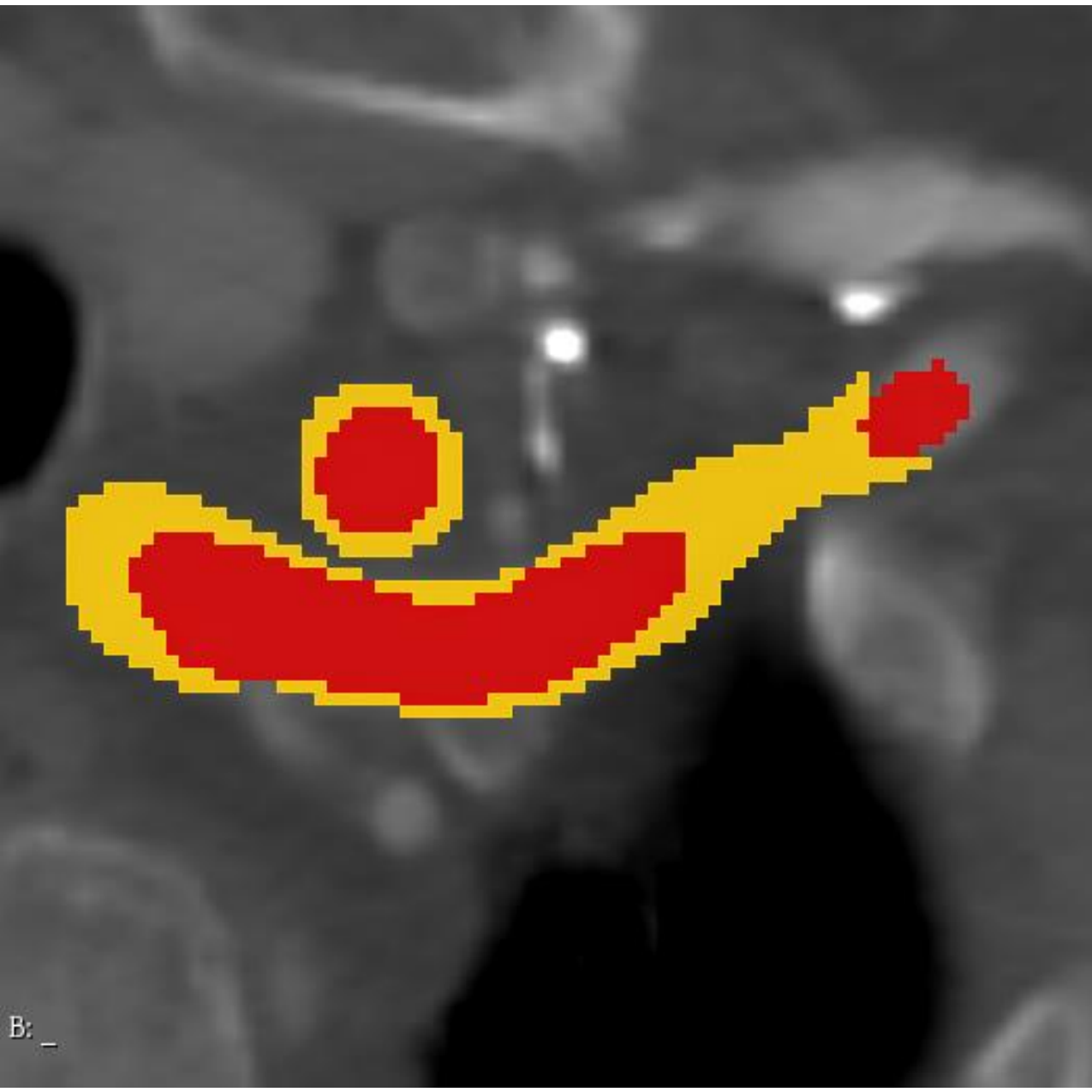}
     \caption{nnUNet}

  \end{subfigure}
      \begin{subfigure}{0.14\linewidth}
     \includegraphics[width=1\textwidth]{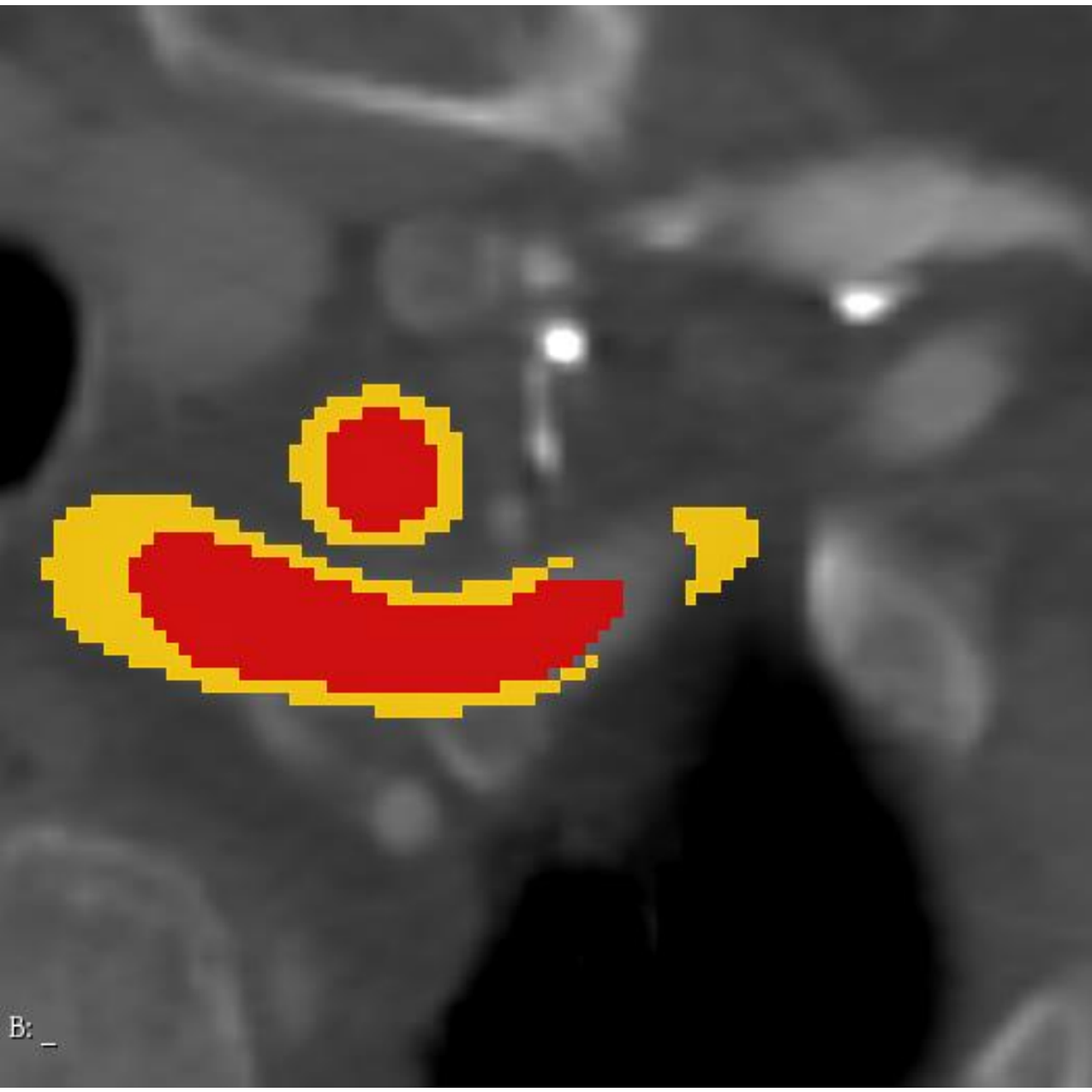}
     \caption{CRF}
  \end{subfigure}
    \begin{subfigure}{0.15\linewidth}
     \includegraphics[width=1\textwidth]{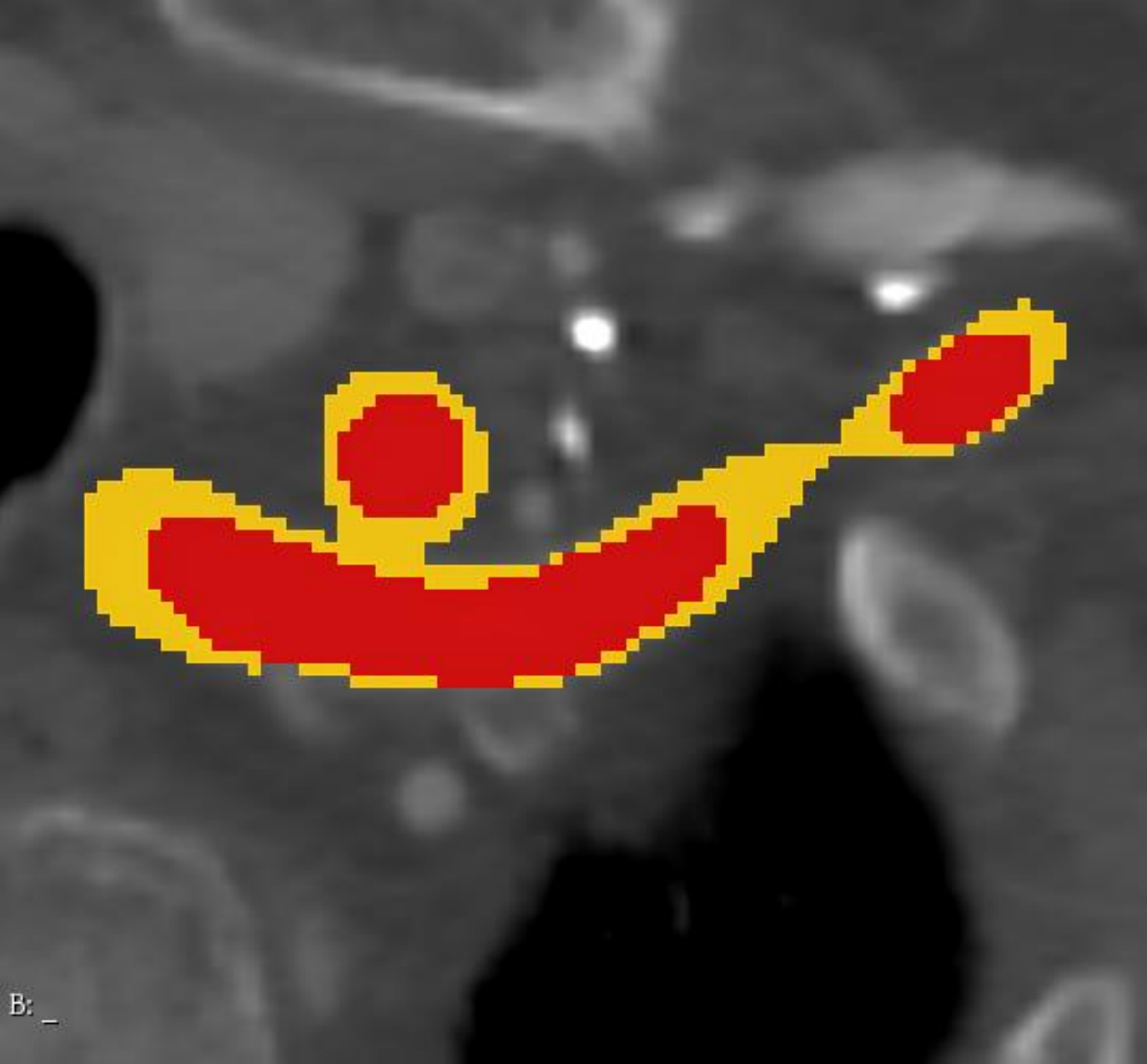}
     \caption{MIDL}
  \end{subfigure}
      \begin{subfigure}{0.15\linewidth}
     \includegraphics[width=1\textwidth]{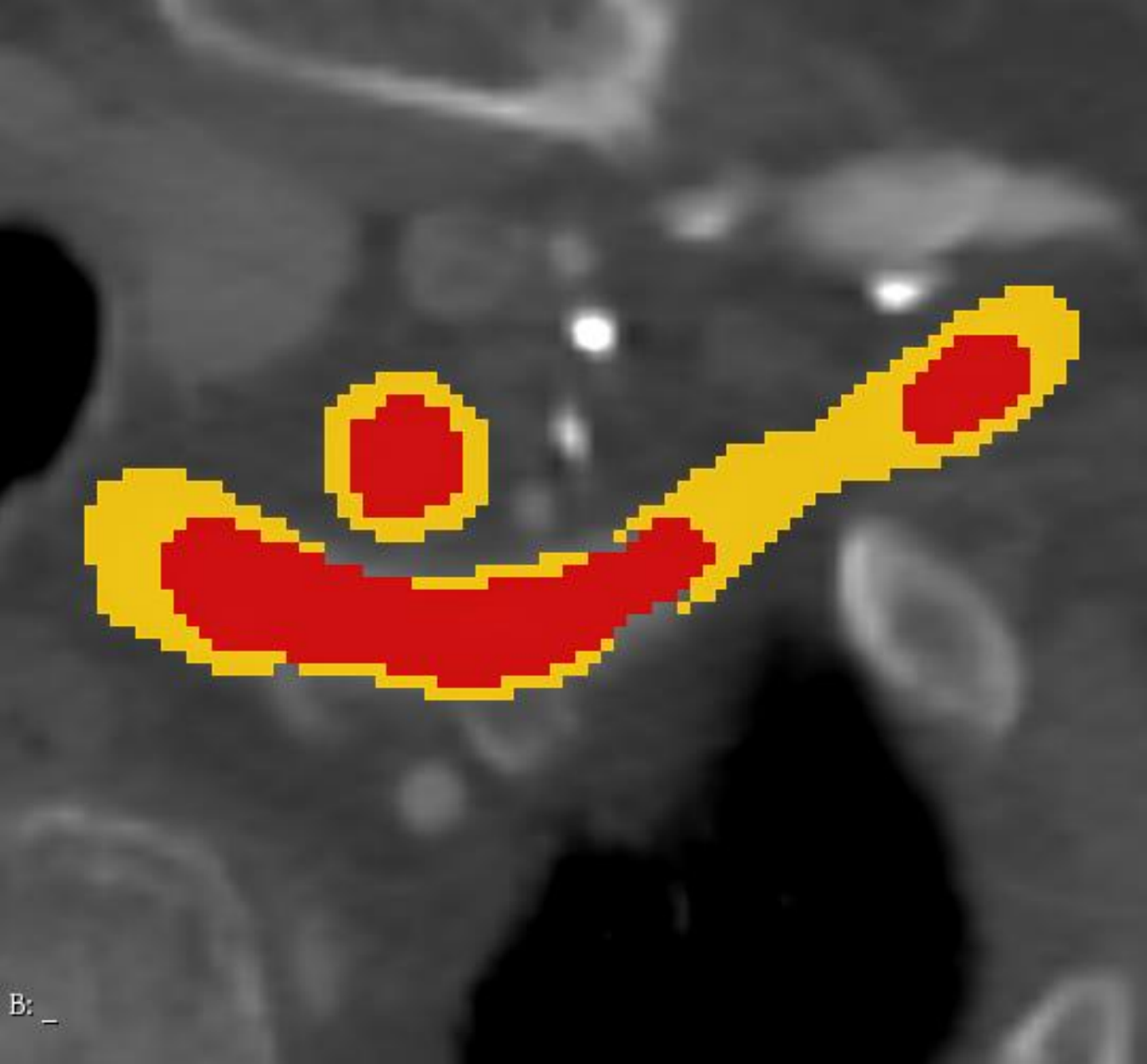}
     \caption{NonAdj}
  \end{subfigure}
      \begin{subfigure}{0.15\linewidth}
     \includegraphics[width=1\textwidth]{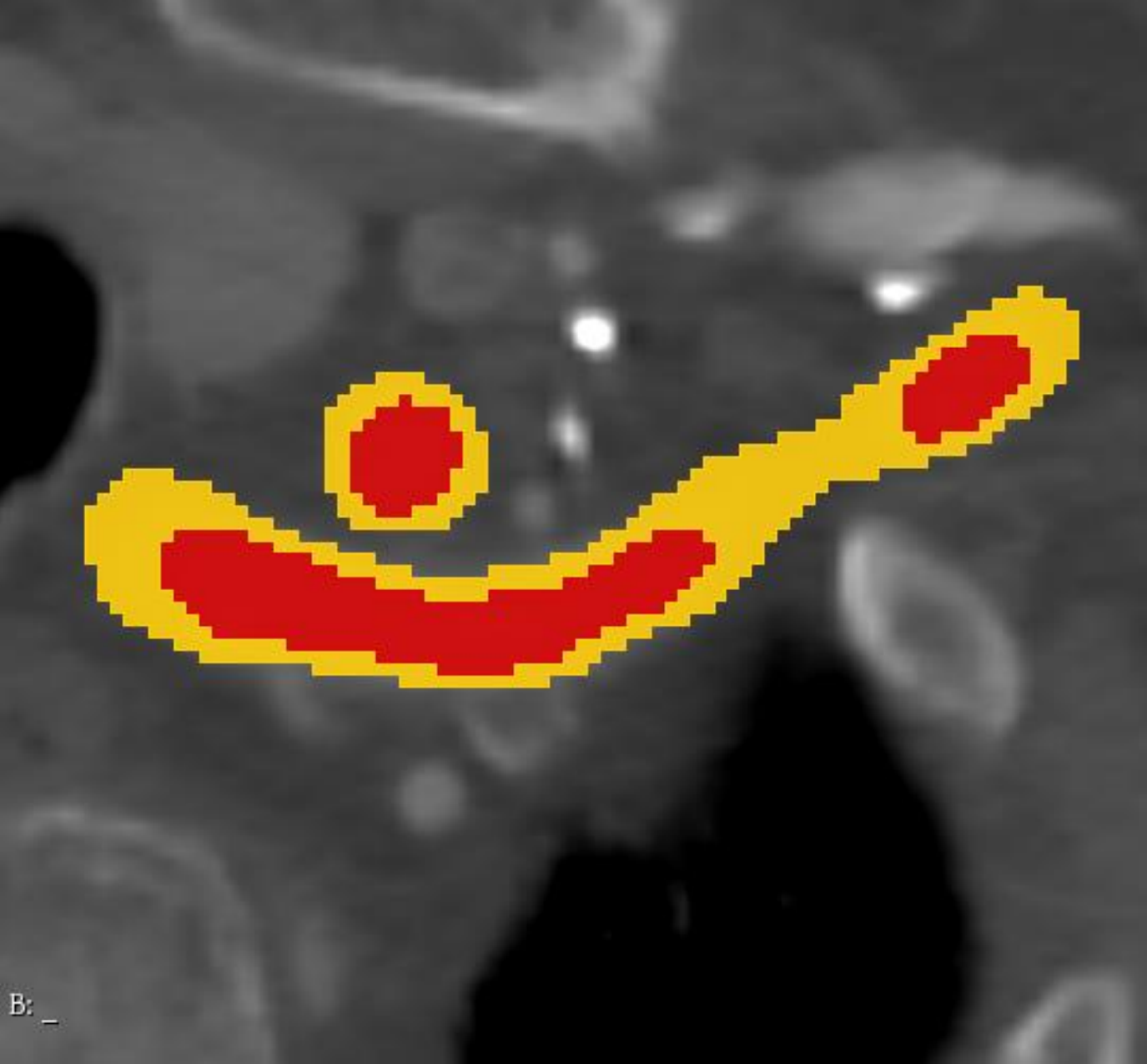}
     \caption{Ours6C}
  \end{subfigure}
        \begin{subfigure}{0.14\linewidth}
     \includegraphics[width=1\textwidth]{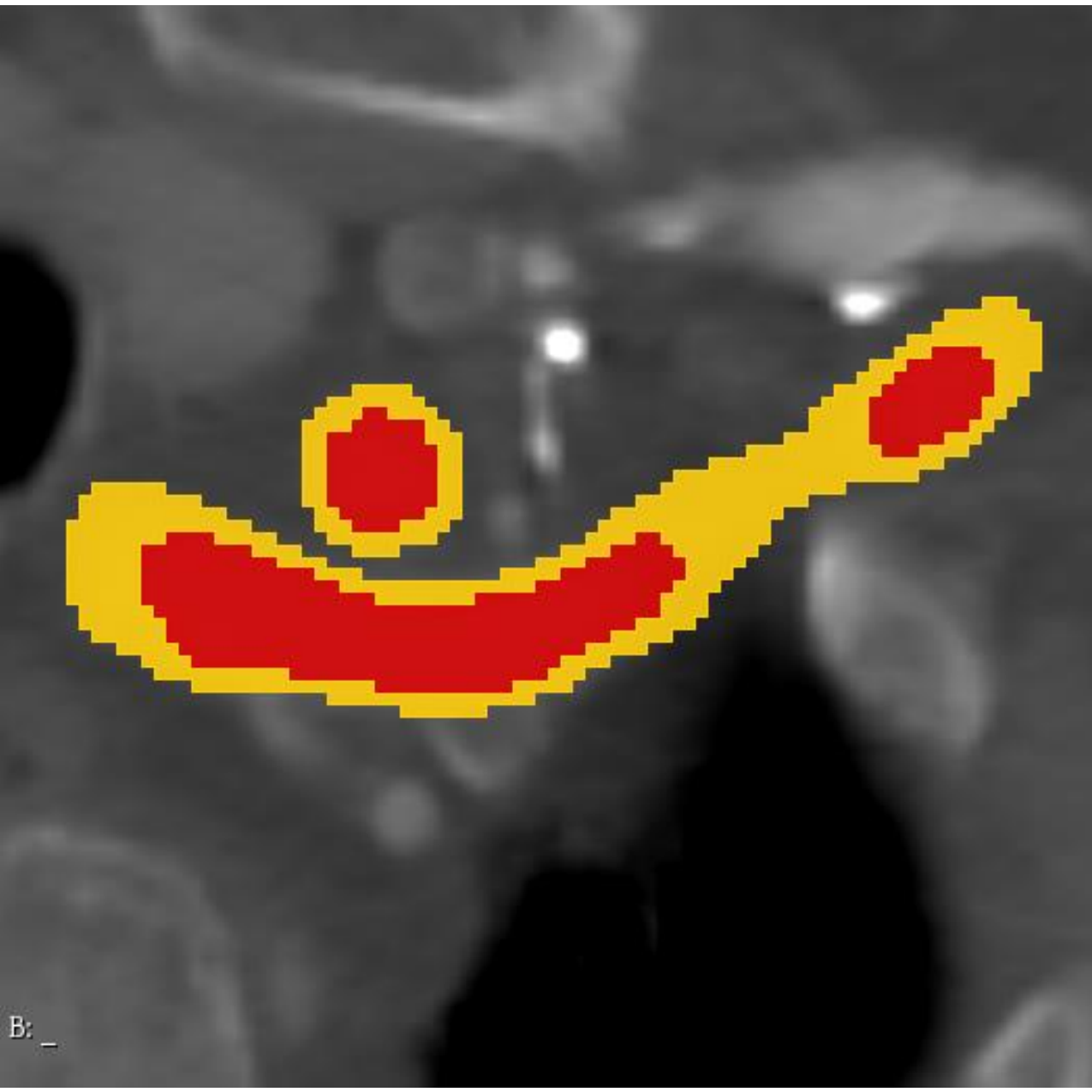}
     \caption{Ours}
  \end{subfigure}
      \begin{subfigure}{0.14\linewidth}
     \includegraphics[width=1\textwidth]{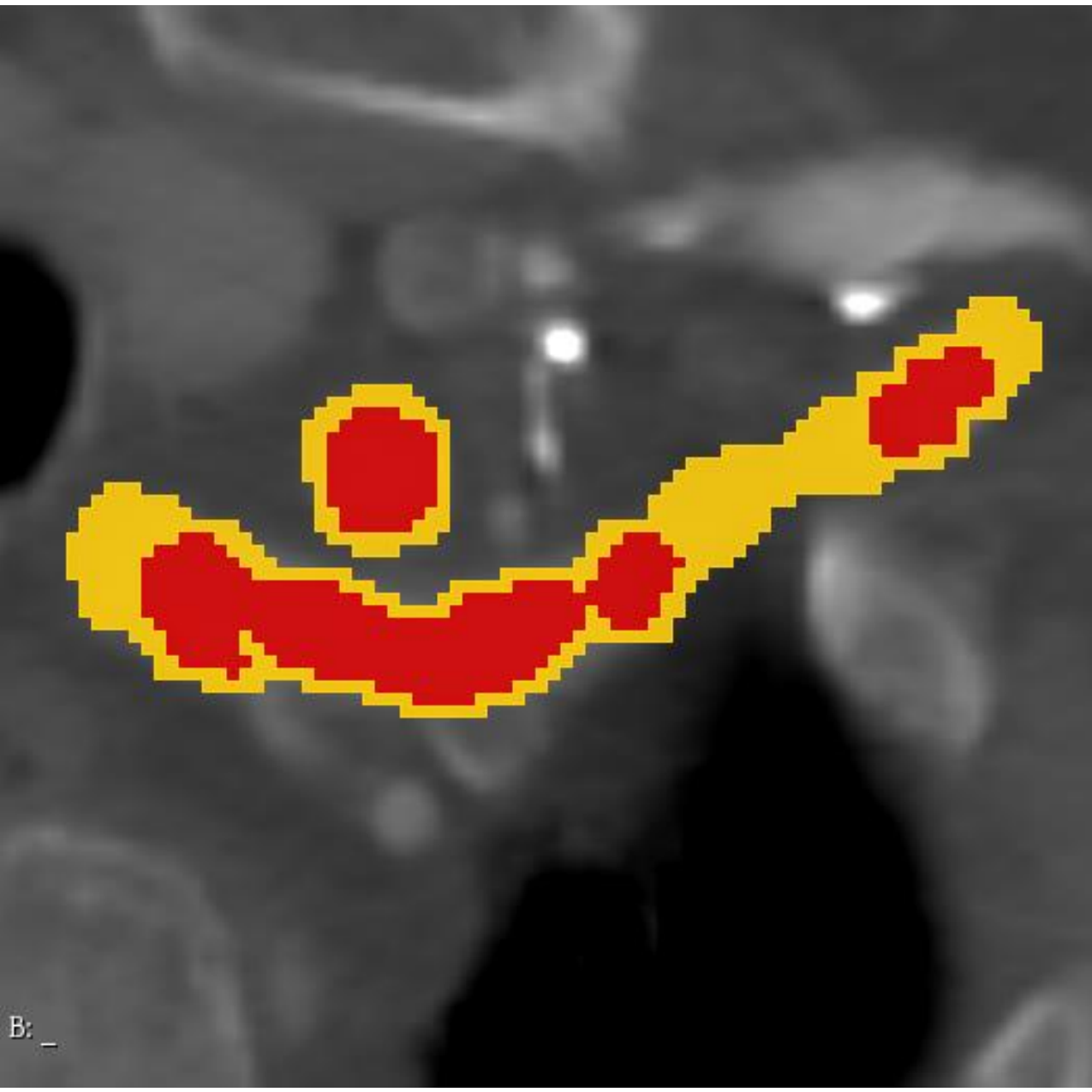}
     \caption{GT}
  \end{subfigure}

        \begin{subfigure}{0.15\linewidth}
  \includegraphics[width=1\textwidth]{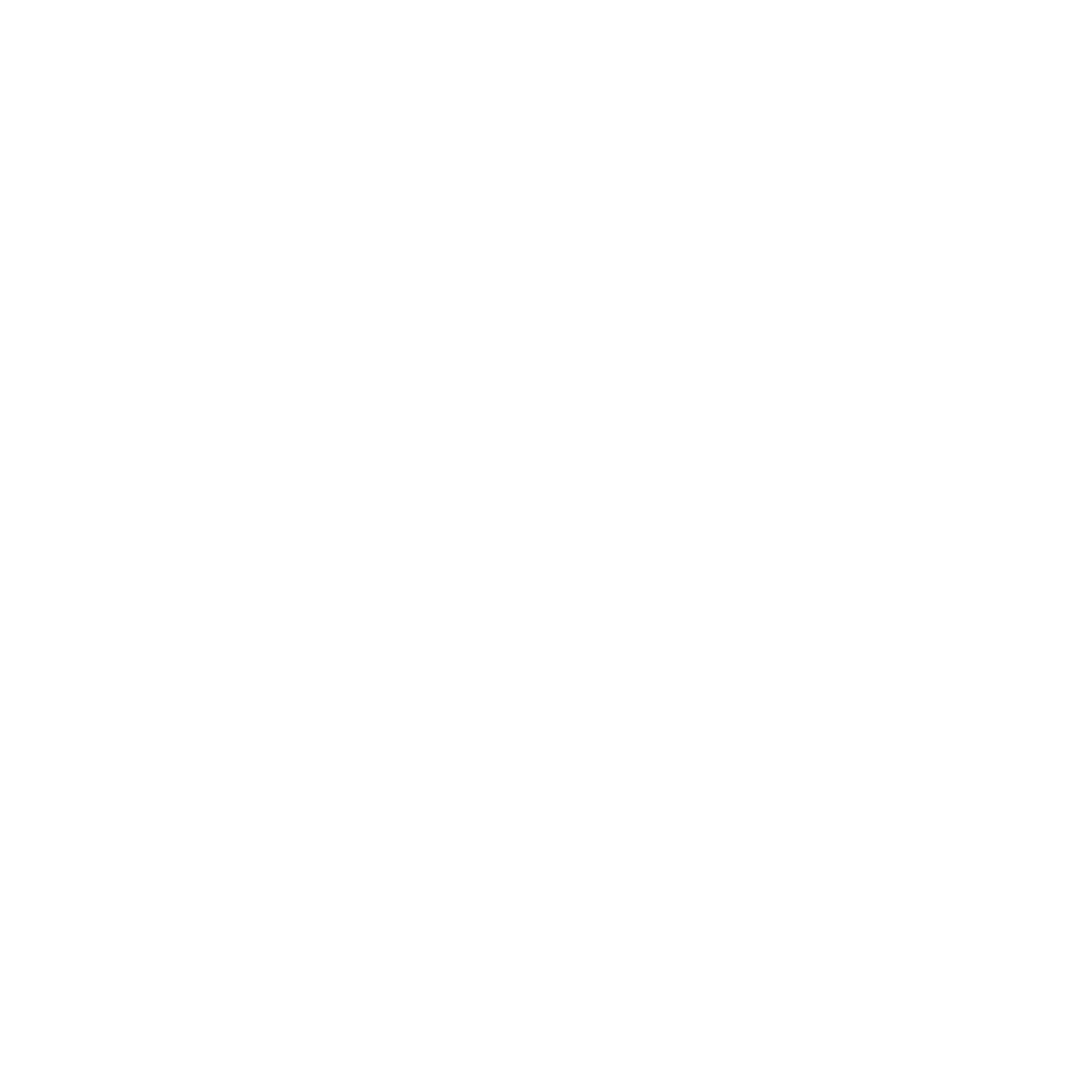}
  \end{subfigure}
  \begin{subfigure}{0.14\linewidth}
     \includegraphics[width=1\textwidth]{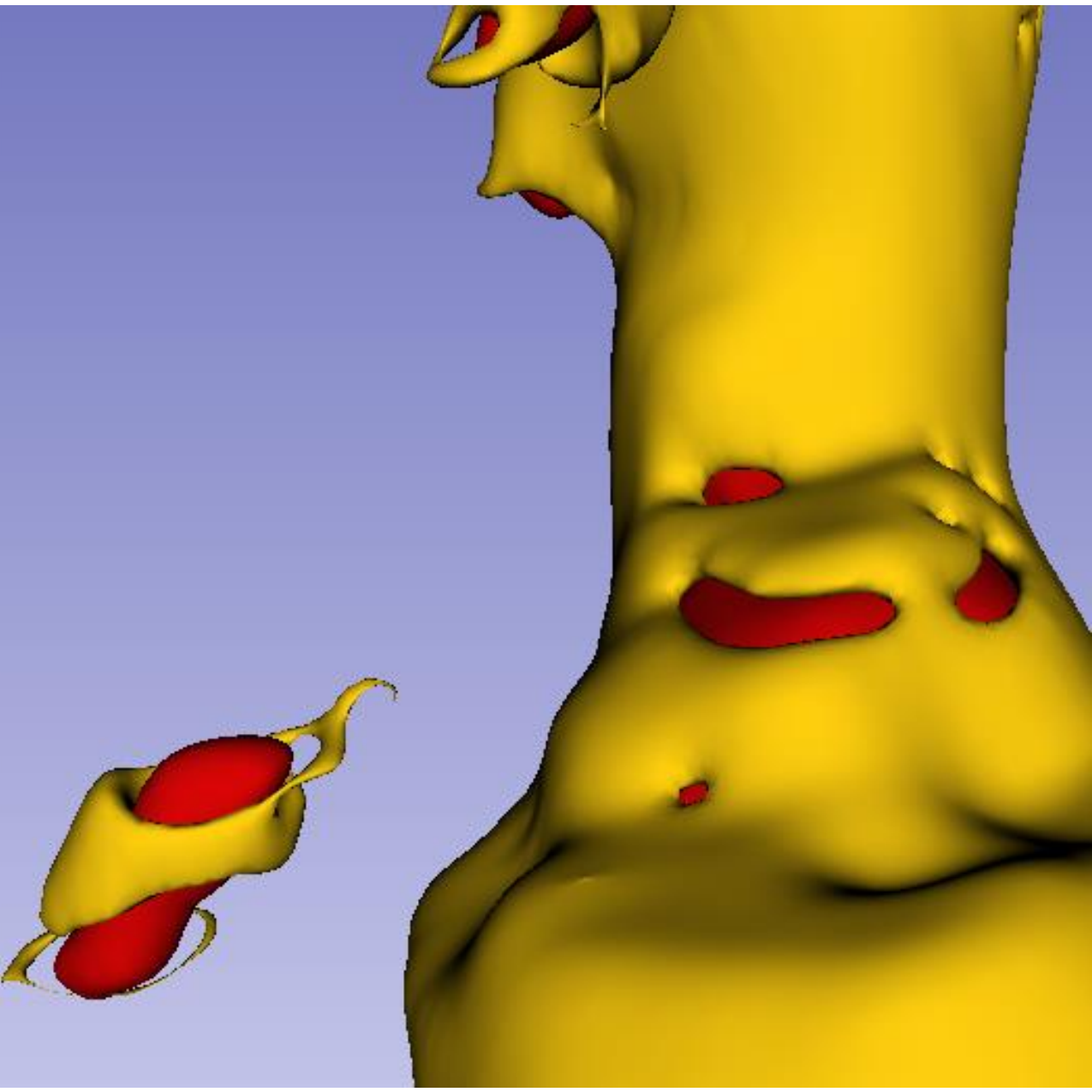}
     \caption{UNet}
  \end{subfigure}
    \begin{subfigure}{0.15\linewidth}
     \includegraphics[width=1\textwidth]{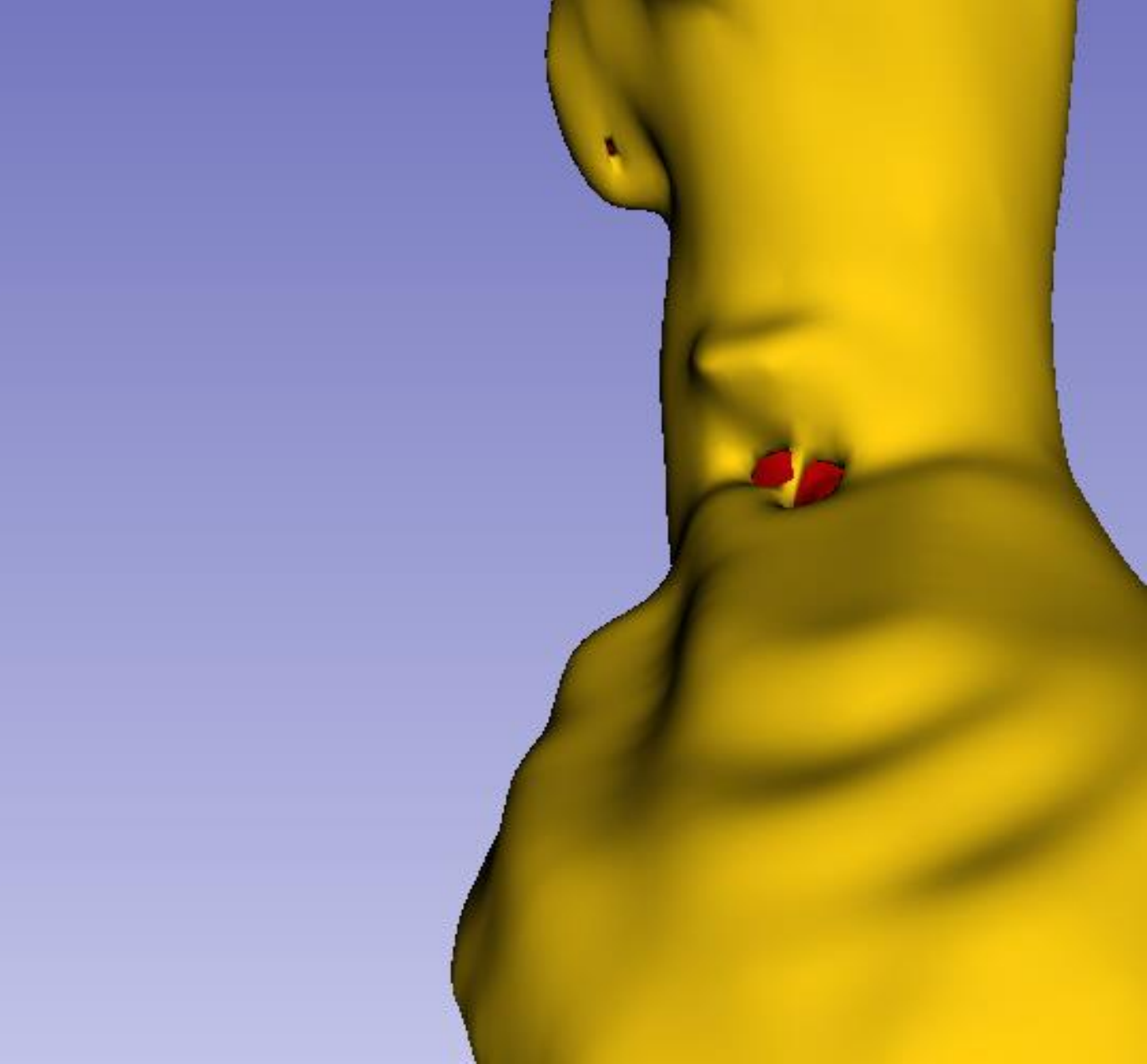}
     \caption{UNet+O}
  \end{subfigure}
    \begin{subfigure}{0.14\linewidth}
     \includegraphics[width=1\textwidth]{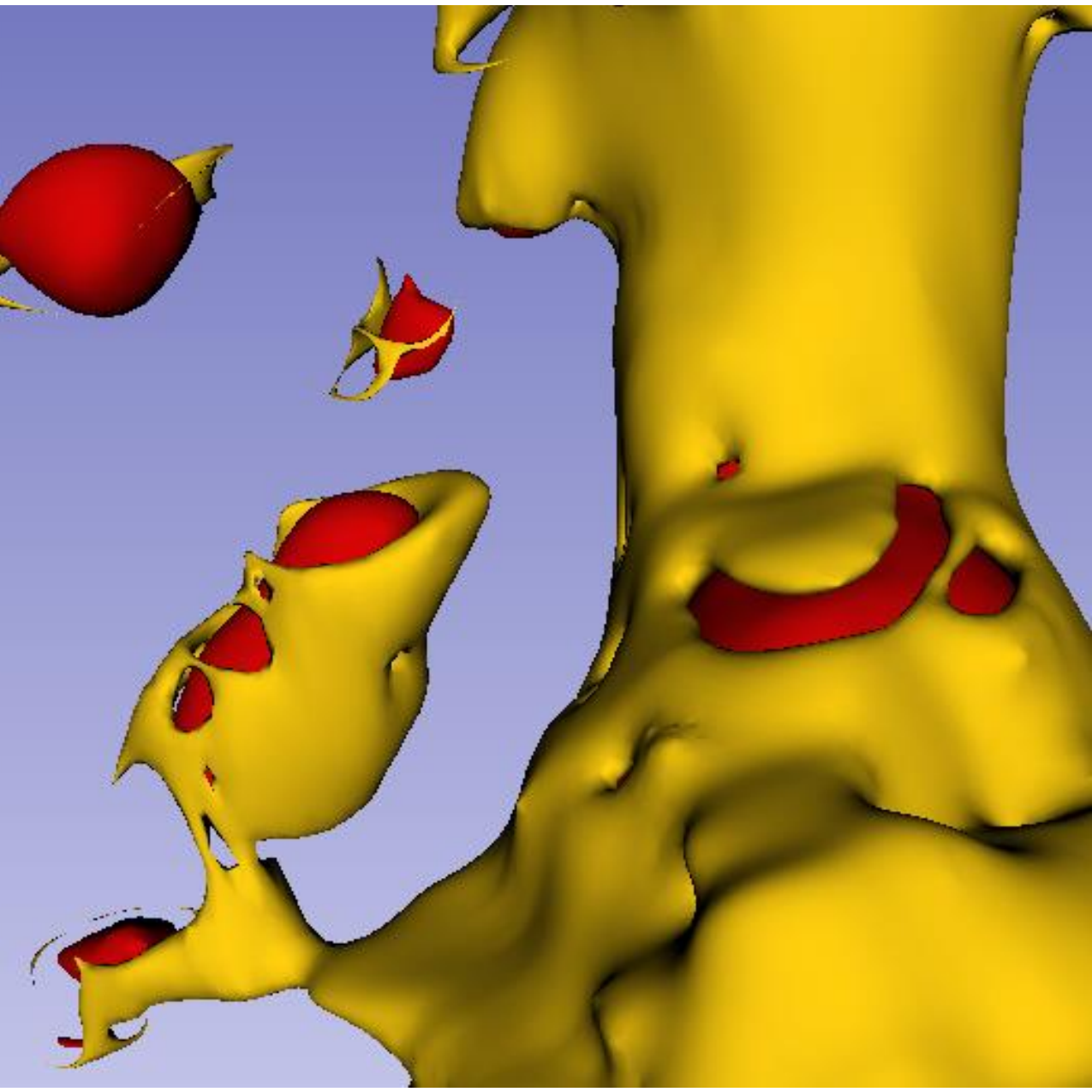}
     \caption{FCN}
  \end{subfigure}
      \begin{subfigure}{0.15\linewidth}
     \includegraphics[width=1\textwidth]{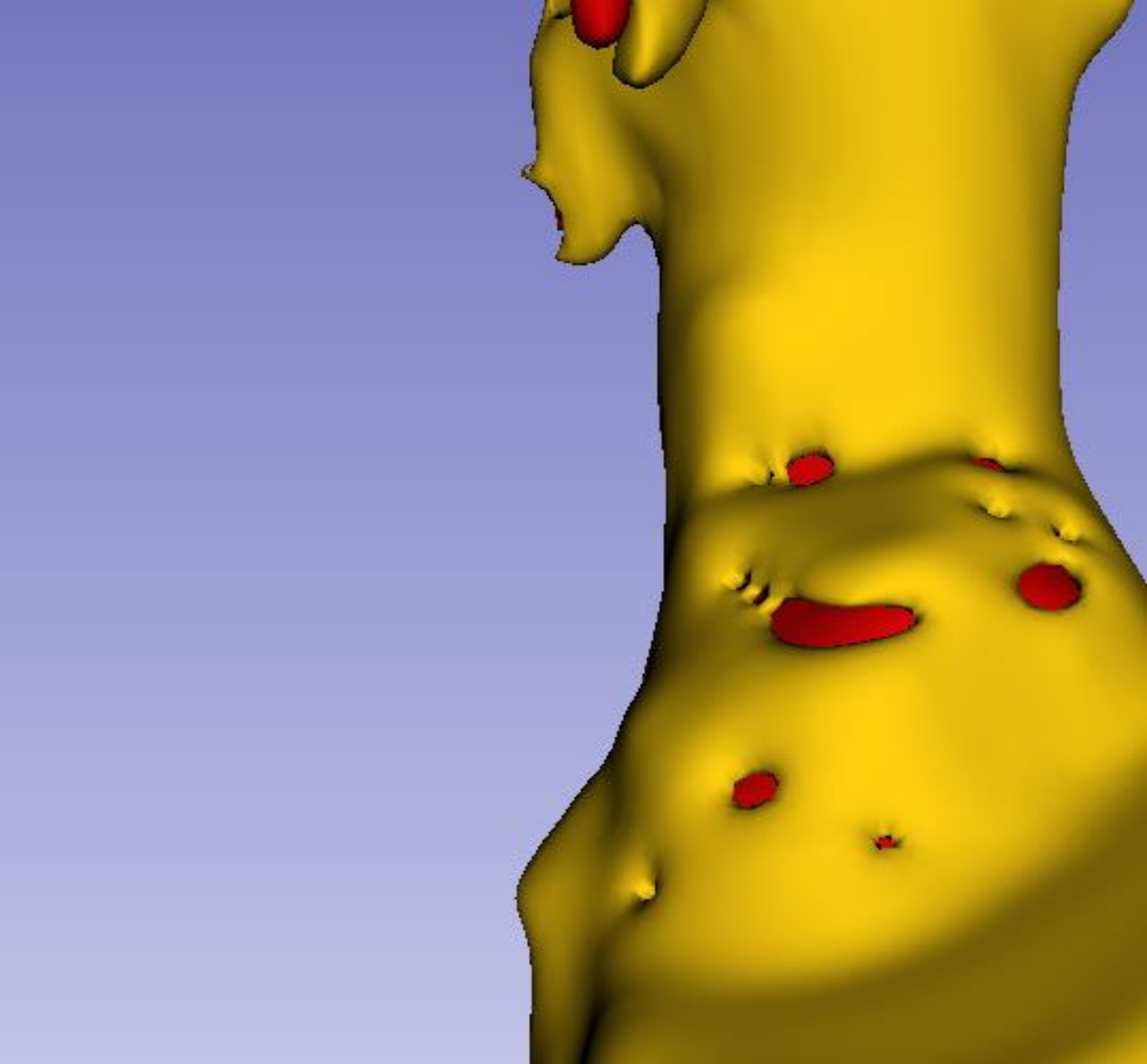}
     \caption{FCN+O}
  \end{subfigure}
    \begin{subfigure}{0.14\linewidth}
     \includegraphics[width=1\textwidth]{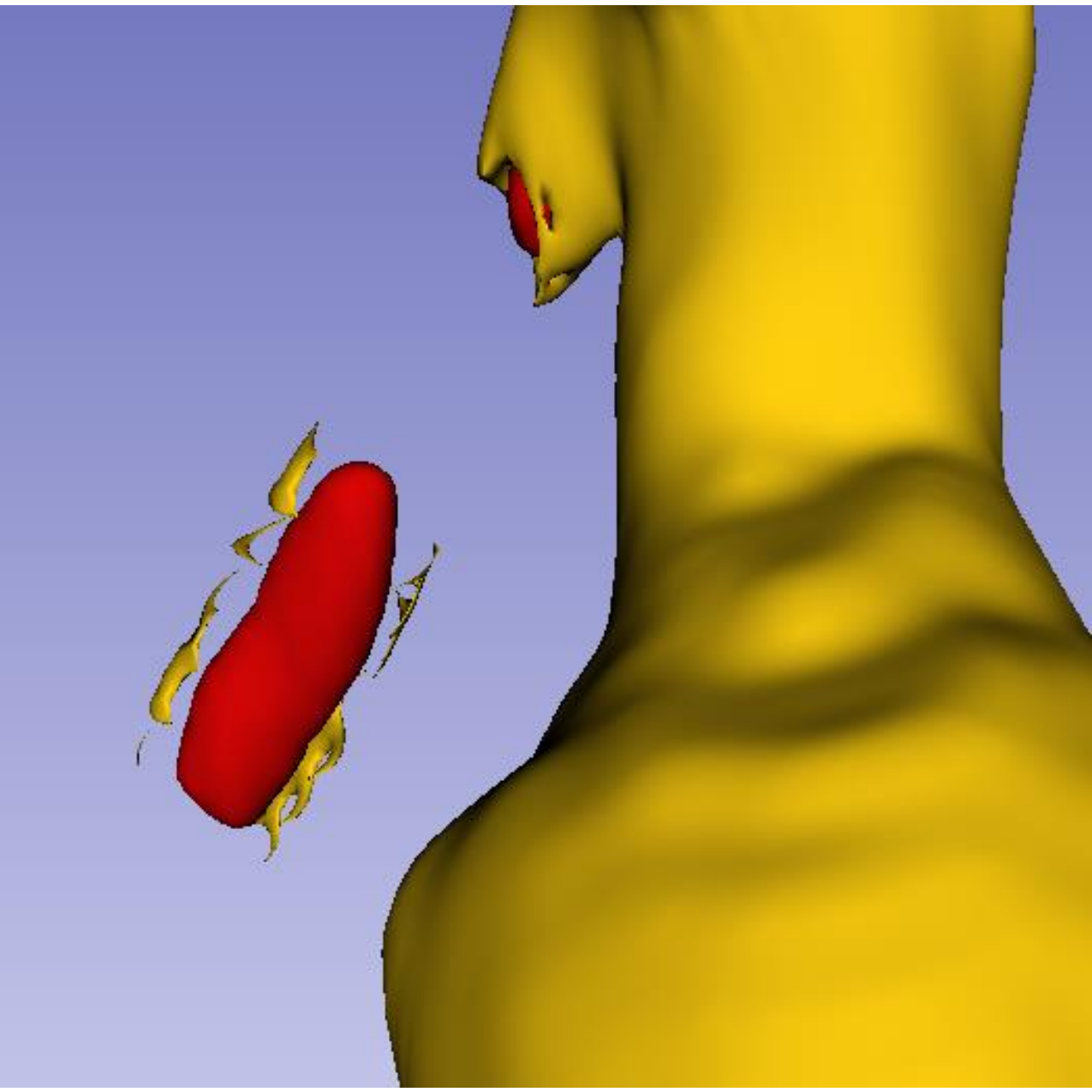}
     \caption{nnUNet}
  \end{subfigure}

      \begin{subfigure}{0.14\linewidth}
     \includegraphics[width=1\textwidth]{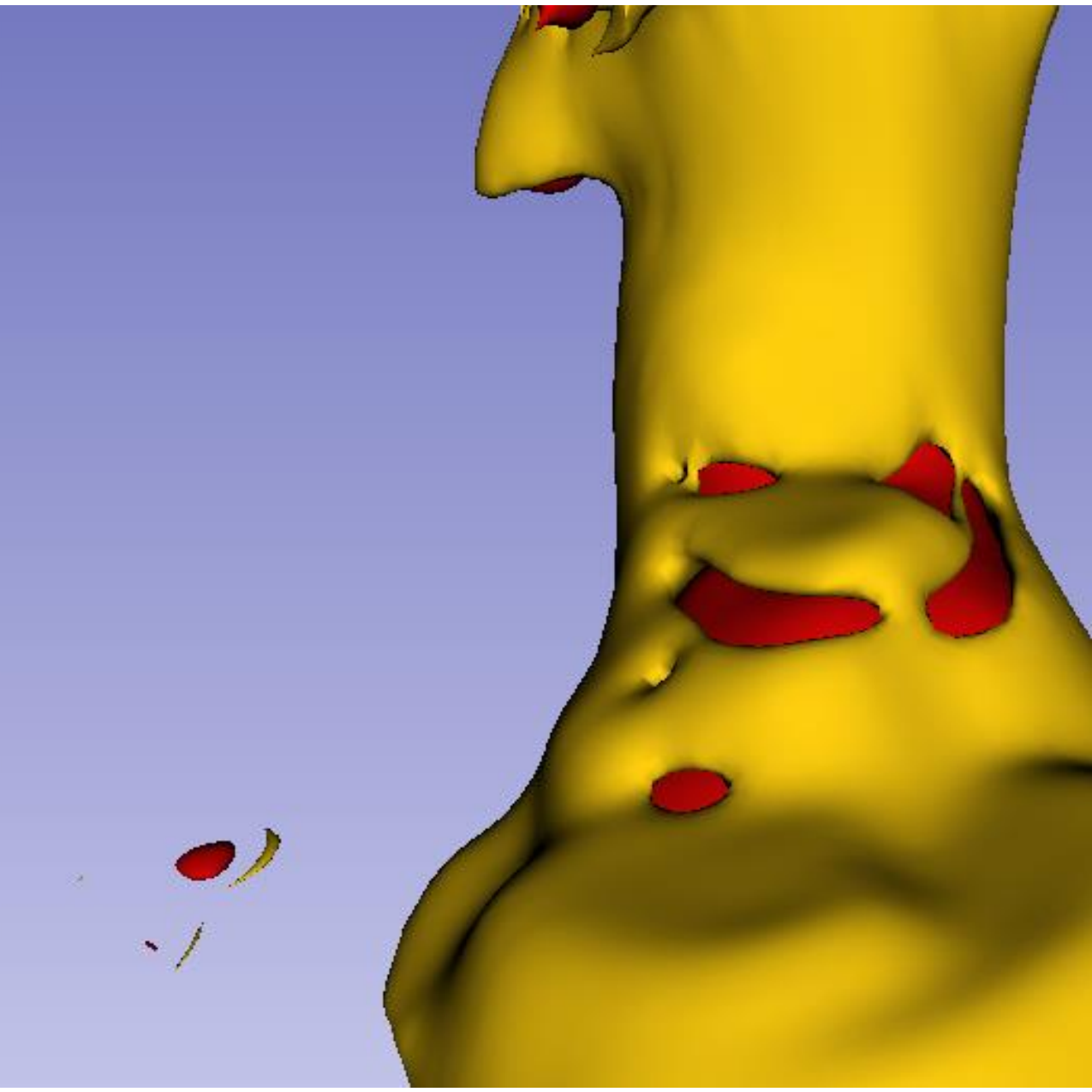}
     \caption{CRF}
  \end{subfigure}
    \begin{subfigure}{0.15\linewidth}
     \includegraphics[width=1\textwidth]{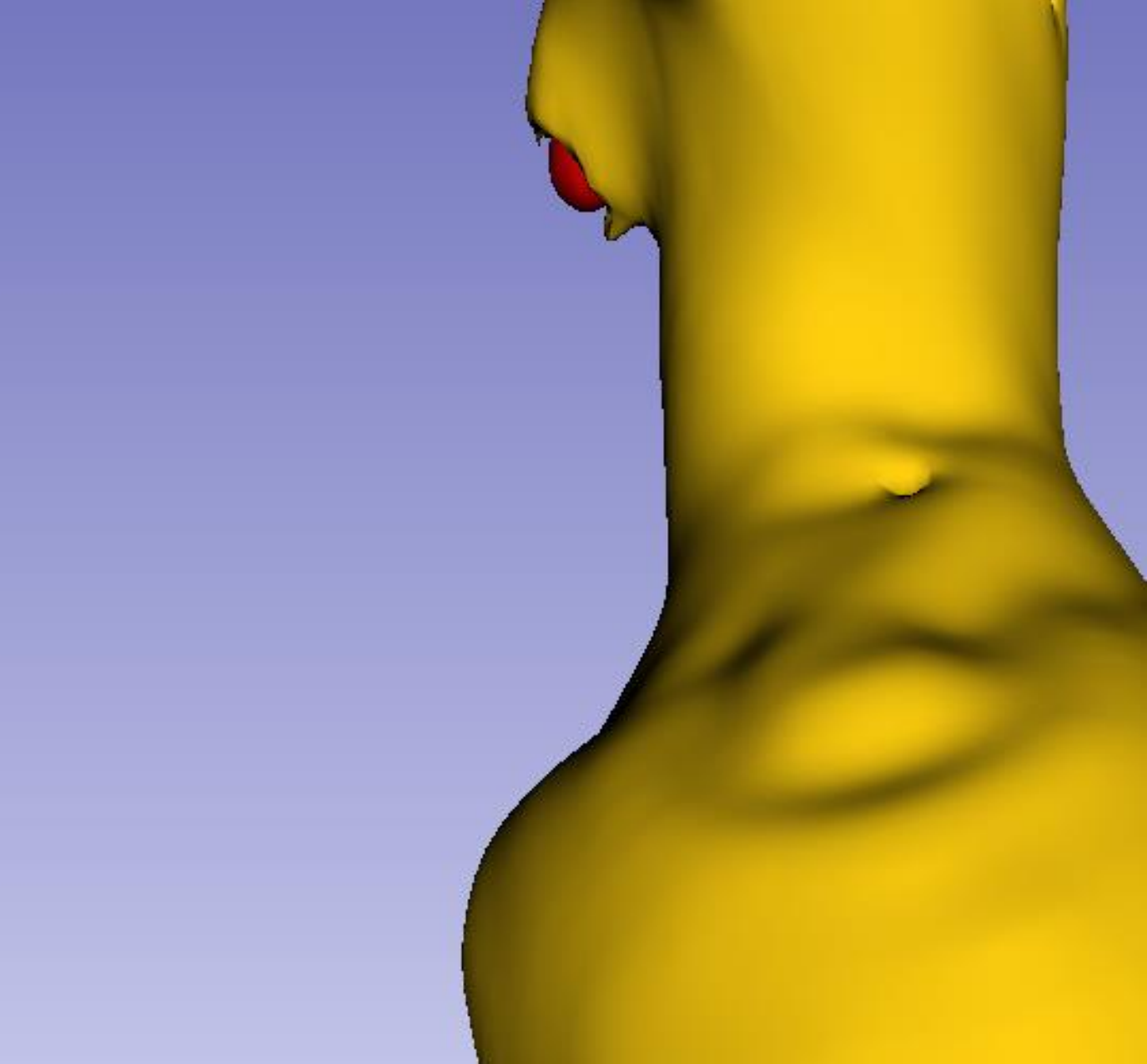}
     \caption{MIDL}
  \end{subfigure}
      \begin{subfigure}{0.15\linewidth}
     \includegraphics[width=1\textwidth]{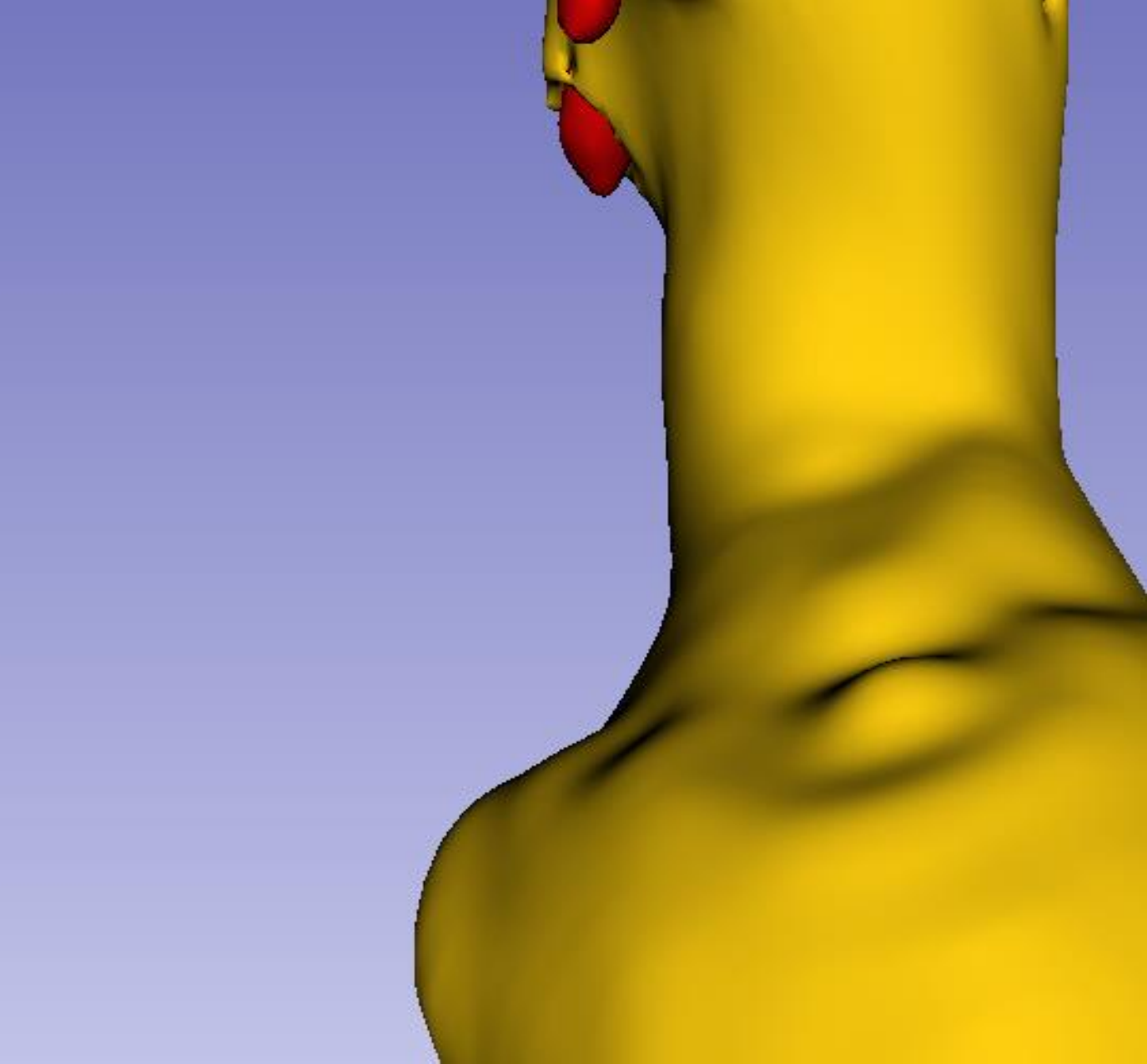}
     \caption{NonAdj}
  \end{subfigure}
      \begin{subfigure}{0.15\linewidth}
     \includegraphics[width=1\textwidth]{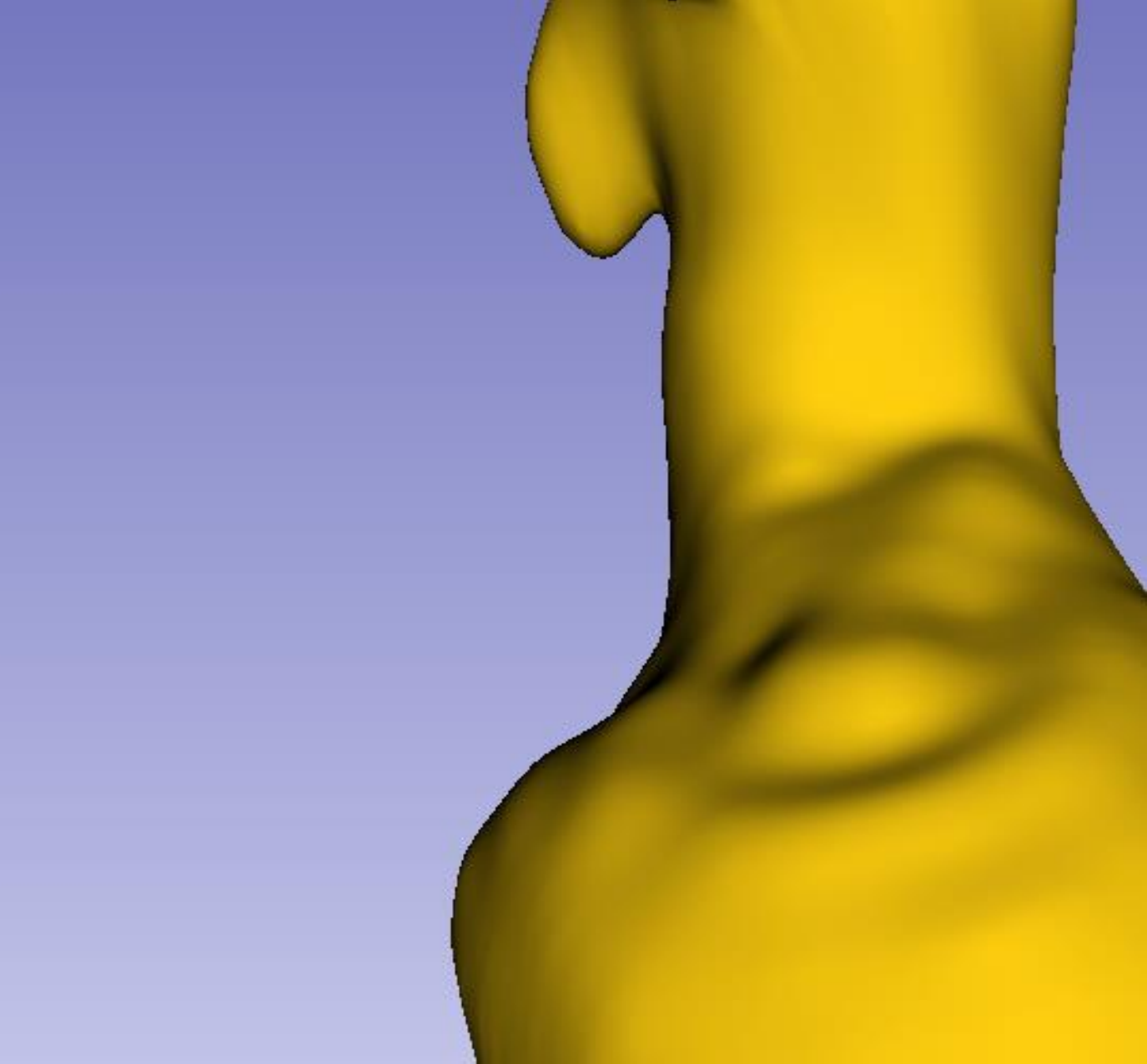}
     \caption{Ours6C}
  \end{subfigure} 
        \begin{subfigure}{0.14\linewidth}
     \includegraphics[width=1\textwidth]{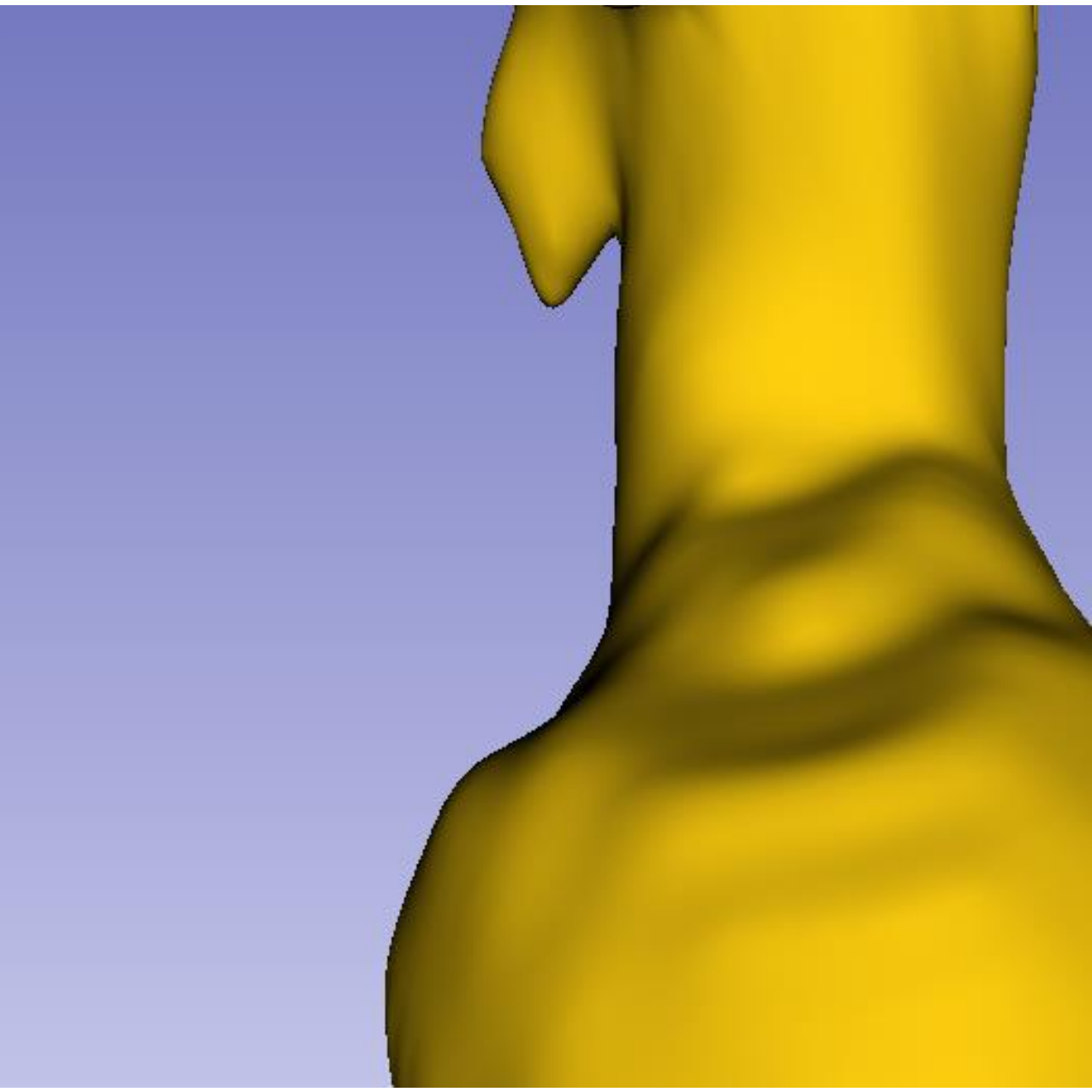}
     \caption{Ours}
  \end{subfigure}
      \begin{subfigure}{0.14\linewidth}
     \includegraphics[width=1\textwidth]{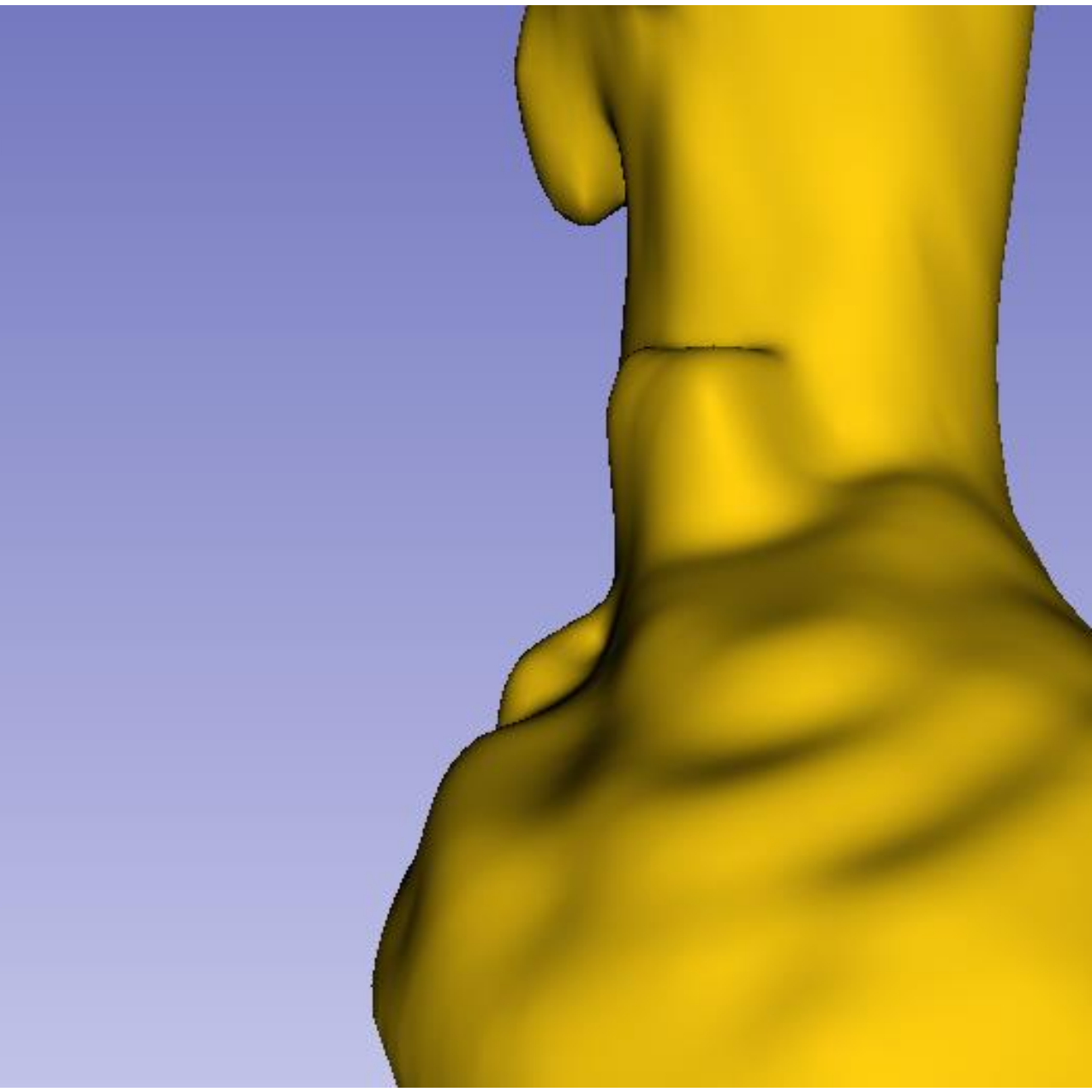}
     \caption{GT}
  \end{subfigure}

\caption{Additional qualitative Aorta results compared with the baselines. Rows 3-4 are corresponding 3D renderings. It is hard to visualize the input 3D volumetric image and so we leave it blank in the third row. Colors for the classes correspond to the ones used in Fig.~\ref{fig:data-interactions}.}
\label{fig:aorta-add-2}
\end{figure}

\begin{figure}[t]
\centering 

      \begin{subfigure}{0.15\linewidth}
  \includegraphics[width=1\textwidth]{figures/ivus/sample7/input-slice.pdf}
  \caption{Input}
  \end{subfigure}
  \begin{subfigure}{0.15\linewidth}
     \includegraphics[width=1\textwidth]{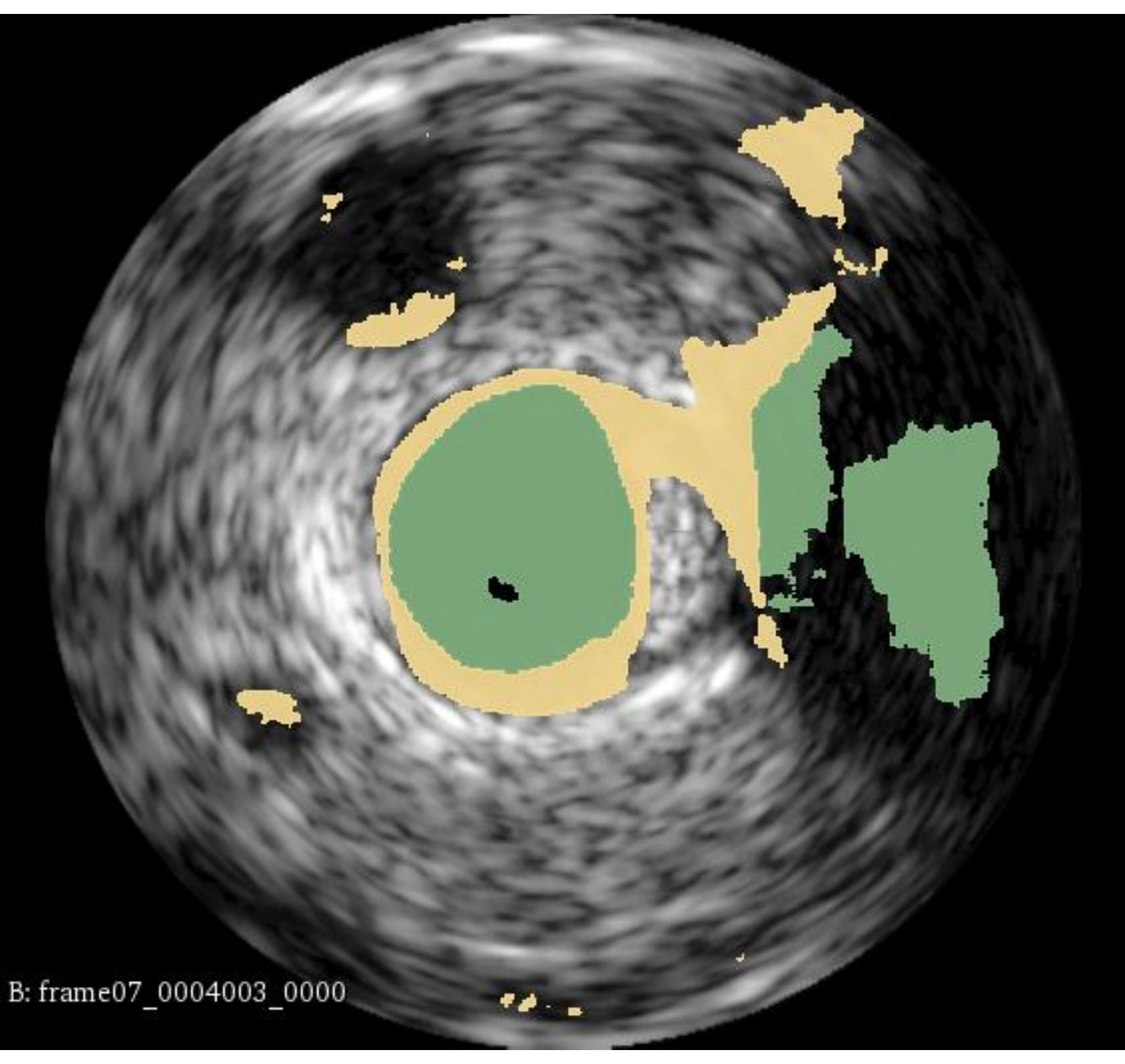}
     \caption{UNet}
  \end{subfigure}
          \begin{subfigure}{0.15\linewidth}
  \includegraphics[width=1\textwidth]{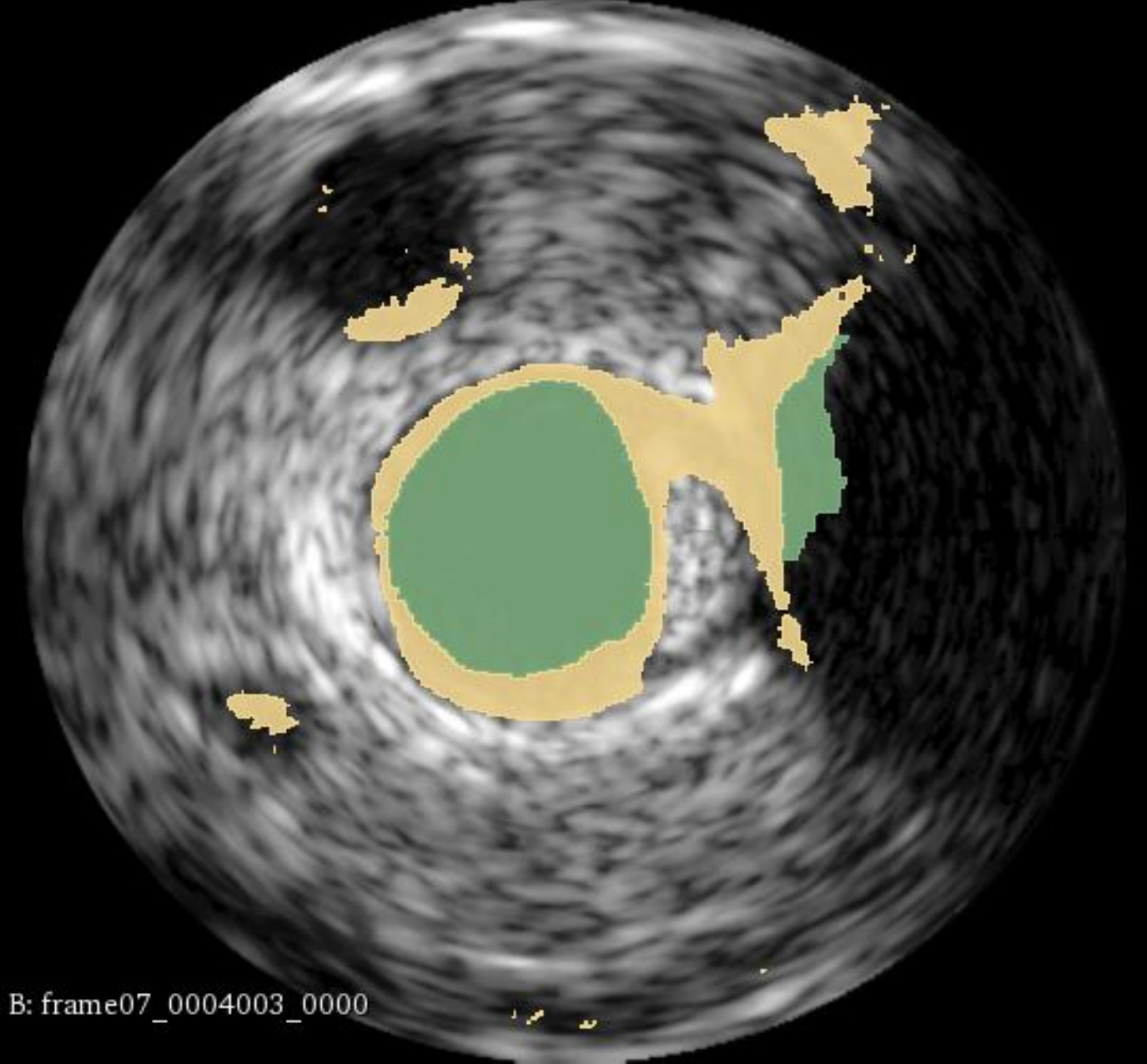}
  \caption{UNet+O}
  \end{subfigure}
    \begin{subfigure}{0.15\linewidth}
     \includegraphics[width=1\textwidth]{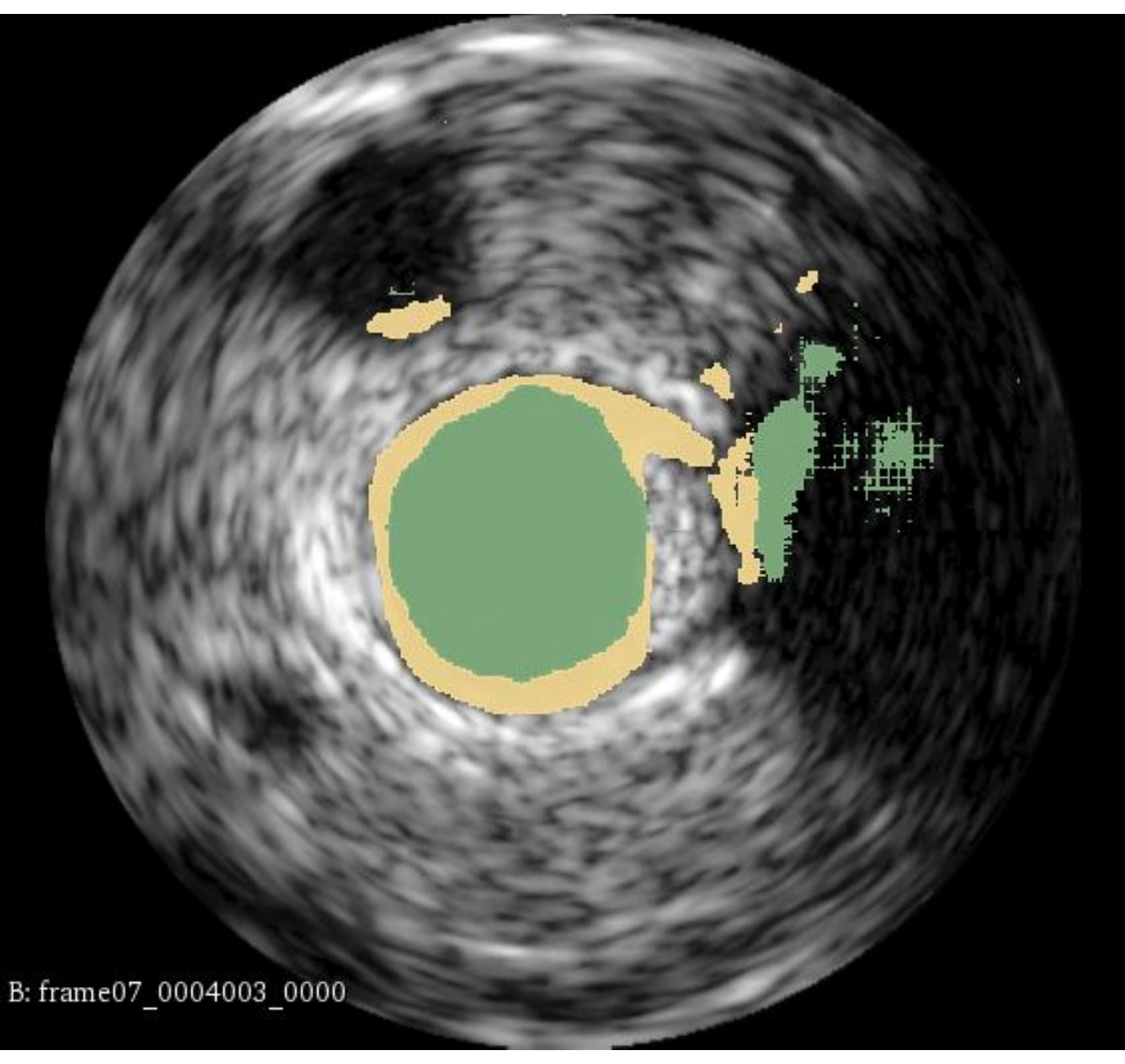}
     \caption{FCN}
  \end{subfigure}
    \begin{subfigure}{0.15\linewidth}
     \includegraphics[width=1\textwidth]{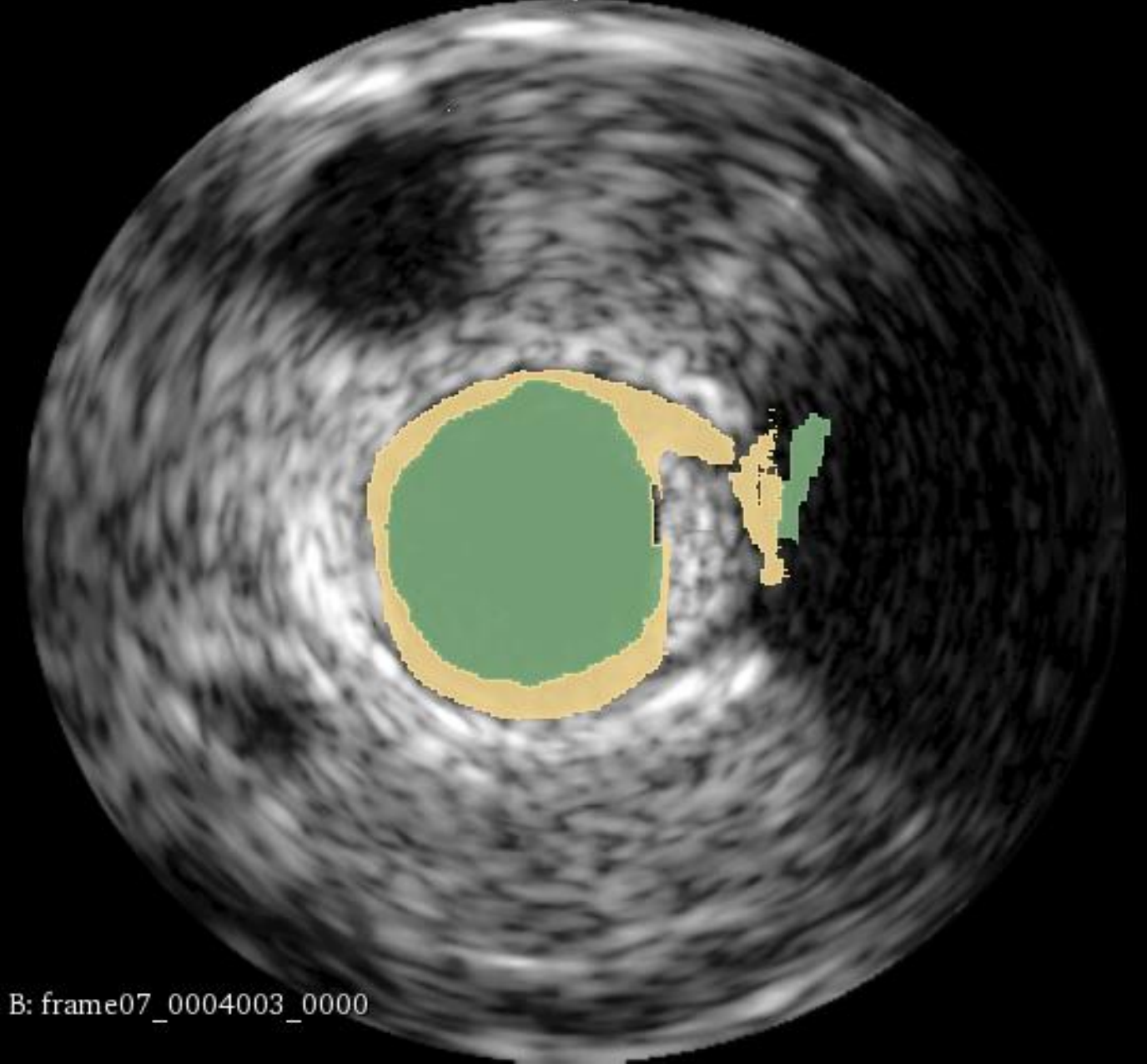}
     \caption{FCN+O}
  \end{subfigure}
    \begin{subfigure}{0.15\linewidth}
     \includegraphics[width=1\textwidth]{figures/ivus/sample7/nnunet-slice.pdf}
     \caption{nnUNet}
  \end{subfigure}

      \begin{subfigure}{0.15\linewidth}
     \includegraphics[width=1\textwidth]{figures/ivus/sample7/crf-slice.pdf}
     \caption{CRF}
  \end{subfigure}
      \begin{subfigure}{0.15\linewidth}
     \includegraphics[width=1\textwidth]{figures/ivus/sample7-rebut/midl-fin.pdf}
     \caption{MIDL}
  \end{subfigure}
    \begin{subfigure}{0.15\linewidth}
     \includegraphics[width=1\textwidth]{figures/ivus/sample7-rebut/nonadj-fin.pdf}
     \caption{NonAdj}
  \end{subfigure}
      \begin{subfigure}{0.15\linewidth}
     \includegraphics[width=1\textwidth]{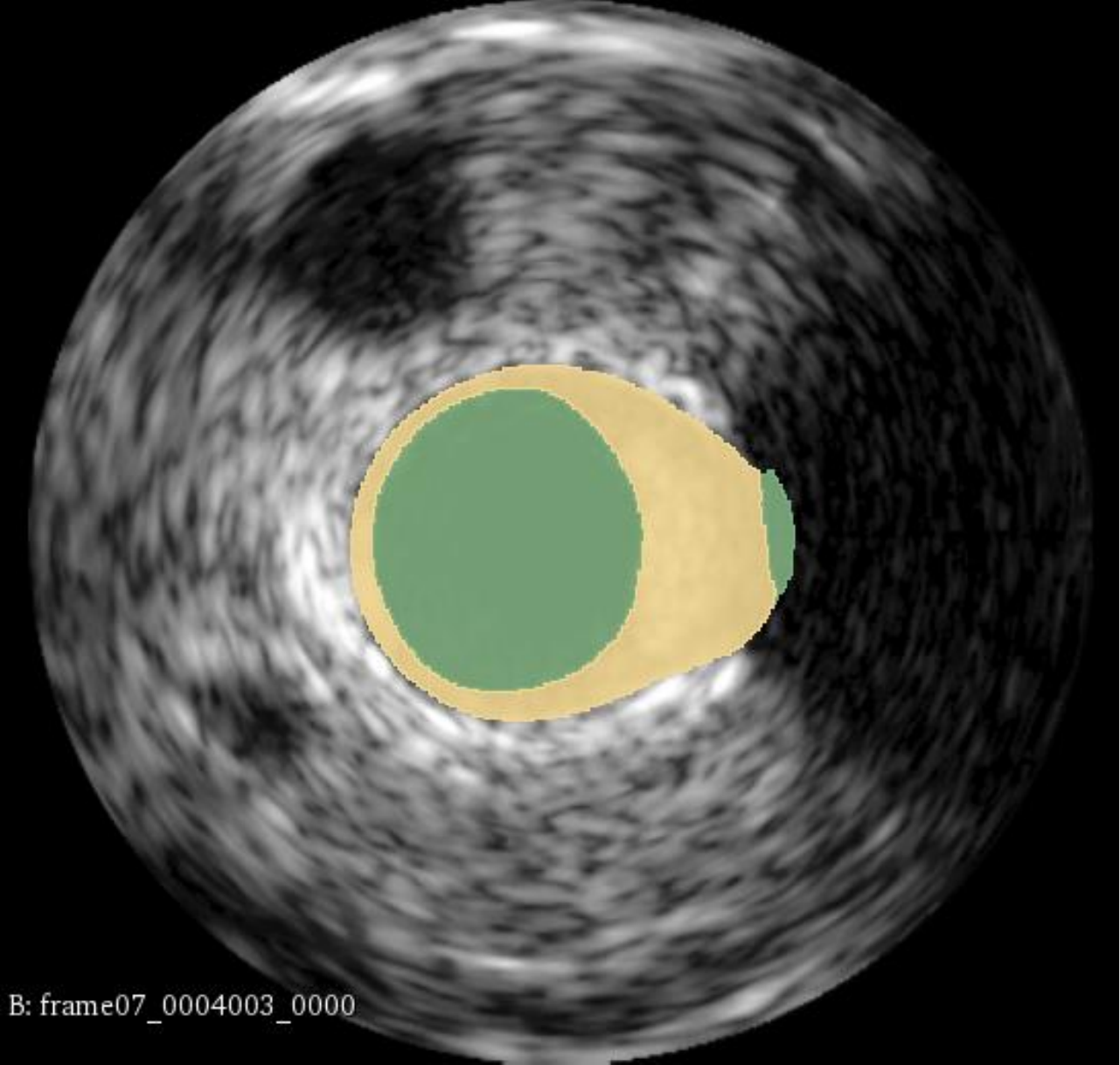}
     \caption{Ours4C}
  \end{subfigure}
      \begin{subfigure}{0.15\linewidth}
     \includegraphics[width=1\textwidth]{figures/ivus/sample7/topo-slice.pdf}
     \caption{Ours}
  \end{subfigure}
      \begin{subfigure}{0.15\linewidth}
     \includegraphics[width=1\textwidth]{figures/ivus/sample7/gt-slice.pdf}
     \caption{GT}
  \end{subfigure}

\caption{Qualitative IVUS results compared with the baselines. Colors for the classes correspond to the ones used in Fig.~\ref{fig:data-interactions}.}
\label{fig:ivus-add-1}
\end{figure}

\begin{figure}[t]
\centering 

  \begin{subfigure}{0.15\linewidth}
  \includegraphics[width=1\textwidth]{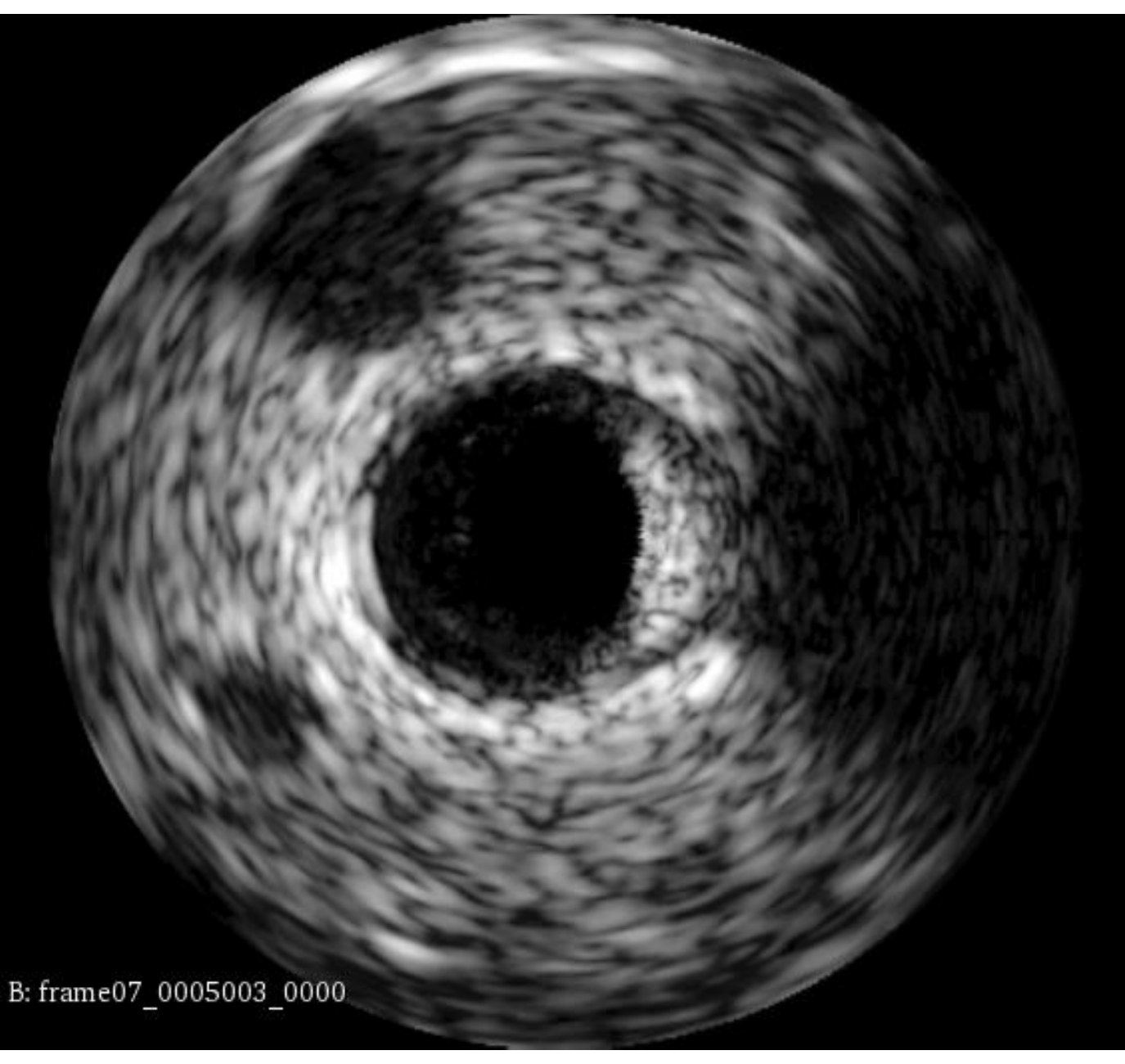}
  \caption{Input}
  \end{subfigure}
  \begin{subfigure}{0.15\linewidth}
     \includegraphics[width=1\textwidth]{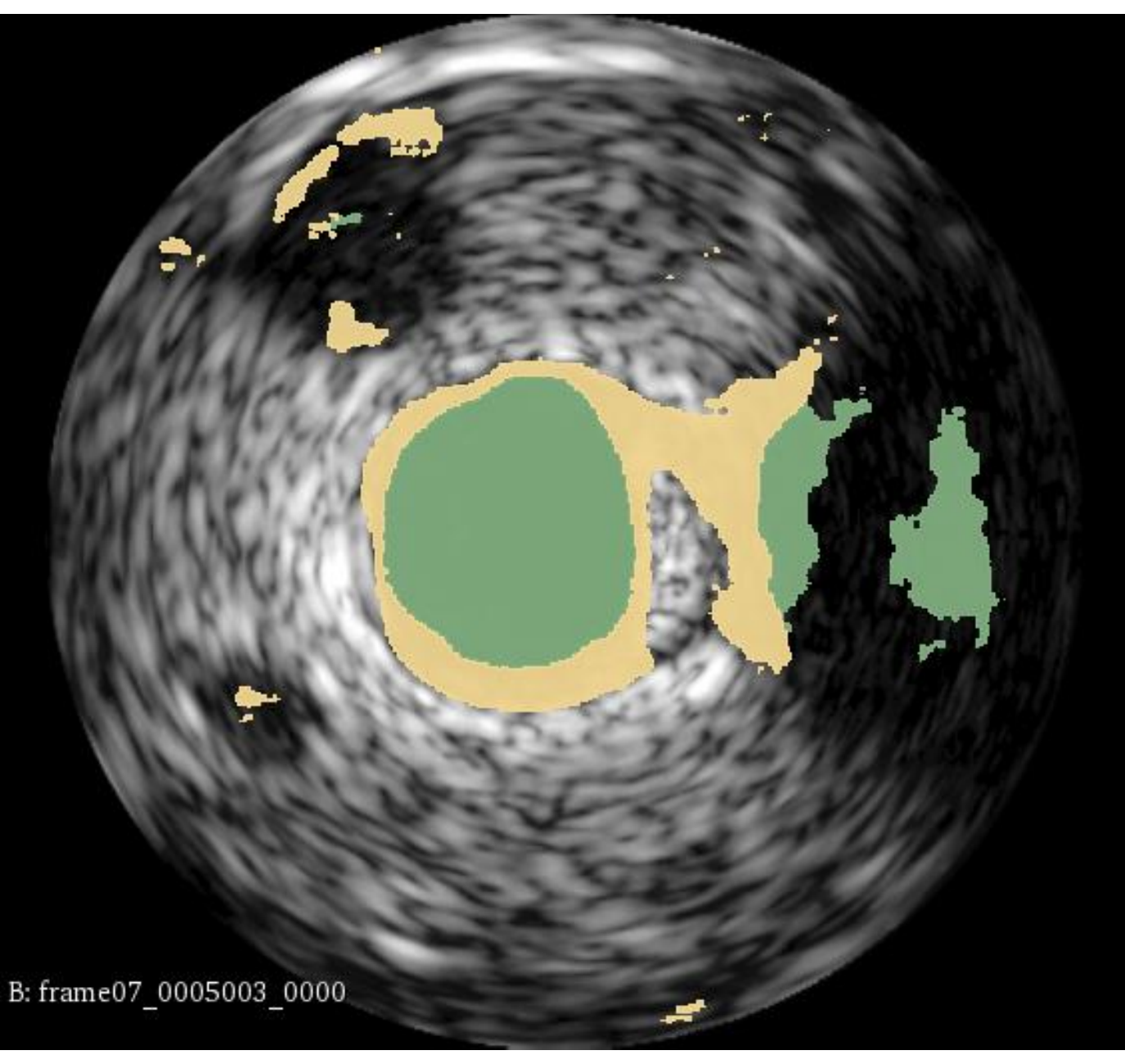}
     \caption{UNet}
  \end{subfigure}
    \begin{subfigure}{0.15\linewidth}
     \includegraphics[width=1\textwidth]{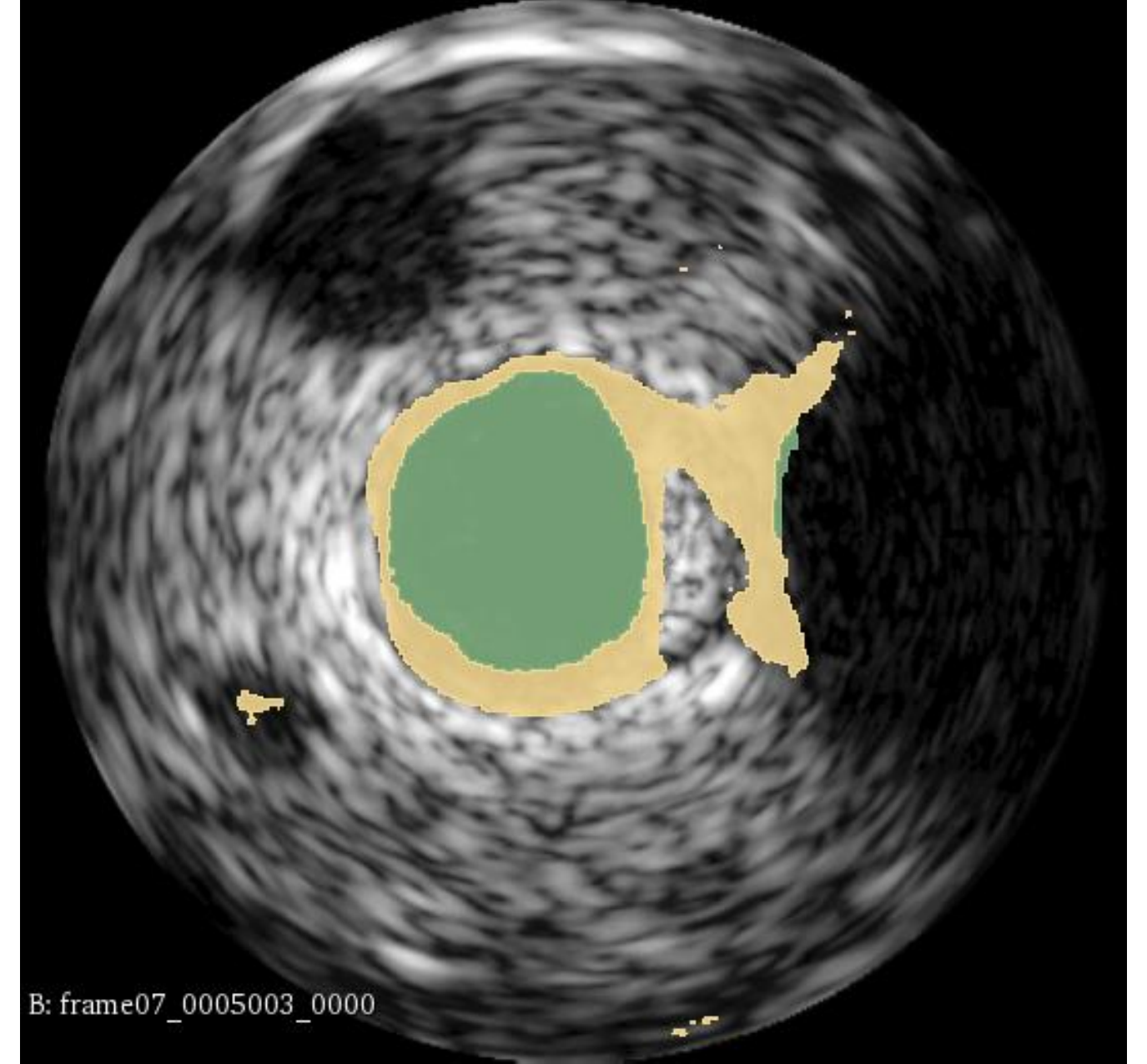}
     \caption{UNet+O}
  \end{subfigure}
    \begin{subfigure}{0.15\linewidth}
     \includegraphics[width=1\textwidth]{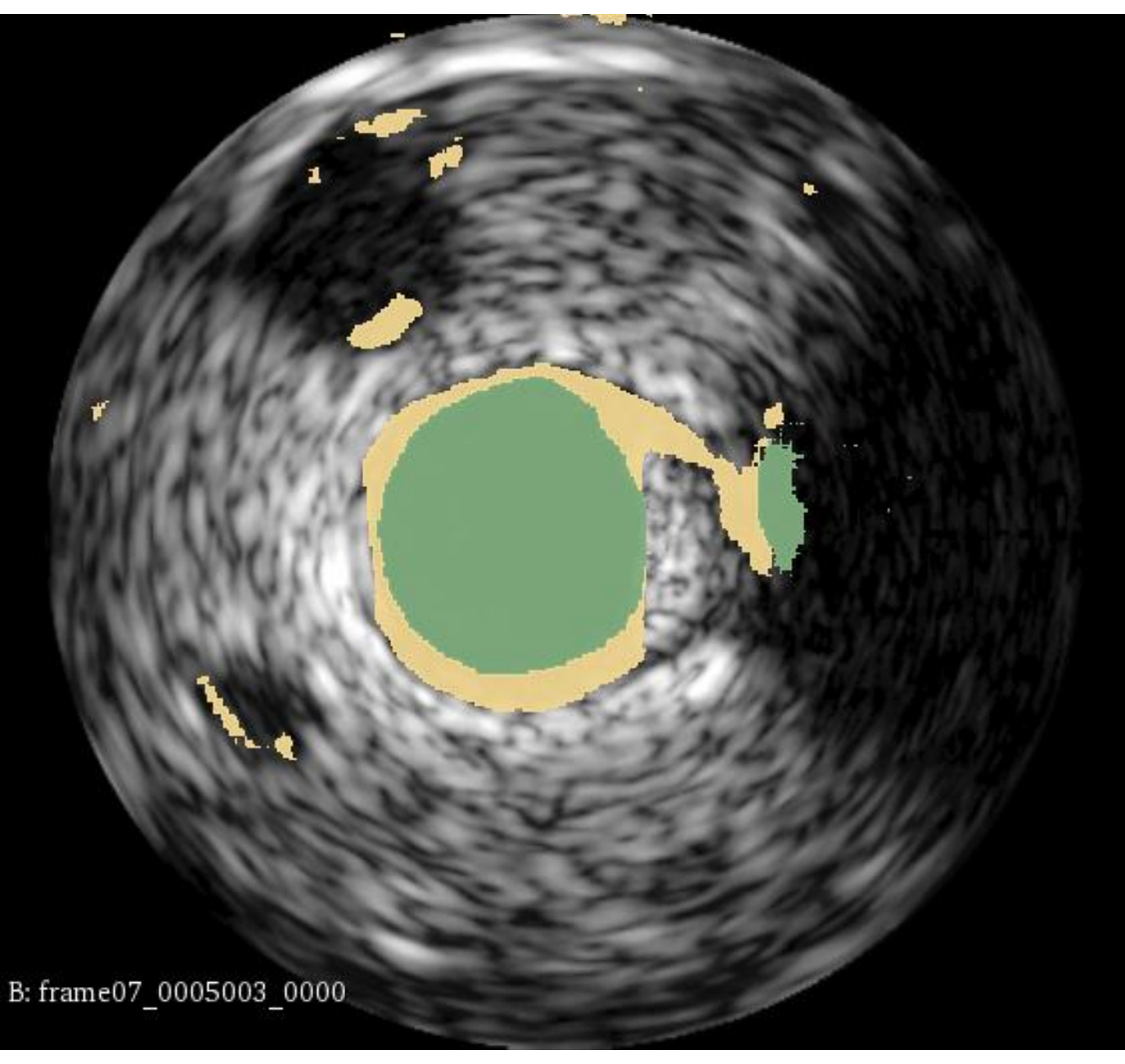}
     \caption{FCN}
  \end{subfigure}
      \begin{subfigure}{0.15\linewidth}
     \includegraphics[width=1\textwidth]{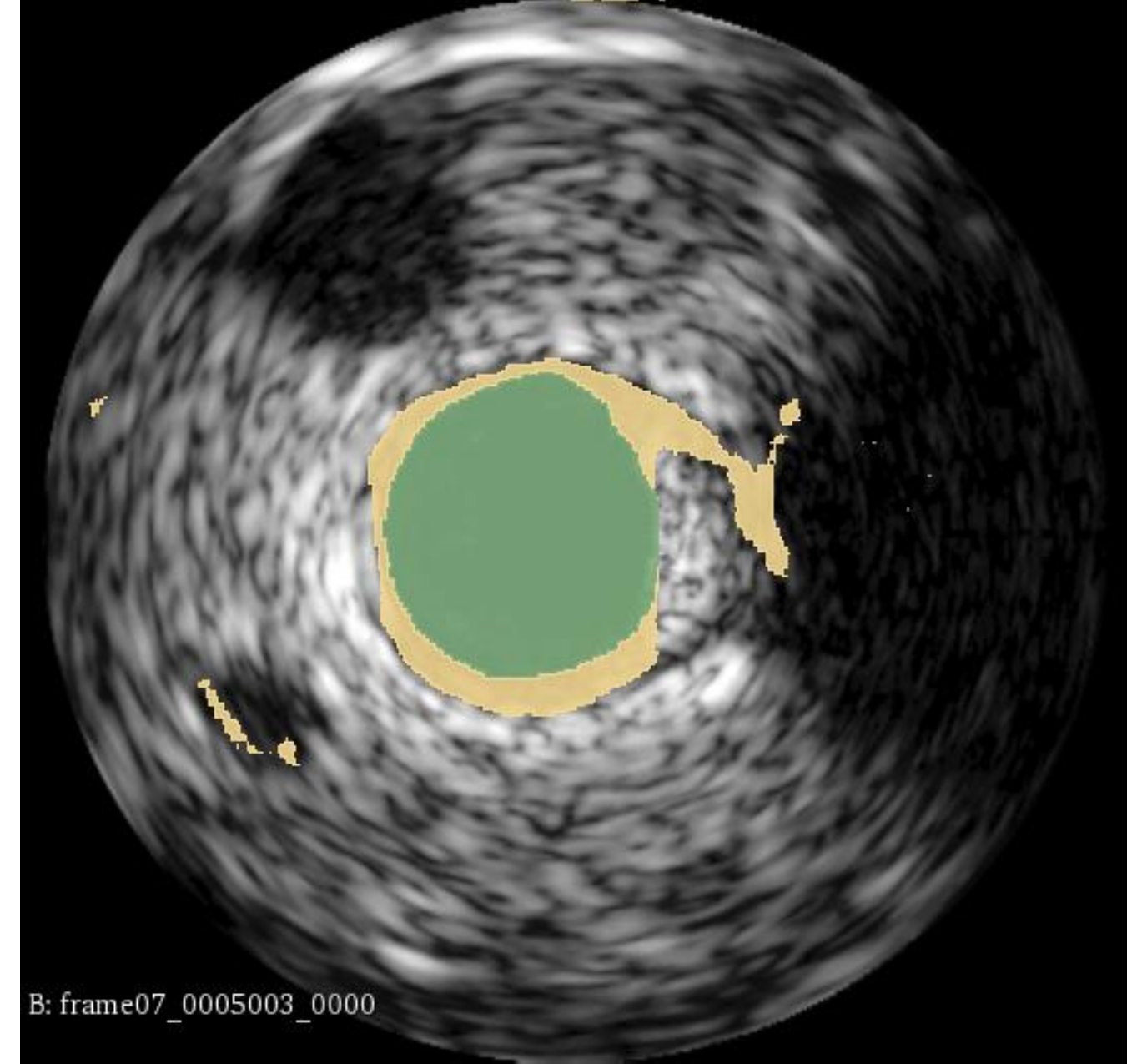}
     \caption{FCN+O}
  \end{subfigure}
    \begin{subfigure}{0.15\linewidth}
     \includegraphics[width=1\textwidth]{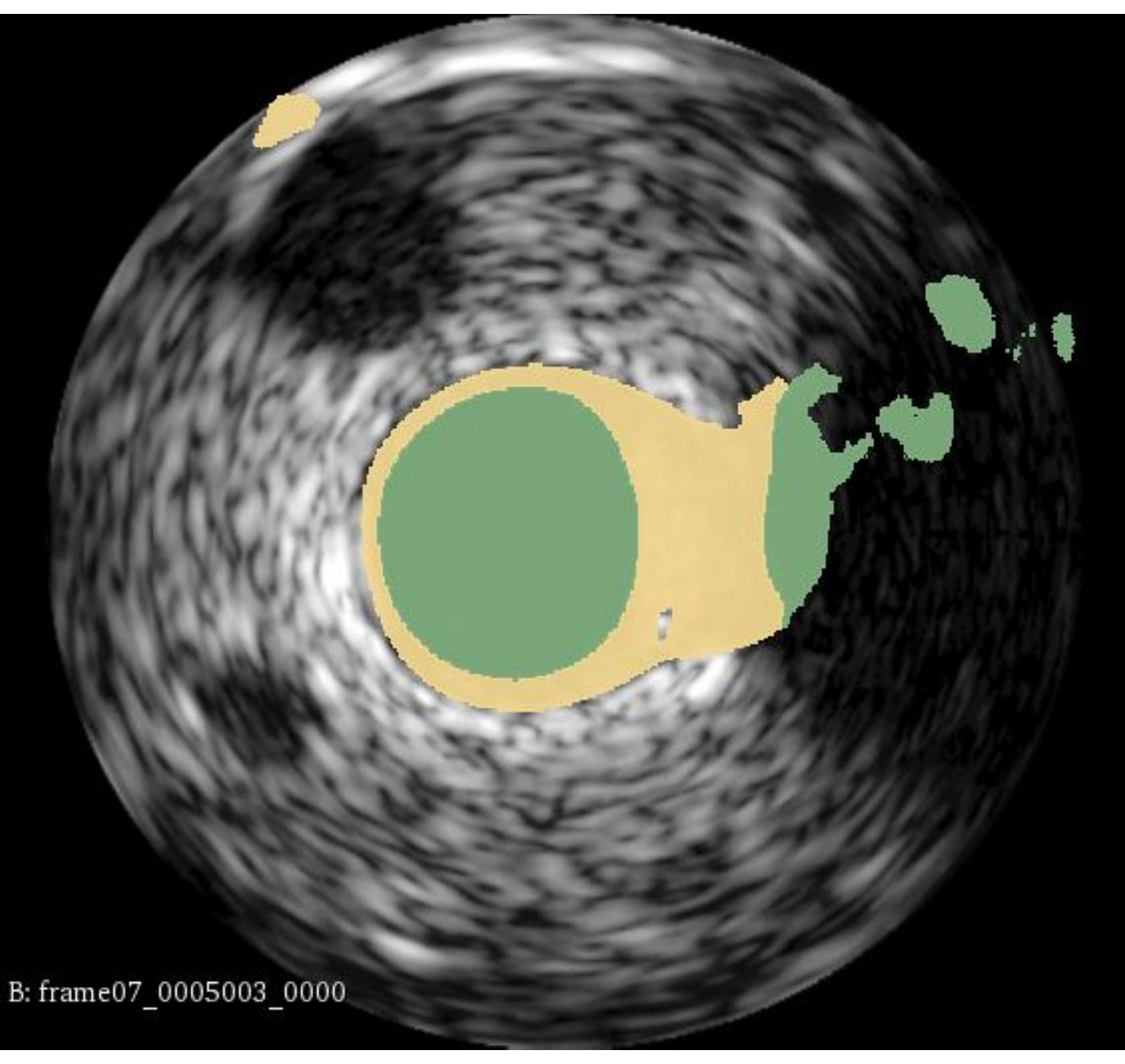}
     \caption{nnUNet}
  \end{subfigure}

      \begin{subfigure}{0.15\linewidth}
     \includegraphics[width=1\textwidth]{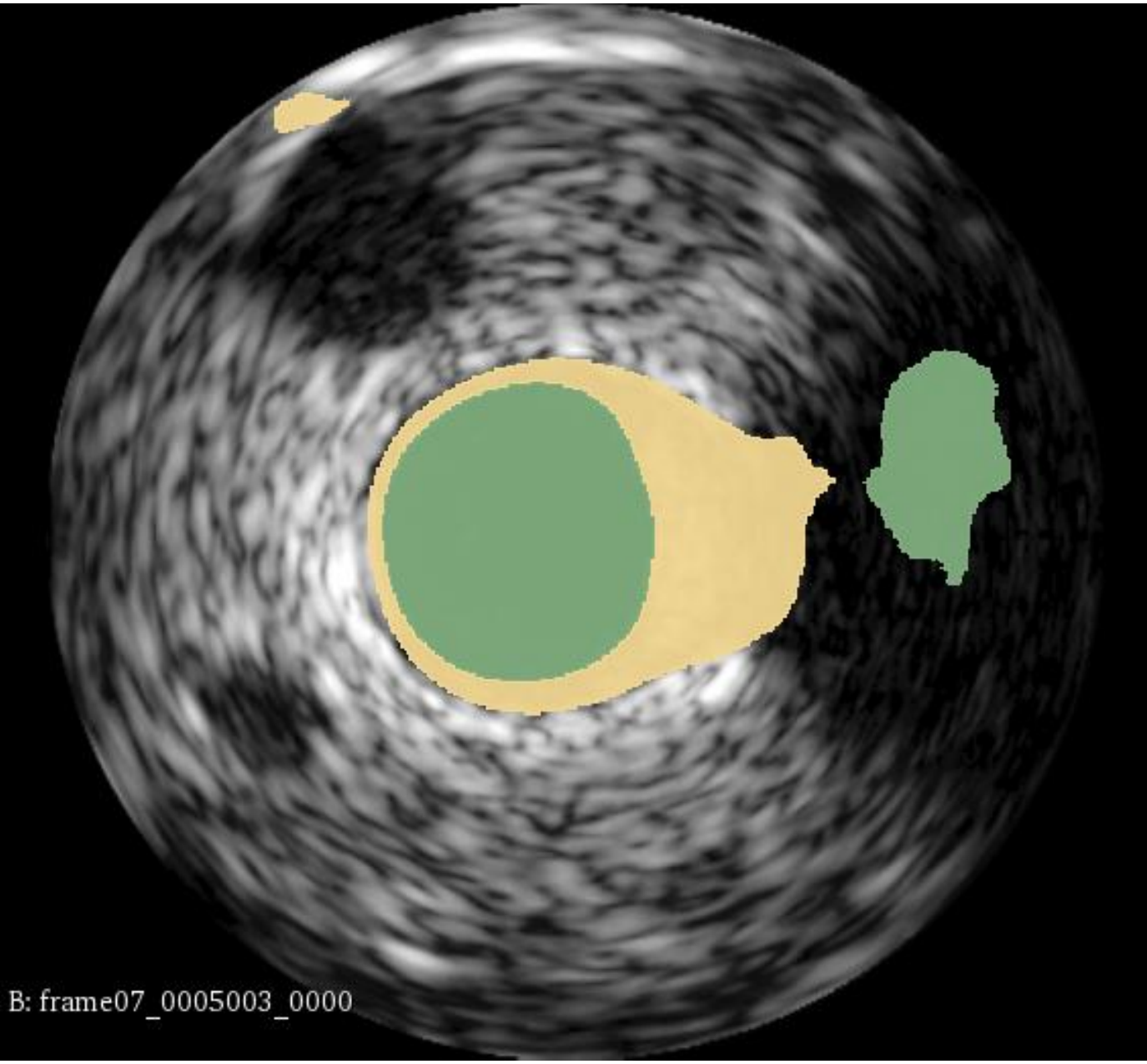}
     \caption{CRF}
  \end{subfigure}
    \begin{subfigure}{0.15\linewidth}
     \includegraphics[width=1\textwidth]{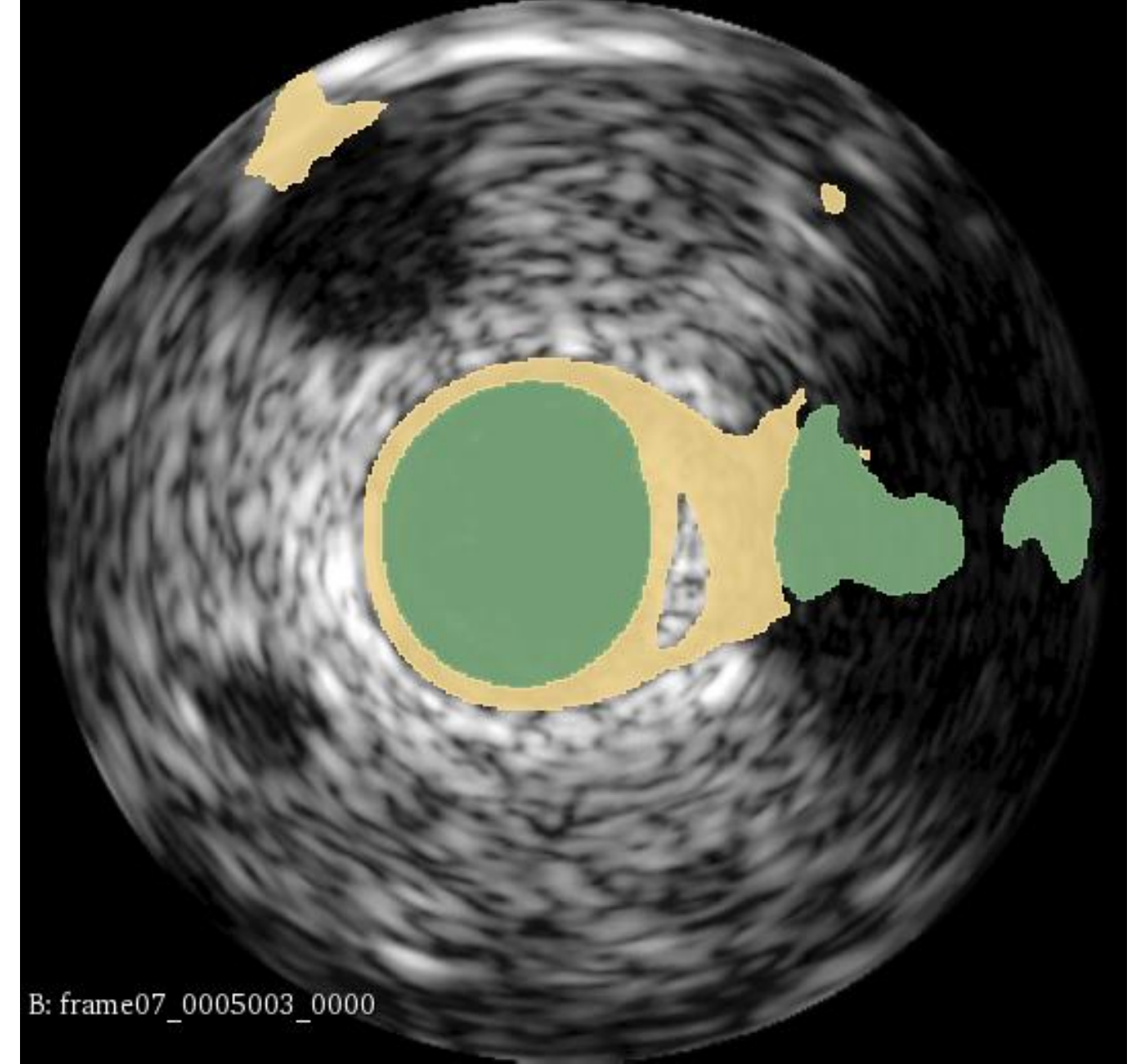}
     \caption{MIDL}
  \end{subfigure}
      \begin{subfigure}{0.15\linewidth}
     \includegraphics[width=1\textwidth]{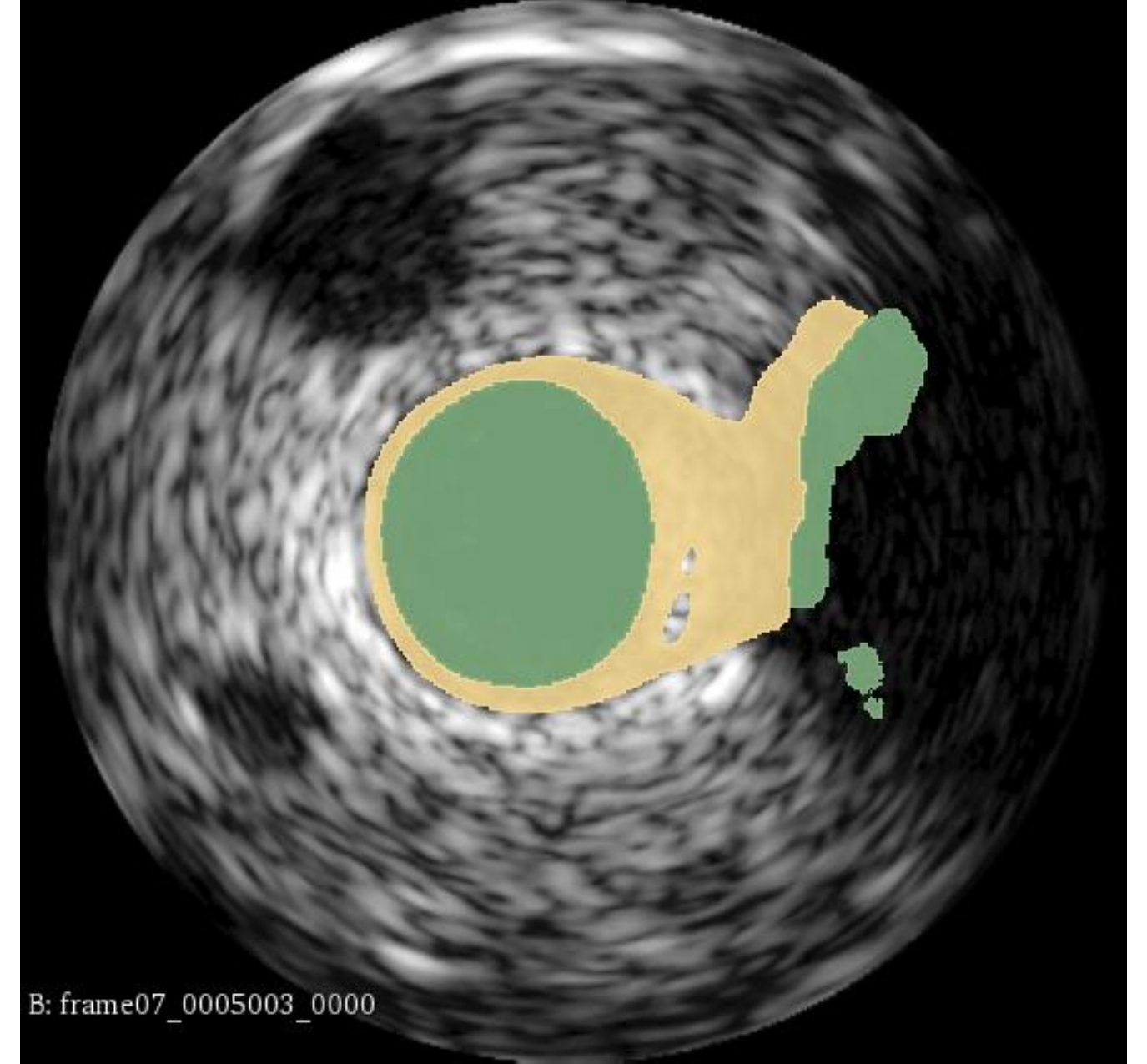}
     \caption{NonAdj}
  \end{subfigure}
      \begin{subfigure}{0.15\linewidth}
     \includegraphics[width=1\textwidth]{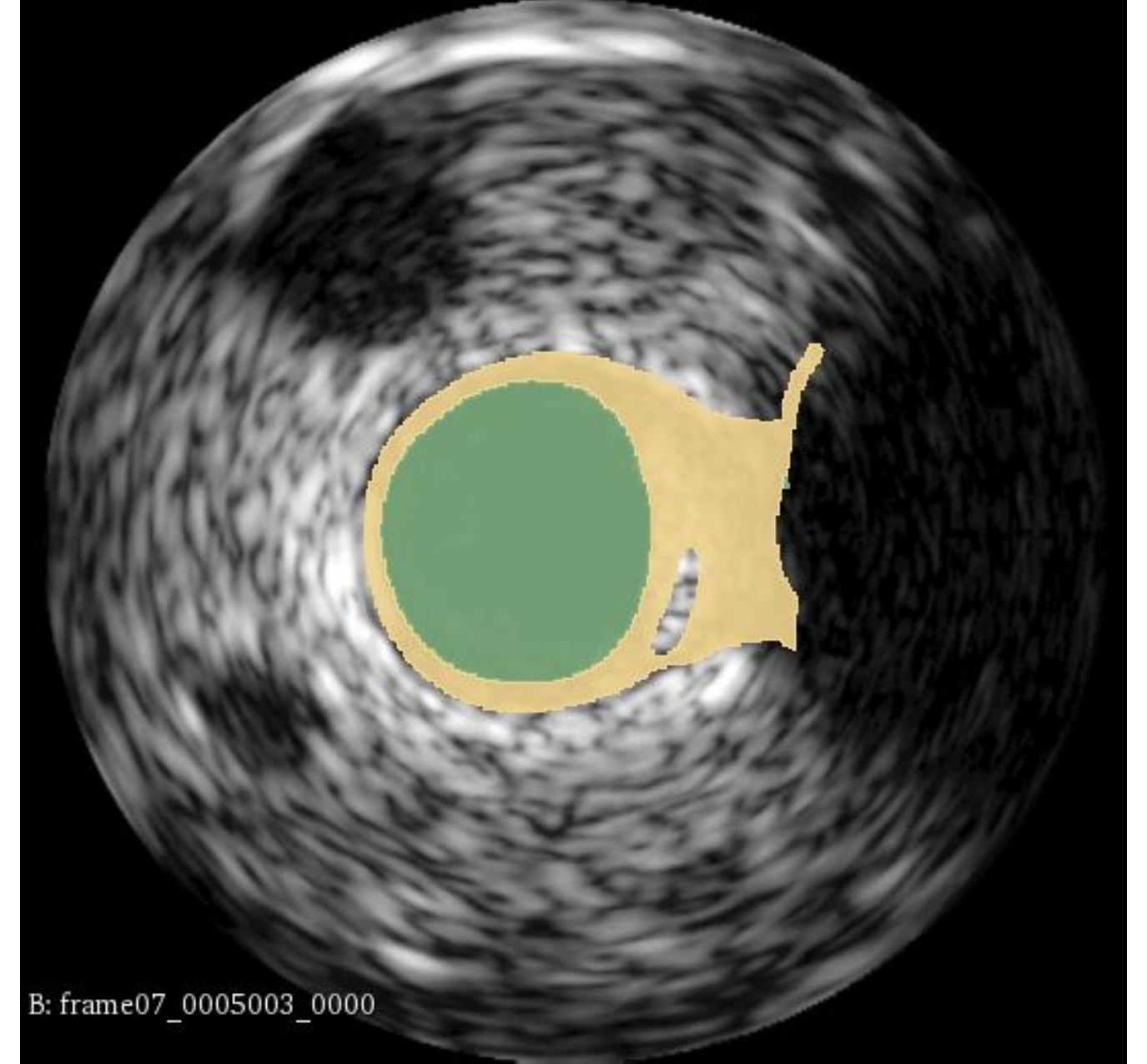}
     \caption{Ours4C}
  \end{subfigure}
        \begin{subfigure}{0.15\linewidth}
     \includegraphics[width=1\textwidth]{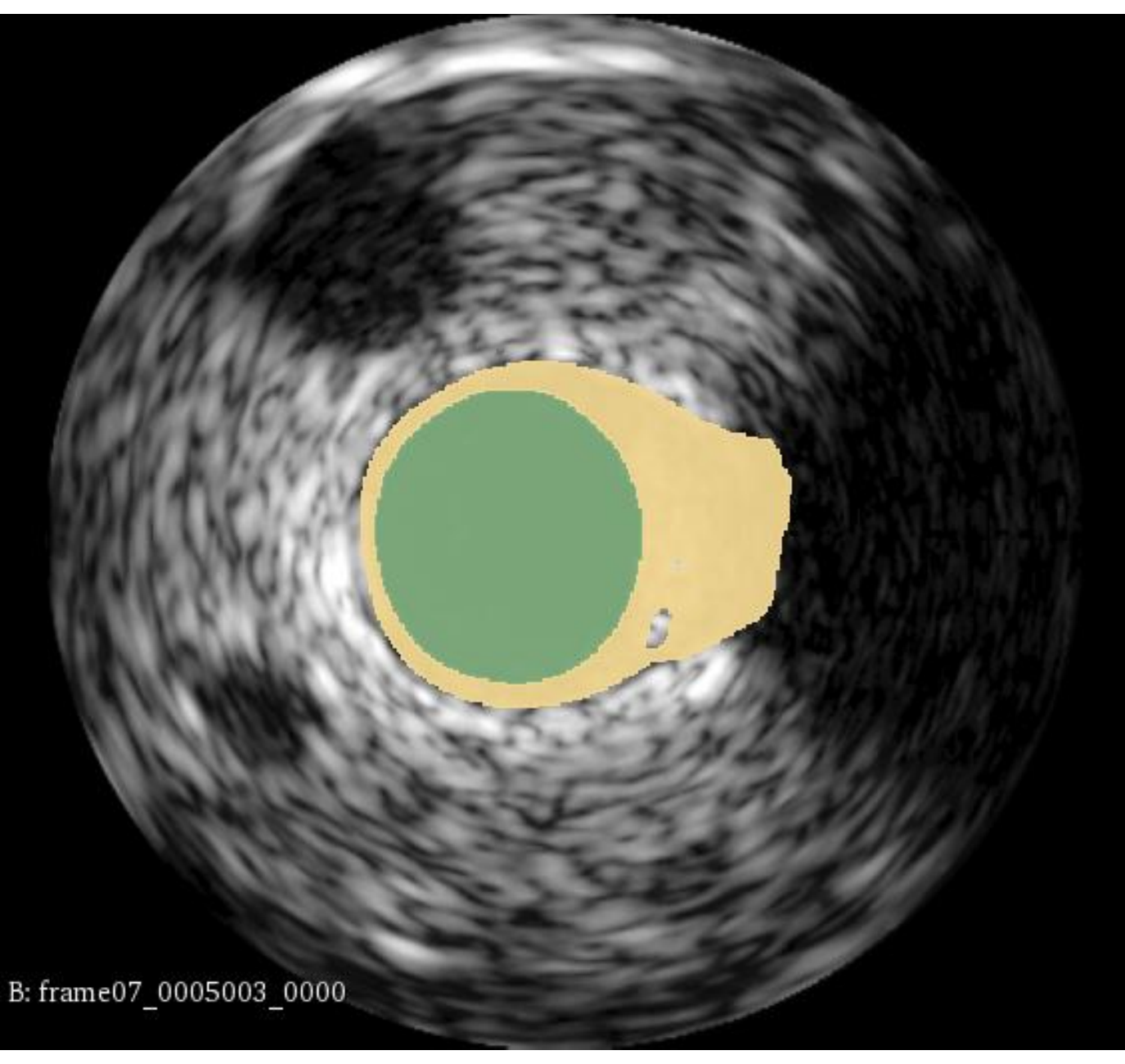}
     \caption{Ours}
  \end{subfigure}
      \begin{subfigure}{0.15\linewidth}
     \includegraphics[width=1\textwidth]{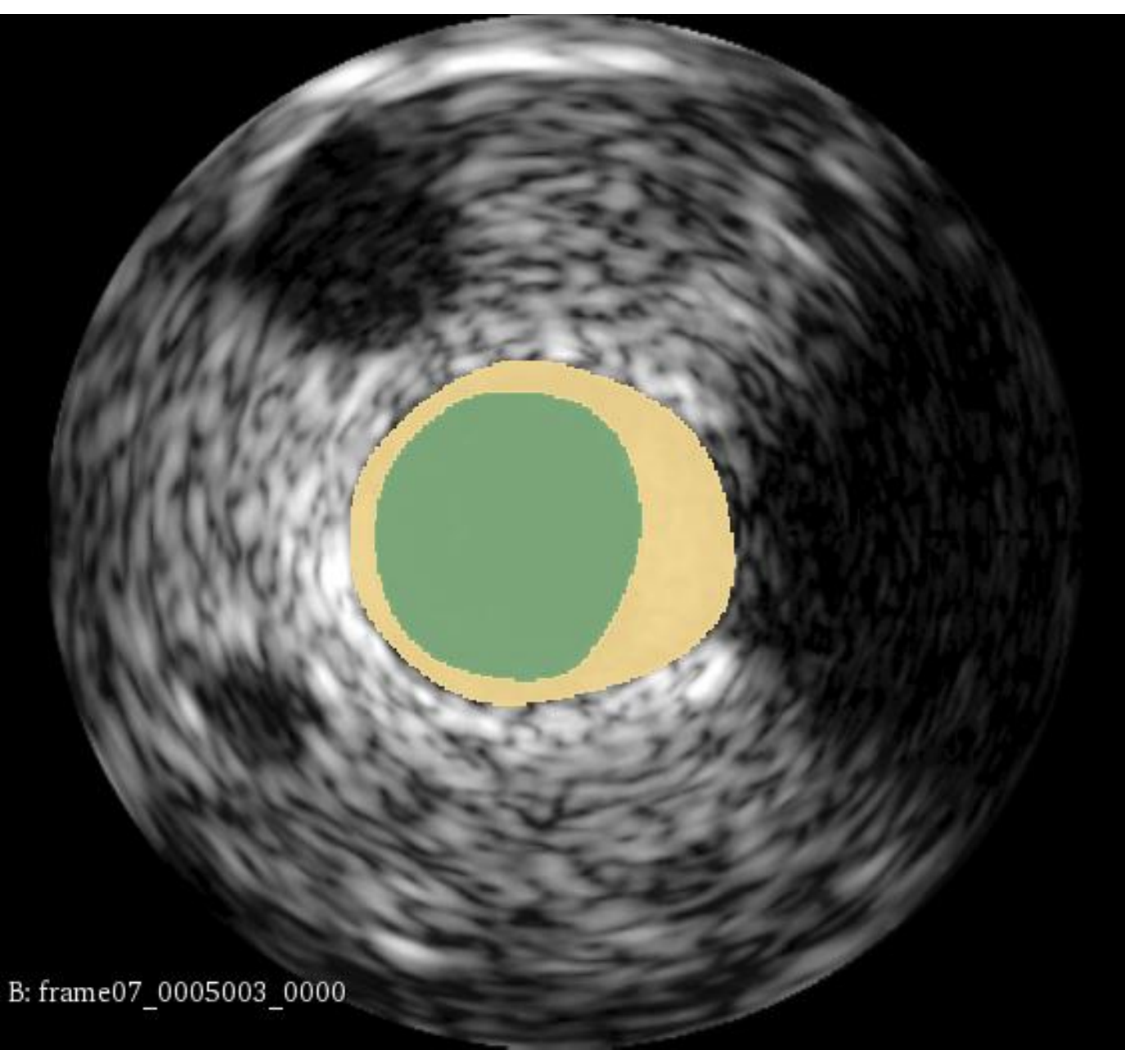}
     \caption{GT}
  \end{subfigure}
  
  \caption{Additional qualitative IVUS results compared with the baselines. Colors for the classes correspond to the ones used in Fig.~\ref{fig:data-interactions}.}
\label{fig:ivus-add-2}
\end{figure}

\begin{figure}[t]
\centering 

    \begin{subfigure}{0.14\linewidth}
  \includegraphics[width=1\textwidth]{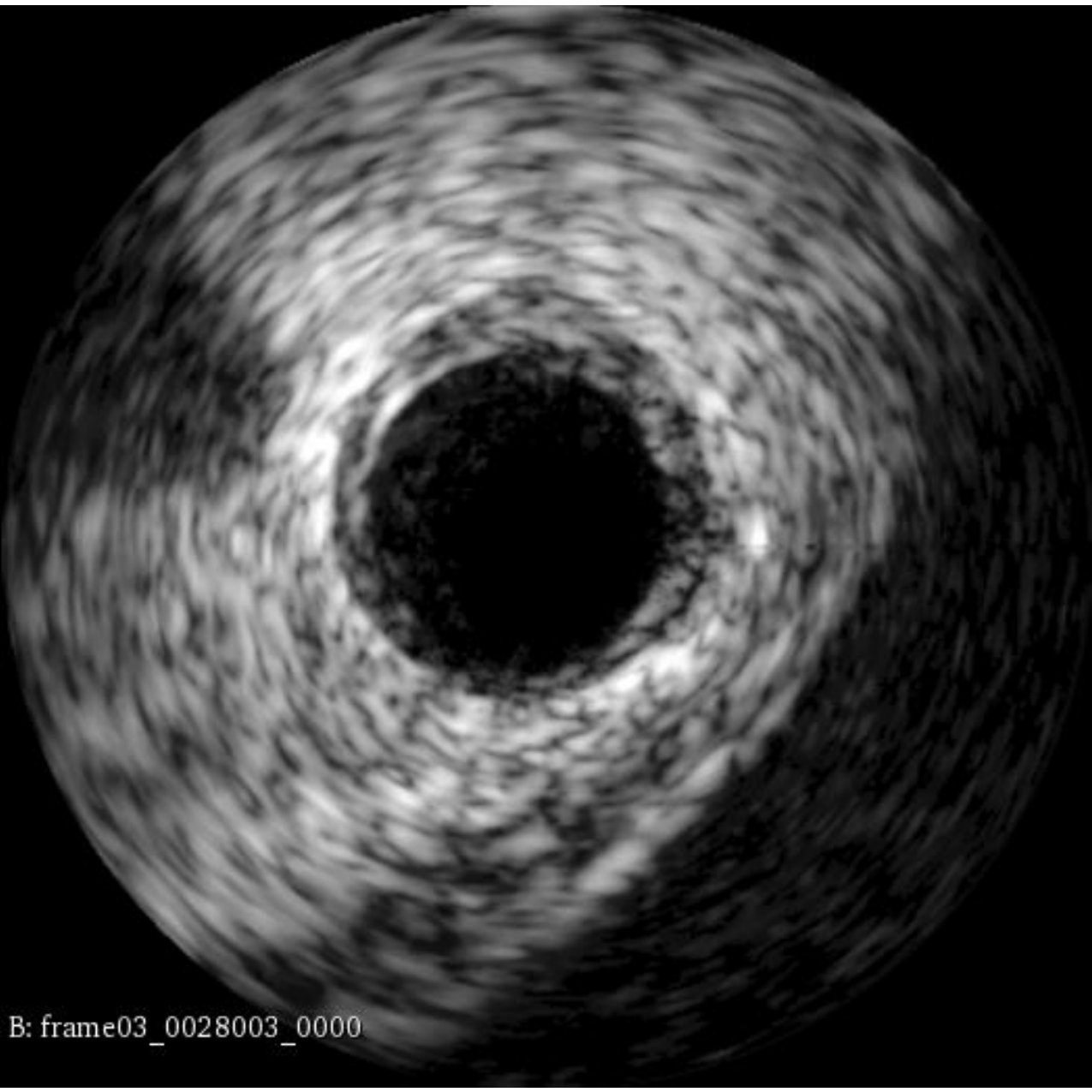}
  \caption{Input}
  \end{subfigure}
  \begin{subfigure}{0.14\linewidth}
     \includegraphics[width=1\textwidth]{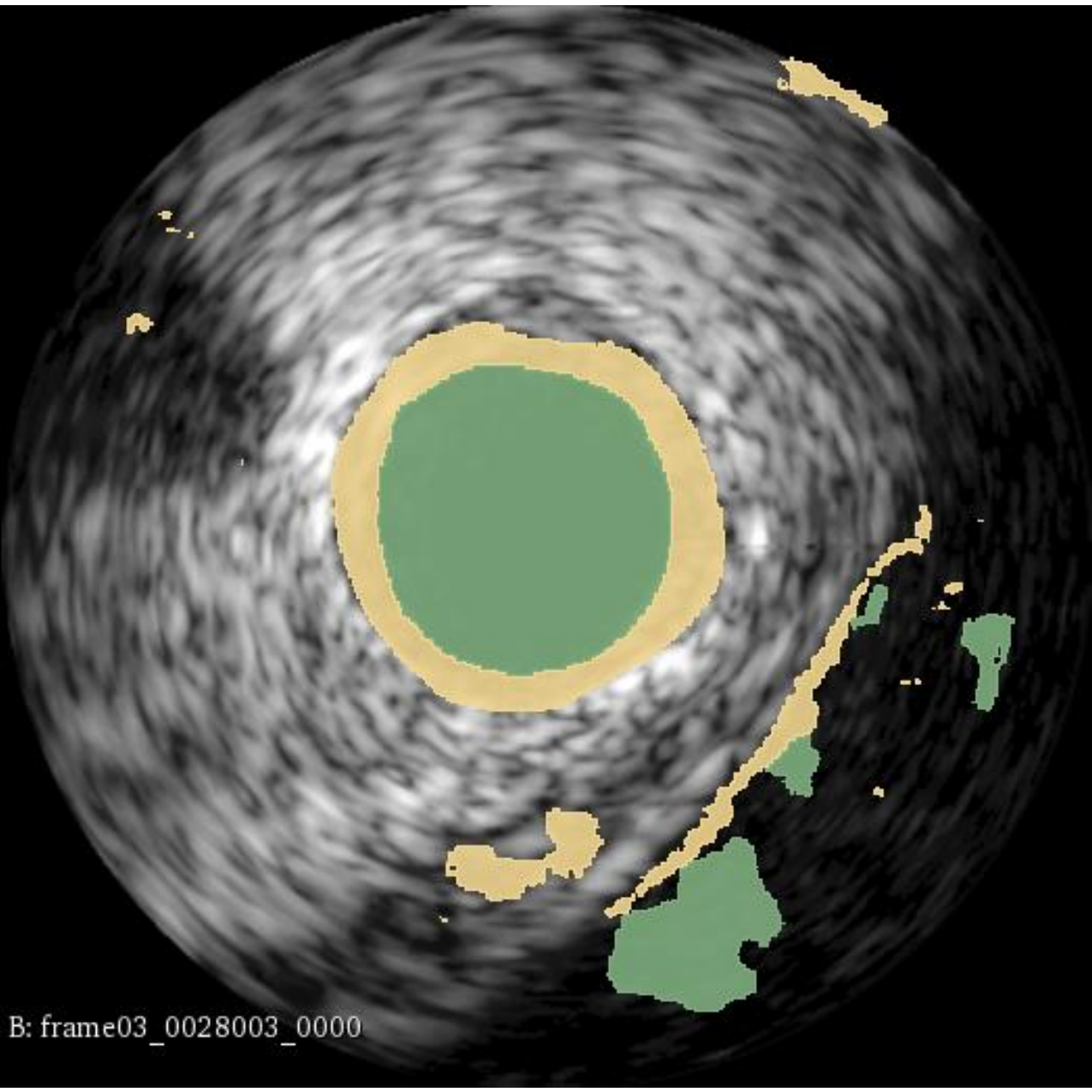}
     \caption{UNet}
  \end{subfigure}
     \begin{subfigure}{0.15\linewidth}
     \includegraphics[width=1\textwidth]{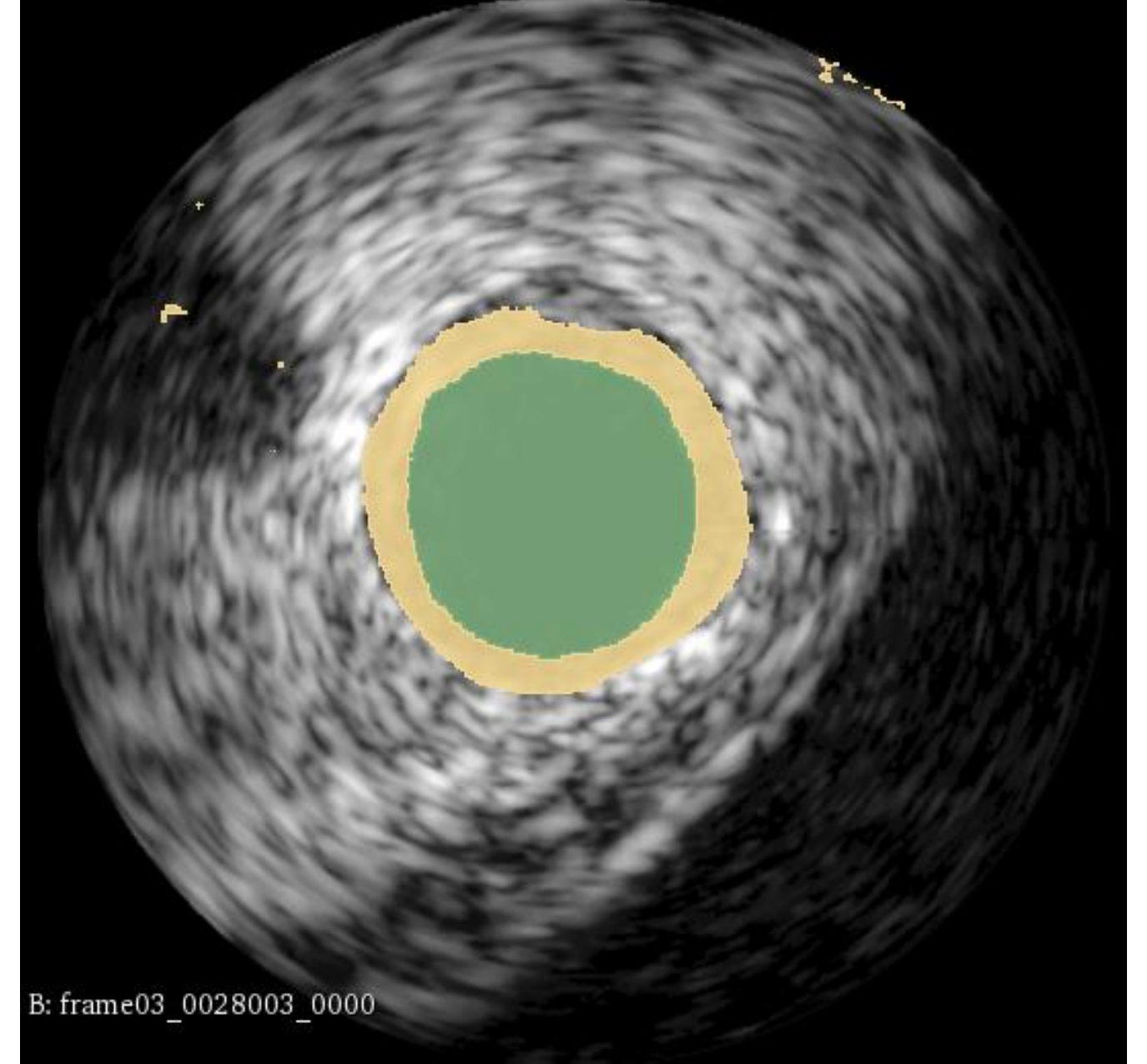}
     \caption{UNet+O}
  \end{subfigure}
    \begin{subfigure}{0.14\linewidth}
     \includegraphics[width=1\textwidth]{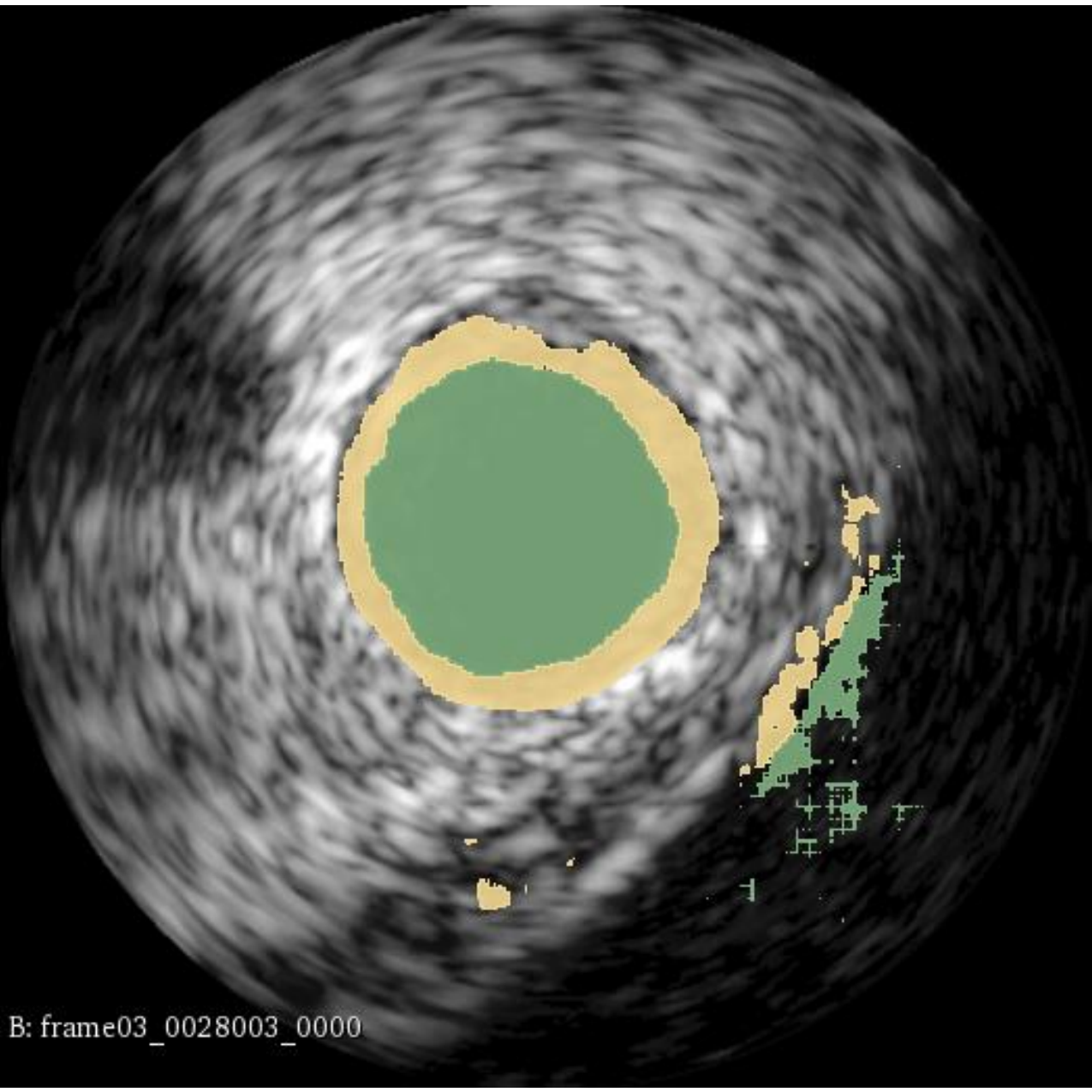}
     \caption{FCN}
  \end{subfigure}
      \begin{subfigure}{0.15\linewidth}
     \includegraphics[width=1\textwidth]{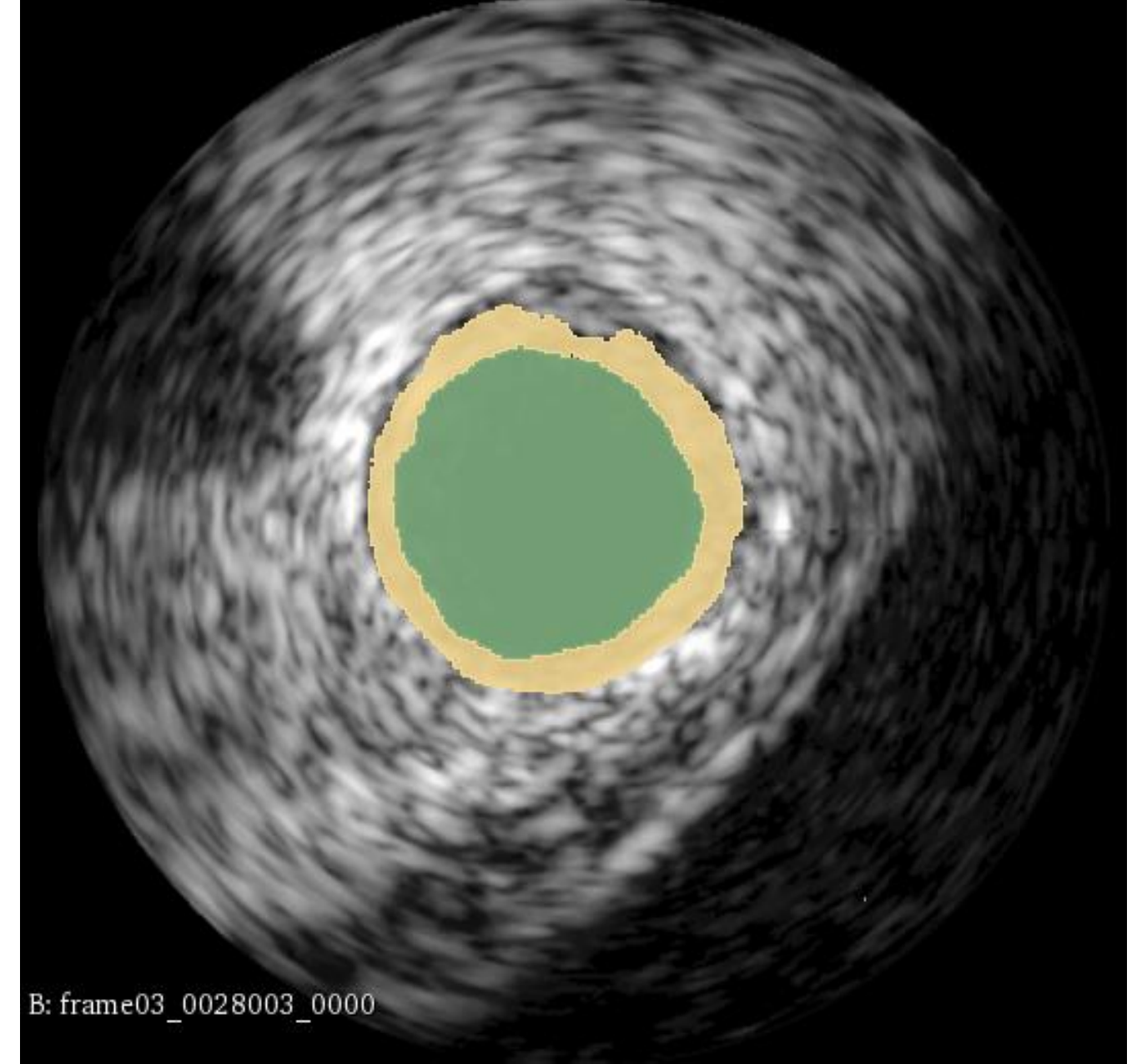}
     \caption{FCN+O}
  \end{subfigure}
    \begin{subfigure}{0.14\linewidth}
     \includegraphics[width=1\textwidth]{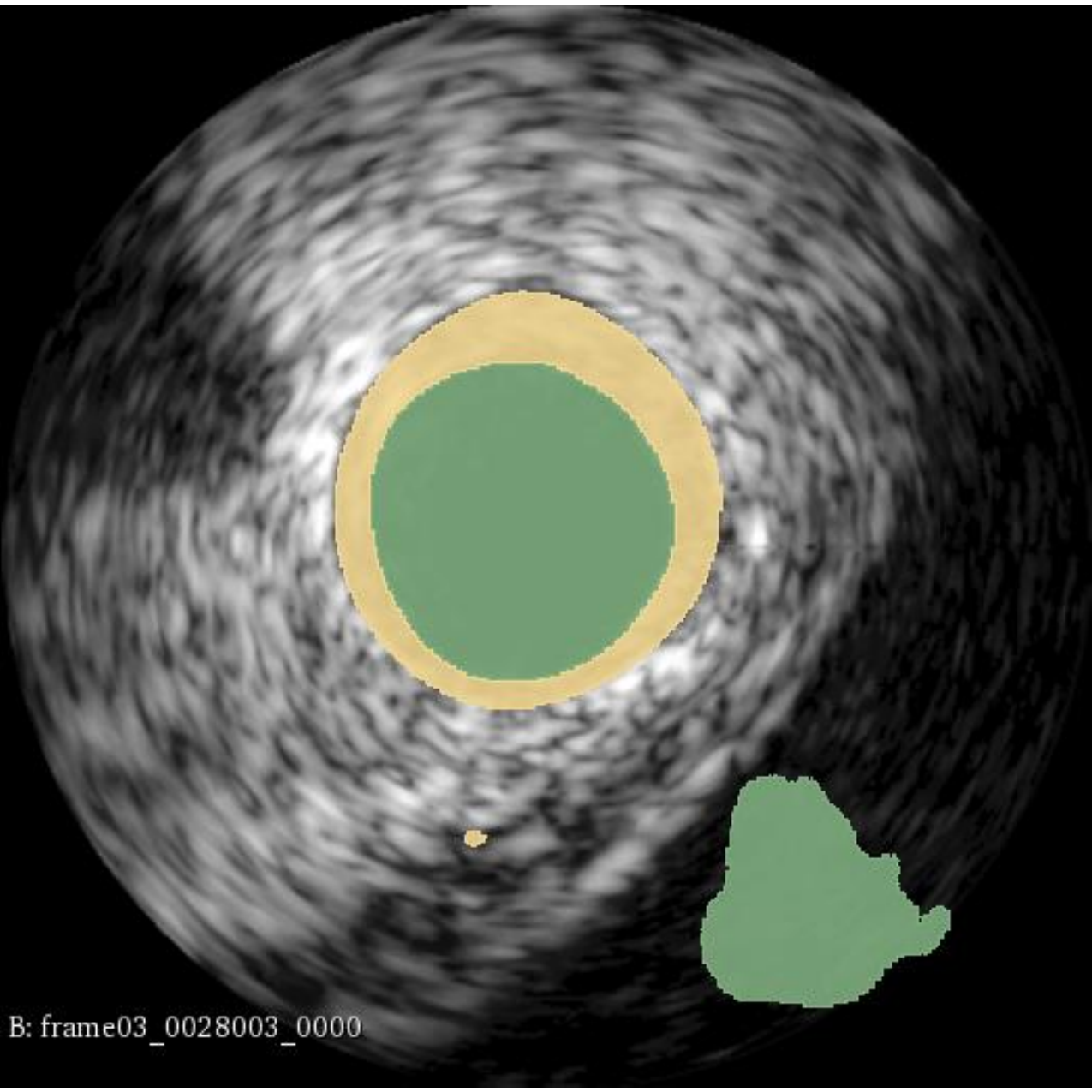}
     \caption{nnUNet}
  \end{subfigure}

      \begin{subfigure}{0.14\linewidth}
     \includegraphics[width=1\textwidth]{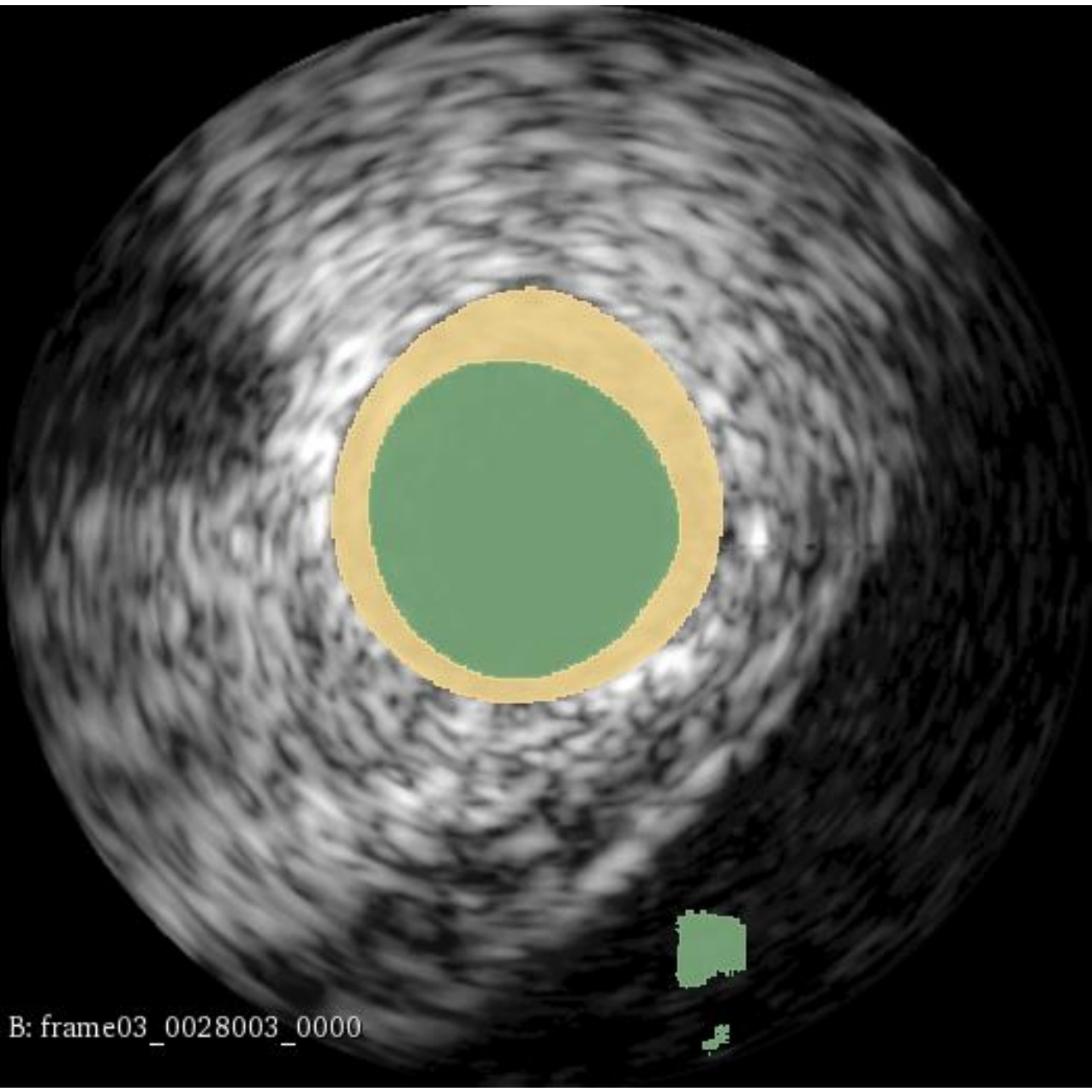}
     \caption{CRF}
  \end{subfigure}
      \begin{subfigure}{0.15\linewidth}
     \includegraphics[width=1\textwidth]{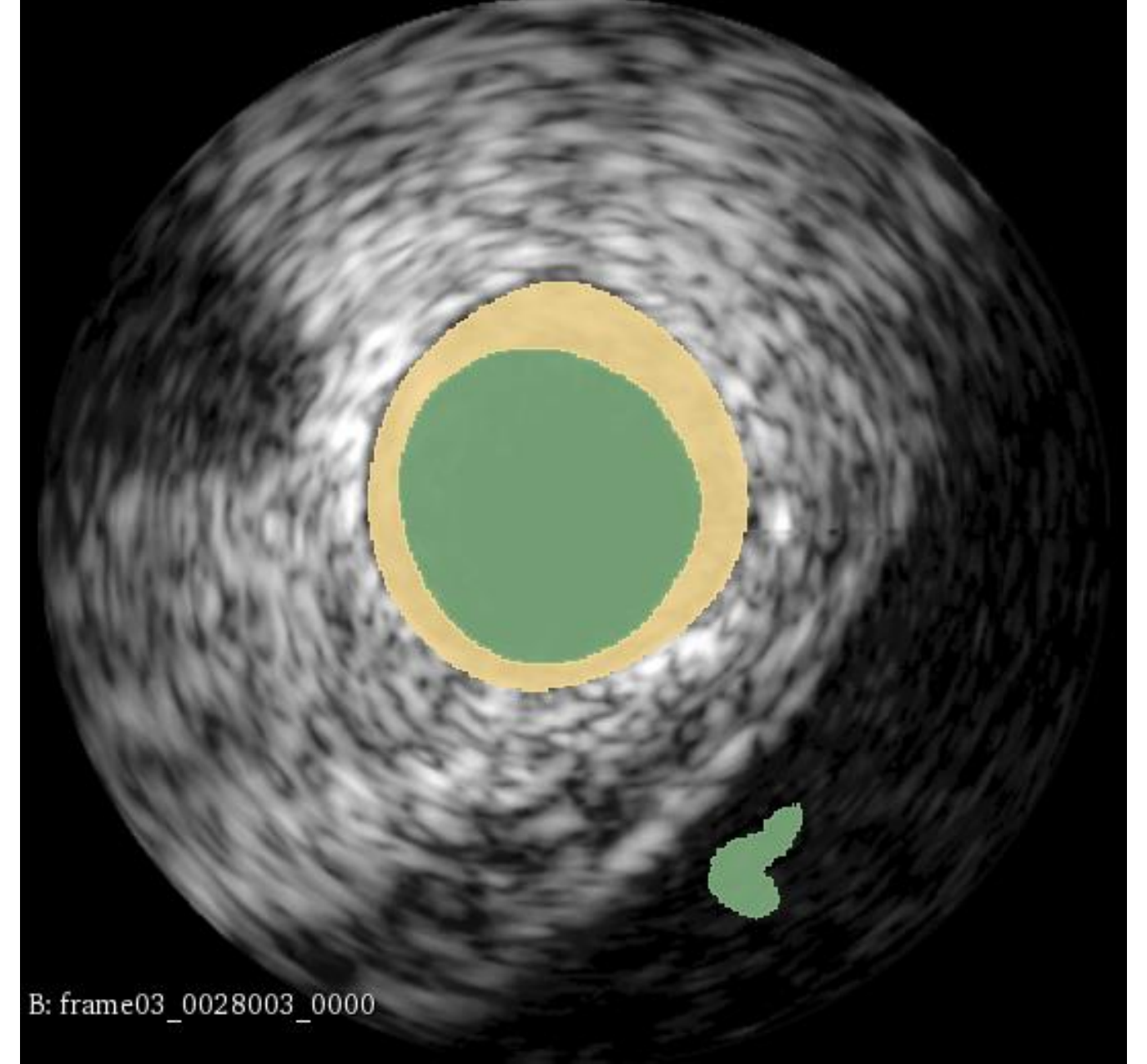}
     \caption{MIDL}
  \end{subfigure}
      \begin{subfigure}{0.15\linewidth}
     \includegraphics[width=1\textwidth]{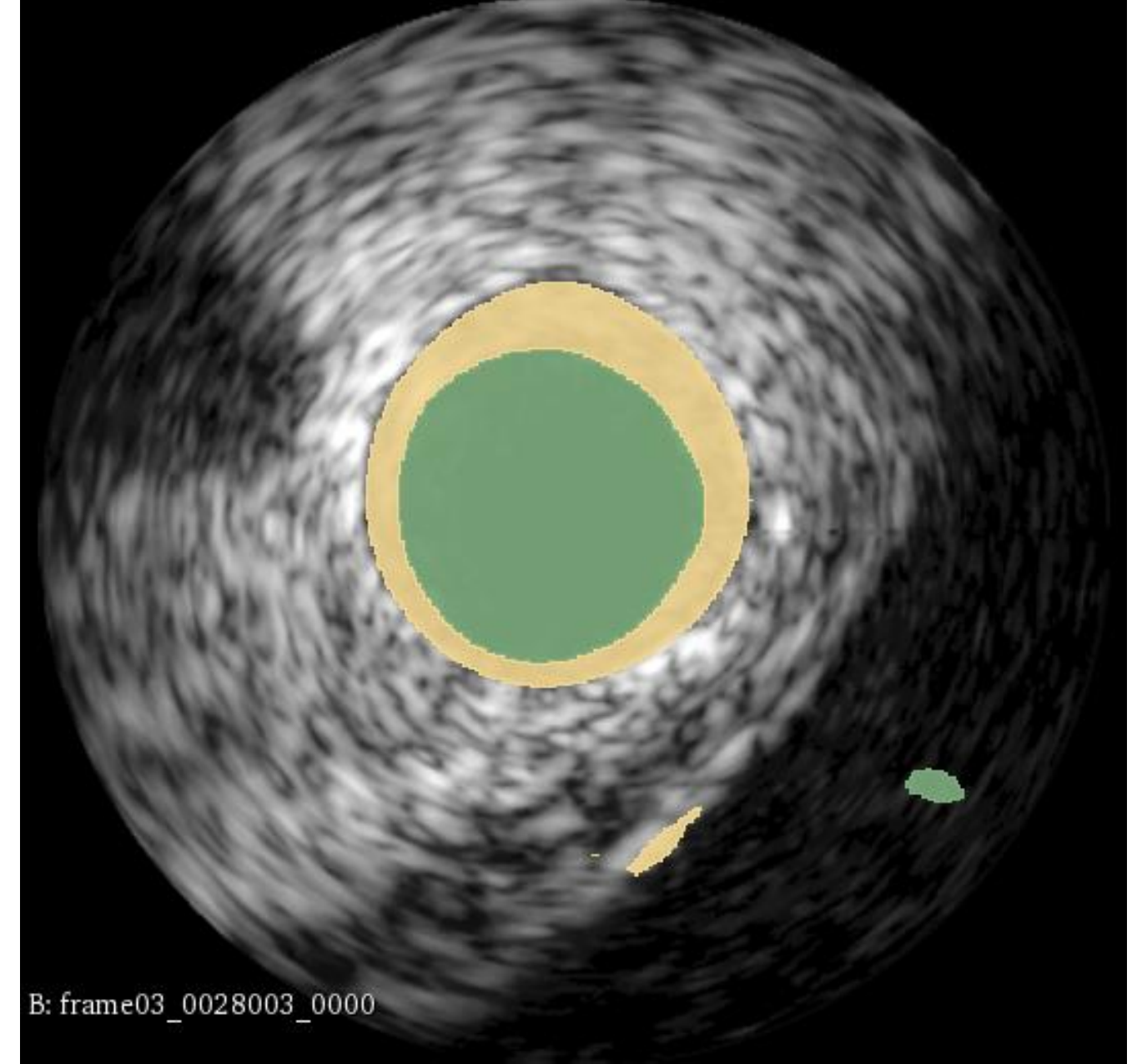}
     \caption{NonAdj}
  \end{subfigure} 
        \begin{subfigure}{0.15\linewidth}
     \includegraphics[width=1\textwidth]{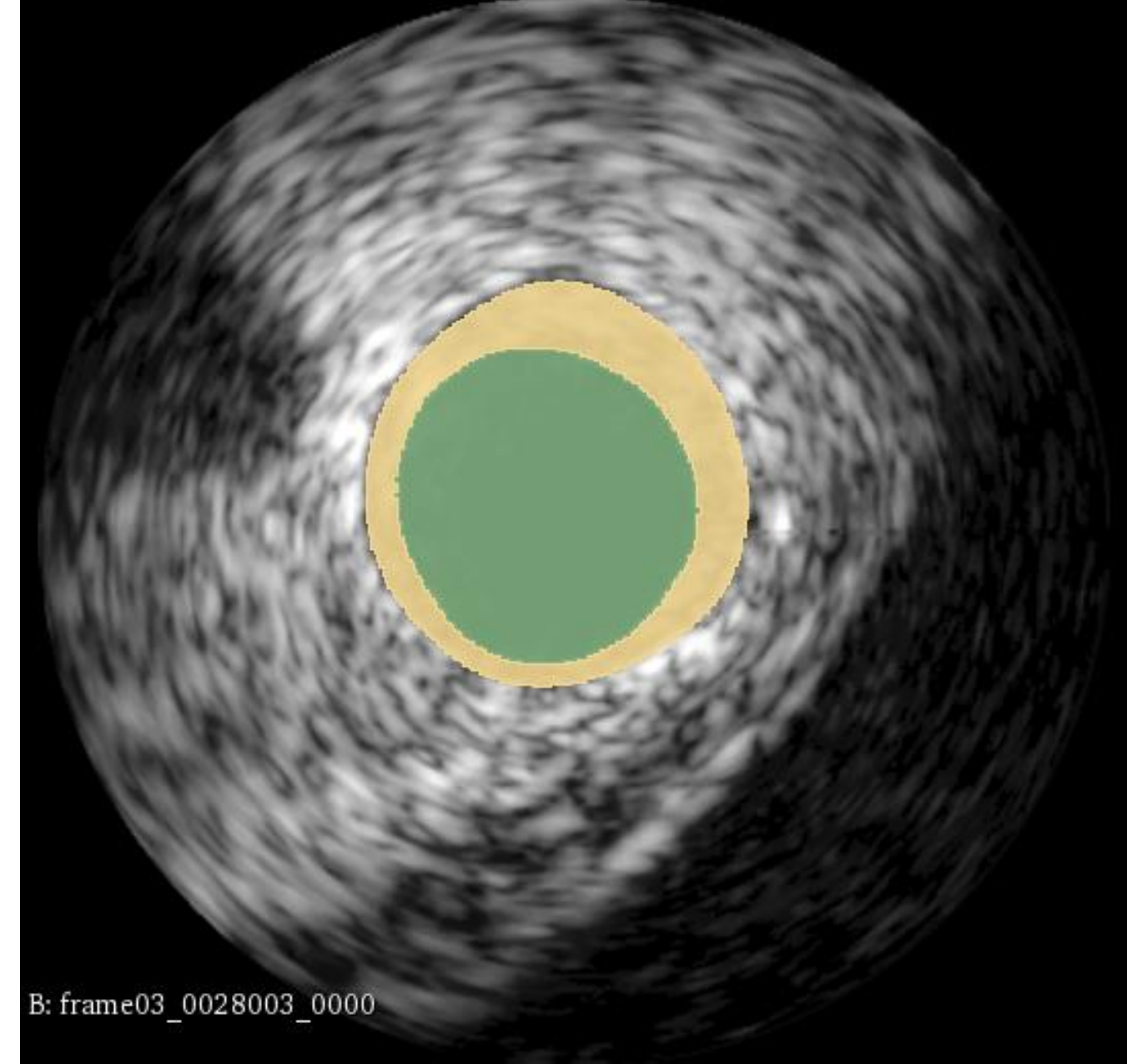}
     \caption{Ous4C}
  \end{subfigure} 
        \begin{subfigure}{0.14\linewidth}
     \includegraphics[width=1\textwidth]{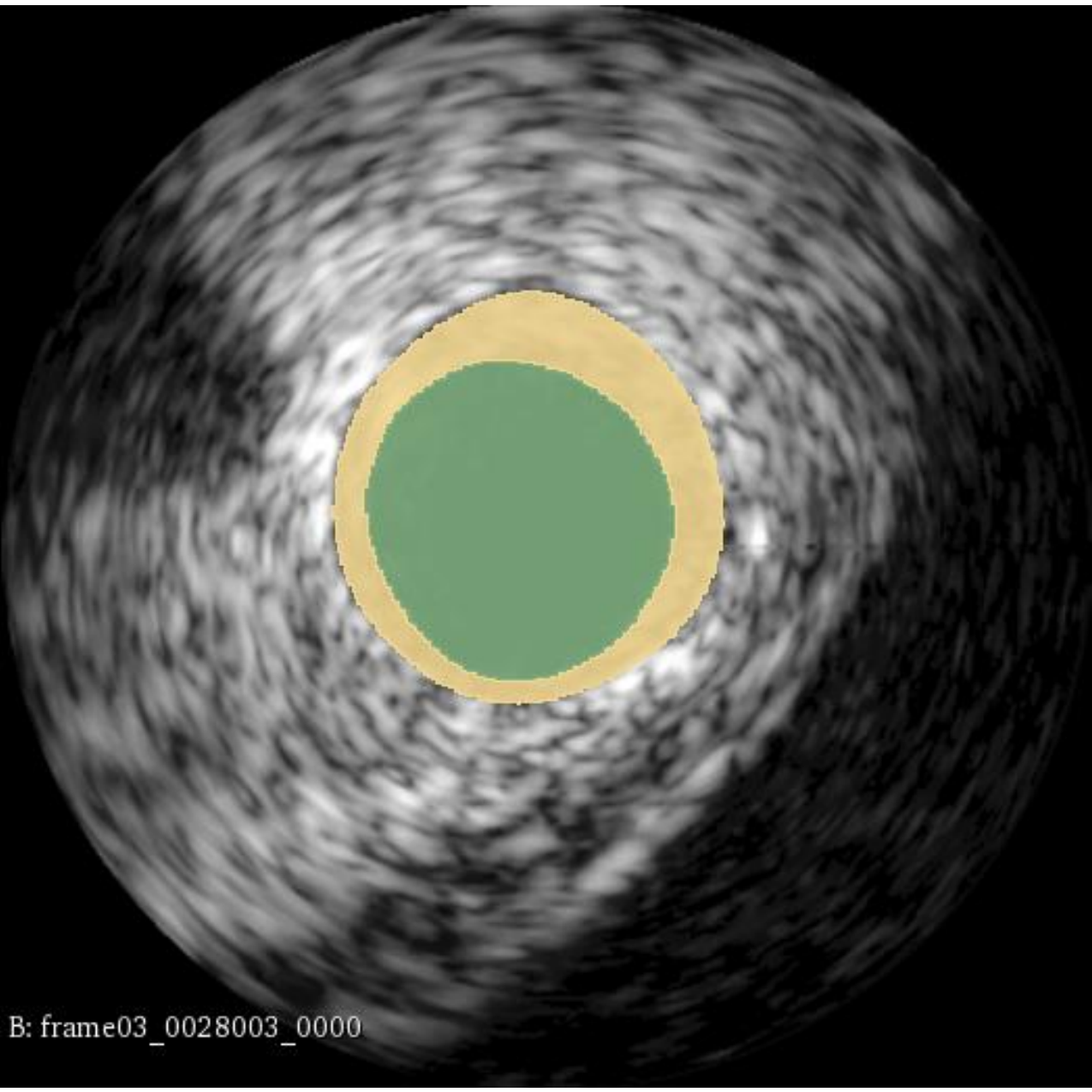}
     \caption{Ours}
  \end{subfigure}
      \begin{subfigure}{0.14\linewidth}
     \includegraphics[width=1\textwidth]{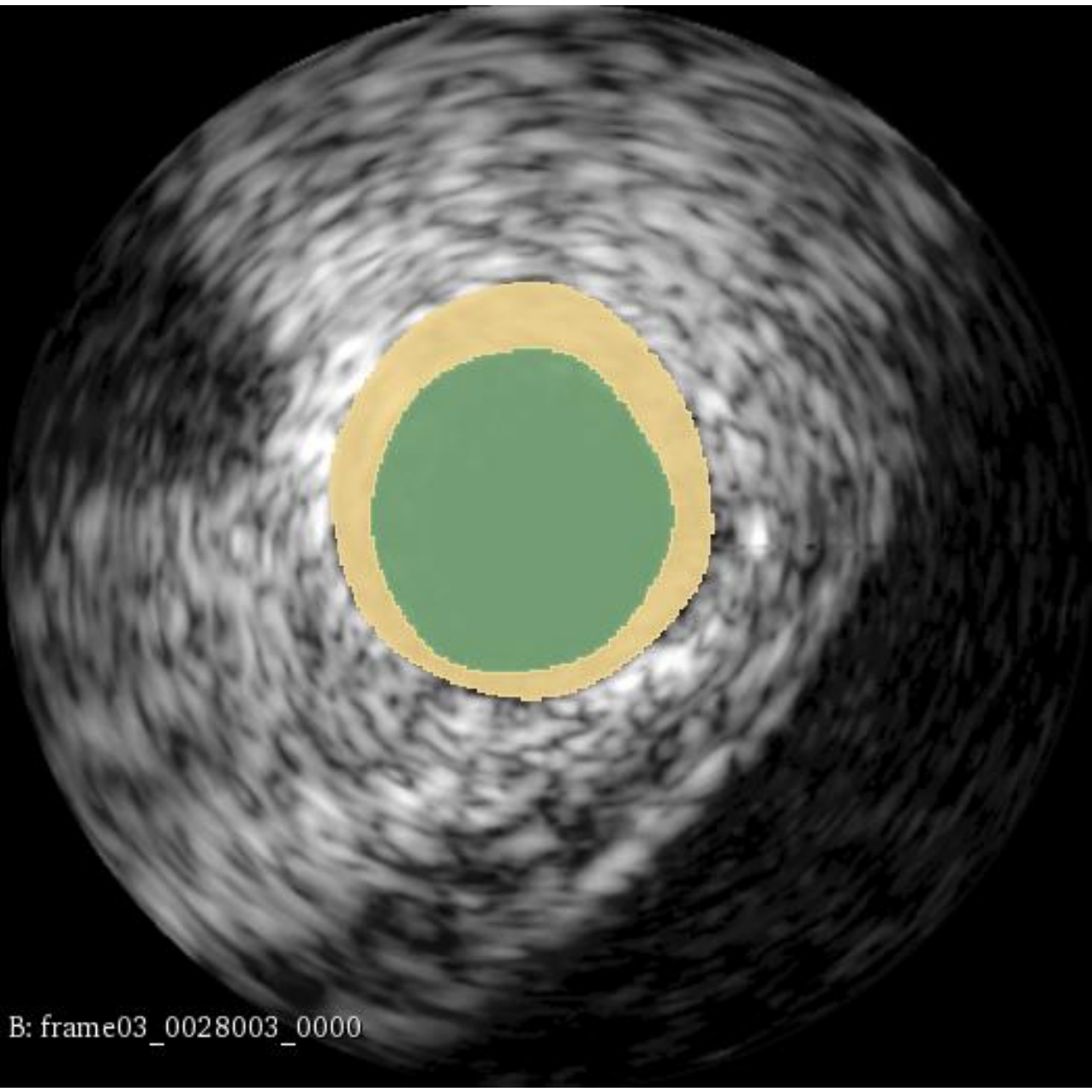}
     \caption{GT}
  \end{subfigure}

\caption{Additional qualitative IVUS results compared with the baselines. Colors for the classes correspond to the ones used in Fig.~\ref{fig:data-interactions}.}
\label{fig:ivus-add-3}
\end{figure}

\begin{figure}[t]
\centering 

 \begin{subfigure}{0.15\linewidth}
  \includegraphics[width=1\textwidth]{figures/multi-organ/sample5/input-slice.pdf}
  \caption{Input}
  \end{subfigure}
  \begin{subfigure}{0.15\linewidth}
     \includegraphics[width=1\textwidth]{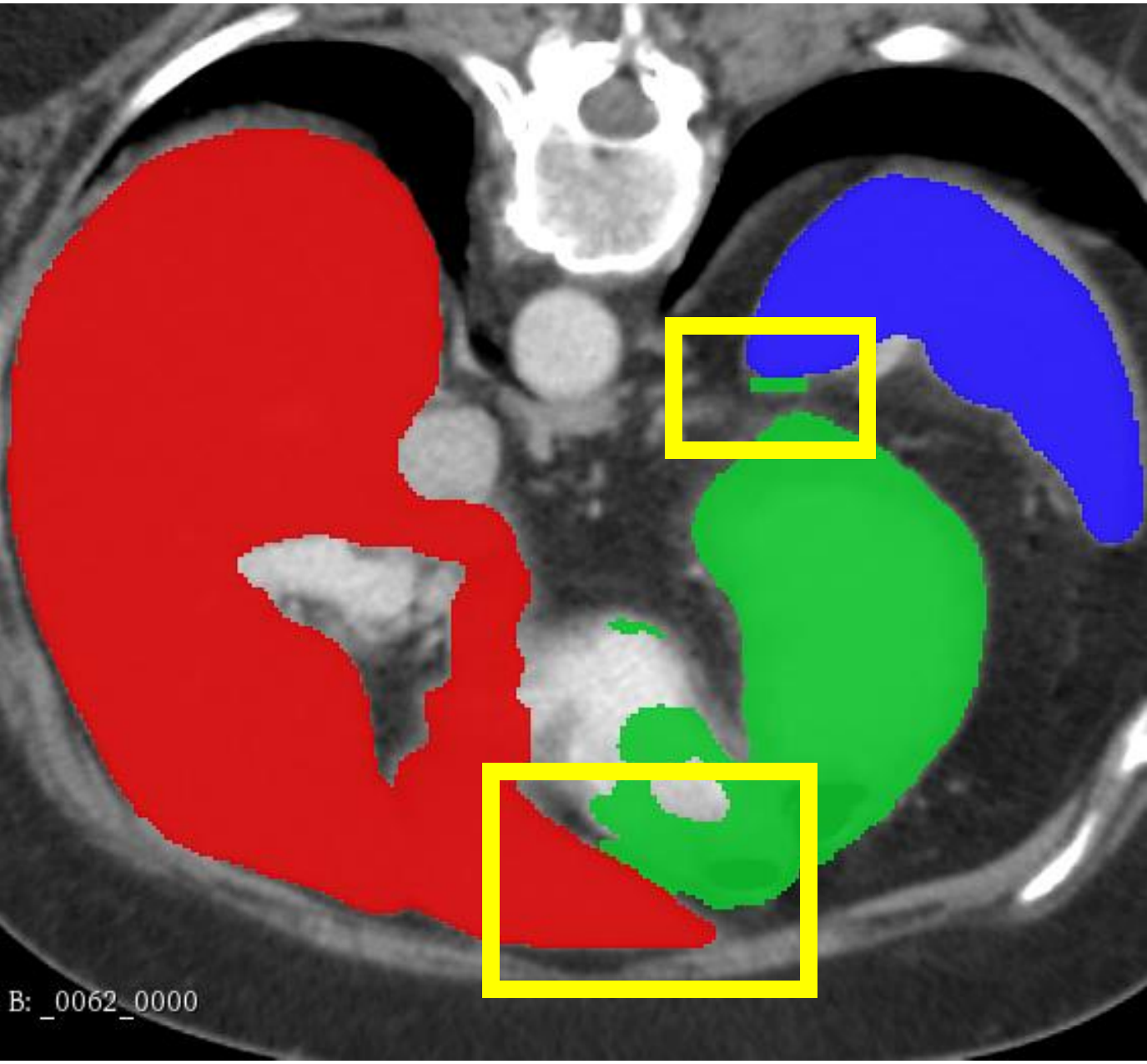}
     \caption{UNet}
  \end{subfigure}
    \begin{subfigure}{0.15\linewidth}
     \includegraphics[width=1\textwidth]{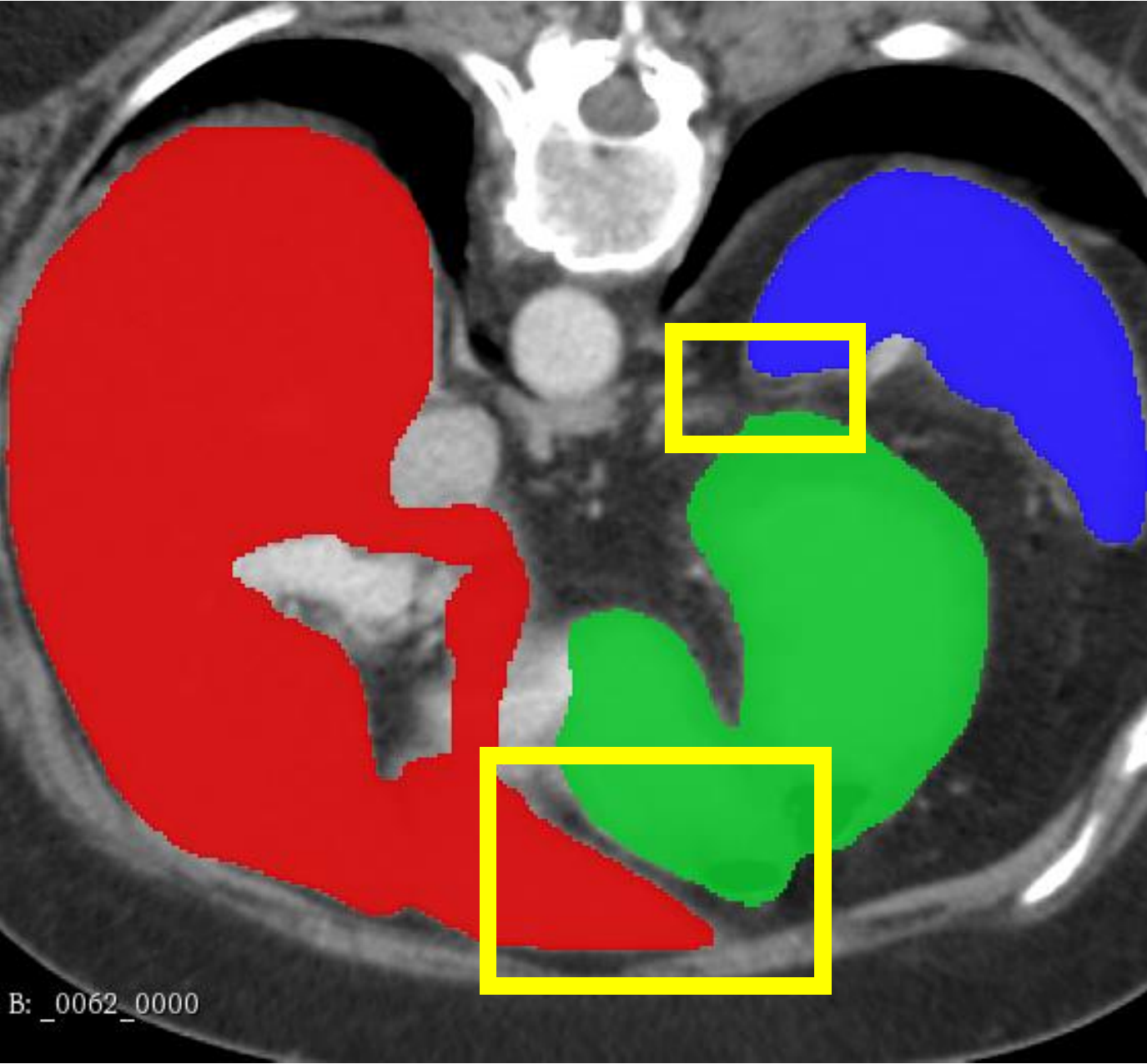}
     \caption{UNet+O}
  \end{subfigure}
    \begin{subfigure}{0.15\linewidth}
     \includegraphics[width=1\textwidth]{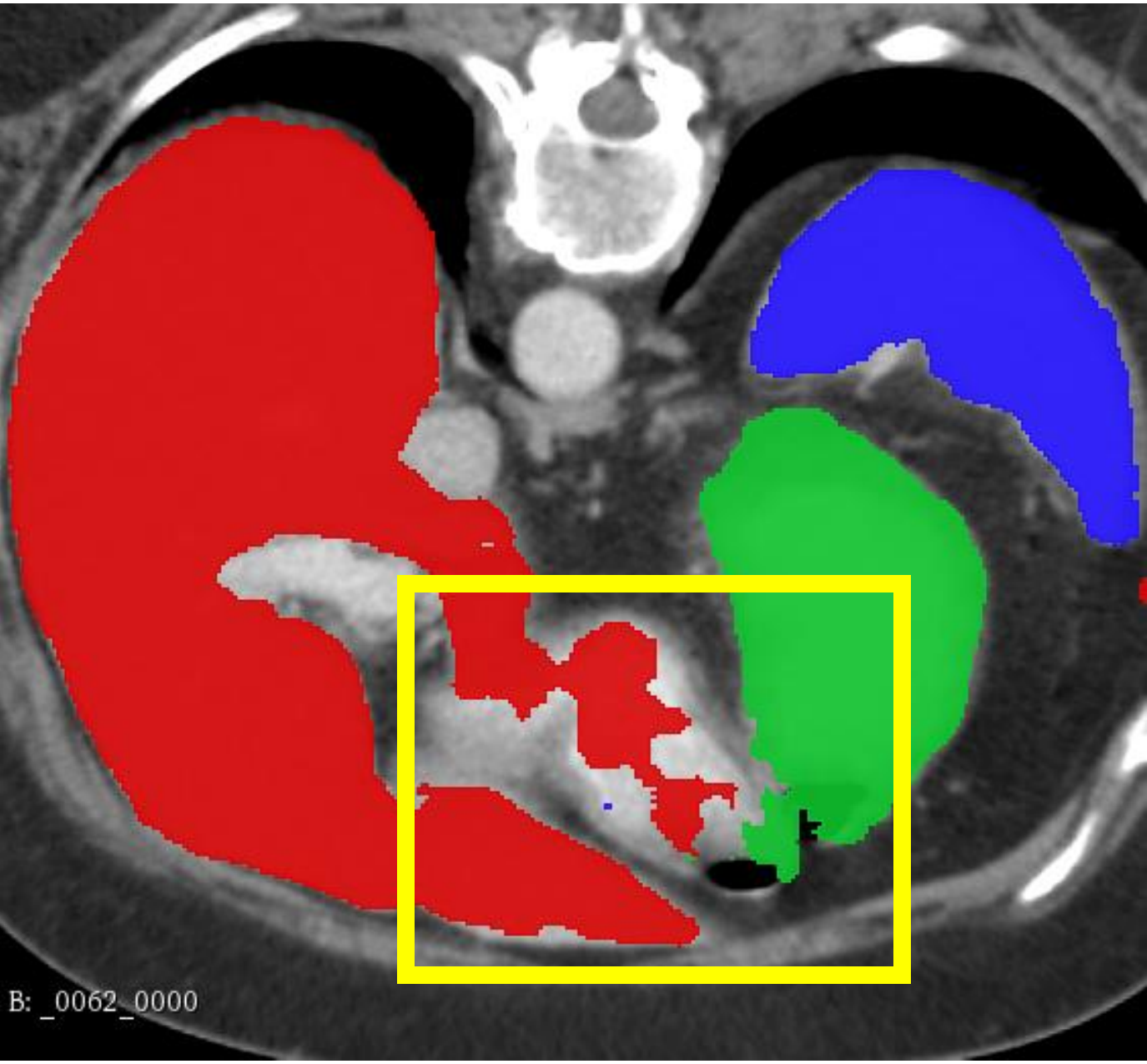}
     \caption{FCN}
  \end{subfigure}
      \begin{subfigure}{0.15\linewidth}
     \includegraphics[width=1\textwidth]{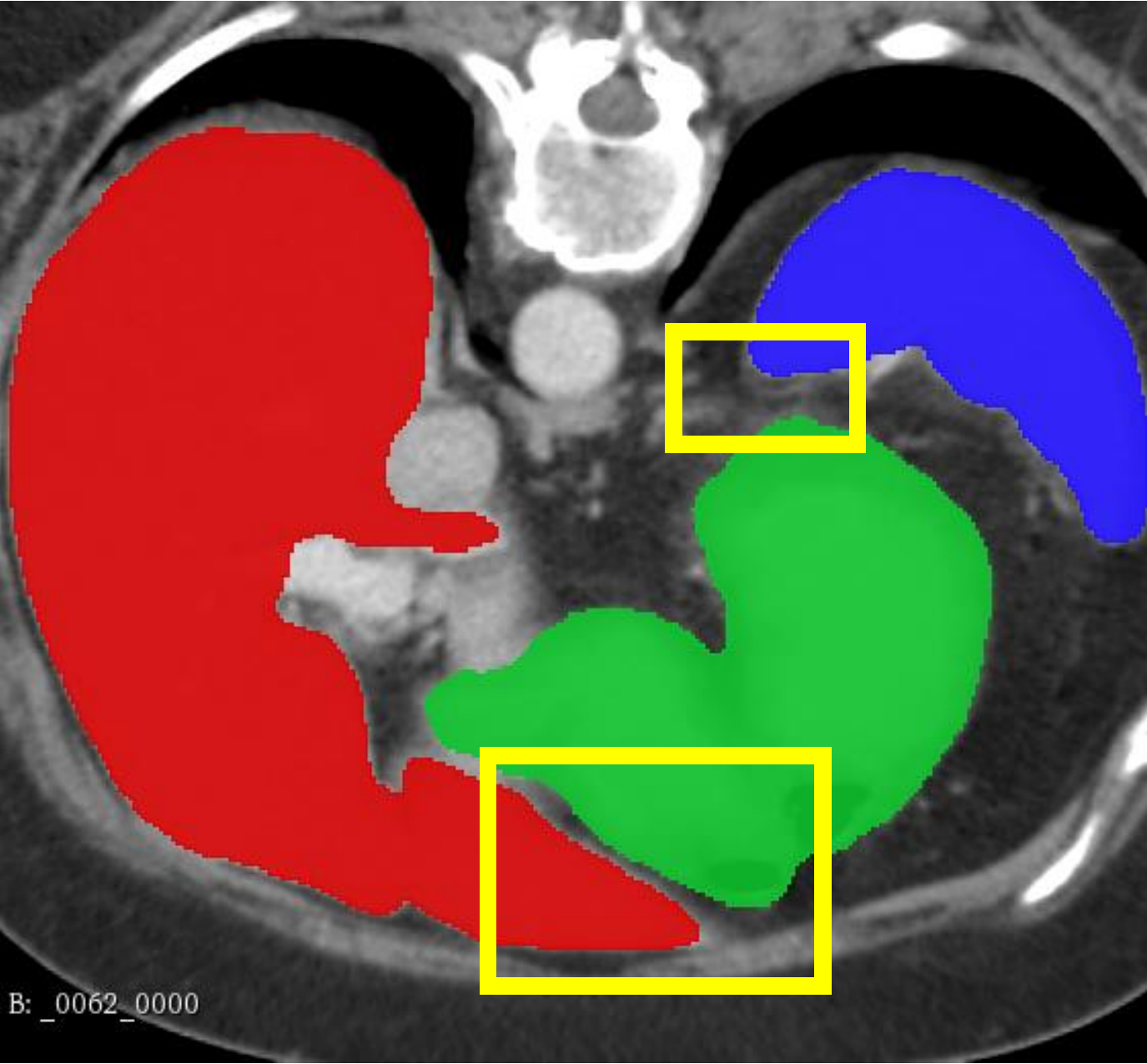}
     \caption{FCN+O}
  \end{subfigure}
    \begin{subfigure}{0.15\linewidth}
     \includegraphics[width=1\textwidth]{figures/multi-organ/sample6/nnunet-slice.pdf}
     \caption{nnUNet}
  \end{subfigure}

  \begin{subfigure}{0.15\linewidth}
     \includegraphics[width=1\textwidth]{figures/multi-organ/sample6/crf-slice.pdf}
     \caption{CRF}
  \end{subfigure}
      \begin{subfigure}{0.15\linewidth}
     \includegraphics[width=1\textwidth]{figures/multi-organ/sample6/midl-slice.pdf}
     \caption{MIDL}
  \end{subfigure}
  \begin{subfigure}{0.15\linewidth}
     \includegraphics[width=1\textwidth]{figures/multi-organ/sample6/nonadj-slice.pdf}
     \caption{NonAdj}
  \end{subfigure}
        \begin{subfigure}{0.15\linewidth}
     \includegraphics[width=1\textwidth]{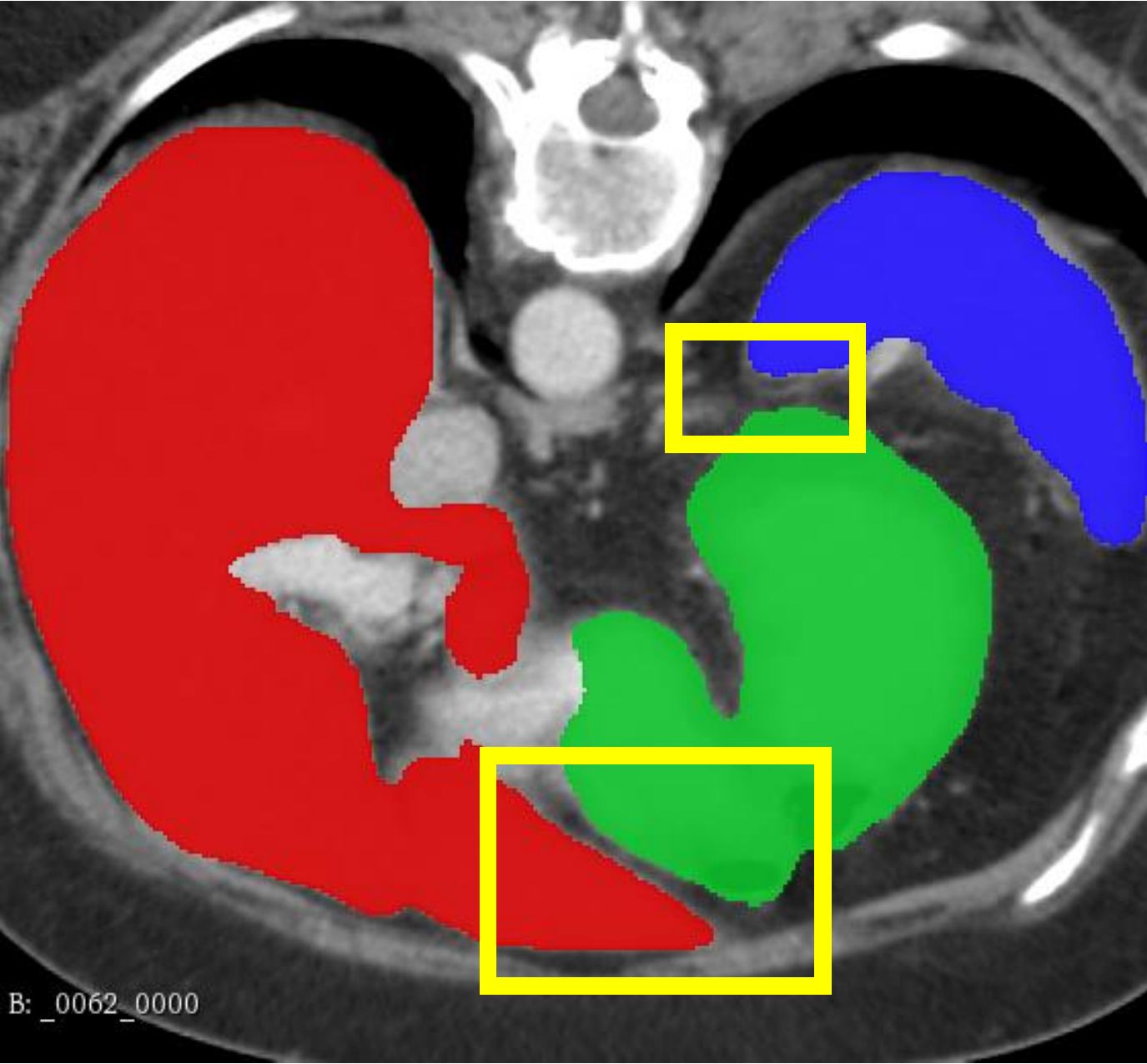}
     \caption{Ours6C}
  \end{subfigure}
      \begin{subfigure}{0.15\linewidth}
     \includegraphics[width=1\textwidth]{figures/multi-organ/sample6/topo-slice.pdf}
     \caption{Ours}
  \end{subfigure}
      \begin{subfigure}{0.15\linewidth}
     \includegraphics[width=1\textwidth]{figures/multi-organ/sample6/gt-slice.pdf}
     \caption{GT}
  \end{subfigure}

\caption{Qualitative Multi-Atlas results compared with the baselines. Colors for the classes correspond to the ones used in Fig.~\ref{fig:data-interactions}.}
\label{fig:multi-add-1}
\end{figure}

\begin{figure}[t]
\centering 

   \begin{subfigure}{0.14\linewidth}
  \includegraphics[width=1\textwidth]{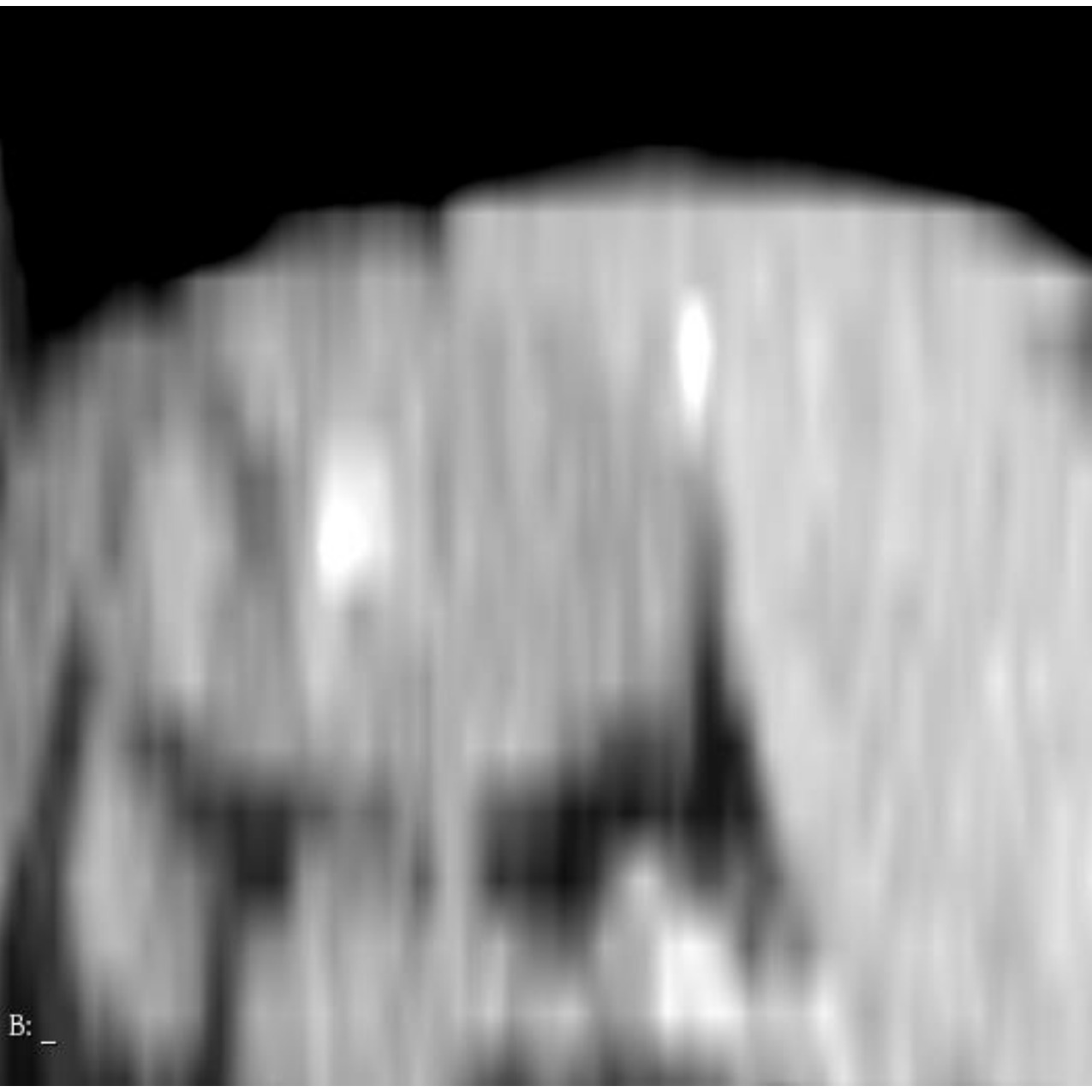}
  \caption{Input}
  \end{subfigure}
  \begin{subfigure}{0.14\linewidth}
     \includegraphics[width=1\textwidth]{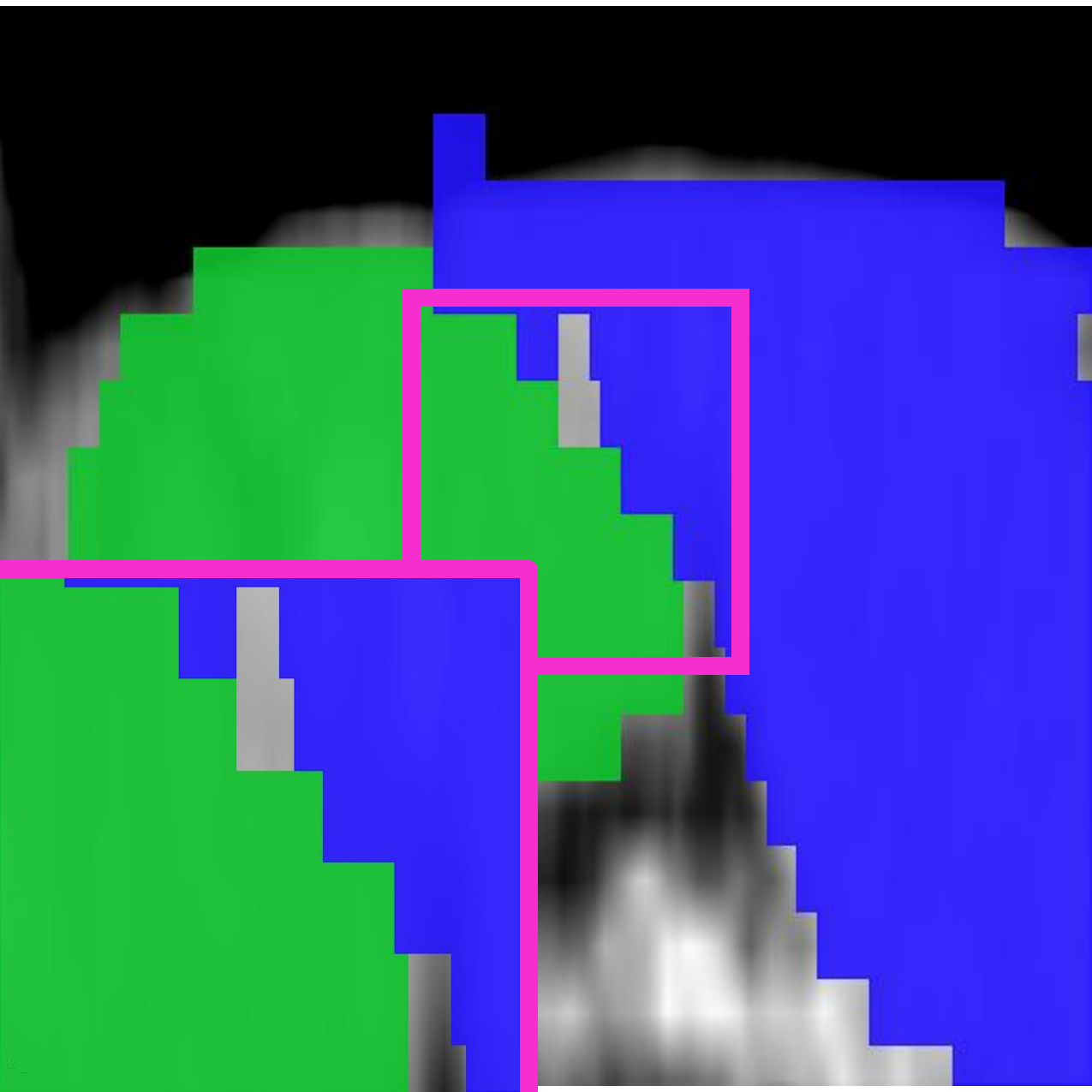}
     \caption{UNet}
  \end{subfigure}
      \begin{subfigure}{0.15\linewidth}
     \includegraphics[width=1\textwidth]{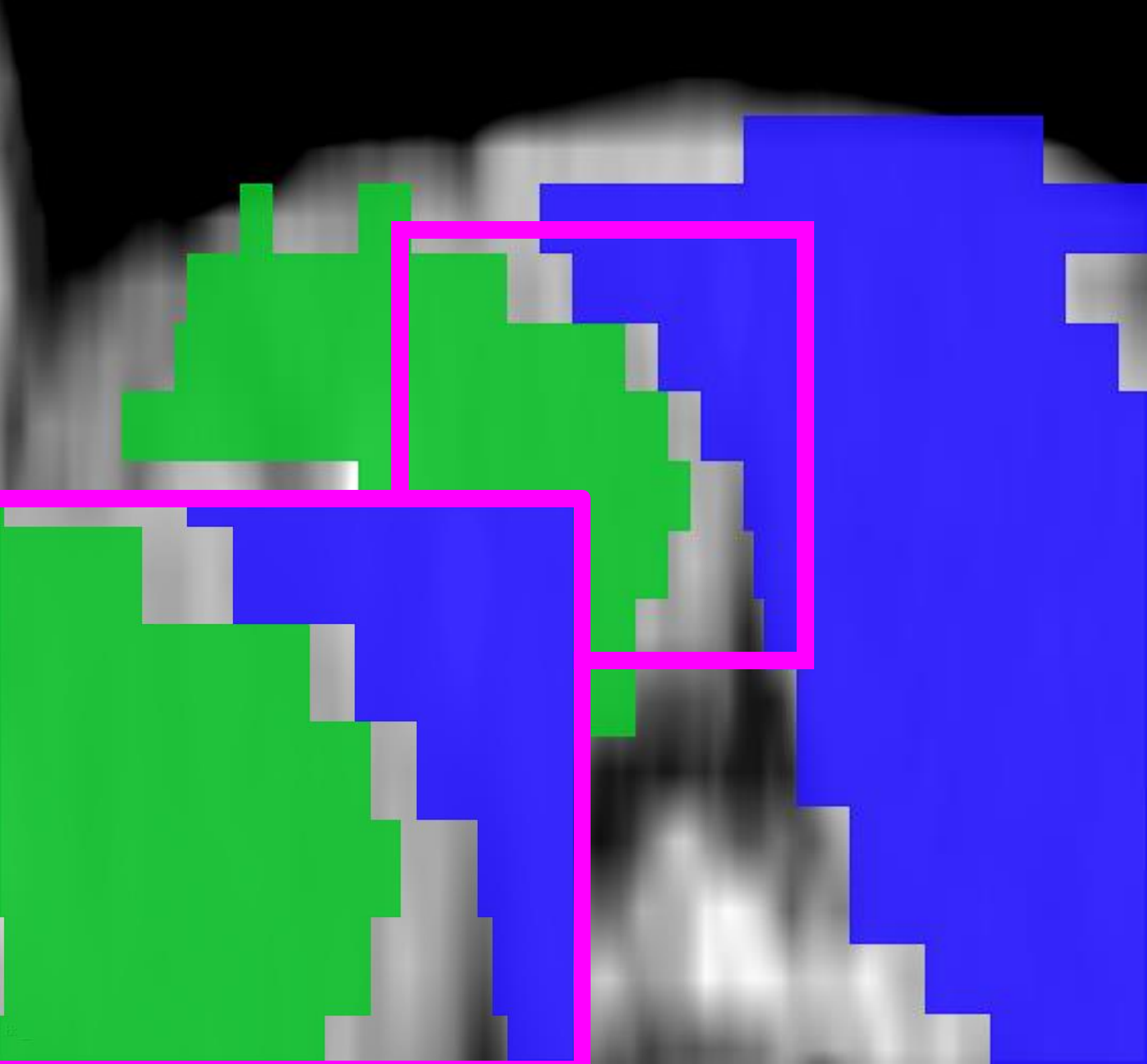}
     \caption{UNet+O}
  \end{subfigure}
    \begin{subfigure}{0.14\linewidth}
     \includegraphics[width=1\textwidth]{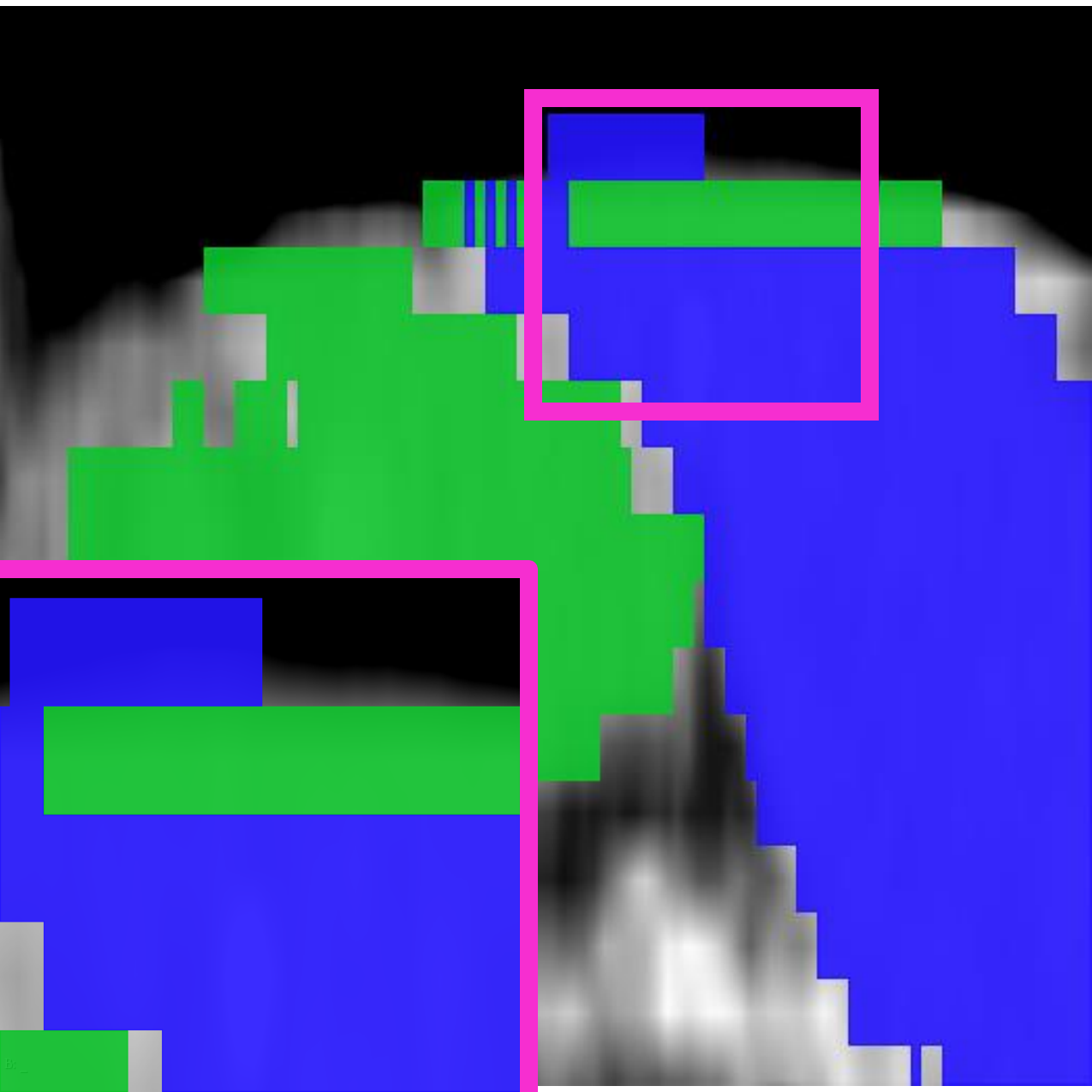}
     \caption{FCN}
  \end{subfigure}
      \begin{subfigure}{0.15\linewidth}
     \includegraphics[width=1\textwidth]{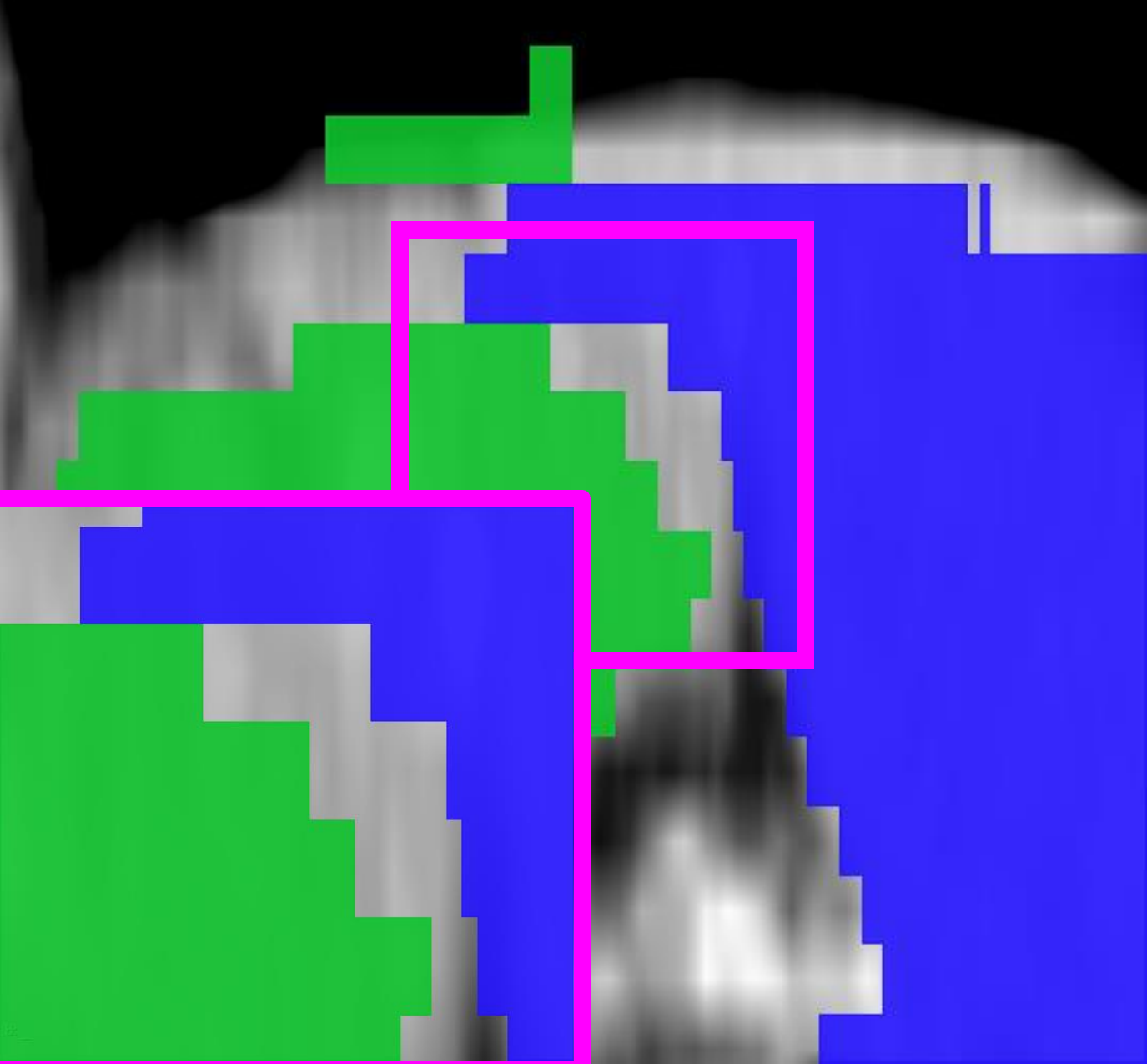}
     \caption{FCN+O}
  \end{subfigure}
    \begin{subfigure}{0.14\linewidth}
     \includegraphics[width=1\textwidth]{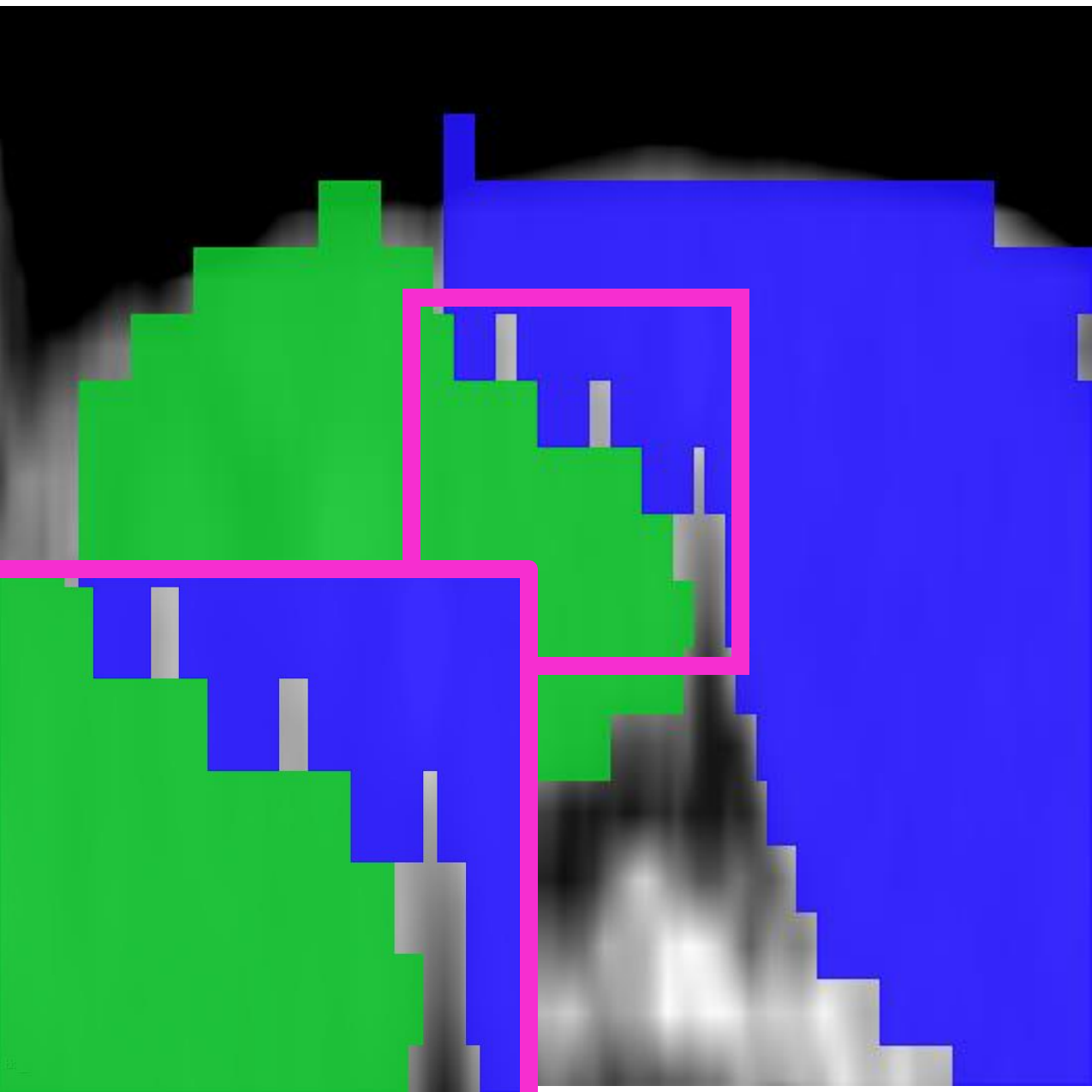}
     \caption{nnUNet}
  \end{subfigure}

      \begin{subfigure}{0.14\linewidth}
     \includegraphics[width=1\textwidth]{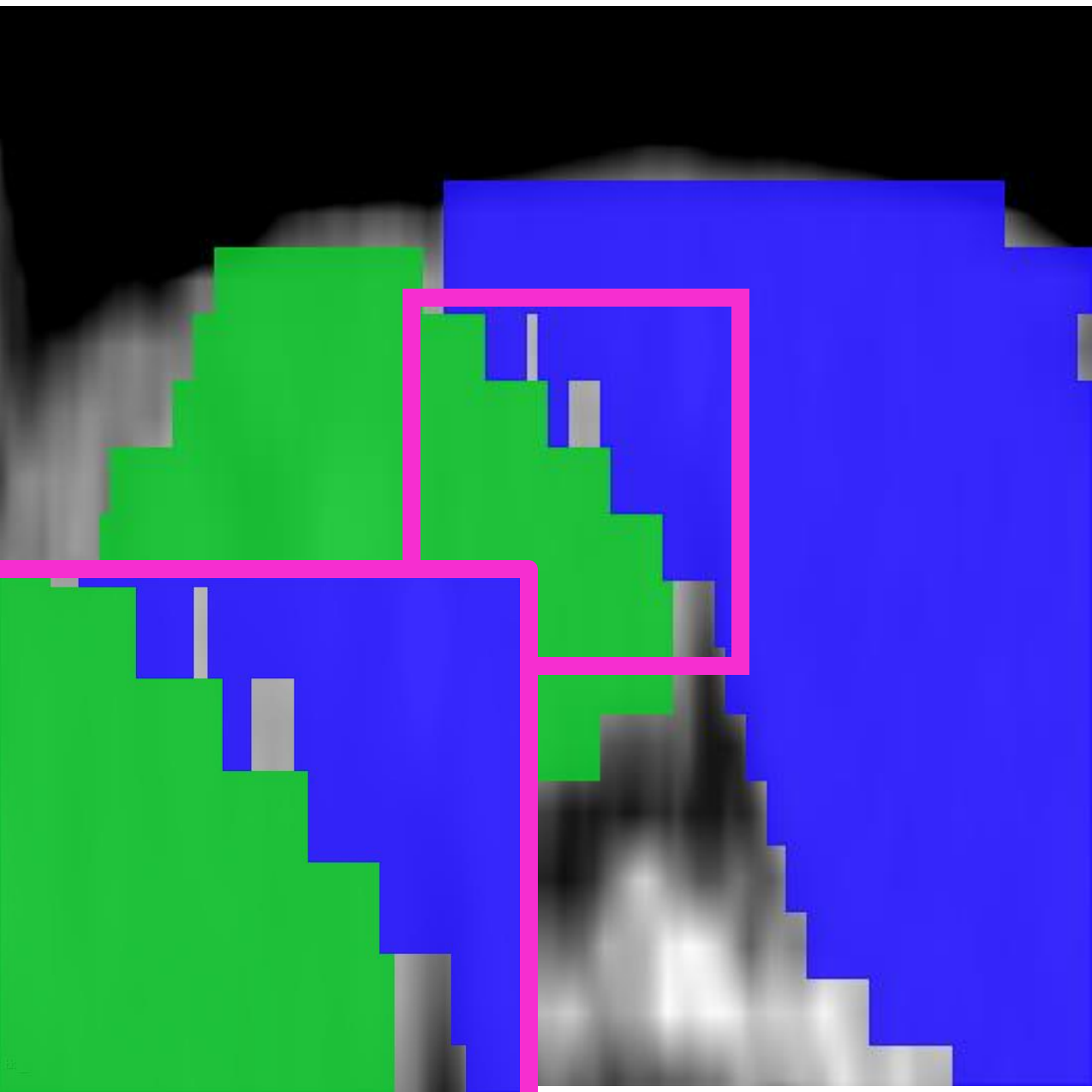}
     \caption{CRF}
  \end{subfigure}
      \begin{subfigure}{0.15\linewidth}
     \includegraphics[width=1\textwidth]{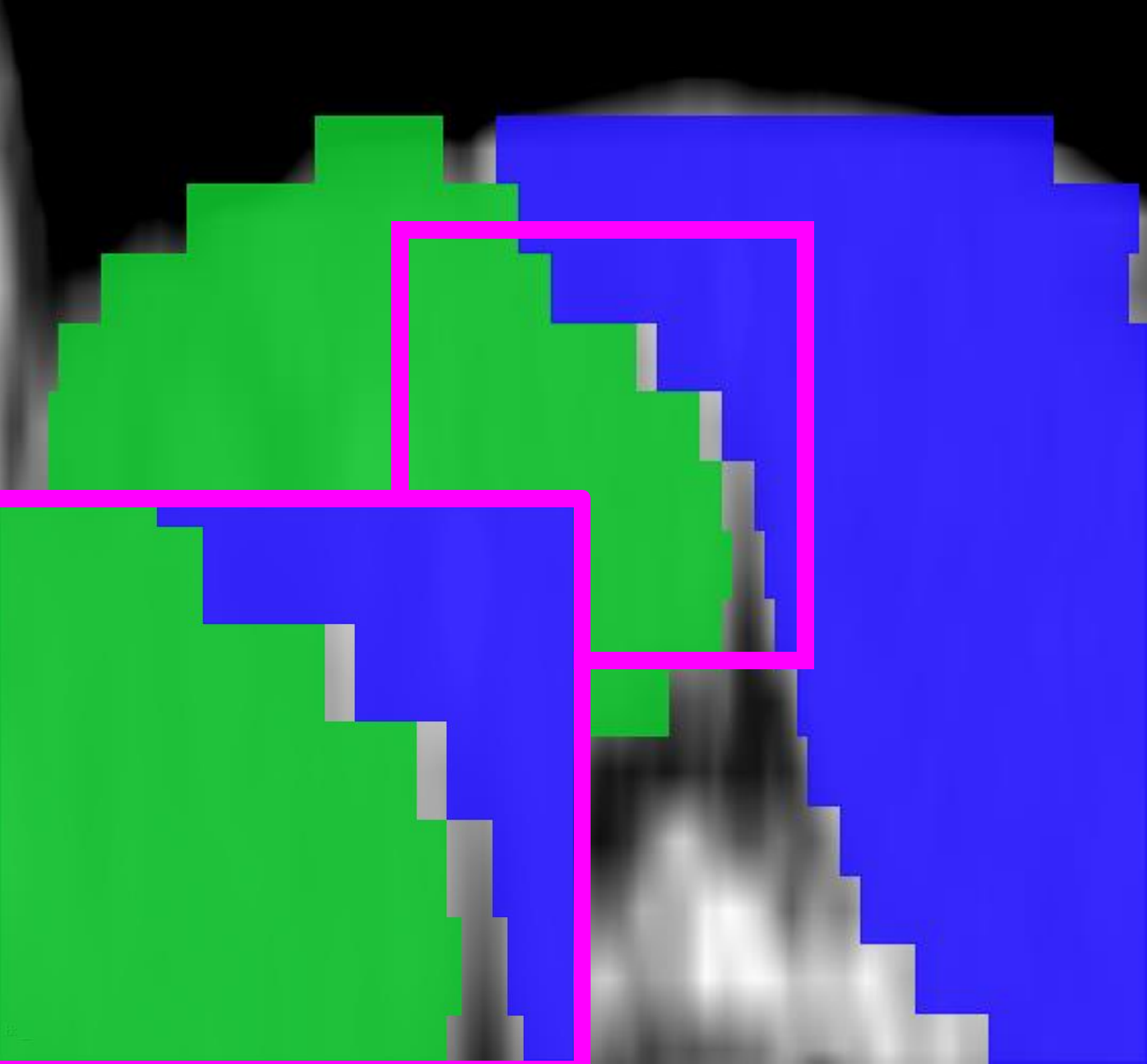}
     \caption{MIDL}
  \end{subfigure}
      \begin{subfigure}{0.15\linewidth}
     \includegraphics[width=1\textwidth]{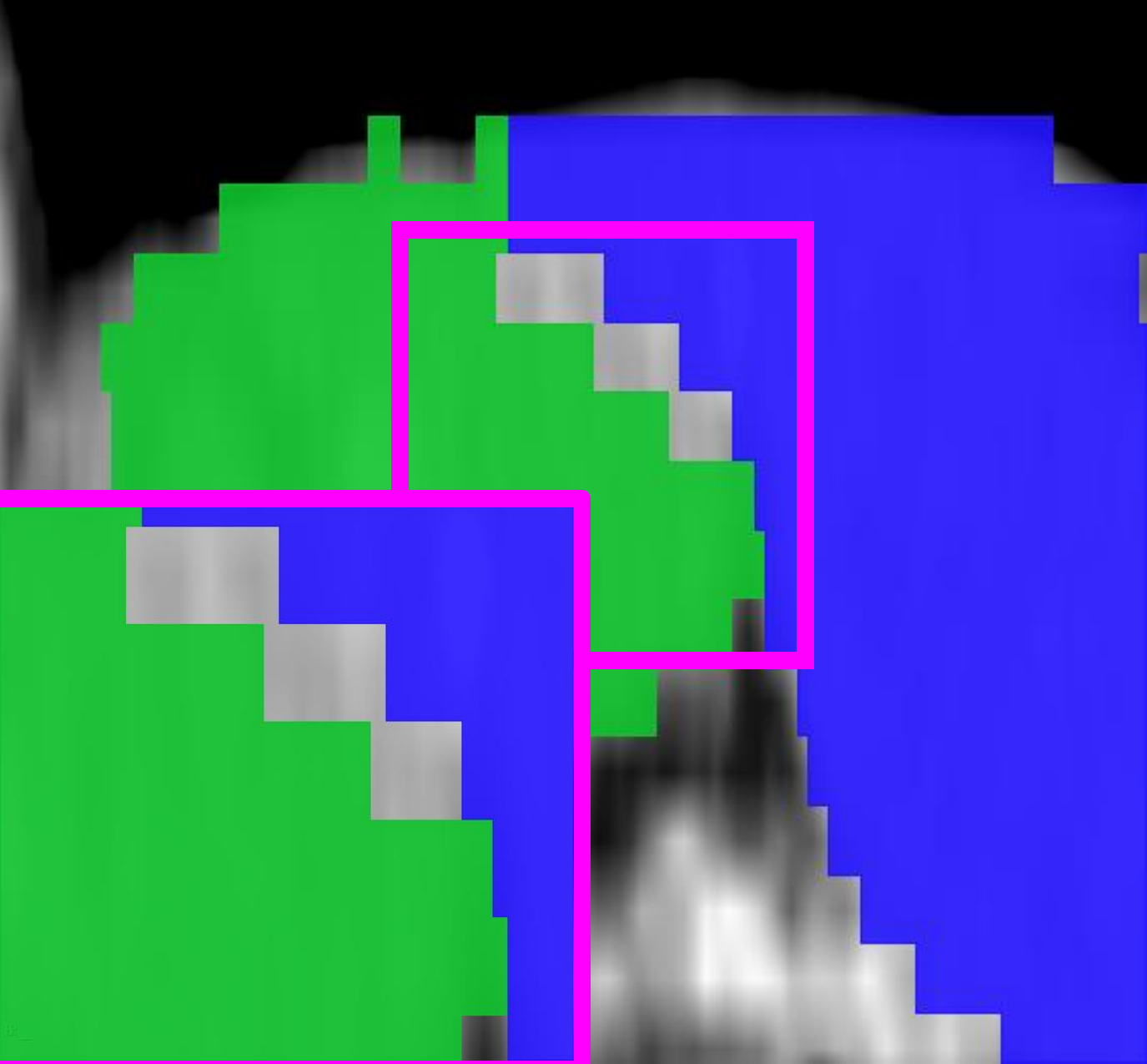}
     \caption{NonAdj}
  \end{subfigure}
        \begin{subfigure}{0.15\linewidth}
     \includegraphics[width=1\textwidth]{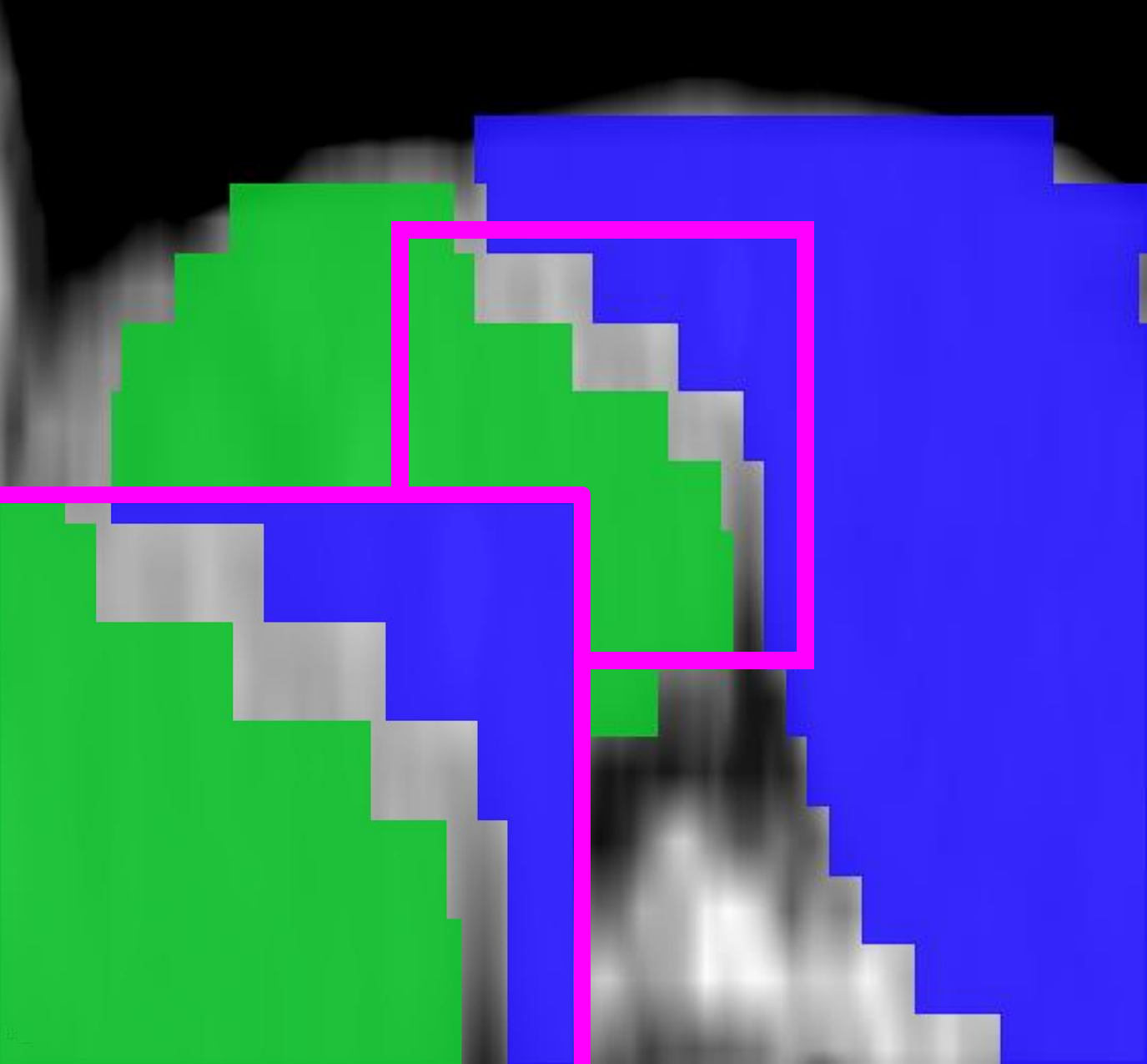}
     \caption{Ours6C}
  \end{subfigure}
        \begin{subfigure}{0.14\linewidth}
     \includegraphics[width=1\textwidth]{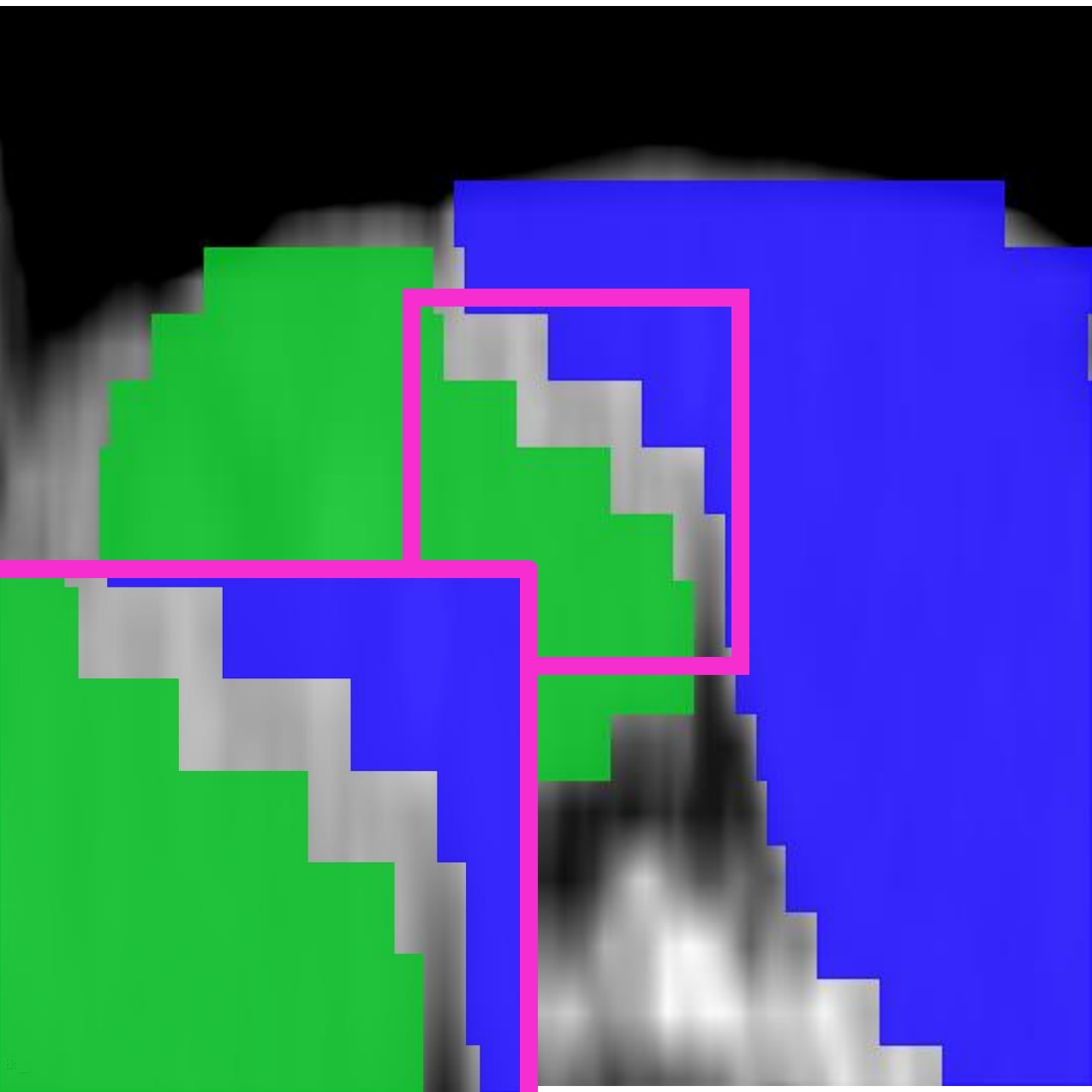}
     \caption{Ours}
  \end{subfigure}
      \begin{subfigure}{0.14\linewidth}
     \includegraphics[width=1\textwidth]{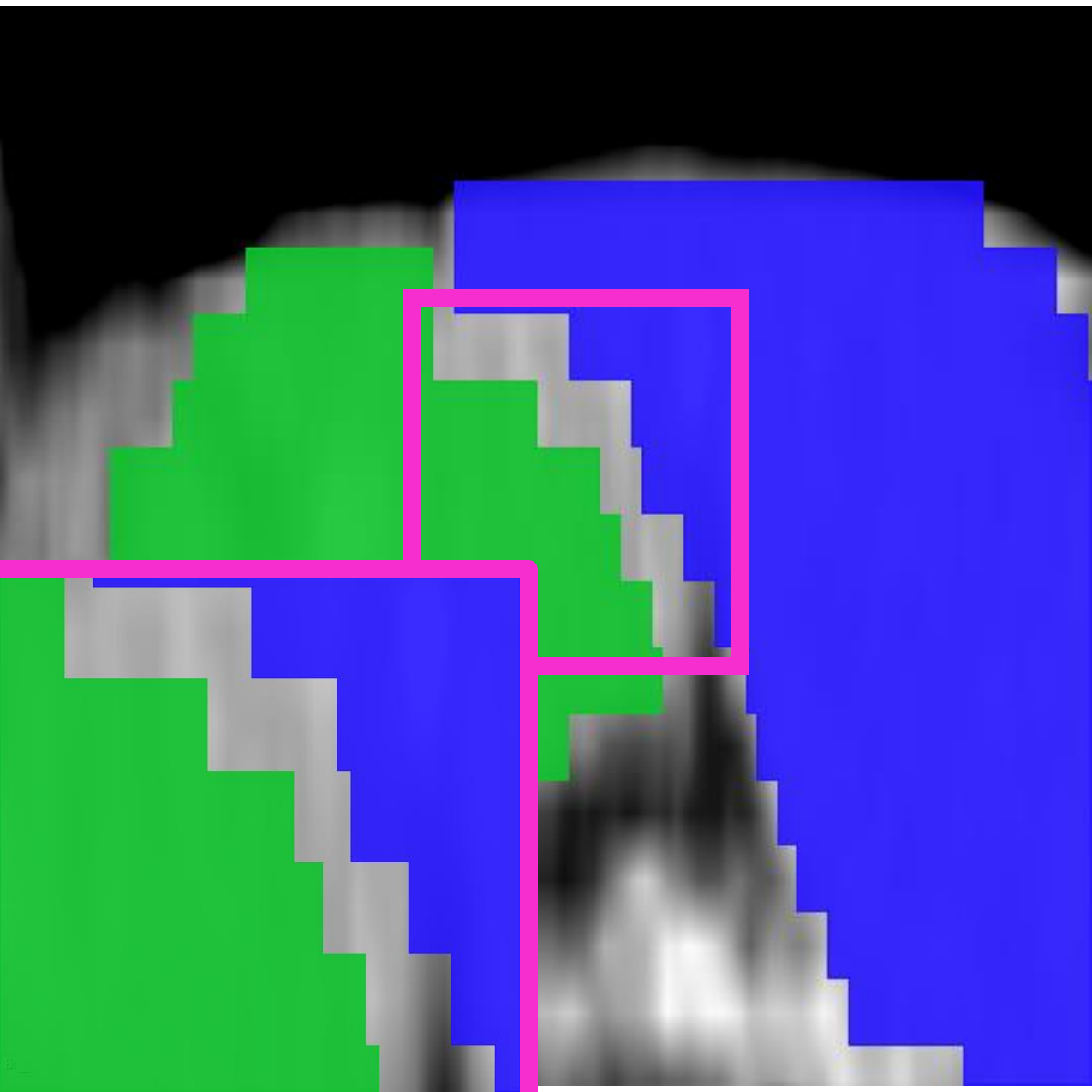}
     \caption{GT}
  \end{subfigure}

        \begin{subfigure}{0.14\linewidth}
  \includegraphics[width=1\textwidth]{figures/empty3.pdf}
  \end{subfigure}
  \begin{subfigure}{0.14\linewidth}
     \includegraphics[width=1\textwidth]{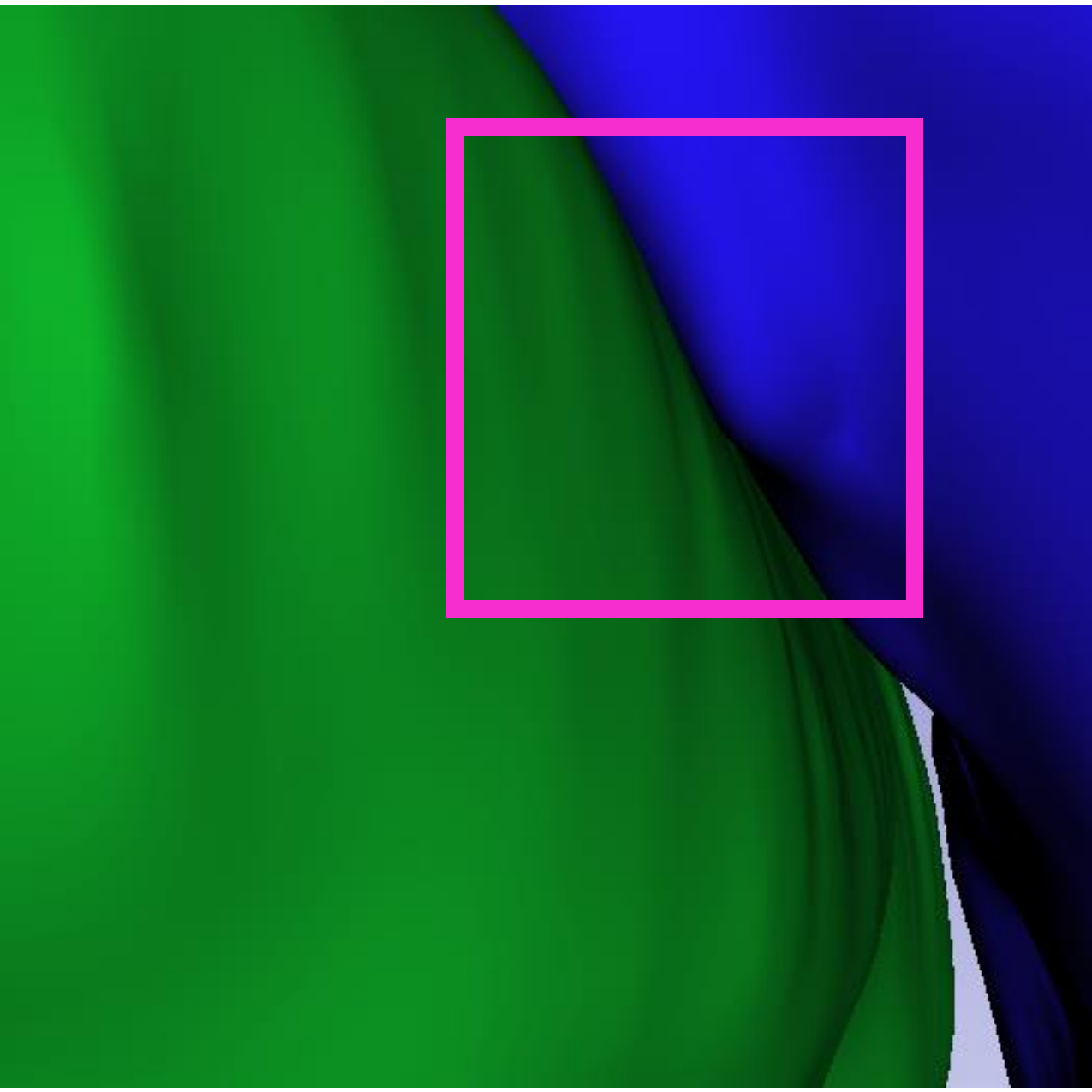}
     \caption{UNet}
  \end{subfigure}
      \begin{subfigure}{0.15\linewidth}
     \includegraphics[width=1\textwidth]{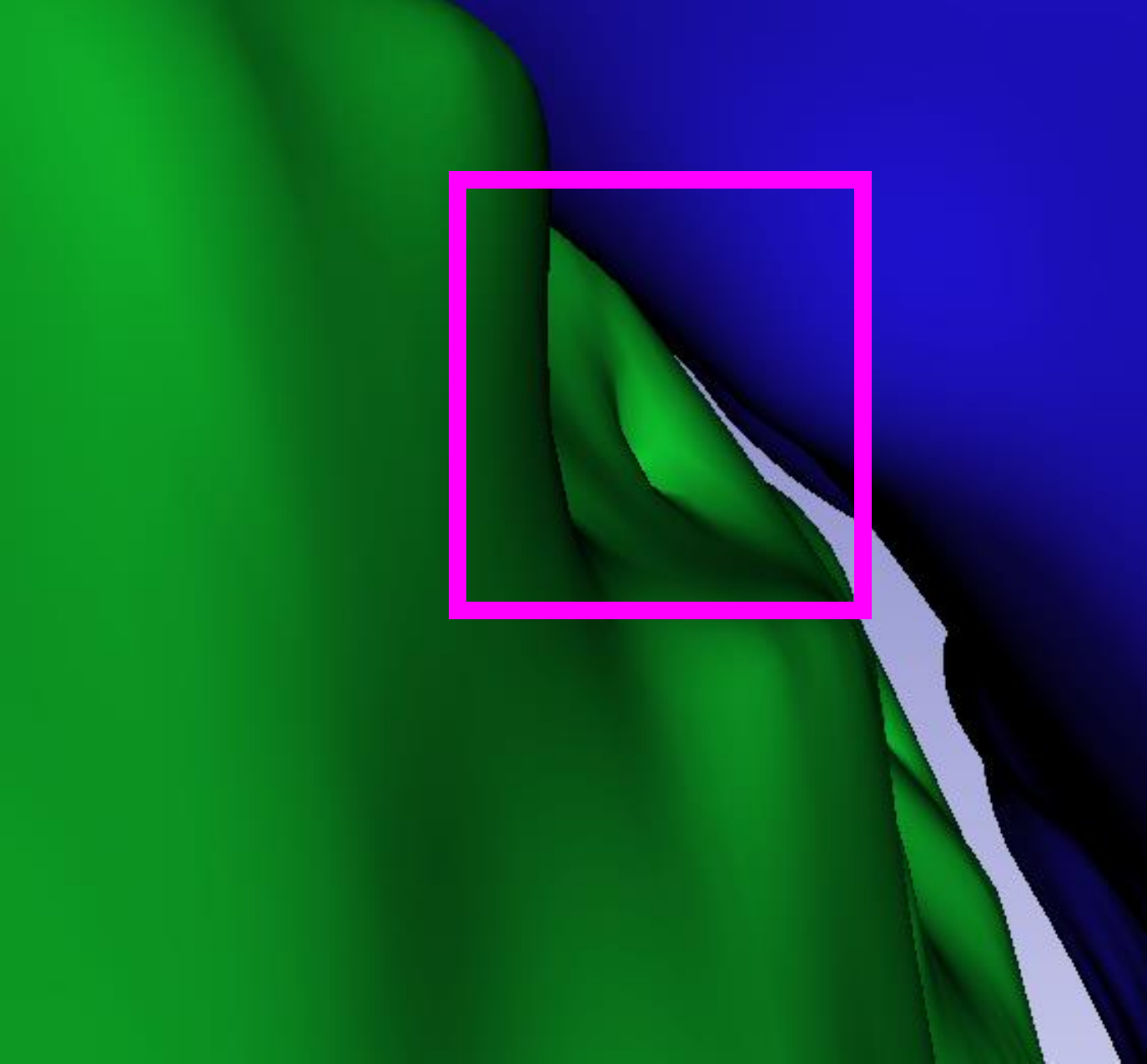}
     \caption{UNet+O}
  \end{subfigure}
    \begin{subfigure}{0.14\linewidth}
     \includegraphics[width=1\textwidth]{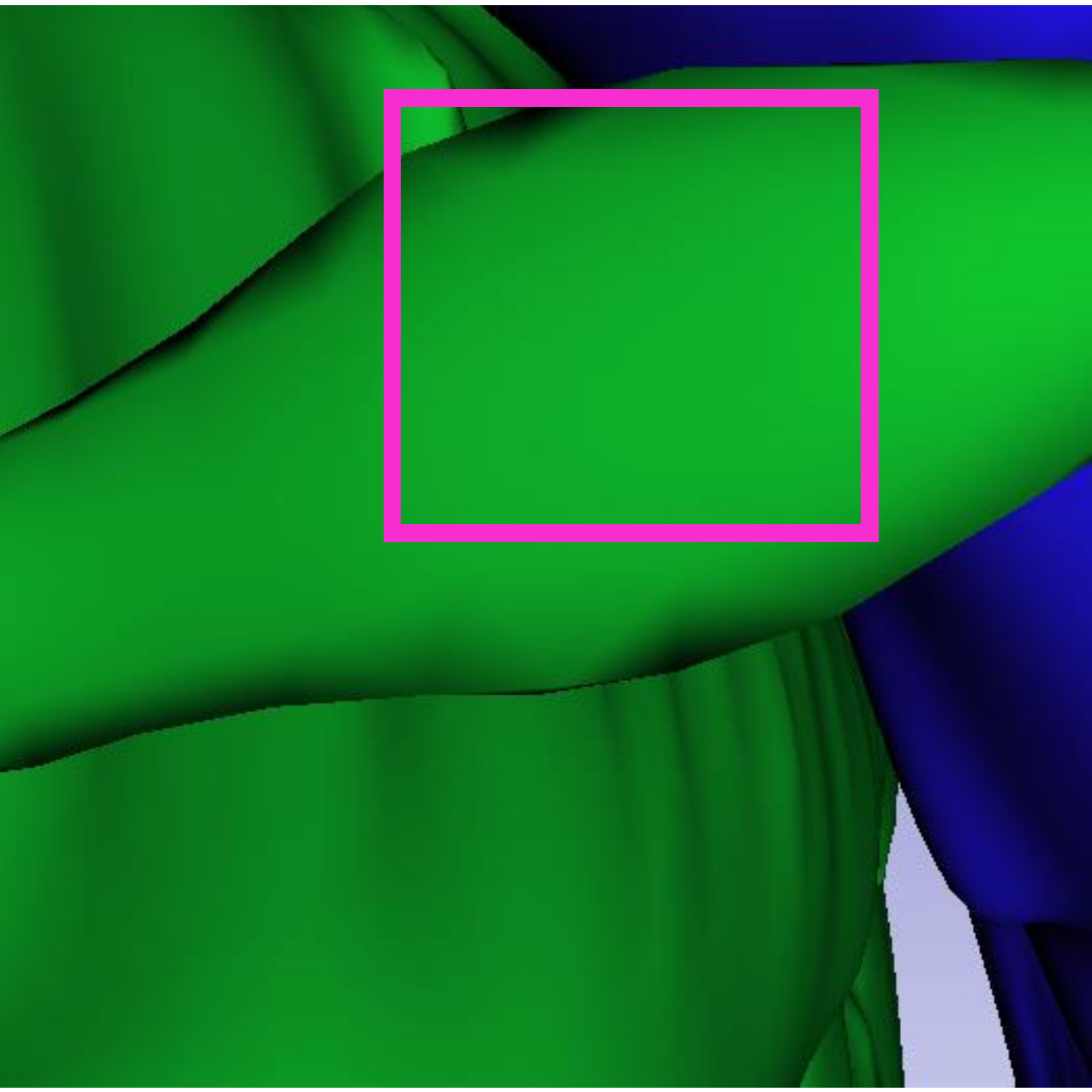}
     \caption{FCN}
  \end{subfigure}
      \begin{subfigure}{0.15\linewidth}
     \includegraphics[width=1\textwidth]{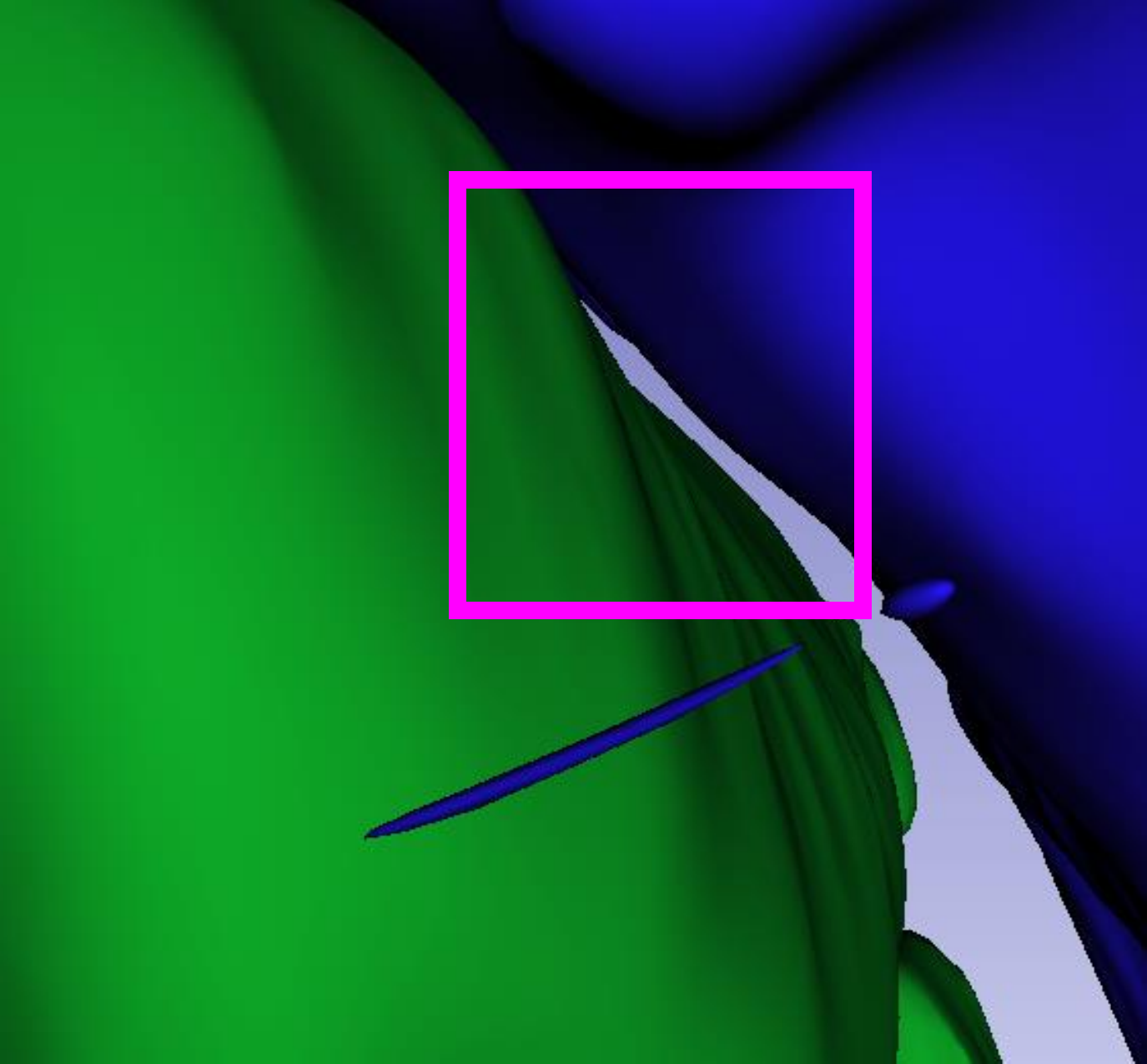}
     \caption{FCN+O}
  \end{subfigure}
    \begin{subfigure}{0.14\linewidth}
     \includegraphics[width=1\textwidth]{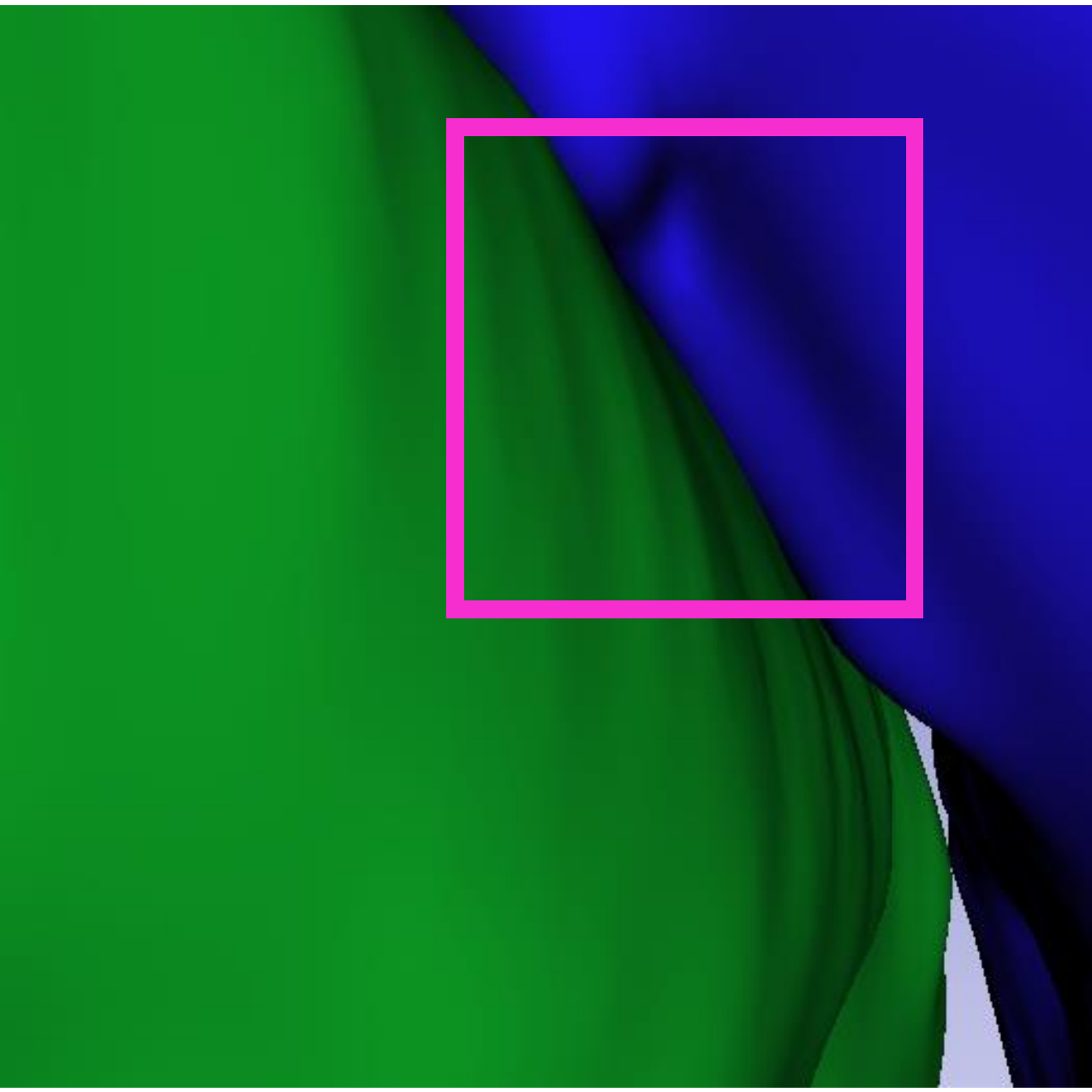}
     \caption{nnUNet}
  \end{subfigure}

      \begin{subfigure}{0.14\linewidth}
     \includegraphics[width=1\textwidth]{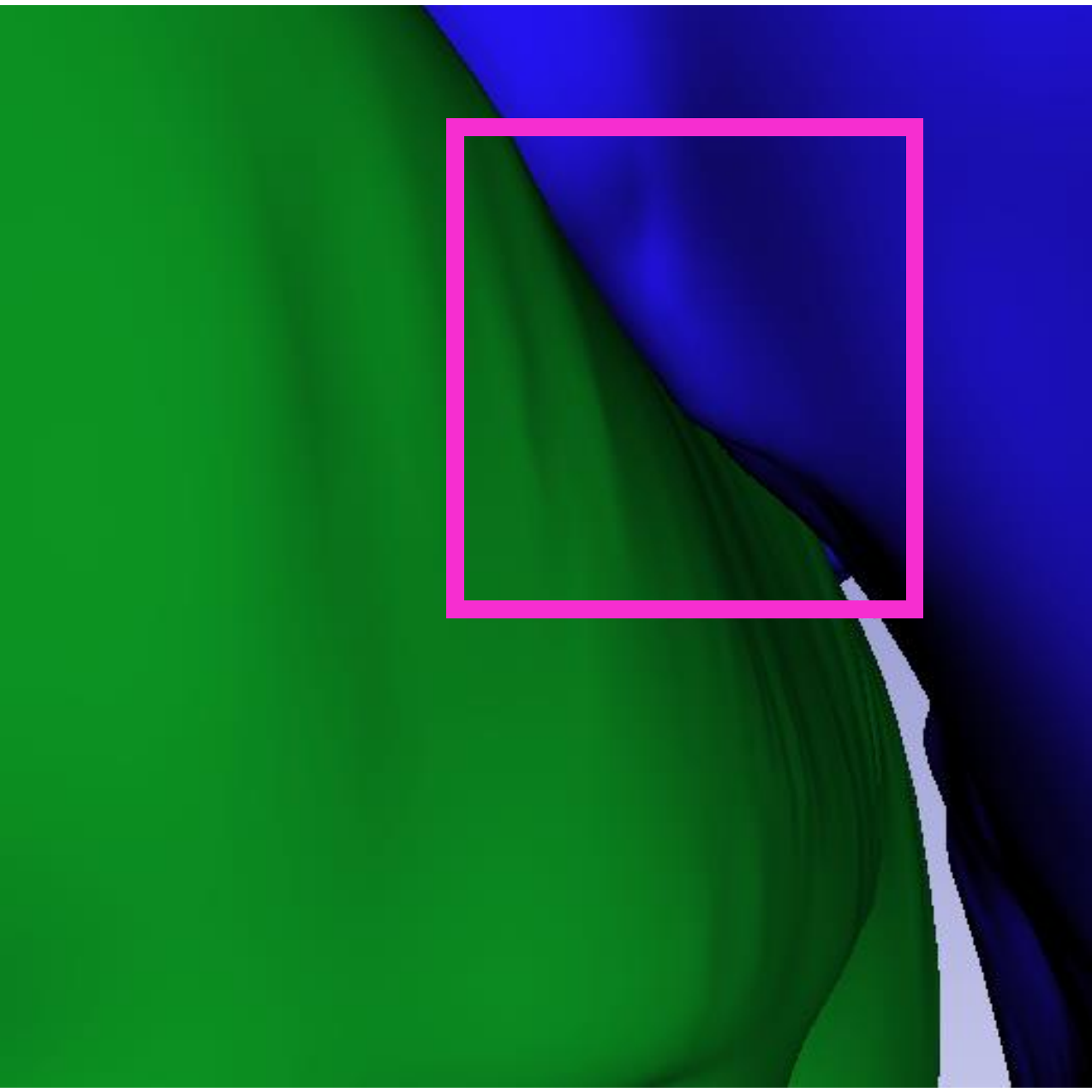}
     \caption{CRF}
  \end{subfigure}
    \begin{subfigure}{0.15\linewidth}
     \includegraphics[width=1\textwidth]{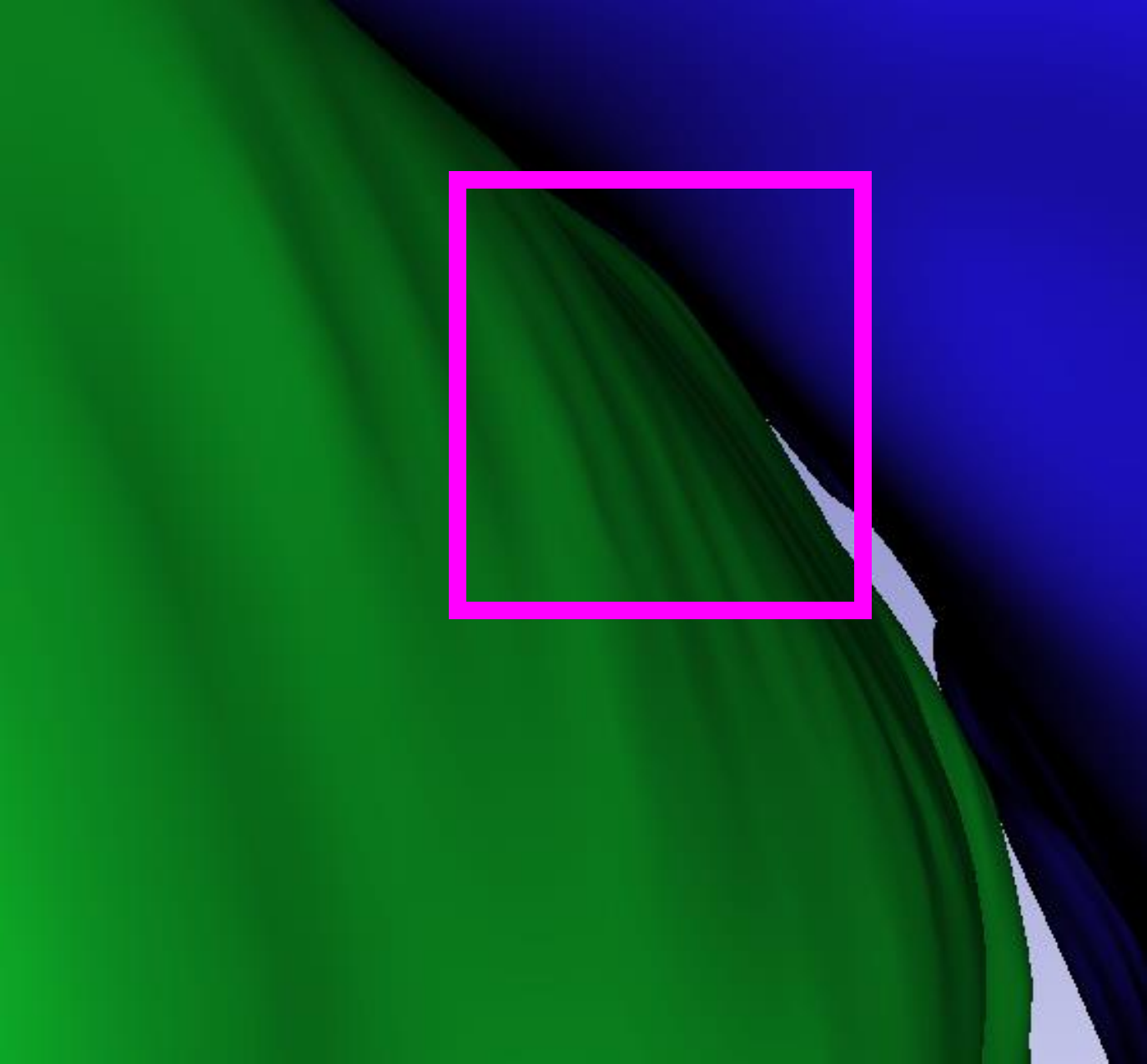}
     \caption{MIDL}
  \end{subfigure}
      \begin{subfigure}{0.15\linewidth}
     \includegraphics[width=1\textwidth]{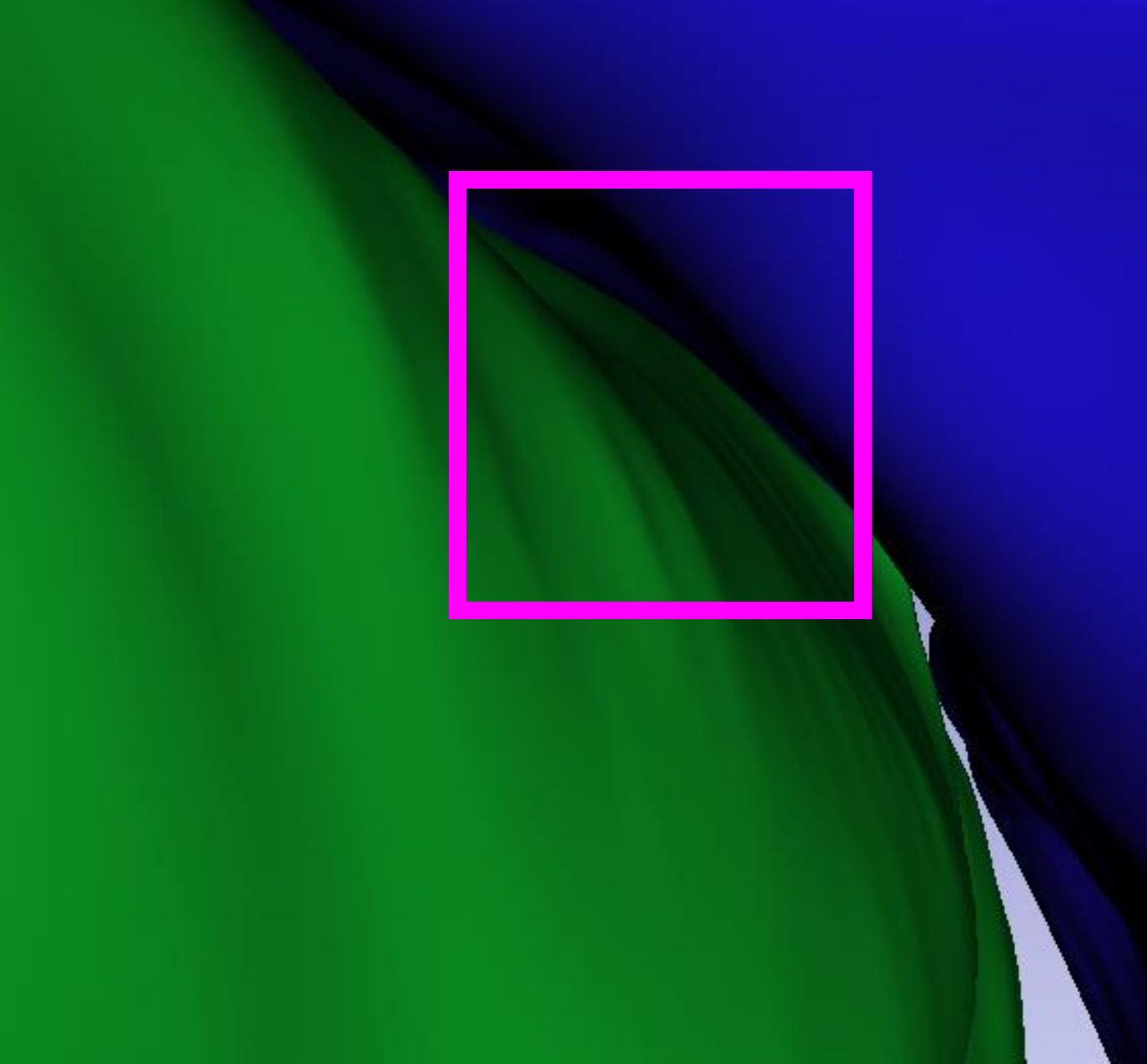}
     \caption{NonAdj}
  \end{subfigure}
      \begin{subfigure}{0.15\linewidth}
     \includegraphics[width=1\textwidth]{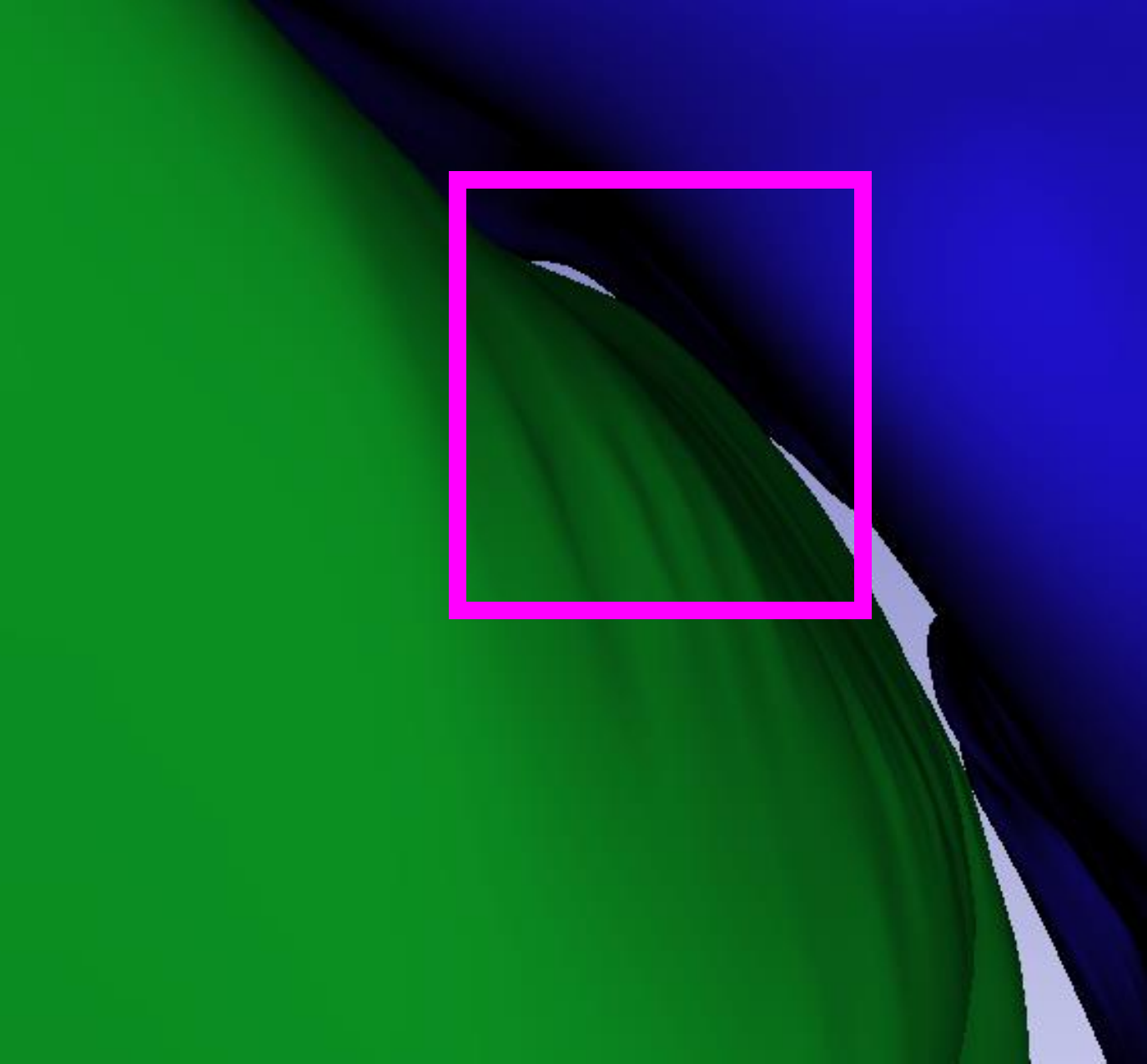}
     \caption{Ours6C}
  \end{subfigure}
        \begin{subfigure}{0.14\linewidth}
     \includegraphics[width=1\textwidth]{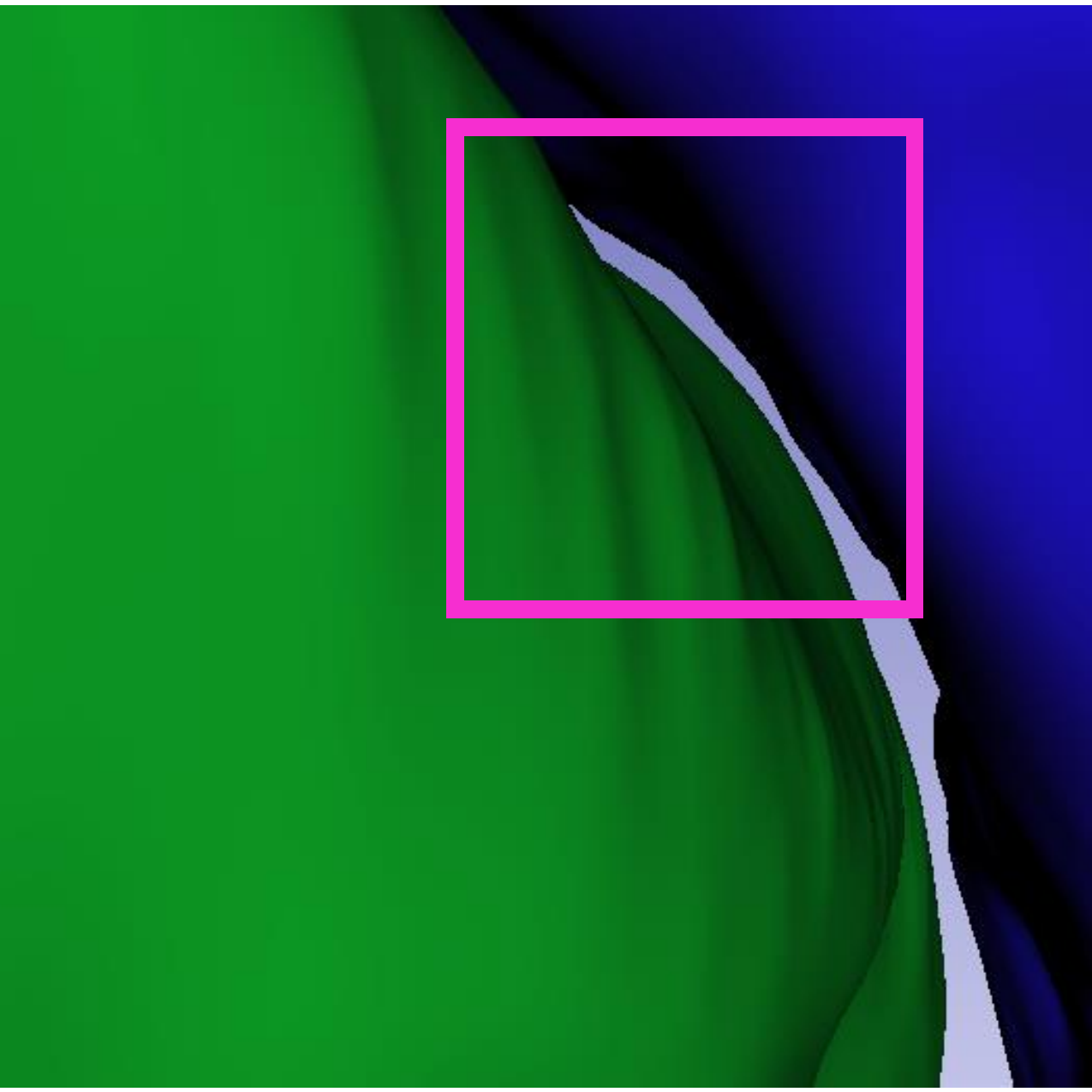}
     \caption{Ours}
  \end{subfigure}
      \begin{subfigure}{0.14\linewidth}
     \includegraphics[width=1\textwidth]{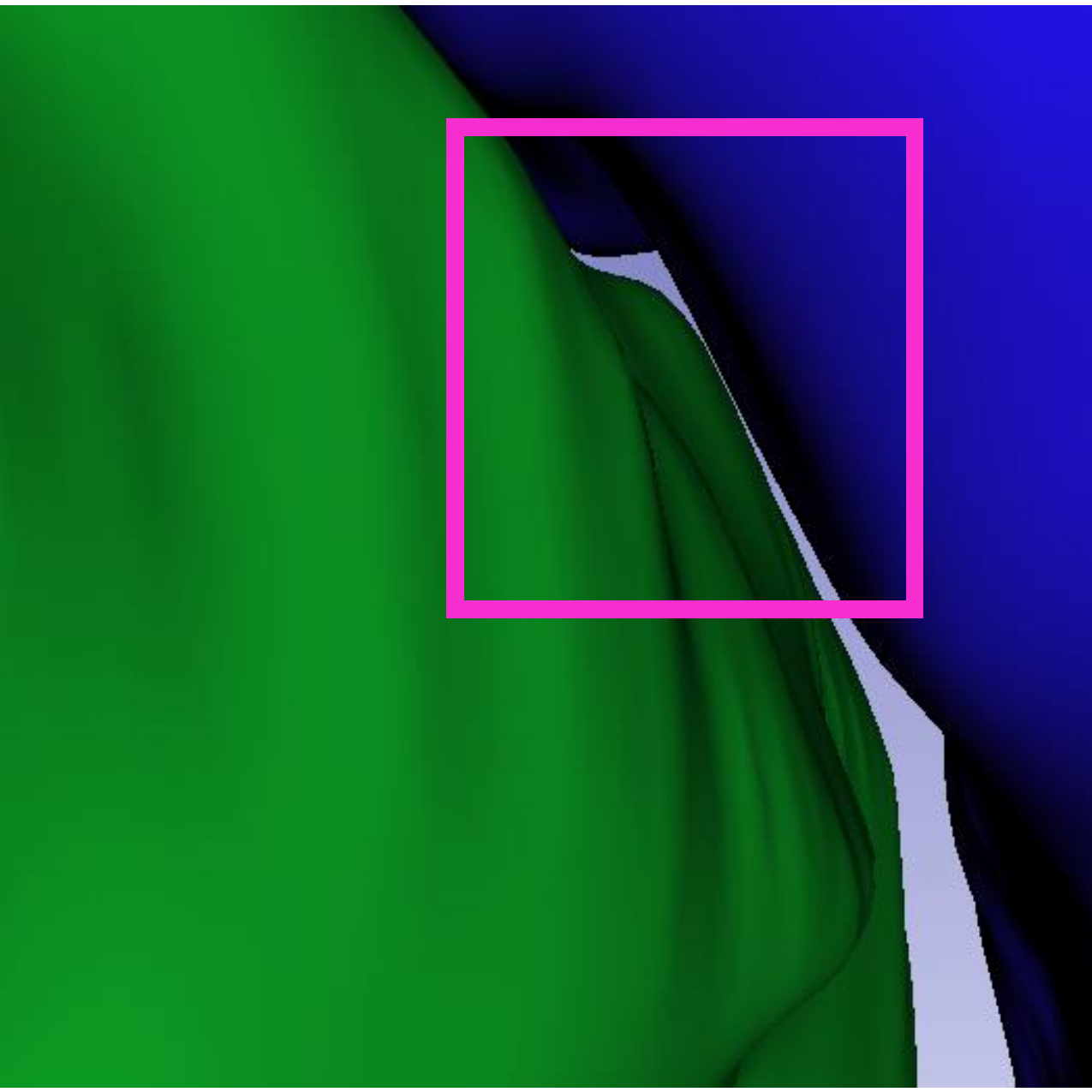}
     \caption{GT}
  \end{subfigure}

\caption{Additional qualitative Multi-Atlas results compared with the baselines. Rows 3-4 are corresponding 3D renderings. It is hard to visualize the input 3D volumetric image and so we leave it blank in the third row.  Colors for the classes correspond to the ones used in Fig.~\ref{fig:data-interactions}.}
\label{fig:multi-add-2}
\end{figure}

\begin{figure}[t]
\centering 

        \begin{subfigure}{0.15\linewidth}
  \includegraphics[width=1\textwidth]{figures/thor/sample2/input-slice.pdf}
      \caption{Input}
  \end{subfigure}
  \begin{subfigure}{0.15\linewidth}
     \includegraphics[width=1\textwidth]{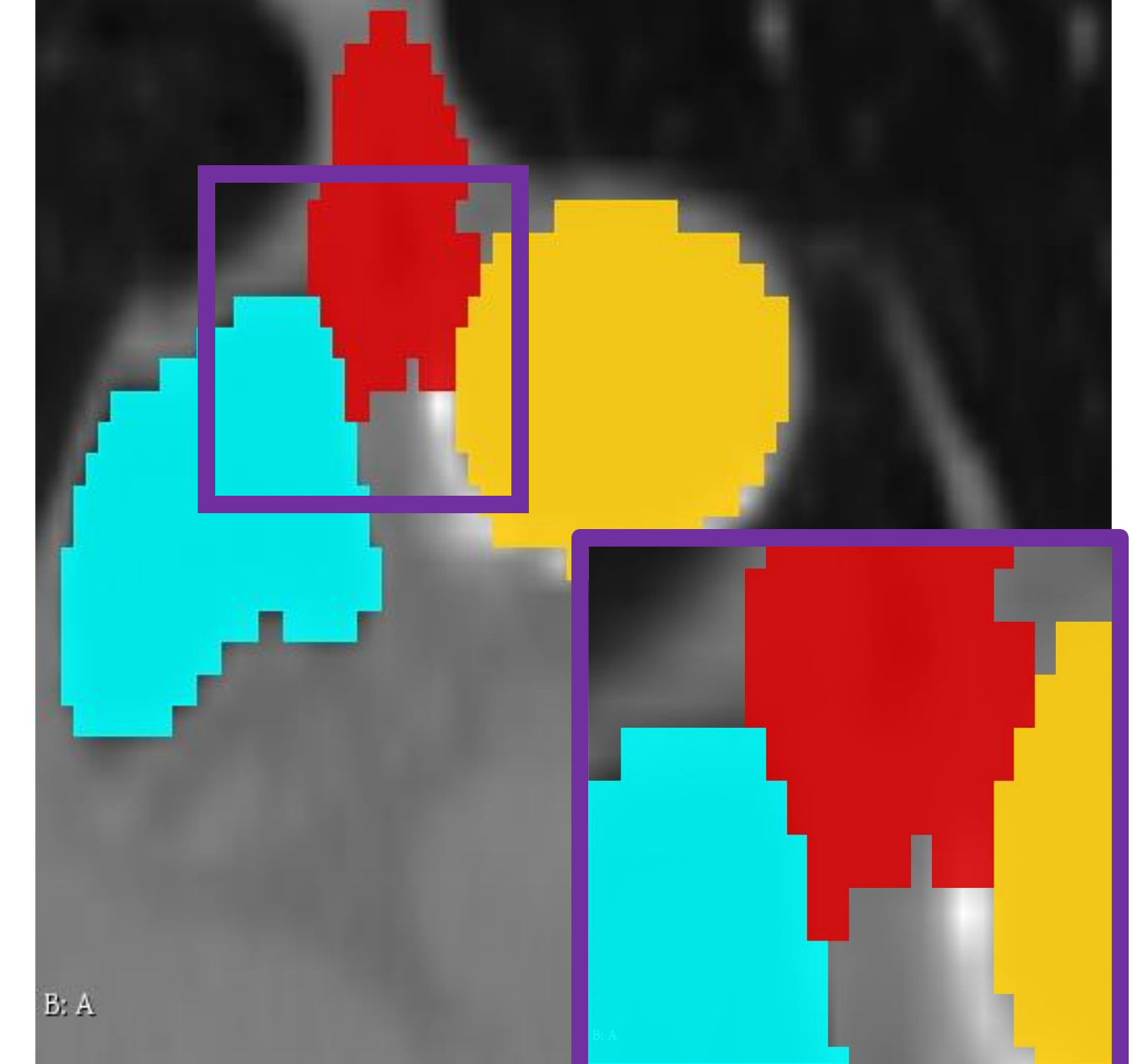}
         \caption{UNet}
  \end{subfigure}
    \begin{subfigure}{0.15\linewidth}
     \includegraphics[width=1\textwidth]{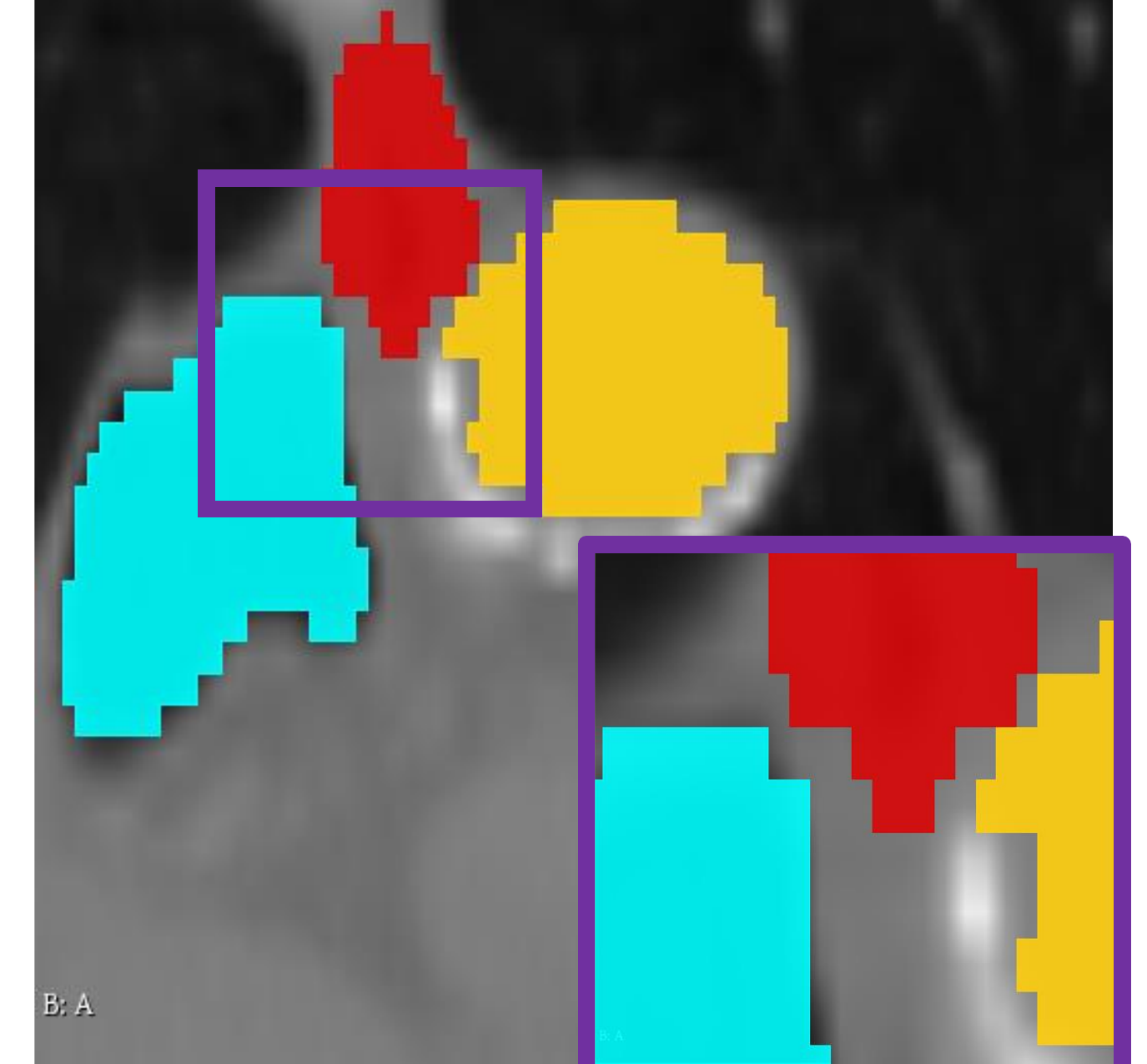}
         \caption{UNet+O}
  \end{subfigure}
    \begin{subfigure}{0.15\linewidth}
     \includegraphics[width=1\textwidth]{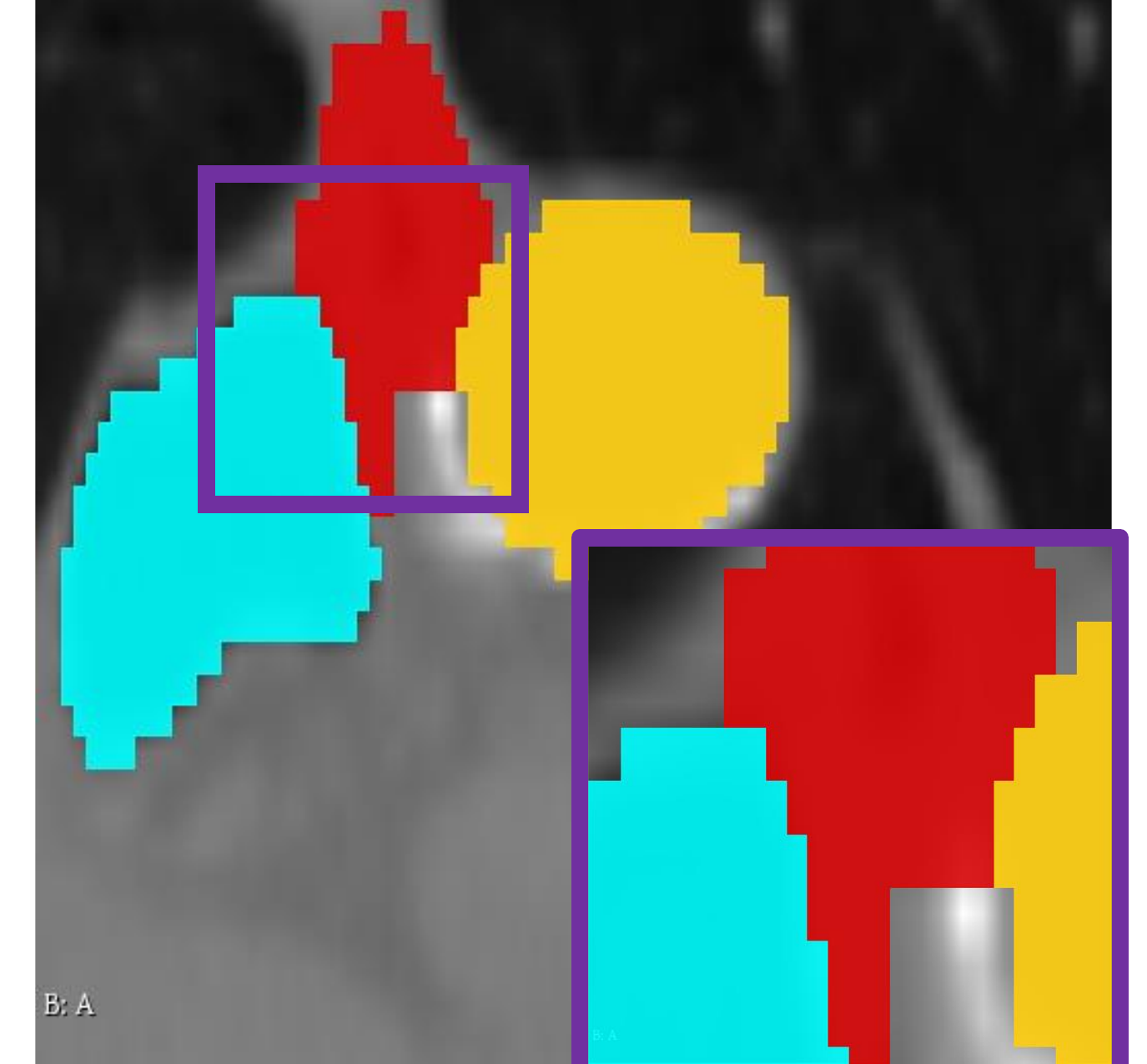}
         \caption{FCN}
  \end{subfigure}
      \begin{subfigure}{0.15\linewidth}
     \includegraphics[width=1\textwidth]{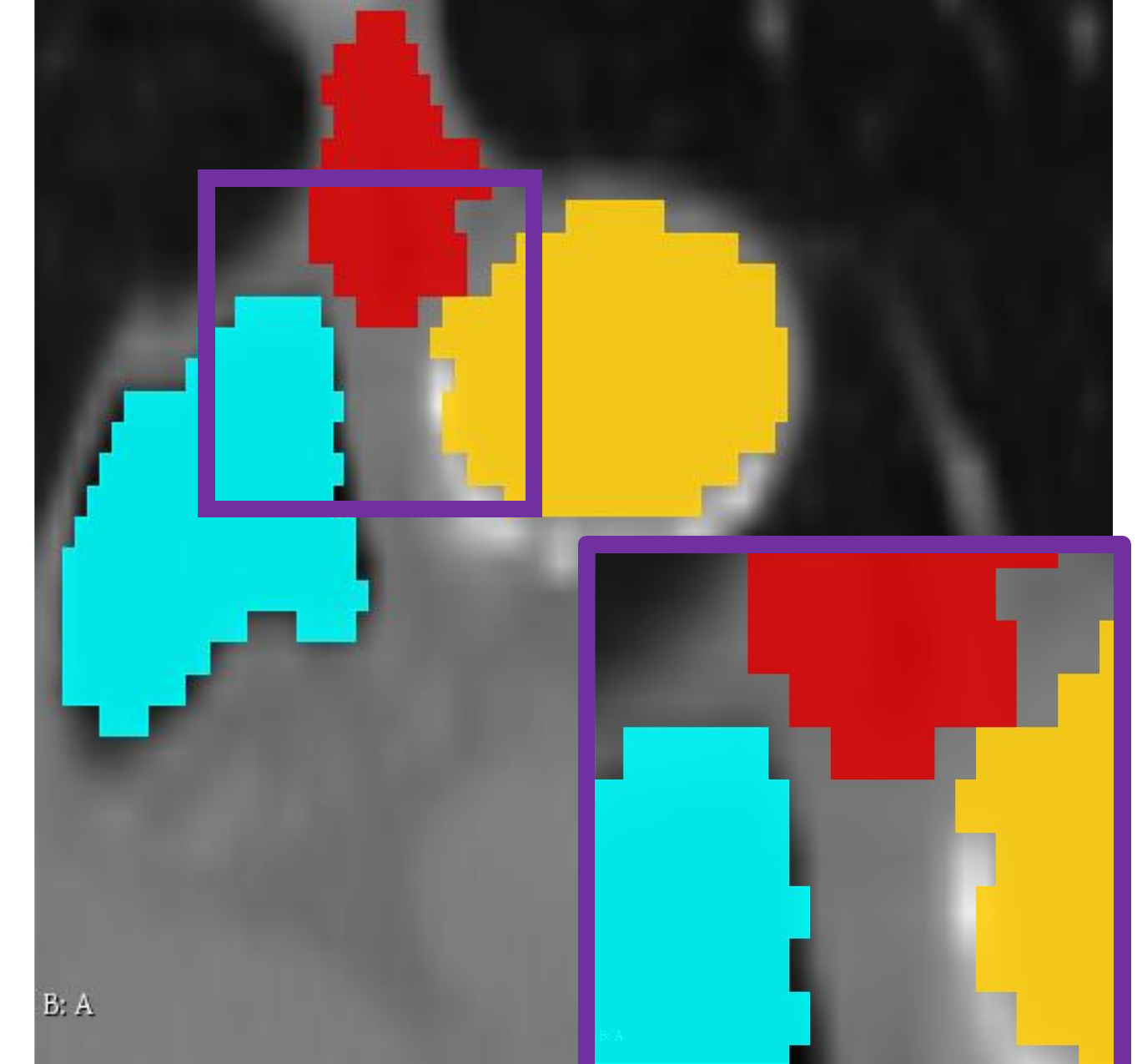}
         \caption{FCN+O}
  \end{subfigure}
    \begin{subfigure}{0.15\linewidth}
     \includegraphics[width=1\textwidth]{figures/thor/sample2/nnunet-slice.pdf}
         \caption{nnUNet}
  \end{subfigure}

      \begin{subfigure}{0.15\linewidth}
     \includegraphics[width=1\textwidth]{figures/thor/sample2/crf-slice.pdf}
     \caption{CRF}
  \end{subfigure}
      \begin{subfigure}{0.15\linewidth}
     \includegraphics[width=1\textwidth]{figures/thor/sample2/midl-slice.pdf}
         \caption{MIDL}
  \end{subfigure}
      \begin{subfigure}{0.15\linewidth}
     \includegraphics[width=1\textwidth]{figures/thor/sample2/nonadj-slice.pdf}
     \caption{NonAdj}
  \end{subfigure}
      \begin{subfigure}{0.15\linewidth}
     \includegraphics[width=1\textwidth]{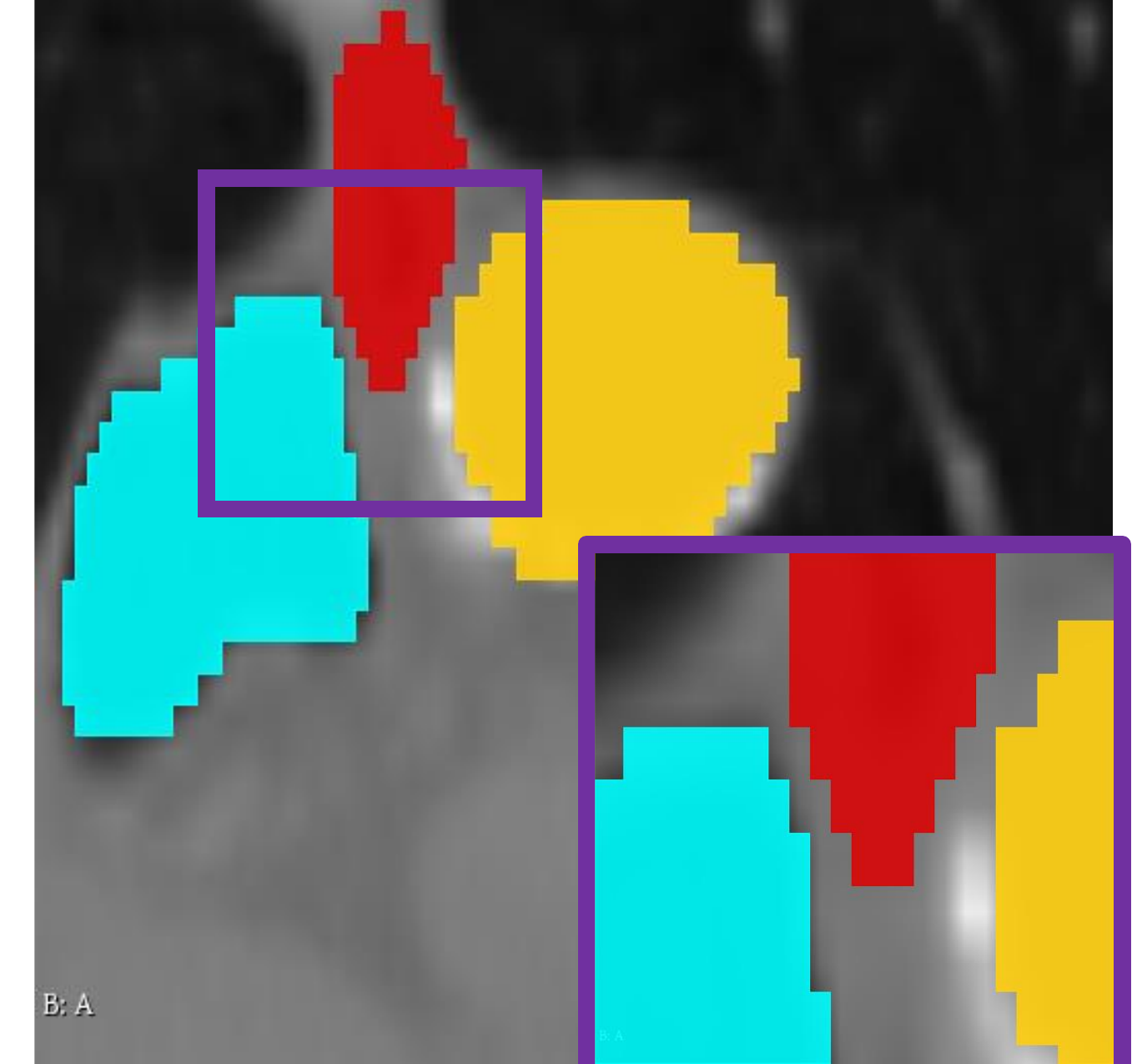}
         \caption{Ours6C}
  \end{subfigure}
        \begin{subfigure}{0.15\linewidth}
     \includegraphics[width=1\textwidth]{figures/thor/sample2/topo-slice.pdf}
         \caption{Ours}
  \end{subfigure}
      \begin{subfigure}{0.15\linewidth}
     \includegraphics[width=1\textwidth]{figures/thor/sample2/gt-slice.pdf}
         \caption{GT}
  \end{subfigure}
\caption{Qualitative SegTHOR results compared with the baselines. Colors for the classes correspond to the ones used in Fig.~\ref{fig:data-interactions}.}
\label{fig:seg-add-1}
\end{figure}

\begin{figure}[t]
\centering 
        \begin{subfigure}{0.14\linewidth}
  \includegraphics[width=1\textwidth]{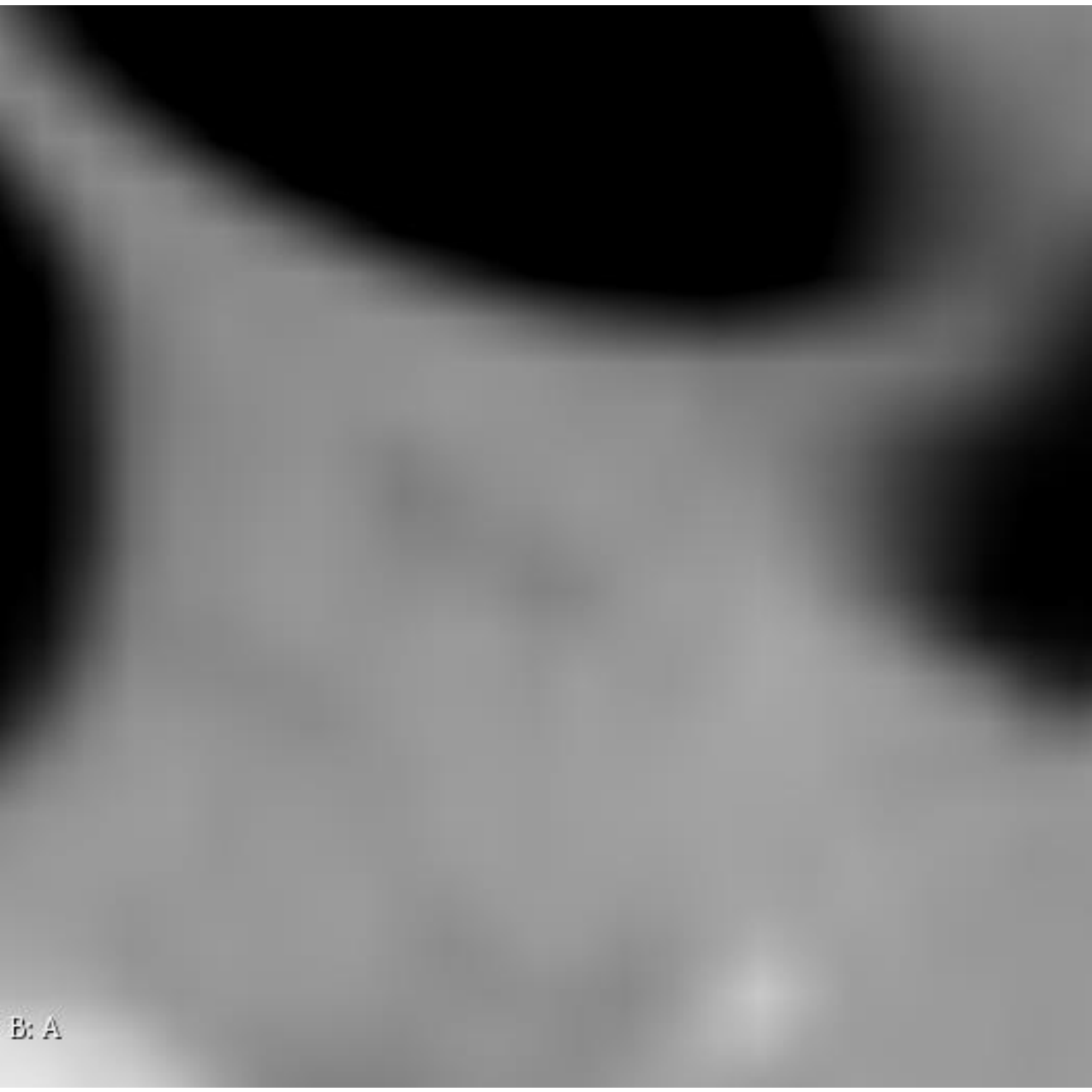}
  \caption{Input}
  \end{subfigure}
  \begin{subfigure}{0.14\linewidth}
     \includegraphics[width=1\textwidth]{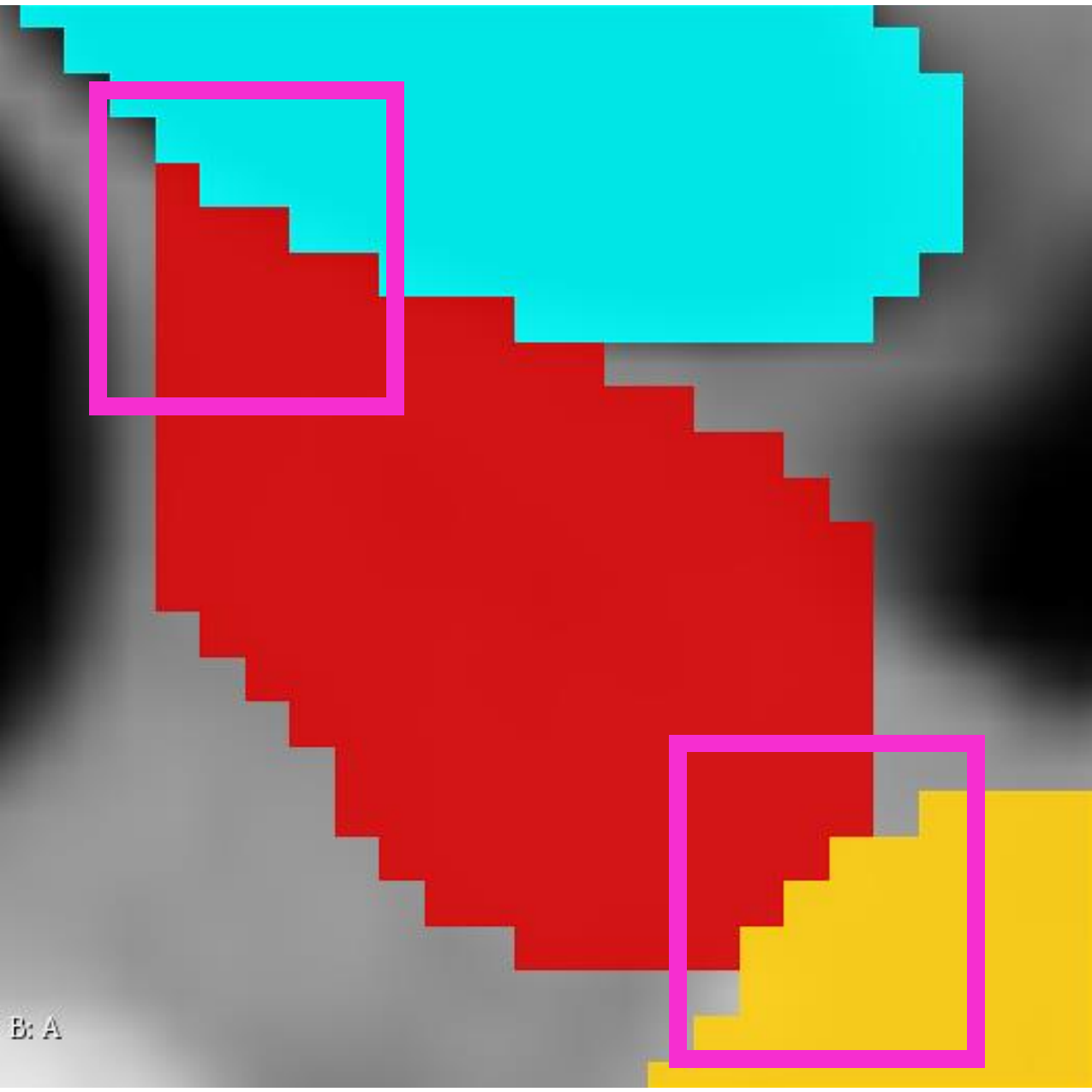}
     \caption{UNet}
  \end{subfigure}
      \begin{subfigure}{0.15\linewidth}
     \includegraphics[width=1\textwidth]{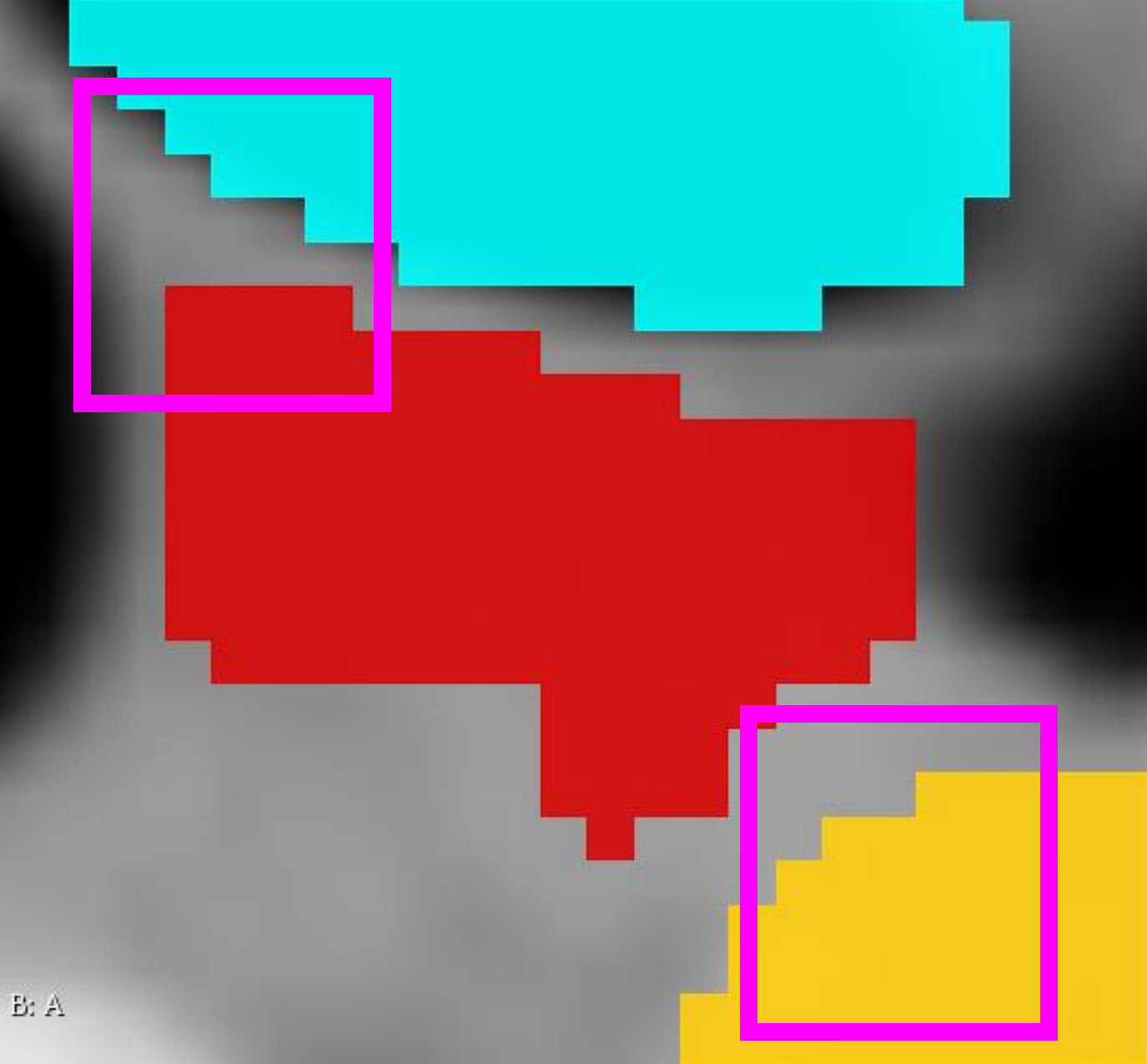}
     \caption{UNet+O}
  \end{subfigure}
    \begin{subfigure}{0.14\linewidth}
     \includegraphics[width=1\textwidth]{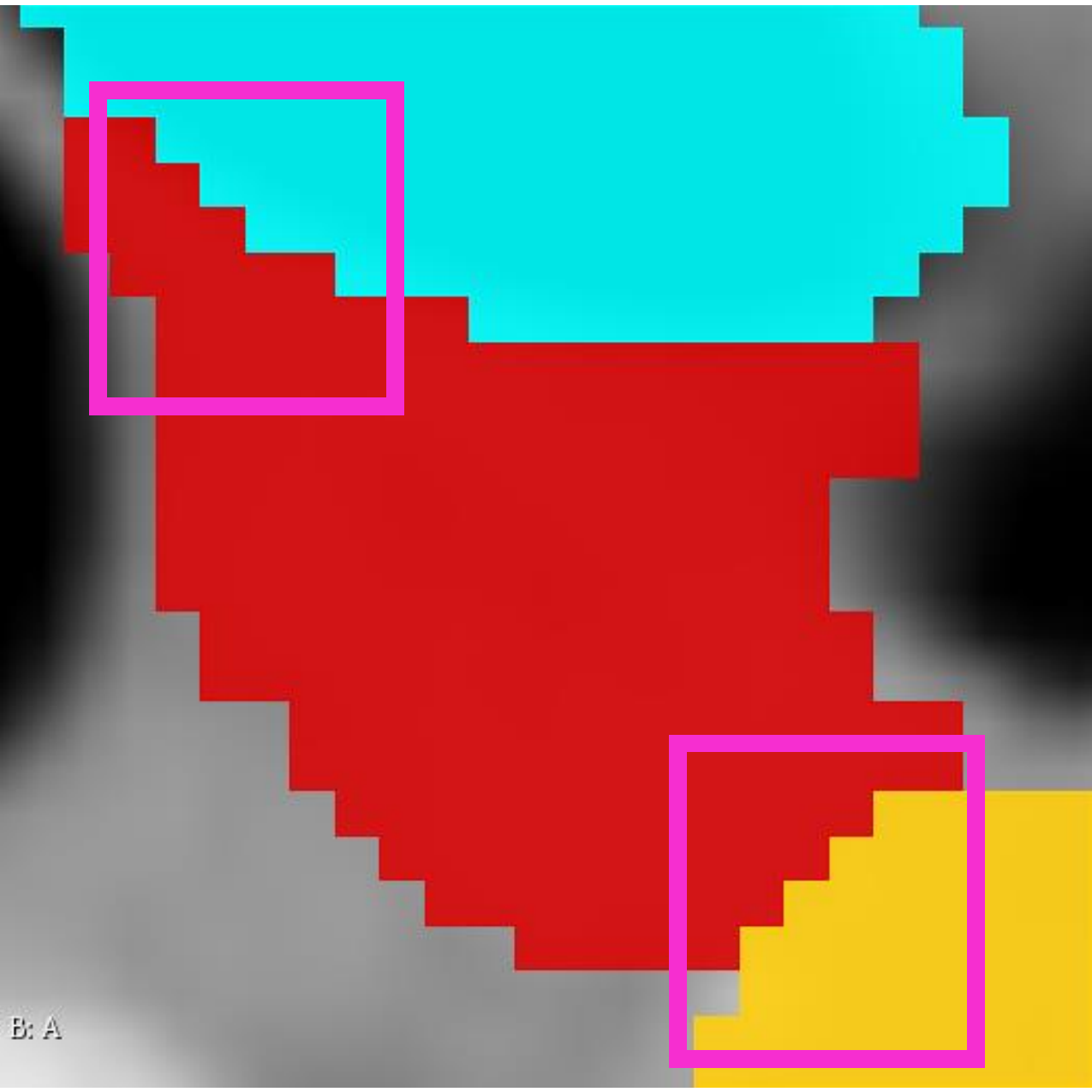}
     \caption{FCN}
  \end{subfigure}
      \begin{subfigure}{0.15\linewidth}
     \includegraphics[width=1\textwidth]{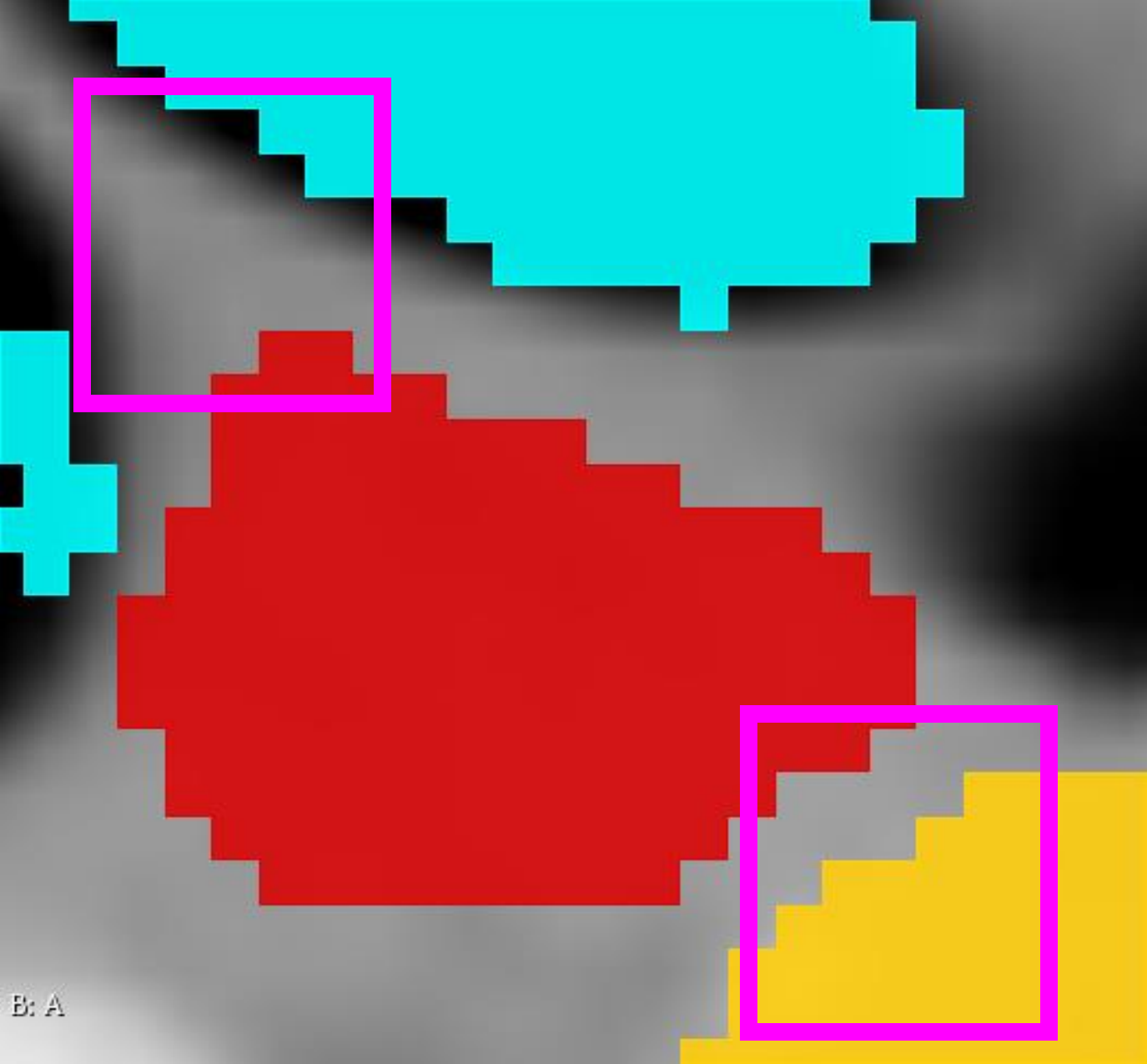}
     \caption{FCN+O}
  \end{subfigure}
    \begin{subfigure}{0.14\linewidth}
     \includegraphics[width=1\textwidth]{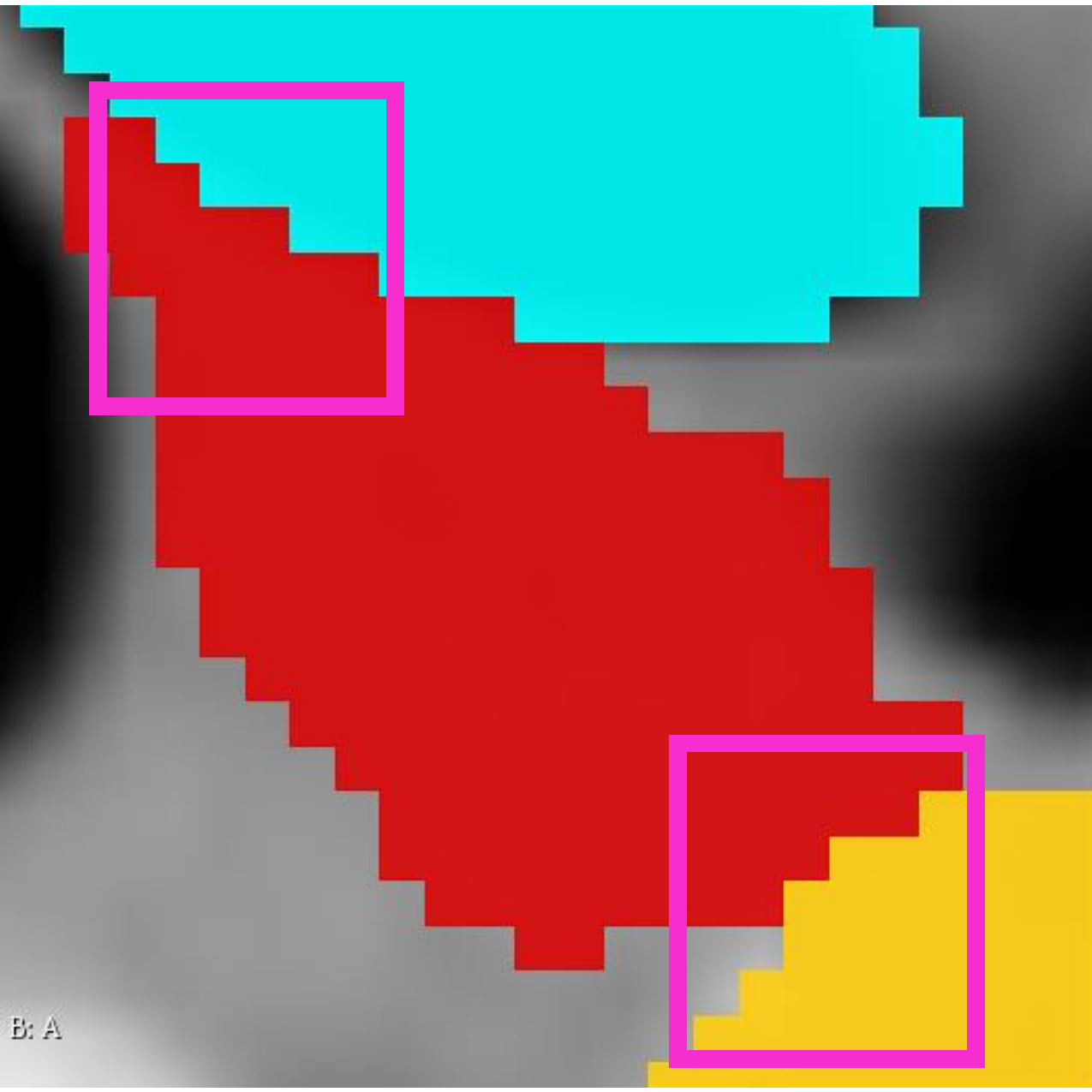}
     \caption{nnUNet}
  \end{subfigure}

      \begin{subfigure}{0.14\linewidth}
     \includegraphics[width=1\textwidth]{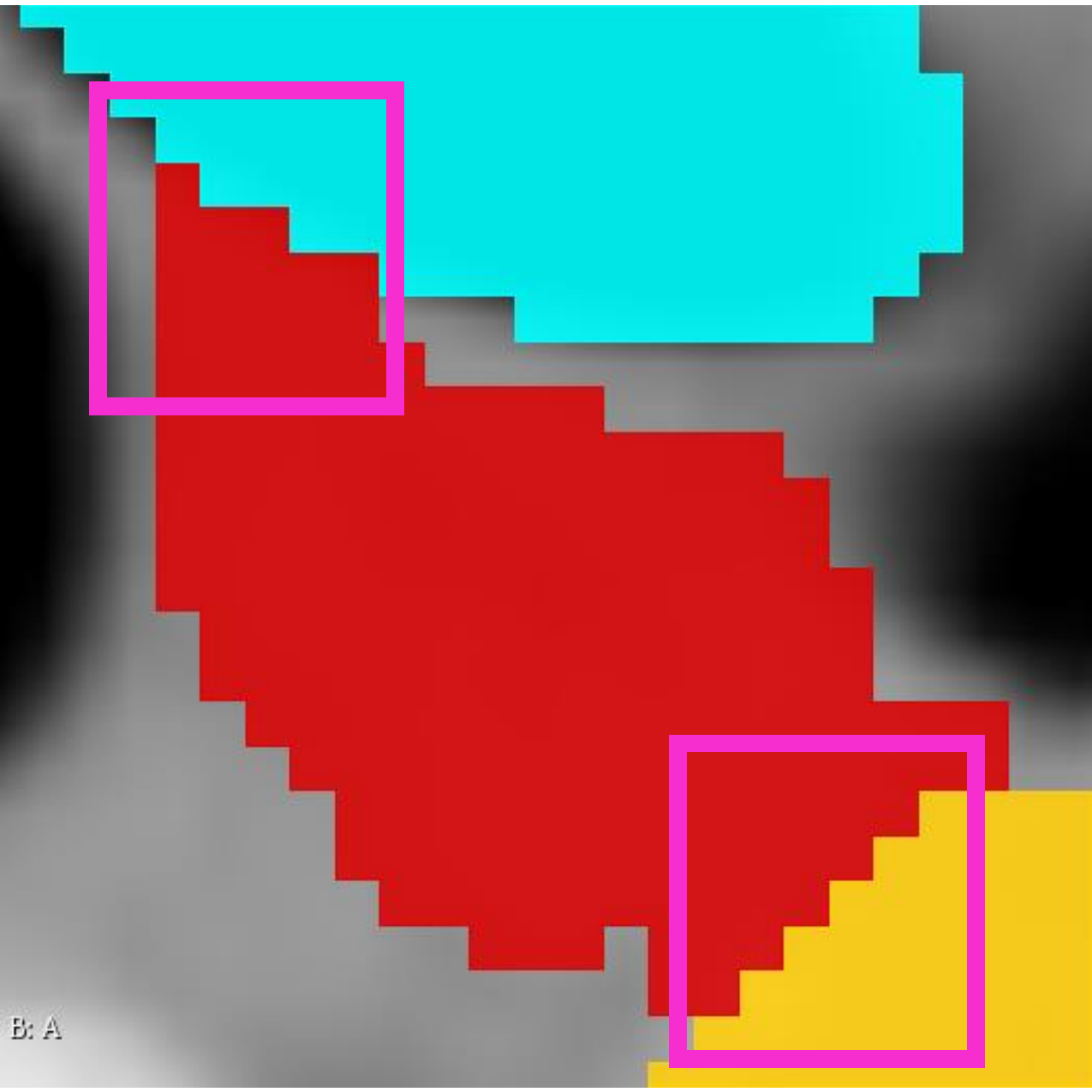}
     \caption{CRF}
  \end{subfigure}
    \begin{subfigure}{0.15\linewidth}
     \includegraphics[width=1\textwidth]{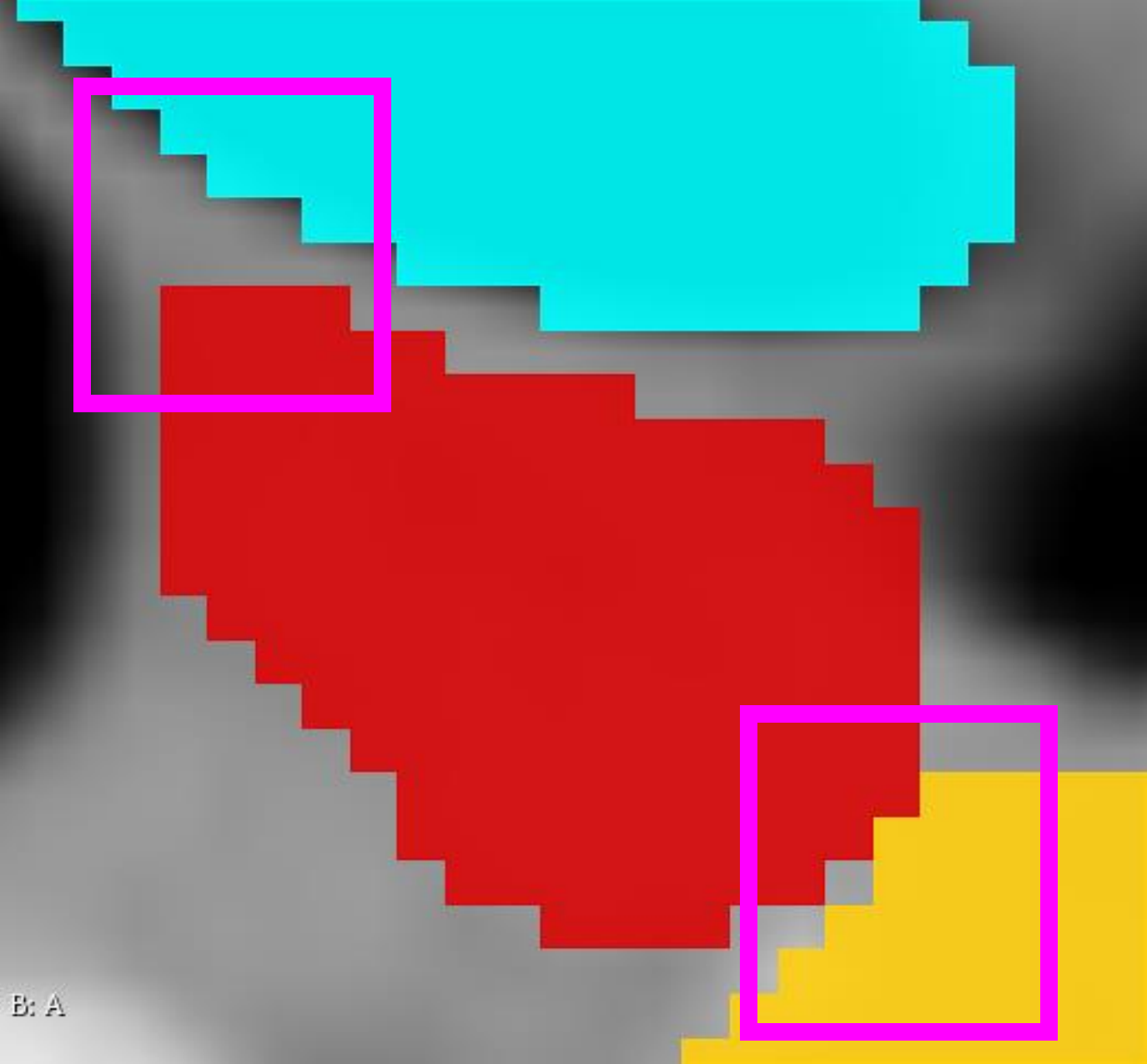}
     \caption{MIDL}
  \end{subfigure}
      \begin{subfigure}{0.15\linewidth}
     \includegraphics[width=1\textwidth]{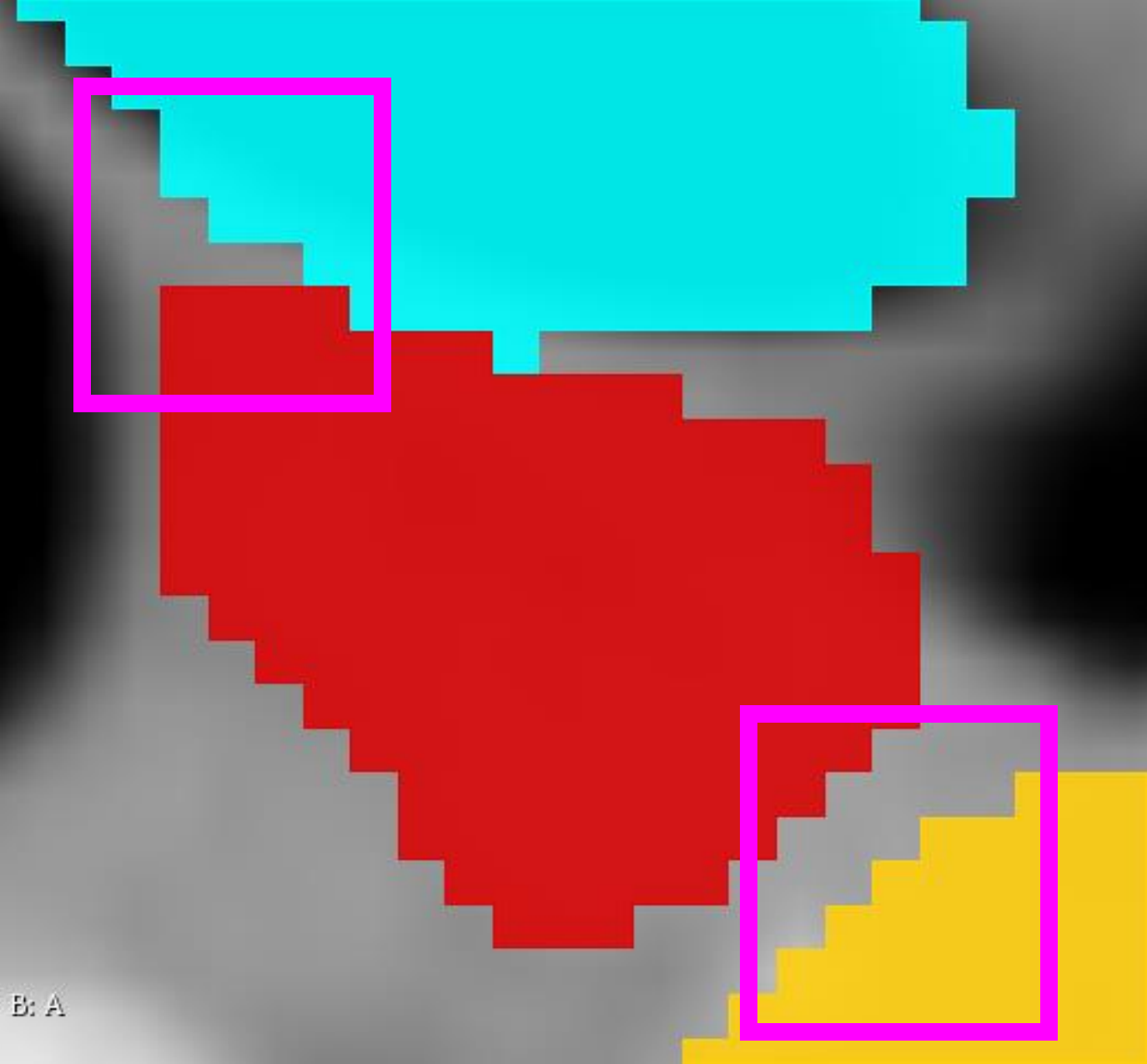}
     \caption{NonAdj}
  \end{subfigure}
      \begin{subfigure}{0.15\linewidth}
     \includegraphics[width=1\textwidth]{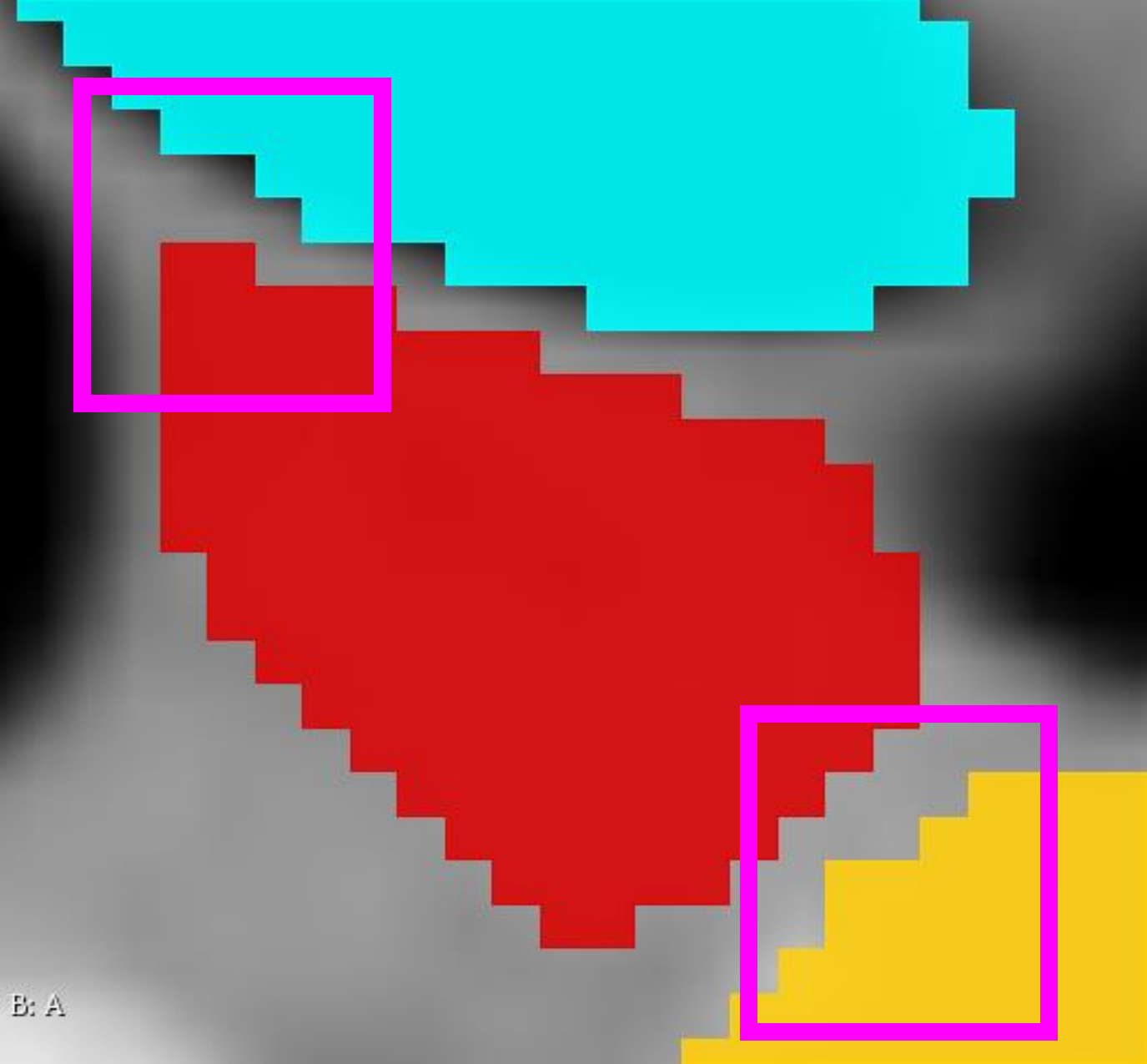}
     \caption{Ours6C}
  \end{subfigure}
        \begin{subfigure}{0.14\linewidth}
     \includegraphics[width=1\textwidth]{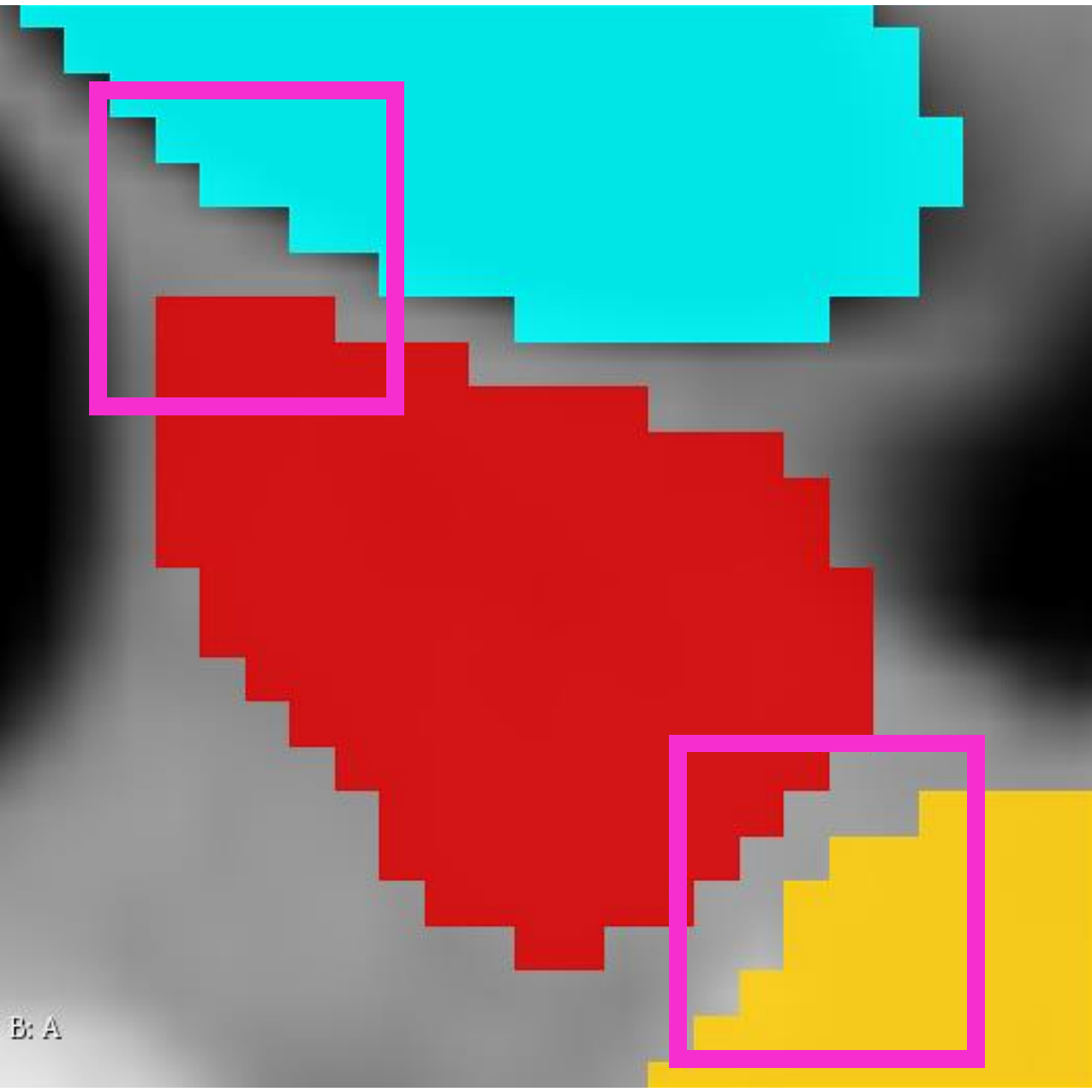}
     \caption{Ours}
  \end{subfigure}
      \begin{subfigure}{0.14\linewidth}
     \includegraphics[width=1\textwidth]{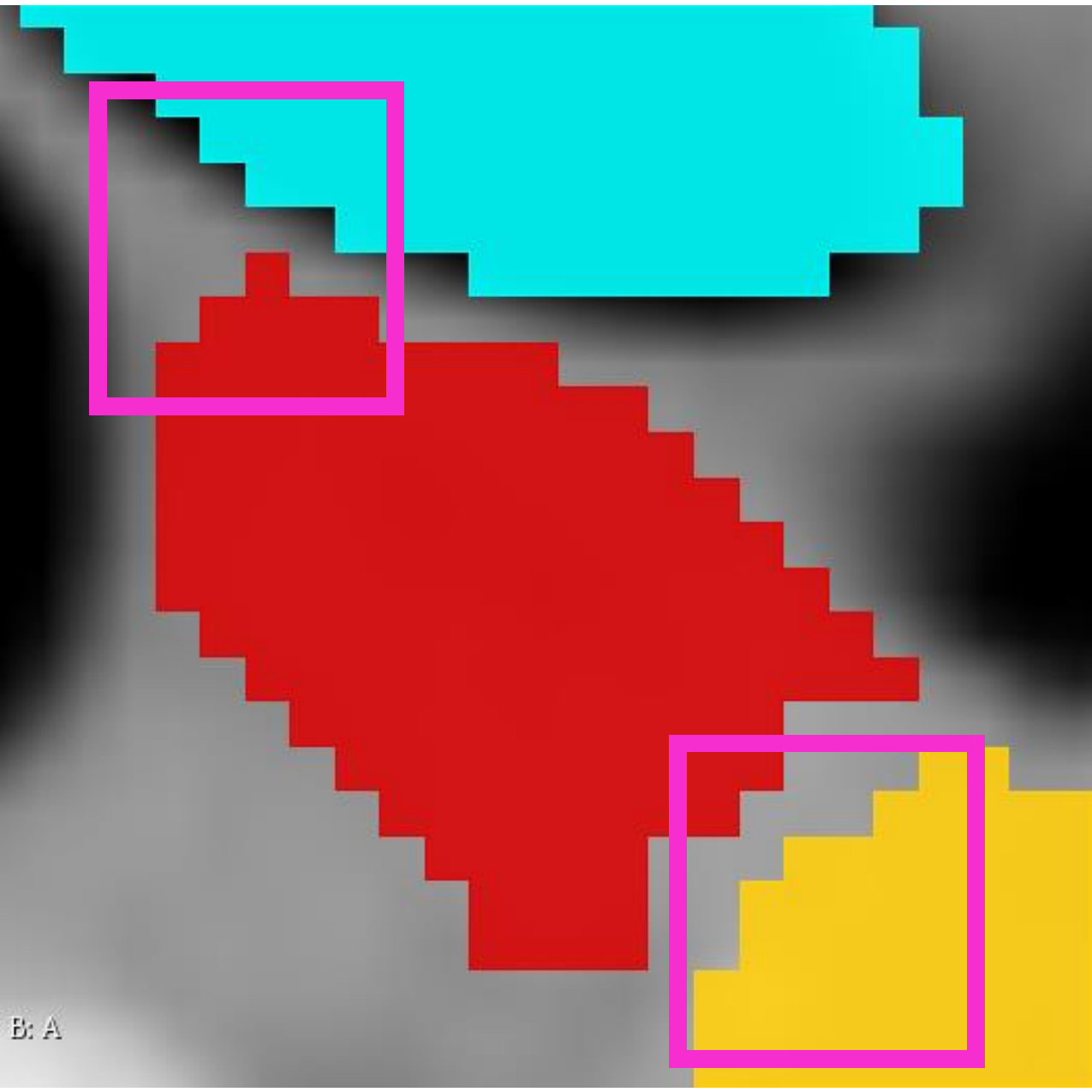}
     \caption{GT}
  \end{subfigure}

        \begin{subfigure}{0.14\linewidth}
  \includegraphics[width=1\textwidth]{figures/empty3.pdf}
  \end{subfigure}
  \begin{subfigure}{0.14\linewidth}
     \includegraphics[width=1\textwidth]{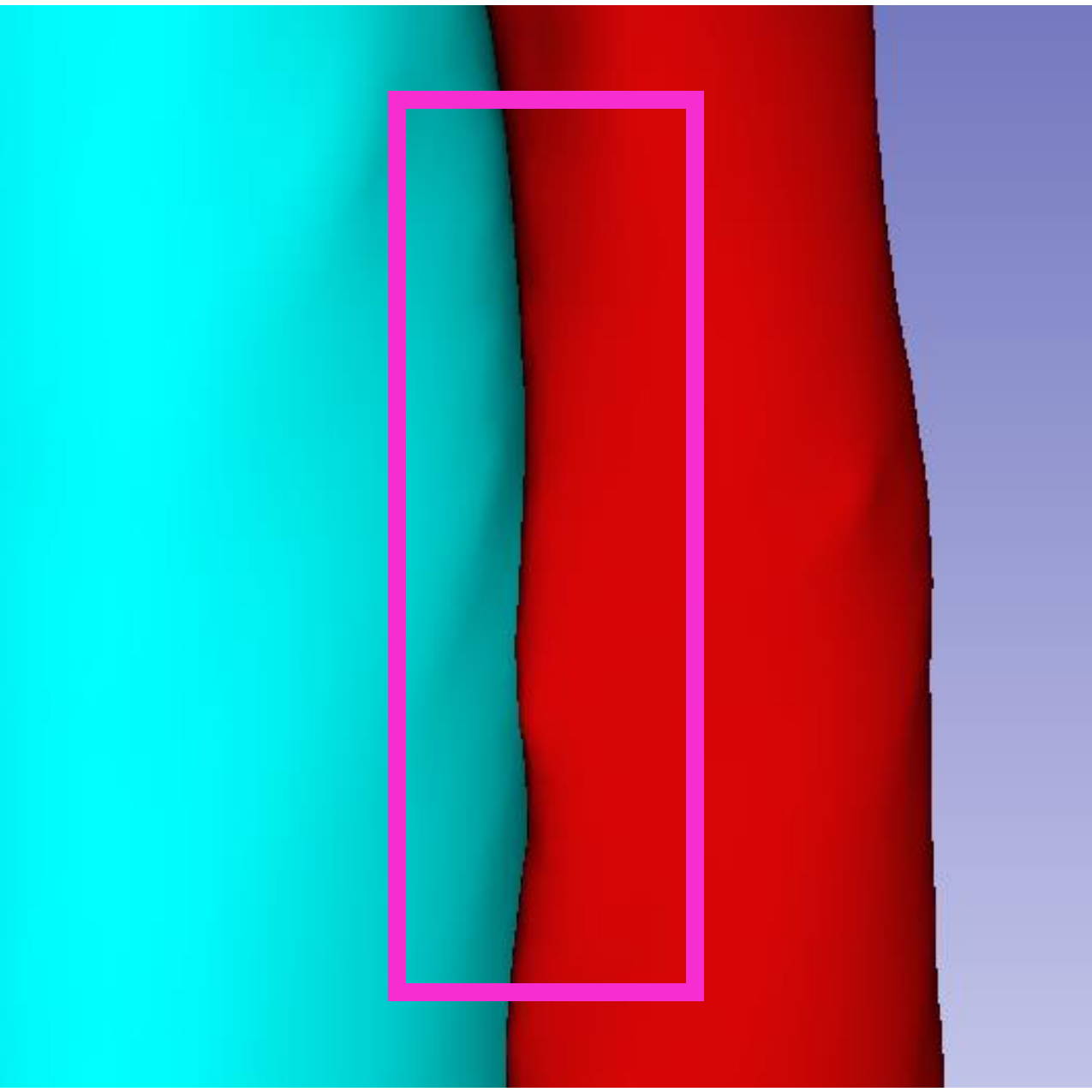}
     \caption{UNet}
  \end{subfigure}
      \begin{subfigure}{0.15\linewidth}
     \includegraphics[width=1\textwidth]{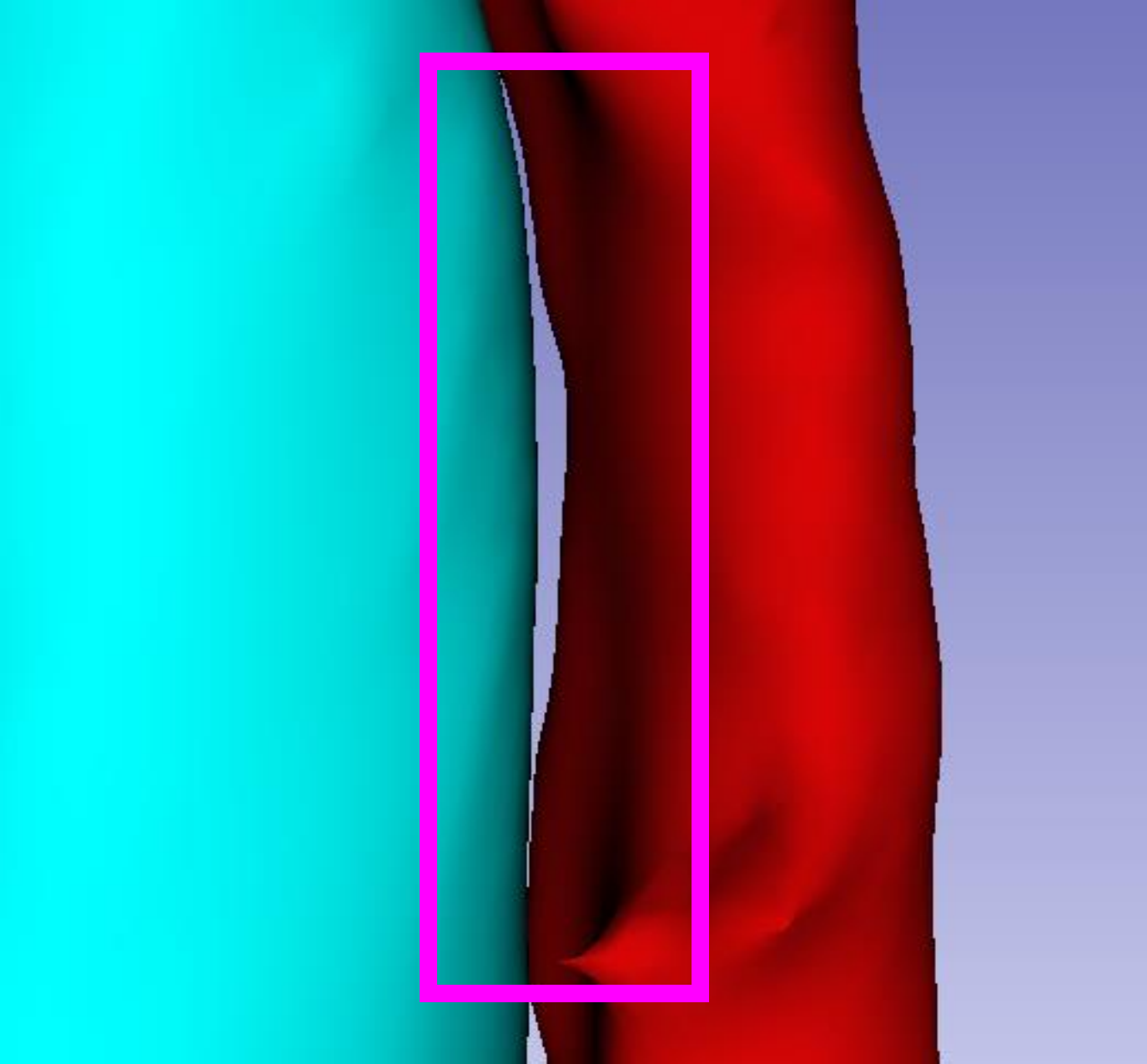}
     \caption{UNet+O}
  \end{subfigure}
    \begin{subfigure}{0.14\linewidth}
     \includegraphics[width=1\textwidth]{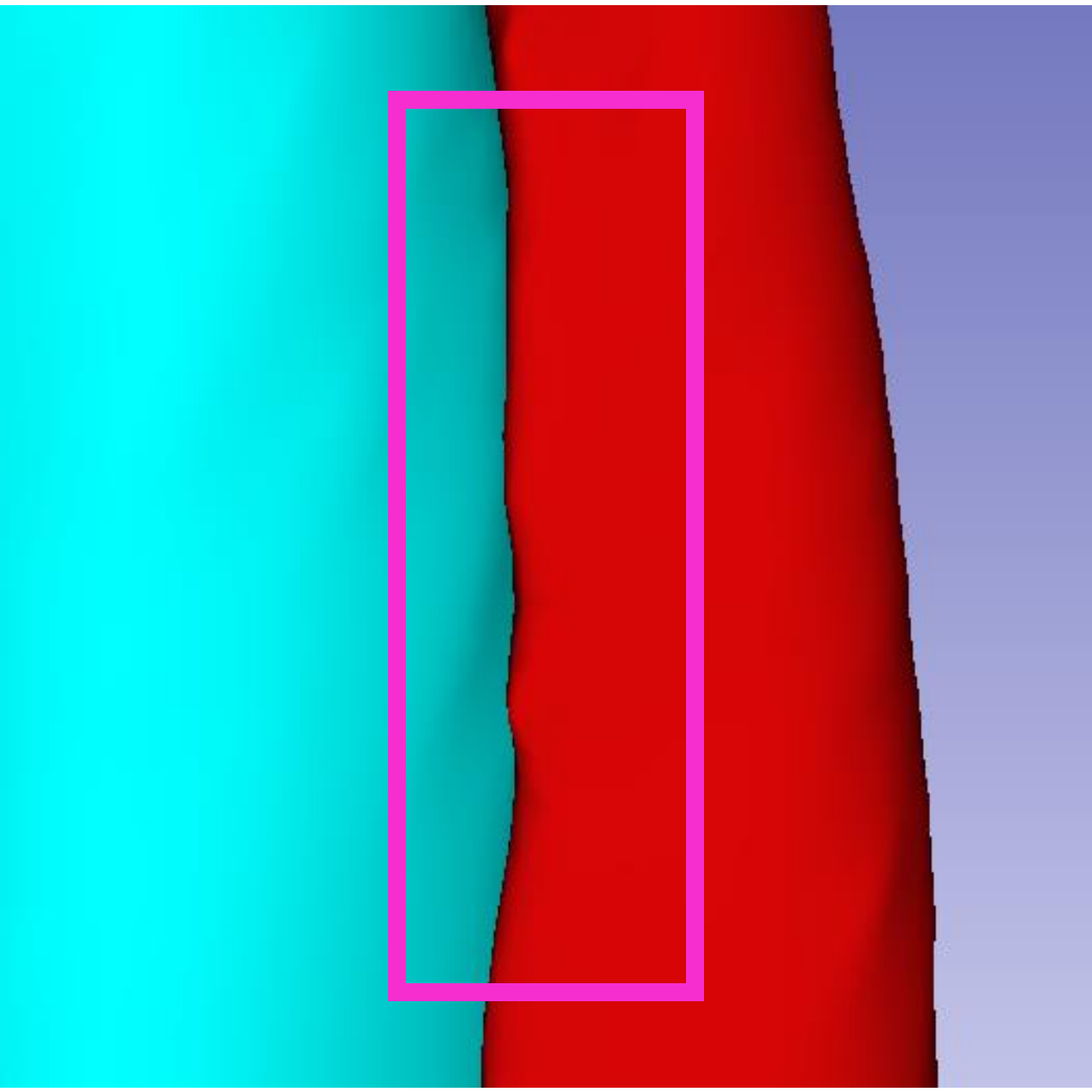}
     \caption{FCN}
  \end{subfigure}
      \begin{subfigure}{0.15\linewidth}
     \includegraphics[width=1\textwidth]{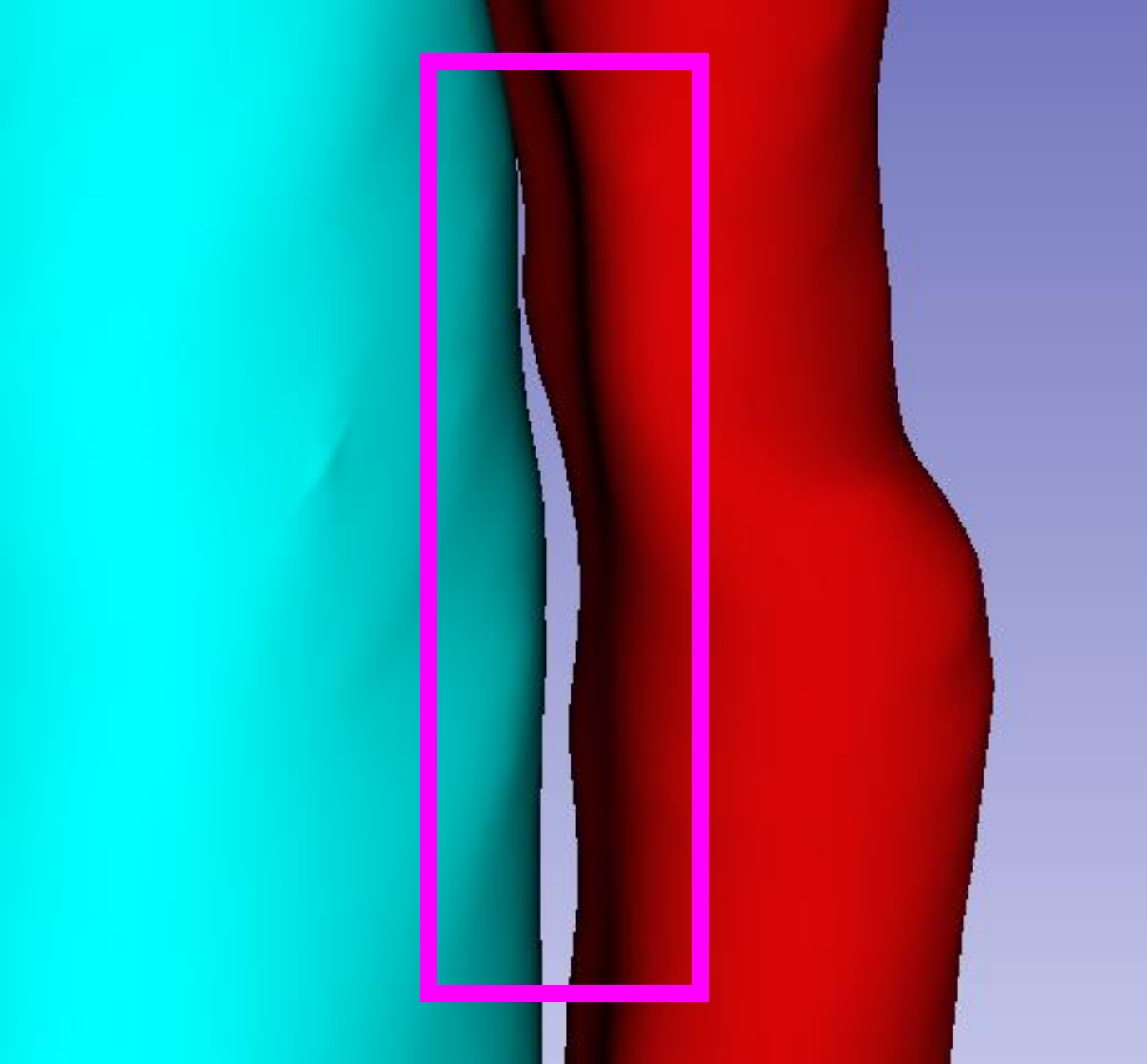}
     \caption{FCN+O}
  \end{subfigure}
    \begin{subfigure}{0.14\linewidth}
     \includegraphics[width=1\textwidth]{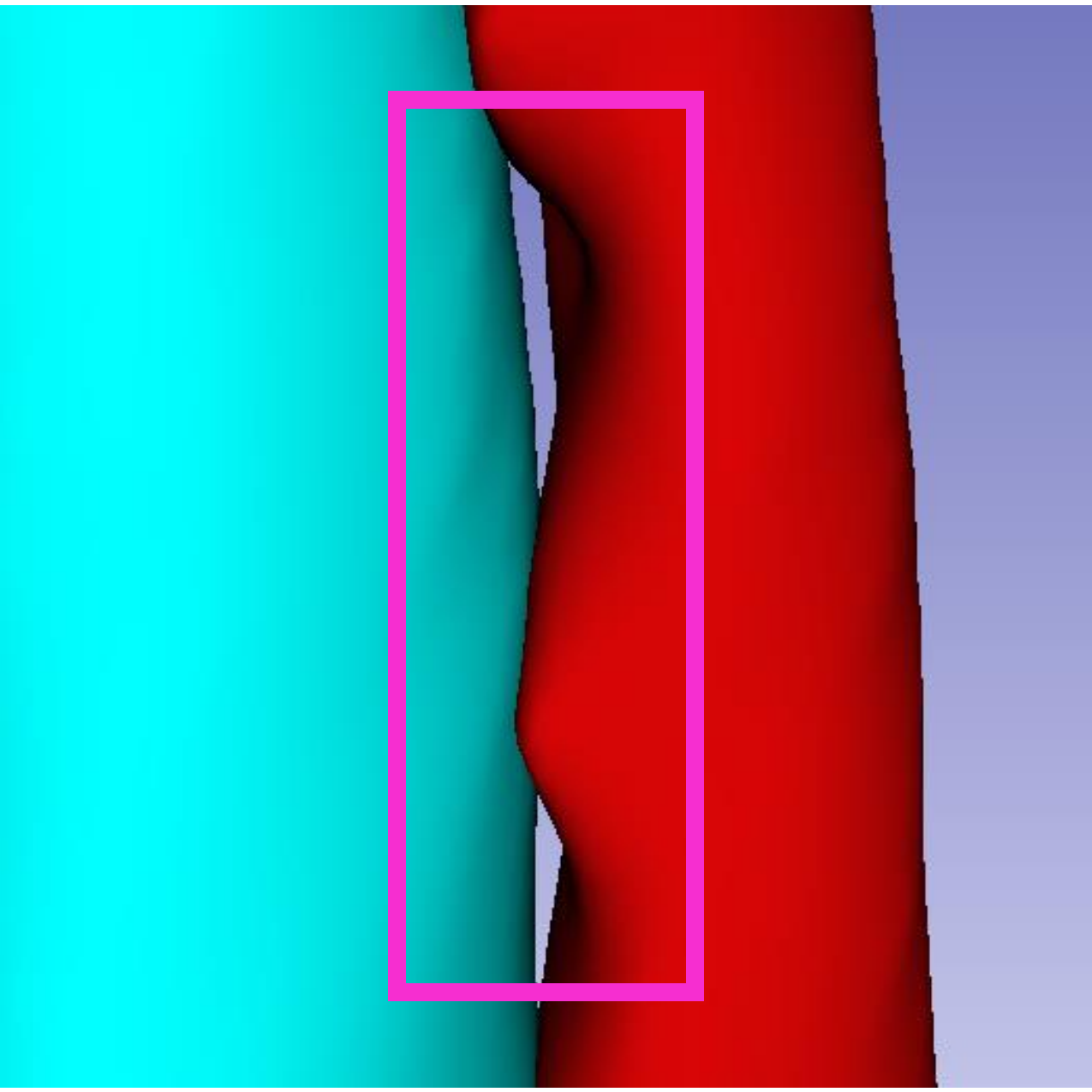}
     \caption{nnUNet}
  \end{subfigure}

      \begin{subfigure}{0.14\linewidth}
     \includegraphics[width=1\textwidth]{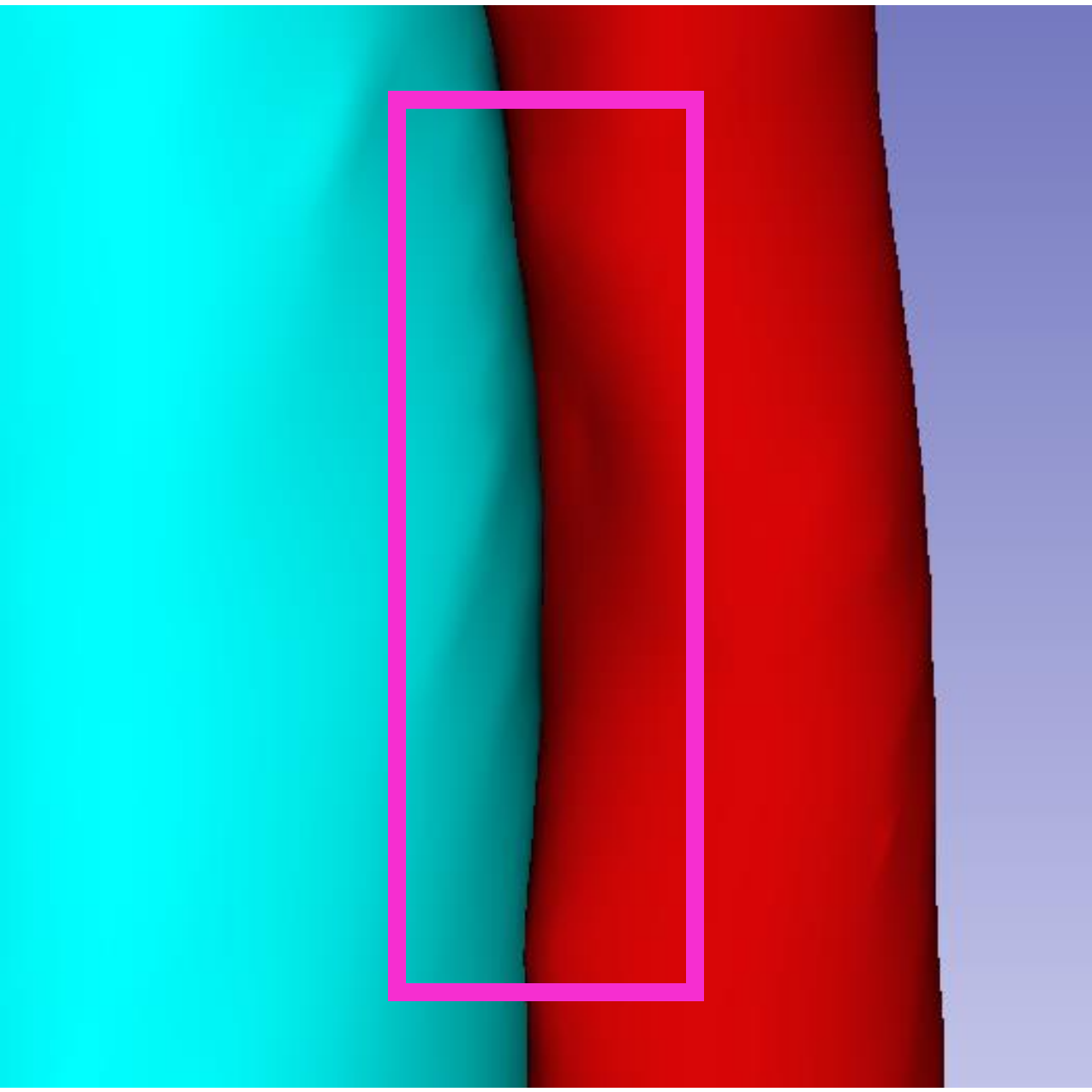}
     \caption{CRF}
  \end{subfigure}
      \begin{subfigure}{0.15\linewidth}
     \includegraphics[width=1\textwidth]{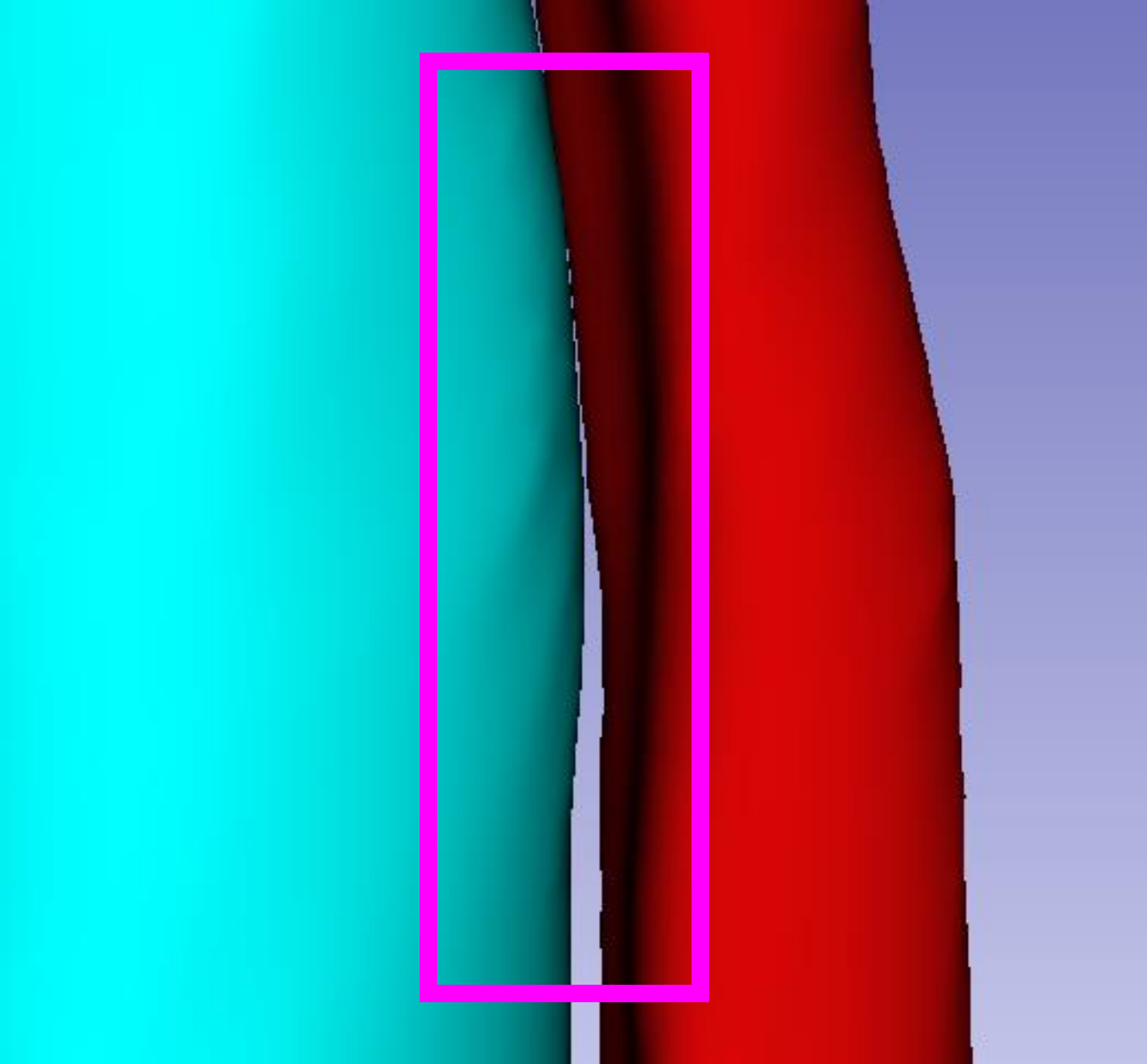}
     \caption{MIDL}
  \end{subfigure}
      \begin{subfigure}{0.15\linewidth}
     \includegraphics[width=1\textwidth]{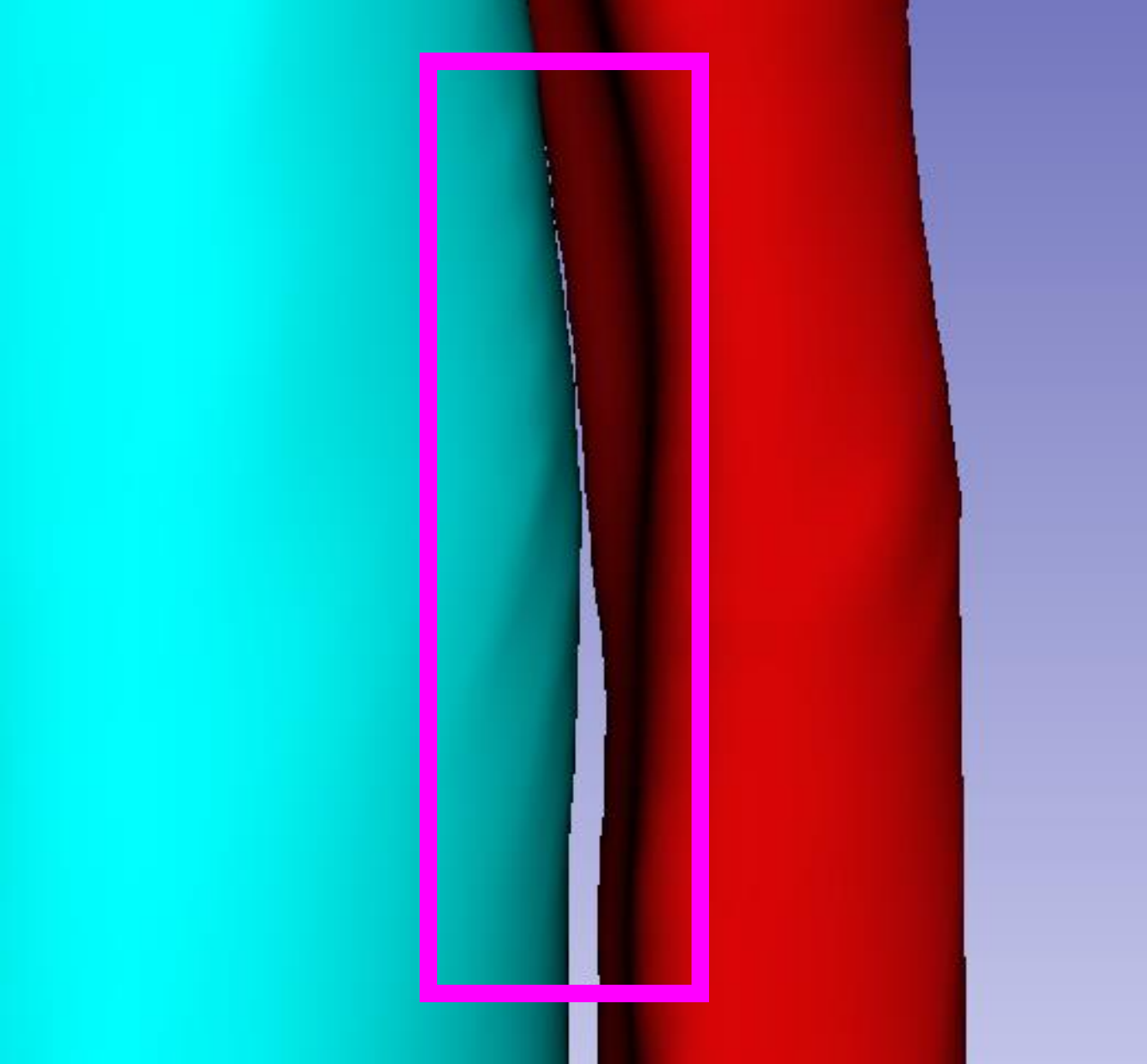}
     \caption{NonAdj}
  \end{subfigure}
      \begin{subfigure}{0.15\linewidth}
     \includegraphics[width=1\textwidth]{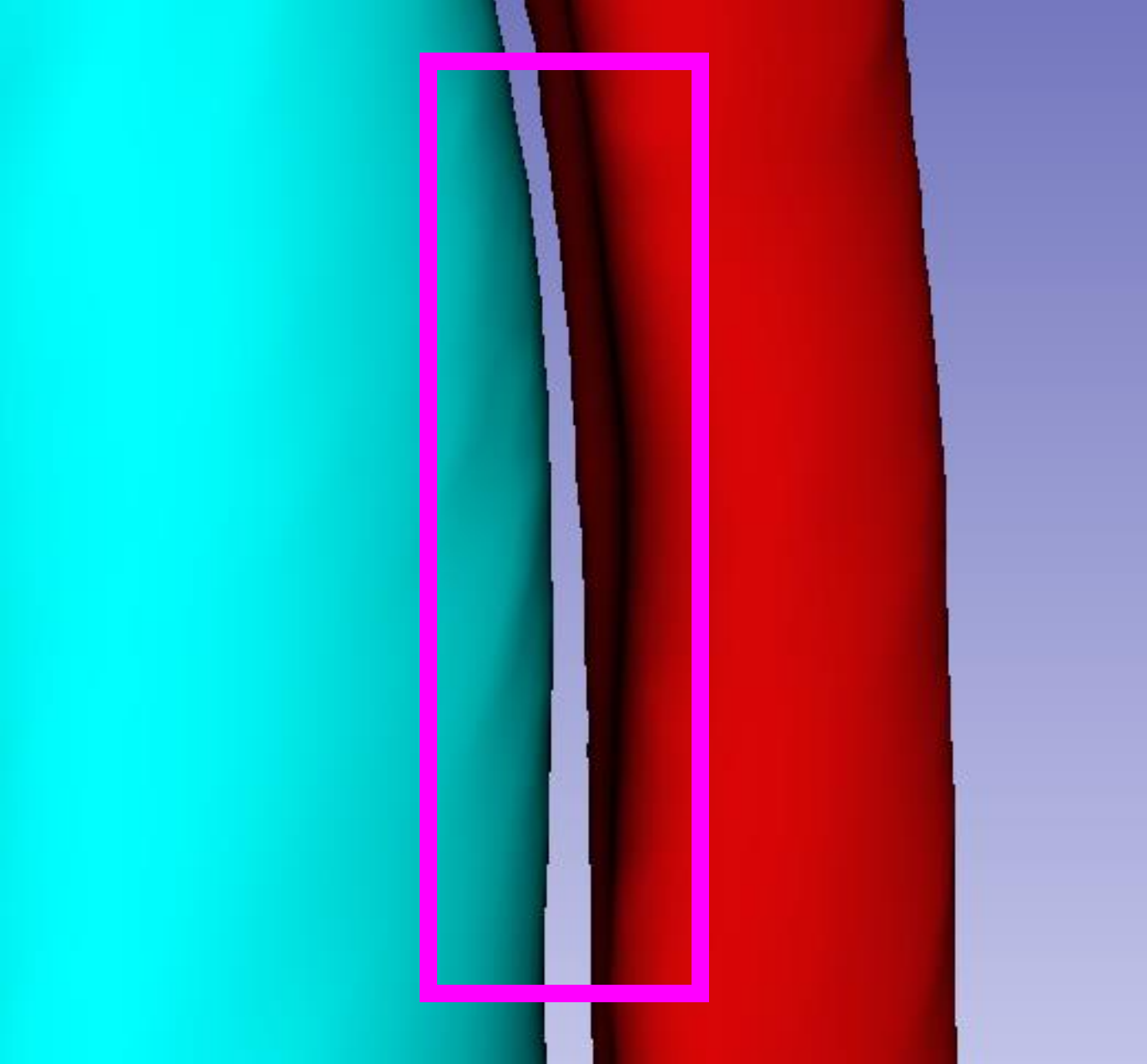}
     \caption{Ours6C}
  \end{subfigure}
        \begin{subfigure}{0.14\linewidth}
     \includegraphics[width=1\textwidth]{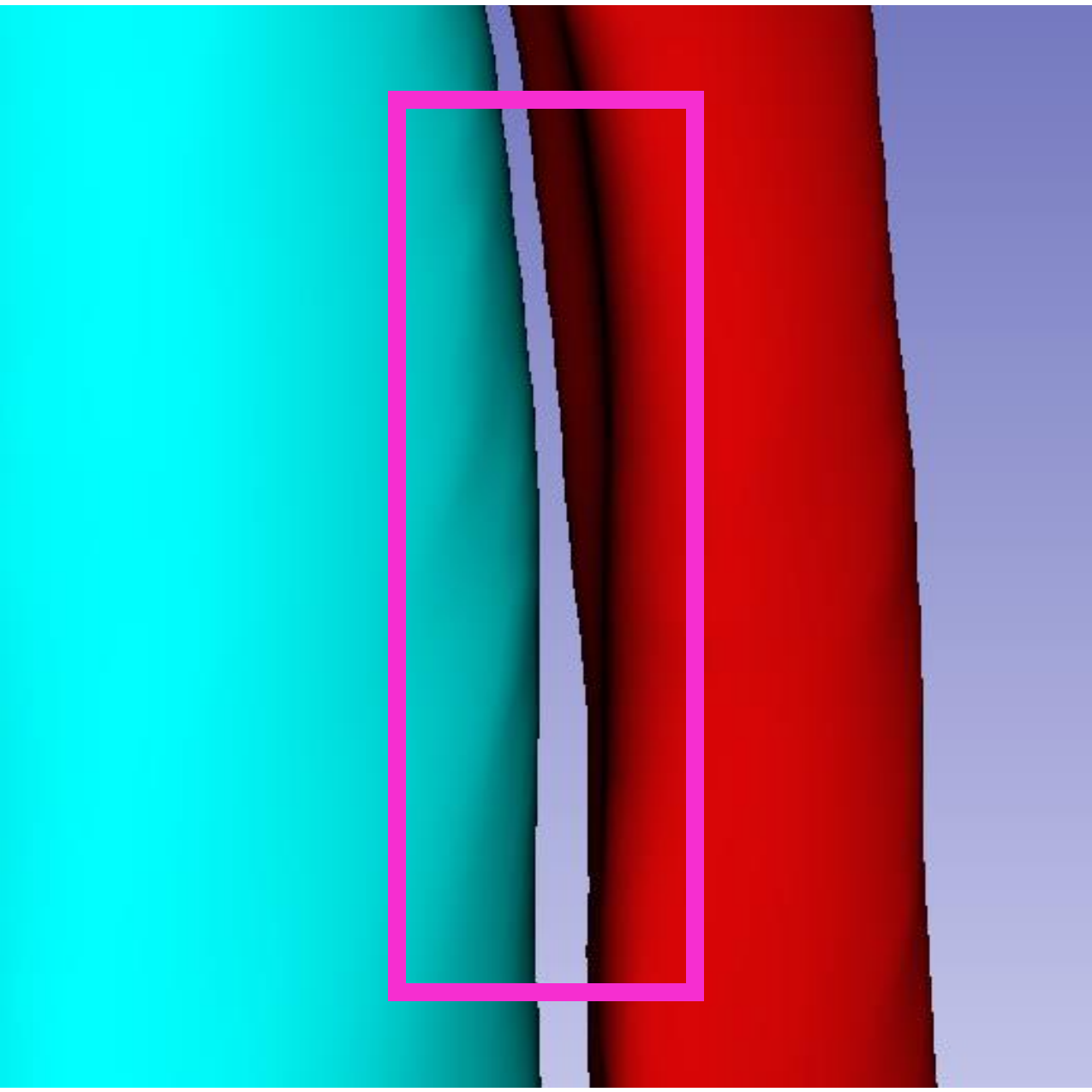}
     \caption{Ours}
  \end{subfigure}
      \begin{subfigure}{0.14\linewidth}
     \includegraphics[width=1\textwidth]{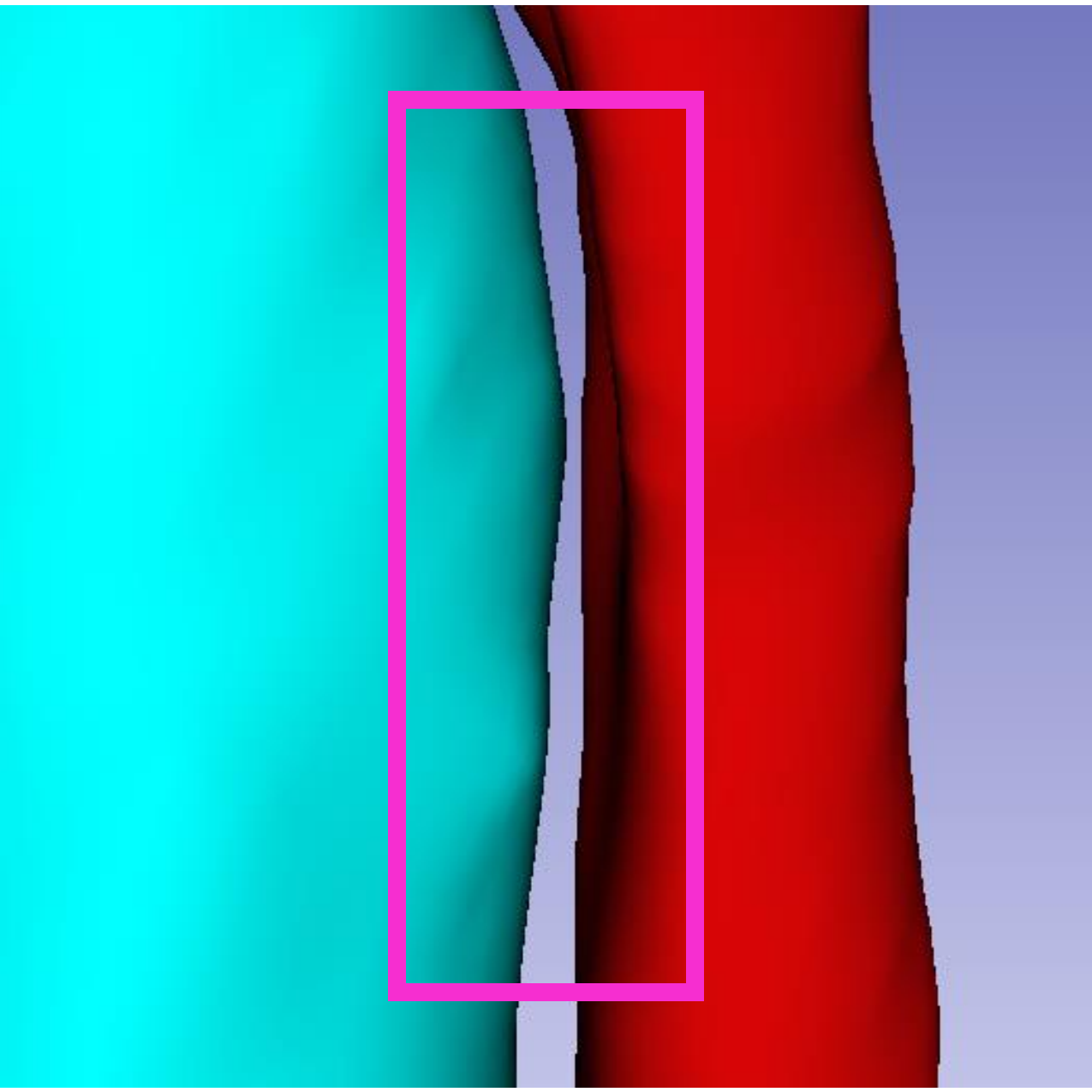}
     \caption{GT}
  \end{subfigure}

\caption{Additional qualitative SegTHOR results compared with the baselines. Rows 3-4 are corresponding 3D renderings. It is hard to visualize the input 3D volumetric image and so we leave it blank in the third row. Colors for the classes correspond to the ones used in Fig.~\ref{fig:data-interactions}.}
\label{fig:seg-add-2}
\end{figure}

We provide additional qualitative results in Fig.~\ref{fig:aorta-add-1}, ~\ref{fig:aorta-add-2}, ~\ref{fig:ivus-add-1}, ~\ref{fig:ivus-add-2}, ~\ref{fig:ivus-add-3}, ~\ref{fig:multi-add-1}, ~\ref{fig:multi-add-2}, ~\ref{fig:seg-add-1}, and ~\ref{fig:seg-add-2}. In the figure sub-captions, `UNet+O' denotes our method trained on the UNet backbone. Similarly, `FCN+O' denotes our method trained on the FCN backbone. `Ours' denotes our method trained on nnUNet with the default connectivity kernel. `Ours4C' and `Ours6C' denotes our method trained on nnUNet with the $4$-connectivity and $6$-connectivity kernel respectively.


 \setlength{\tabcolsep}{2pt}
\begin{table*}
  \centering
  \scriptsize
\caption{Quantitative comparison for Aorta dataset (containment constraint)}
\label{comp-aorta}
\begin{tabular}{c c c c c c} 
 \hline
 \textbf{Class} & \textbf{Model} & \textbf{Dice}$\uparrow$ & \textbf{HD}$\downarrow$ & \textbf{ASSD}$\downarrow$ & \textbf{\% Violations}$\downarrow$ \\
 \hline \hline
\multirow{10}{*}{\textbf{{\rotatebox[origin=c]{90}{Lumen}}}}  
& UNet~\cite{cciccek20163d} & 0.900 $\pm$ 0.016 & 64.392 $\pm$ 16.874 & 9.315 $\pm$	1.749 & 13.994 $\pm$ 1.809 \\
~& UNet~\cite{cciccek20163d} + Ours &  \textbf{0.918 $\pm$ 0.012}  &  \textbf{41.039 $\pm$ 10.952}  &  \textbf{6.415 $\pm$ 1.403}  & \textbf{7.734 $\pm$ 2.174} \\\cline{2-6}
~& FCN~\cite{FCN8s}  & \textit{0.894 $\pm$ 0.013} & 57.974 $\pm$ 19.756 & 9.77 $\pm$	1.421 & 15.675 $\pm$ 2.409\\
~& FCN~\cite{FCN8s} + Ours &  0.892 $\pm$ 0.031 & \textbf{ 47.772 $\pm$ 14.571 } &  \textbf{7.741 $\pm$ 1.385}  & \textbf{9.797 $\pm$ 1.707} \\\cline{2-6}
~& nnUNet~\cite{nnUNet} & 0.906 $\pm$ 0.020 & 36.368 $\pm$ 12.559 & 4.563 $\pm$ 0.675 &  5.424 $\pm$  2.461\\
~& Topo-CRF~\cite{bentaieb2016topology} &  0.897 $\pm$ 0.057 &  40.162 $\pm$ 18.687  & 5.952  $\pm$ 0.999 & 8.358 $\pm$ 2.151 \\
~& MIDL~\cite{reddy2019brain} &  0.912 $\pm$ 0.008  & 32.157 $\pm$ 16.270  &  6.405 $\pm$ 0.524  & 6.377 $\pm$ 1.661 \\
~& NonAdj~\cite{ganaye2019removing} &  0.916 $\pm$ 0.030  &  32.465 $\pm$ 18.848  &  4.771 $\pm$ 1.129  & 4.932 $\pm$ 1.479 \\
~& Ours (6conn) &  0.920 $\pm$ 0.006  & 29.693 $\pm$ 15.746  & 4.269 $\pm$ 0.995  & 3.706 $\pm$ 1.274 \\
~& Ours (26conn) &  \textbf{0.922 $\pm$ 0.009} & \textbf{25.959 $\pm$ 13.574} & \textbf{3.920 $\pm$	0.765} & \textbf{3.526 $\pm$ 1.244}\\
 \hline
\multirow{10}{*}{\textbf{{\rotatebox[origin=c]{90}{Wall}}}}
& UNet~\cite{cciccek20163d} & 0.677 $\pm$ 0.015 & 71.109 $\pm$ 24.653 & 12.497 $\pm$	1.372 & /\\
~& UNet~\cite{cciccek20163d} + Ours & \textbf{0.737 $\pm$ 0.024}  & \textbf{44.372 $\pm$ 11.702}  & \textbf{7.289 $\pm$ 0.792}  & / \\\cline{2-6}
~& FCN~\cite{FCN8s}  & 0.651 $\pm$ 0.015 & 66.059 $\pm$ 17.188 & 12.339 $\pm$ 0.959 & /\\
~& FCN~\cite{FCN8s} + Ours &\textbf{ 0.681 $\pm$ 0.023}  & \textbf{50.068 $\pm$ 4.469}  & \textbf{9.530 $\pm$ 1.275}  & / \\\cline{2-6}
~& nnUNet~\cite{nnUNet} & 0.741 $\pm$ 0.026 & 42.486 $\pm$ 15.139 & 8.005 $\pm$	0.811 & /\\
~& Topo-CRF~\cite{bentaieb2016topology} &  0.739 $\pm$ 0.010 &  46.873 $\pm$ 17.636  &  7.914 $\pm$ 0.877 & / \\
~& MIDL~\cite{reddy2019brain} & 0.742 $\pm$ 0.028  & 43.132 $\pm$ 15.624  & 6.420  $\pm$ 1.242  & / \\
~& NonAdj~\cite{ganaye2019removing} & 0.748 $\pm$ 0.017  & 38.197 $\pm$ 19.598  & 4.887 $\pm$ 0.702  & / \\
~& Ours (6conn) & 0.753 $\pm$ 0.015  & 35.977 $\pm$ 17.358  & 4.200 $\pm$ 0.738  & / \\
~& Ours (26conn) &  \textbf{0.758 $\pm$ 0.017} & \textbf{31.137 $\pm$ 17.772} & \textbf{5.799 $\pm$ 0.737 }& /\\
 \hline
\end{tabular}
\end{table*}

  \setlength{\tabcolsep}{2pt}
\begin{table*}
  \centering
  \scriptsize
\caption{Quantitative comparison for IVUS dataset (containment constraint)}
\label{comp-ivus}
\begin{tabular}{c c c c c c} 
 \hline
 \textbf{Class} & \textbf{Model} & \textbf{Dice}$\uparrow$ & \textbf{HD}$\downarrow$ & \textbf{ASSD}$\downarrow$ & \textbf{\% Violations}$\downarrow$ \\
 \hline \hline
\multirow{10}{*}{\textbf{{\rotatebox[origin=c]{90}{Lumen}}}}
 & UNet~\cite{unet2d} & 0.786 $\pm$ 0.144 & 6.643 $\pm$ 1.936 & 30.944 $\pm$ 11.631 & 5.970 $\pm$ 2.141  \\
 ~& UNet~\cite{unet2d} + Ours &  \textbf{0.843 $\pm$ 0.128}  & \textbf{ 4.258 $\pm$ 1.612}  & \textbf{ 21.597 $\pm$ 9.138}  &  \textbf{2.042 $\pm$ 1.320}    \\\cline{2-6}
~ & FCN~\cite{FCN8s}  &  0.824 $\pm$ 0.071 & 5.319 $\pm$ 1.519 & 22.551 $\pm$ 7.882 & 3.766 $\pm$ 1.444\\
~ & FCN~\cite{FCN8s} + Ours &  \textbf{0.871 $\pm$ 0.082}  &  \textbf{3.976 $\pm$ 1.207}  &  \textbf{11.531 $\pm$ 4.736 } & \textbf{ 1.752 $\pm$ 1.105}    \\\cline{2-6}
~ & nnUNet~\cite{nnUNet} & 0.893 $\pm$ 0.066 & 3.464 $\pm$ 0.917 & 11.152 $\pm$ 3.954 & 2.708 $\pm$ 1.032\\ 
~ & Topo-CRF~\cite{bentaieb2016topology} &  0.887 $\pm$ 0.096  &  4.138 $\pm$ 1.454  &  10.497 $\pm$ 2.487 & 2.371 $\pm$ 0.960 \\
~& MIDL~\cite{reddy2019brain} &  0.891 $\pm$ 0.073  &  4.226 $\pm$ 1.390  &  10.641 $\pm$ 2.322  &  2.394 $\pm$ 0.918    \\
~& NonAdj~\cite{ganaye2019removing} &  0.897 $\pm$ 0.081  & 3.140  $\pm$ 1.154  &  9.628 $\pm$ 3.221  & 2.173 $\pm$ 0.994    \\
~& Ours (4conn) &  0.912 $\pm$ 0.087  &  2.857 $\pm$ 0.949  &  6.710 $\pm$ 3.186  &  0.311 $\pm$ 0.927    \\
~ & Ours (8conn) &  \textbf{0.949 $\pm$ 0.070} & \textbf{2.046 $\pm$ 1.079} & \textbf{6.057 $\pm$ 2.746} & \textbf{0.157 $\pm$ 0.808}\\
 \hline
\multirow{10}{*}{\textbf{{\rotatebox[origin=c]{90}{Media}}}}
 & UNet~\cite{unet2d} &  0.651 $\pm$ 0.130 & 7.391 $\pm$ 1.072 & 21.984 $\pm$ 6.634 & /\\
  ~& UNet~\cite{unet2d} + Ours &  \textbf{0.688 $\pm$ 0.115}  &  \textbf{7.012 $\pm$ 0.983}  &  \textbf{18.651 $\pm$ 5.776}  & / \\\cline{2-6}
~ & FCN~\cite{FCN8s}  &  0.782 $\pm$ 0.144 & 6.806 $\pm$ 1.147 & 13.863 $\pm$ 4.511 & / \\
~& FCN~\cite{FCN8s} + Ours & \textbf{0.809 $\pm$ 0.127}  & \textbf{ 6.137 $\pm$ 1.093} & \textbf{9.115 $\pm$ 3.689}  &  / \\\cline{2-6}
~ & nnUNet~\cite{nnUNet} & 0.856 $\pm$ 0.090 & 5.646 $\pm$ 1.228 & 6.491 $\pm$ 2.314 & /\\ 
~ & Topo-CRF~\cite{bentaieb2016topology} &  0.843 $\pm$ 0.106  &  5.409 $\pm$ 1.166  &  5.929 $\pm$ 1.785 & / \\
~& MIDL~\cite{reddy2019brain} &   0.841 $\pm$ 0.121  &  5.461 $\pm$ 1.214  &  6.071 $\pm$ 1.837  & / \\
~& NonAdj~\cite{ganaye2019removing} &  0.848$\pm$ 0.117  & 5.983  $\pm$ 1.342  &  6.615 $\pm$ 1.937  & / \\
~& Ours (4conn) & 0.884 $\pm$ 0.094  & 4.188 $\pm$ 1.156  &  3.622 $\pm$ 2.008  & /  \\
~ &Ours (8conn) &  \textbf{0.910 $\pm$ 0.089} & \textbf{3.873 $\pm$ 0.933} & \textbf{3.171 $\pm$ 1.871} & /\\
 \hline
\end{tabular}
\end{table*}


\setlength{\tabcolsep}{2pt}
\begin{table}[t]
  \centering
  \scriptsize
\caption{Quantitative comparison for Multi-Atlas dataset (exclusion constraint)}
\label{comp-multi}
\begin{tabular}{c c c c c c} 
 \hline
 \textbf{Class} & \textbf{Model} & \textbf{Dice}$\uparrow$ & \textbf{HD}$\downarrow$ & \textbf{ASSD}$\downarrow$ & \textbf{\% Violations}$\downarrow$ \\
 \hline \hline
\multirow{10}{*}{\textbf{{\rotatebox[origin=c]{90}{Spleen}}}}
& UNet~\cite{cciccek20163d} &  $0.919 \pm 0.041$ & $47.037 \pm 17.365$ & $4.323 \pm 0.367$  &  $1.857 \pm 0.123$  \\
~& UNet~\cite{cciccek20163d} + Ours &  \textit{0.932 $\pm$ 0.059}  &  \textbf{34.445 $\pm$ 10.684}  &  \textbf{2.020 $\pm$ 0.218}  & \textbf{1.256 $\pm$ 0.153} \\\cline{2-6}
~& FCN~\cite{FCN8s}  & $0.909 \pm 0.037$ & $134.915 \pm 65.623$ & $17.646 \pm 10.604$ &  $3.041 \pm 0.181$  \\
~& FCN~\cite{FCN8s} + Ours &  \textbf{0.927 $\pm$ 0.011}  & \textbf{ 66.407 $\pm$ 9.946}  & \textbf{9.038 $\pm$ 2.146}  & \textbf{2.680 $\pm$ 0.128 }\\\cline{2-6}
~& nnUNet~\cite{nnUNet} &  $0.950 \pm 0.041$ & $6.084 \pm 1.078$ & $0.573 \pm 0.131$ & $0.819 \pm 0.064$ \\
~& Topo-CRF~\cite{bentaieb2016topology} & 0.947  $\pm$ 0.028  & 6.403  $\pm$ 1.039  & 1.844   $\pm$ 0.517 & 0.934 $\pm$ 0.032 \\
~& MIDL~\cite{reddy2019brain} &  0.944 $\pm$ 0.015  &  5.597 $\pm$ 1.374  & 0.565  $\pm$ 0.124  & 0.725 $\pm$ 0.151 \\
~& NonAdj~\cite{ganaye2019removing} & 0.952 $\pm$ 0.058  & 5.621 $\pm$ 1.065  &  0.513 $\pm$ 0.175  & 0.521 $\pm$ 0.082  \\
~& Ours (6conn) &  0.957 $\pm$ 0.023  & 5.395 $\pm$ 1.057  &  0.498 $\pm$ 0.127  & 0.486 $\pm$ 0.075 \\
~& Ours (26conn) & \textit{0.960 $\pm$ 0.009} &  \textbf{5.340 $\pm$ 1.049} & \textbf{0.484 $\pm$ 0.109} & \textbf{0.464 $\pm$ 0.043}\\
 \hline

\multirow{10}{*}{\textbf{{\rotatebox[origin=c]{90}{Kidney}}}}
& UNet~\cite{cciccek20163d} & $0.908 \pm 0.079$ & $61.602 \pm 13.168$ & $9.992 \pm 2.461$ &/\\
~& UNet~\cite{cciccek20163d} + Ours &  \textit{0.921 $\pm$ 0.023}  &  \textbf{42.525 $\pm$ 10.103}  &  \textbf{6.446 $\pm$ 1.404}  & / \\\cline{2-6}
~& FCN~\cite{FCN8s}  & $0.892 \pm 0.018$ &	$187.472 \pm 36.096$ & $11.583 \pm 2.396$ &/\\
~& FCN~\cite{FCN8s} + Ours & \textbf{0.916 $\pm$ 0.014}  & \textbf{93.283 $\pm$ 10.293}  & \textbf{8.675  $\pm$ 1.129}  & / \\\cline{2-6}
~& nnUNet~\cite{nnUNet} & $0.931 \pm 0.018$ &	$27.252 \pm 5.406$ &	$5.352 \pm 0.199$ &/\\
~& Topo-CRF~\cite{bentaieb2016topology} & 0.928 $\pm$ 0.059  &  30.209 $\pm$ 5.317  & 6.308  $\pm$ 0.905 & / \\
~& MIDL~\cite{reddy2019brain} & 0.935  $\pm$ 0.071  & 25.208  $\pm$ 5.440  &  4.885 $\pm$ 0.421  & / \\
~& NonAdj~\cite{ganaye2019removing} & 0.934 $\pm$ 0.012  & 24.182  $\pm$ 5.561  & 4.692 $\pm$ 0.657  & / \\
~& Ours (6conn) &  0.932 $\pm$ 0.013  & 23.176 $\pm$ 3.593  & 4.540 $\pm$ 0.883  & / \\
~& Ours (26conn) & \textit{0.936 $\pm$ 0.026} & \textbf{20.013 $\pm$ 2.785} & \textbf{4.298 $\pm$ 0.798} &/\\
 \hline

\multirow{10}{*}{\textbf{{\rotatebox[origin=c]{90}{Liver}}}}
& UNet~\cite{cciccek20163d} & $0.912 \pm 0.016$ & $64.556 \pm 13.894$ & $2.324 \pm 0.513$ &/\\
~& UNet~\cite{cciccek20163d} + Ours & \textbf{0.941  $\pm$ 0.038}  & \textbf{46.174  $\pm$ 11.744}  & \textbf{1.452  $\pm$ 0.717}  & / \\\cline{2-6}
~& FCN~\cite{FCN8s} & $0.885 \pm 0.034$	& $183.870 \pm 49.796$ & $29.061 \pm 13.484$ &/\\
~& FCN~\cite{FCN8s} + Ours & \textbf{ 0.937 $\pm$ 0.013 } & \textbf{117.200  $\pm$ 16.663}  & \textbf{ 7.324  $\pm$ 5.201}  & / \\\cline{2-6}
~& nnUNet~\cite{nnUNet} & $0.951 \pm 0.008$ & $38.931 \pm 12.161$ & $1.922 \pm 0.506$ &/\\
~& Topo-CRF~\cite{bentaieb2016topology} & 0.949  $\pm$ 0.006  & 46.449 $\pm$ 14.188  &  2.072 $\pm$ 0.313 & / \\
~& MIDL~\cite{reddy2019brain} & 0.955  $\pm$ 0.005  & 34.276 $\pm$ 11.253  & 1.344  $\pm$ 0.431  & / \\
~& NonAdj~\cite{ganaye2019removing} & 0.957  $\pm$ 0.003  & 33.671  $\pm$ 13.543  & 1.185  $\pm$ 0.372  & / \\
~& Ours (6conn) & 0.958  $\pm$ 0.006  &  32.674 $\pm$ 12.566  & 1.098  $\pm$ 0.405  & / \\
~& Ours (26conn) & \textbf{0.962 $\pm$ 0.005} & \textbf{30.341 $\pm$ 9.111} & \textbf{0.985 $\pm$ 0.386} &/\\
 \hline

\multirow{10}{*}{\textbf{{\rotatebox[origin=c]{90}{Stomach}}}}
& UNet~\cite{cciccek20163d} & $0.846 \pm 0.084$ & $76.000 \pm 24.352$ & $5.023 \pm 1.508$ &/\\
~& UNet~\cite{cciccek20163d} + Ours & \textit{0.872  $\pm$ 0.074}  & \textbf{54.039  $\pm$ 19.131}  & \textbf{3.611  $\pm$ 1.301}  & / \\\cline{2-6}
~& FCN~\cite{FCN8s}   &  $0.708 \pm 0.156$ & $172.855 \pm 43.735$ & $11.328 \pm 3.178$ &/\\
~& FCN~\cite{FCN8s} + Ours &  \textit{0.799 $\pm$ 0.127}  & \textbf{104.331  $\pm$ 10.276}  & \textbf{6.892  $\pm$ 1.905}  & / \\\cline{2-6}
~& nnUNet~\cite{nnUNet} & $0.895 \pm 0.015$ & $45.767 \pm 7.960$ & $2.720 \pm 0.430$ &/\\
~& Topo-CRF~\cite{bentaieb2016topology} & 0.888 $\pm$ 0.015  & 46.877  $\pm$ 9.861  &  3.675 $\pm$ 0.358 & / \\
~& MIDL~\cite{reddy2019brain} & 0.899  $\pm$ 0.012  & 40.282  $\pm$ 6.437  &  2.567 $\pm$ 0.431  & / \\
~& NonAdj~\cite{ganaye2019removing} & 0.907  $\pm$ 0.028  & 41.749 $\pm$ 8.630  &  2.184 $\pm$ 0.325  & / \\
~& Ours (6conn) &  0.908 $\pm$ 0.017  & 39.853 $\pm$ 9.544  & 1.879 $\pm$ 0.587  & / \\
~& Ours (26conn) & \textbf{0.910 $\pm$ 0.018} & \textbf{35.514 $\pm$ 10.295} & \textbf{1.644 $\pm$ 0.311} &/\\
 \hline
  
\end{tabular}
\end{table}

 \setlength{\tabcolsep}{2pt}
\begin{table}[t]
  \centering
  \scriptsize
\caption{Quantitative comparison for SegTHOR dataset (exclusion constraint)}
\label{comp-seg}
\begin{tabular}{c c c c c c} 
 \hline
 \textbf{Class} & \textbf{Model} & \textbf{Dice}$\uparrow$ & \textbf{HD}$\downarrow$ & \textbf{ASSD}$\downarrow$ & \textbf{\% Violations}$\downarrow$ \\
 \hline \hline
\multirow{10}{*}{\textbf{{\rotatebox[origin=c]{90}{Esophagus}}}}
& UNet~\cite{cciccek20163d} & 0.827 $\pm$ 0.038 & 11.357	$\pm$ 2.709 & 1.186 $\pm$ 0.113 & 3.212 $\pm$	0.720 \\
~& UNet~\cite{cciccek20163d} + Ours &  \textit{0.841 $\pm$ 0.026}  &  \textbf{8.916 $\pm$ 2.437}  & \textbf{0.970  $\pm$ 0.124}  & \textbf{2.559 $\pm$ 0.412} \\\cline{2-6}
~ &FCN~\cite{FCN8s}  &  0.800 $\pm$ 0.031  &  10.770	$\pm$ 2.085  &  1.303 $\pm$ 0.128 & 3.616	$\pm$ 0.709  \\
~& FCN~\cite{FCN8s} + Ours & \textbf{0.839 $\pm$ 0.027}  & \textbf{9.055  $\pm$ 2.681}  & \textbf{0.986 $\pm$ 0.108}  & \textbf{2.889 $\pm$ 0.618} \\\cline{2-6}
~ &nnUNet~\cite{nnUNet} &  0.841 $\pm$ 0.014  &  8.018	$\pm$ 2.085  &  0.950 $\pm$ 0.070 & 1.947 $\pm$ 0.525  \\ 
~& Topo-CRF~\cite{bentaieb2016topology} &  0.839 $\pm$ 0.029  & 8.602 $\pm$ 2.363  &  0.991 $\pm$	0.081 & 2.070 $\pm$ 0.687 \\
~& MIDL~\cite{reddy2019brain} & 0.840 $\pm$ 0.020  & 7.266 $\pm$ 2.132  & 0.921 $\pm$ 0.136  & 1.271 $\pm$ 0.912 \\
~& NonAdj~\cite{ganaye2019removing} & 0.843 $\pm$ 0.020  & 6.293 $\pm$ 2.703  & 0.897 $\pm$ 0.078  & 1.215 $\pm$ 0.211 \\
~& Ours (6conn) & 0.849  $\pm$ 0.014  & 5.774 $\pm$ 2.371  & 0.832 $\pm$ 0.074  & 0.911 $\pm$ 0.565 \\
~ & Ours (26conn) &  \textbf{0.858 $\pm$ 0.019}  &  \textbf{5.582	$\pm$ 2.250}  &  \textbf{0.798 $\pm$ 0.042} & \textbf{0.749 $\pm$ 0.428}  \\
 \hline
 
 \multirow{10}{*}{\textbf{{\rotatebox[origin=c]{90}{Trachea}}}}
& UNet~\cite{cciccek20163d} &  0.897 $\pm$ 0.027  &  10.656 $\pm$	4.047  &  0.728 $\pm$ 	0.146 & /  \\
~& UNet~\cite{cciccek20163d} + Ours & \textit{0.908 $\pm$ 0.041}  & \textit{8.957 $\pm$ 3.338}  & \textbf{ 0.592  $\pm$ 0.167}  & / \\\cline{2-6}
~ &FCN~\cite{FCN8s}  &  0.891 $\pm$ 	0.031  &  11.789 $\pm$ 	5.291  &  0.953 $\pm$ 0.221 & /  \\
~& FCN~\cite{FCN8s} + Ours & \textit{0.896 $\pm$ 0.035}  & \textit{9.620 $\pm$ 2.805}  & \textbf{0.683  $\pm$ 0.245}  & / \\\cline{2-6}
~ &nnUNet~\cite{nnUNet} &  0.910 $\pm$  0.018  &  9.423 $\pm$ 	2.393  &  0.478 $\pm$ 0.152 & /  \\ 
~& Topo-CRF~\cite{bentaieb2016topology} &  0.909 $\pm$ 0.022  &  10.435 $\pm$ 2.334  &  0.473 $\pm$ 0.167 & / \\
~& MIDL~\cite{reddy2019brain} & 0.914 $\pm$ 0.027  & 7.929 $\pm$ 2.305  & 0.456 $\pm$ 0.144  & / \\
~& NonAdj~\cite{ganaye2019removing} & 0.913 $\pm$ 0.028  & 7.866 $\pm$ 2.343  & 0.440 $\pm$ 0.113  & / \\
~& Ours (6conn) & 0.922 $\pm$ 0.031  & 7.851 $\pm$ 2.846  & 0.417 $\pm$ 0.157  & / \\
~ & Ours (26conn) & \textbf{ 0.929 $\pm$  0.020 }&  \textbf{7.280 $\pm$ 	2.109}  & \textbf{ 0.316 $\pm$ 0.186} & /  \\
 \hline
 
 \multirow{10}{*}{\textbf{{\rotatebox[origin=c]{90}{Aorta}}}}
& UNet~\cite{cciccek20163d} &  0.929 $\pm$ 0.020 & 9.716 $\pm$ 4.032  &  0.714 $\pm$ 0.293 & /  \\
~& UNet~\cite{cciccek20163d} + Ours & \textit{0.932 $\pm$ 0.029}  & \textbf{6.553  $\pm$ 3.932}  & \textit{0.697  $\pm$ 0.218}  & / \\\cline{2-6}
~ &FCN~\cite{FCN8s}  & 0.924  $\pm$ 0.021  &  9.869	$\pm$ 4.739  &  0.726 $\pm$	0.424 & / \\
~& FCN~\cite{FCN8s} + Ours & \textit{0.929 $\pm$ 0.025}  &  \textbf{6.751 $\pm$ 3.810}  & \textit{0.705  $\pm$ 0.263}  & / \\\cline{2-6}
~& nnUNet~\cite{nnUNet} & 0.935 $\pm$	0.017  & 5.353	 $\pm$ 2.698  & 0.658  $\pm$	0.177 & / \\
~& Topo-CRF~\cite{bentaieb2016topology} & 0.932	 $\pm$ 0.018  &  5.361 $\pm$ 2.763 &  0.690 $\pm$ 0.225 & / \\
~& MIDL~\cite{reddy2019brain} & 0.937 $\pm$ 0.016  & 5.349 $\pm$ 2.458  & 0.668 $\pm$ 0.128  & / \\
~& NonAdj~\cite{ganaye2019removing} & 0.939 $\pm$ 0.021  & 5.060 $\pm$ 2.345  &  0.638 $\pm$ 0.192  & / \\
~& Ours (6conn) & 0.940  $\pm$ 0.017  & 4.840  $\pm$ 2.859  & 0.621  $\pm$ 0.175  & / \\
~ & Ours (26conn) & \textit{0.942  $\pm$ 0.018}  &  \textit{4.758 $\pm$ 2.127} &  \textit{0.606 $\pm$ 0.214} & /  \\
 \hline

 \multirow{10}{*}{\textbf{{\rotatebox[origin=c]{90}{Heart}}}}
& UNet~\cite{cciccek20163d} &  0.948 $\pm$ 0.012  &  8.235 $\pm$ 4.382  &  1.158 $\pm$ 0.571 & /  \\
~& UNet~\cite{cciccek20163d} + Ours &  \textit{0.953 $\pm$ 0.013}  & \textit{7.454  $\pm$ 4.602}  & \textit{1.022  $\pm$ 0.633}  & / \\\cline{2-6}
~ &FCN~\cite{FCN8s}  & 0.948  $\pm$ 0.014  &  8.556 $\pm$	4.302  &  2.206 $\pm$ 0.905 & / \\
~& FCN~\cite{FCN8s} + Ours & \textit{0.950 $\pm$ 0.018}  & \textit{8.085 $\pm$ 4.637}  & \textbf{1.543  $\pm$ 0.596}  & / \\\cline{2-6}
~& nnUNet~\cite{nnUNet} & 0.956 $\pm$	0.014  &  7.732 $\pm$ 4.327  &  0.895 $\pm$ 0.328 & /  \\
~& Topo-CRF~\cite{bentaieb2016topology} &  0.954 $\pm$ 0.016 &  7.936 $\pm$ 4.665  & 1.022  $\pm$ 0.434 & / \\
~& MIDL~\cite{reddy2019brain} & 0.952 $\pm$ 0.014  & 7.615 $\pm$ 4.991  & 0.889 $\pm$ 0.371  & / \\
~& NonAdj~\cite{ganaye2019removing} & 0.956 $\pm$ 0.016  & 7.363 $\pm$ 4.609  & 0.895 $\pm$  0.382  & / \\
~& Ours (6conn) & 0.958 $\pm$ 0.013  & 7.316 $\pm$ 4.129  & 0.874 $\pm$ 0.372  & / \\
~ & Ours (26conn) &  \textit{0.959 $\pm$ 0.012} &  \textit{7.158 $\pm$ 4.355}  & \textit{0.871 $\pm$	0.363} & /  \\
 \hline 
 
\end{tabular}
\end{table}

\clearpage

\bibliographystyle{splncs04}

\bibliography{refs}

\begin{thebibliography}{10}
\providecommand{\url}[1]{\texttt{#1}}
\providecommand{\urlprefix}{URL }
\providecommand{\doi}[1]{https://doi.org/#1}

\bibitem{balocco2014standardized}
Balocco, S., Gatta, C., Ciompi, F., Wahle, A., Radeva, P., Carlier, S., Unal,
  G., Sanidas, E., Mauri, J., Carillo, X., et~al.: Standardized evaluation
  methodology and reference database for evaluating ivus image segmentation.
  Computerized medical imaging and graphics  \textbf{38}(2),  70--90 (2014)

\bibitem{bentaieb2016topology}
BenTaieb, A., Hamarneh, G.: Topology aware fully convolutional networks for
  histology gland segmentation. In: International conference on medical image
  computing and computer-assisted intervention. pp. 460--468. Springer (2016)

\bibitem{chen2011enforcing}
Chen, C., Freedman, D., Lampert, C.H.: Enforcing topological constraints in
  random field image segmentation. In: CVPR (2011)

\bibitem{chen2014semantic}
Chen, L.C., Papandreou, G., Kokkinos, I., Murphy, K., Yuille, A.L.: Semantic
  image segmentation with deep convolutional nets and fully connected crfs.
  arXiv preprint arXiv:1412.7062  (2014)

\bibitem{chen2017deeplab}
Chen, L.C., Papandreou, G., Kokkinos, I., Murphy, K., Yuille, A.L.: Deeplab:
  Semantic image segmentation with deep convolutional nets, atrous convolution,
  and fully connected crfs. IEEE transactions on pattern analysis and machine
  intelligence  \textbf{40}(4),  834--848 (2017)

\bibitem{chen2017rethinking}
Chen, L.C., Papandreou, G., Schroff, F., Adam, H.: Rethinking atrous
  convolution for semantic image segmentation. arXiv preprint arXiv:1706.05587
  (2017)

\bibitem{cciccek20163d}
{\c{C}}i{\c{c}}ek, {\"O}., Abdulkadir, A., Lienkamp, S.S., Brox, T.,
  Ronneberger, O.: 3d u-net: learning dense volumetric segmentation from sparse
  annotation. In: MICCAI (2016)

\bibitem{clough2020topological}
Clough, J., Byrne, N., Oksuz, I., Zimmer, V., Schnabel, J., King, A.: A
  topological loss function for deep-learning based image segmentation using
  persistent homology. TPAMI  (2020)

\bibitem{colliot2006integration}
Colliot, O., Camara, O., Bloch, I.: Integration of fuzzy spatial relations in
  deformable models—application to brain mri segmentation. Pattern
  recognition  \textbf{39}(8),  1401--1414 (2006)

\bibitem{delong2009globally}
Delong, A., Boykov, Y.: Globally optimal segmentation of multi-region objects.
  In: 2009 IEEE 12th International Conference on Computer Vision. pp. 285--292.
  IEEE (2009)

\bibitem{doweidar2019advances}
Doweidar, M.H.: Advances in Biomechanics and Tissue Regeneration. Academic
  Press (2019)

\bibitem{felzenszwalb2010tiered}
Felzenszwalb, P.F., Veksler, O.: Tiered scene labeling with dynamic
  programming. In: 2010 IEEE Computer Society Conference on Computer Vision and
  Pattern Recognition. pp. 3097--3104. IEEE (2010)

\bibitem{gibsoneli20181169361}
Gibson, E., Giganti, F., Hu, Y., Bonmati, E., Bandula, S., Gurusamy, K.,
  Davidson, B., Pereira, S.P., Clarkson, M.J., Barratt, D.C.: Multi-organ
  abdominal ct reference standard segmentations (feb 2018).
  \doi{10.5281/zenodo.1169361}

\bibitem{han2003topology}
Han, X., Xu, C., Prince, J.L.: A topology preserving level set method for
  geometric deformable models. TPAMI  (2003)

\bibitem{haralick1987image}
Haralick, R.M., Sternberg, S.R., Zhuang, X.: Image analysis using mathematical
  morphology. TPAMI  (1987)

\bibitem{he2017mask}
He, K., Gkioxari, G., Doll{\'a}r, P., Girshick, R.: Mask r-cnn. In: Proceedings
  of the IEEE international conference on computer vision. pp. 2961--2969
  (2017)

\bibitem{heimann2009comparison}
Heimann, T., Van~Ginneken, B., Styner, M.A., Arzhaeva, Y., Aurich, V., Bauer,
  C., Beck, A., Becker, C., Beichel, R., Bekes, G., et~al.: Comparison and
  evaluation of methods for liver segmentation from ct datasets. IEEE
  transactions on medical imaging  \textbf{28}(8),  1251--1265 (2009)

\bibitem{hu2019topology}
Hu, X., Li, F., Samaras, D., Chen, C.: Topology-preserving deep image
  segmentation. NeurIPS  (2019)

\bibitem{hu2021topology}
Hu, X., Wang, Y., Fuxin, L., Samaras, D., Chen, C.: Topology-aware segmentation
  using discrete morse theory. ICLR  (2021)

\bibitem{huttenlocher1993comparing}
Huttenlocher, D.P., Klanderman, G.A., Rucklidge, W.J.: Comparing images using
  the hausdorff distance. IEEE Transactions on pattern analysis and machine
  intelligence  \textbf{15}(9),  850--863 (1993)

\bibitem{nnUNet}
Isensee, F., Jaeger, P.F., Kohl, S.A., Petersen, J., Maier-Hein, K.H.: nnu-net:
  a self-configuring method for deep learning-based biomedical image
  segmentation. Nature methods  (2021)

\bibitem{kappes2016higher}
Kappes, J.H., Speth, M., Reinelt, G., Schn{\"o}rr, C.: Higher-order
  segmentation via multicuts. Computer Vision and Image Understanding
  \textbf{143},  104--119 (2016)

\bibitem{lambert2019segthor}
Lambert, Z., Petitjean, C., Dubray, B., Ruan, S.: Segthor: Segmentation of
  thoracic organs at risk in ct images (2019)

\bibitem{landman2015miccai}
Landman, B., Xu, Z., Igelsias, J., Styner, M., Langerak, T., Klein, A.: Miccai
  multi-atlas labeling beyond the cranial vault--workshop and challenge. In:
  Proc. MICCAI Multi-Atlas Labeling Beyond Cranial Vault—Workshop Challenge.
  vol.~5, p.~12 (2015)

\bibitem{le2008self}
Le~Guyader, C., Vese, L.A.: Self-repelling snakes for topology-preserving
  segmentation models. TIP  (2008)

\bibitem{leon2017multi}
Leon, L.M.C., De~Miranda, P.A.V.: Multi-object segmentation by hierarchical
  layered oriented image foresting transform. In: 2017 30th SIBGRAPI Conference
  on Graphics, Patterns and Images (SIBGRAPI). pp. 79--86. IEEE (2017)

\bibitem{li2005optimal}
Li, K., Wu, X., Chen, D.Z., Sonka, M.: Optimal surface segmentation in
  volumetric images-a graph-theoretic approach. IEEE transactions on pattern
  analysis and machine intelligence  \textbf{28}(1),  119--134 (2005)

\bibitem{litvin2005coupled}
Litvin, A., Karl, W.C.: Coupled shape distribution-based segmentation of
  multiple objects. In: Biennial International Conference on Information
  Processing in Medical Imaging. pp. 345--356. Springer (2005)

\bibitem{long2015fully}
Long, J., Shelhamer, E., Darrell, T.: Fully convolutional networks for semantic
  segmentation. In: Proceedings of the IEEE conference on computer vision and
  pattern recognition. pp. 3431--3440 (2015)

\bibitem{FCN8s}
Long, J., Shelhamer, E., Darrell, T.: Fully convolutional networks for semantic
  segmentation. In: CVPR (2015)

\bibitem{nosrati2014local}
Nosrati, M.S., Hamarneh, G.: Local optimization based segmentation of
  spatially-recurring, multi-region objects with part configuration
  constraints. IEEE transactions on medical imaging  \textbf{33}(9),
  1845--1859 (2014)

\bibitem{ganaye2019removing}
Pierre-Antoine, et~al., G.: Removing segmentation inconsistencies with
  semi-supervised non-adjacency constraint. MedIA  (2019)

\bibitem{reddy2019brain}
Reddy, C., Gopinath, K., Lombaert, H.: Brain tumor segmentation using
  topological loss in convolutional networks. In: MIDL (2019)

\bibitem{unet2d}
Ronneberger, O., Fischer, P., Brox, T.: U-net: Convolutional networks for
  biomedical image segmentation. In: MICCAI (2015)

\bibitem{rosenfeld1979digital}
Rosenfeld, A.: Digital topology. The American Mathematical Monthly
  \textbf{86}(8),  621--630 (1979)

\bibitem{shit2021cldice}
Shit, S., Paetzold, J.C., Sekuboyina, A., Ezhov, I., Unger, A., Zhylka, A.,
  Pluim, J.P., Bauer, U., Menze, B.H.: cldice-a novel topology-preserving loss
  function for tubular structure segmentation. In: CVPR (2021)

\bibitem{strekalovskiy2011generalized}
Strekalovskiy, E., Cremers, D.: Generalized ordering constraints for multilabel
  optimization. In: 2011 International Conference on Computer Vision. pp.
  2619--2626. IEEE (2011)

\bibitem{student1908probable}
Student: The probable error of a mean. Biometrika pp. 1--25 (1908)

\bibitem{ulen2012efficient}
Ul{\'e}n, J., Strandmark, P., Kahl, F.: An efficient optimization framework for
  multi-region segmentation based on lagrangian duality. IEEE transactions on
  medical imaging  \textbf{32}(2),  178--188 (2012)

\bibitem{yang2021topological}
Yang, J., Hu, X., Chen, C., Tsai, C.: A topological-attention convlstm network
  and its application to em images. In: International Conference on Medical
  Image Computing and Computer-Assisted Intervention. pp. 217--228. Springer
  (2021)

\bibitem{zou2004statistical}
Zou, K.H., Warfield, S.K., Bharatha, A., Tempany, C.M., Kaus, M.R., Haker,
  S.J., Wells~III, W.M., Jolesz, F.A., Kikinis, R.: Statistical validation of
  image segmentation quality based on a spatial overlap index1: scientific
  reports. Academic radiology  \textbf{11}(2),  178--189 (2004)

\end{thebibliography}
\end{document}